\documentclass[11pt]{report}
\usepackage{thesis}

\renewcommand{\arraystretch}{0.67}
\newcounter{treecount}
\newcounter{branchcount}
\newsavebox{\parentbox}
\newsavebox{\treebox}
\newsavebox{\treeboxone}
\newsavebox{\treeboxtwo}
\newsavebox{\treeboxthree}
\newsavebox{\treeboxfour}
\newsavebox{\treeboxfive}
\newsavebox{\treeboxsix}
\newsavebox{\treeboxseven}
\newsavebox{\treeboxeight}
\newsavebox{\treeboxnine}
\newsavebox{\treeboxten}
\newsavebox{\treeboxeleven}
\newsavebox{\treeboxtwelve}
\newsavebox{\treeboxthirteen}
\newsavebox{\treeboxfourteen}
\newsavebox{\treeboxfifteen}
\newsavebox{\treeboxsixteen}
\newsavebox{\treeboxseventeen}
\newsavebox{\treeboxeighteen}
\newsavebox{\treeboxnineteen}
\newsavebox{\treeboxtwenty}
\newlength{\treeoffsetone}
\newlength{\treeoffsettwo}
\newlength{\treeoffsetthree}
\newlength{\treeoffsetfour}
\newlength{\treeoffsetfive}
\newlength{\treeoffsetsix}
\newlength{\treeoffsetseven}
\newlength{\treeoffseteight}
\newlength{\treeoffsetnine}
\newlength{\treeoffsetten}
\newlength{\treeoffseteleven}
\newlength{\treeoffsettwelve}
\newlength{\treeoffsetthirteen}
\newlength{\treeoffsetfourteen}
\newlength{\treeoffsetfifteen}
\newlength{\treeoffsetsixteen}
\newlength{\treeoffsetseventeen}
\newlength{\treeoffseteighteen}
\newlength{\treeoffsetnineteen}
\newlength{\treeoffsettwenty}

\newlength{\treeshiftone}
\newlength{\treeshifttwo}
\newlength{\treeshiftthree}
\newlength{\treeshiftfour}
\newlength{\treeshiftfive}
\newlength{\treeshiftsix}
\newlength{\treeshiftseven}
\newlength{\treeshifteight}
\newlength{\treeshiftnine}
\newlength{\treeshiftten}
\newlength{\treeshifteleven}
\newlength{\treeshifttwelve}
\newlength{\treeshiftthirteen}
\newlength{\treeshiftfourteen}
\newlength{\treeshiftfifteen}
\newlength{\treeshiftsixteen}
\newlength{\treeshiftseventeen}
\newlength{\treeshifteighteen}
\newlength{\treeshiftnineteen}
\newlength{\treeshifttwenty}
\newlength{\treewidthone}
\newlength{\treewidthtwo}
\newlength{\treewidththree}
\newlength{\treewidthfour}
\newlength{\treewidthfive}
\newlength{\treewidthsix}
\newlength{\treewidthseven}
\newlength{\treewidtheight}
\newlength{\treewidthnine}
\newlength{\treewidthten}
\newlength{\treewidtheleven}
\newlength{\treewidthtwelve}
\newlength{\treewidththirteen}
\newlength{\treewidthfourteen}
\newlength{\treewidthfifteen}
\newlength{\treewidthsixteen}
\newlength{\treewidthseventeen}
\newlength{\treewidtheighteen}
\newlength{\treewidthnineteen}
\newlength{\treewidthtwenty}
\newlength{\daughteroffsetone}
\newlength{\daughteroffsettwo}
\newlength{\daughteroffsetthree}
\newlength{\daughteroffsetfour}
\newlength{\branchwidthone}
\newlength{\branchwidthtwo}
\newlength{\branchwidththree}
\newlength{\branchwidthfour}
\newlength{\parentoffset}
\newlength{\treeoffset}
\newlength{\daughteroffset}
\newlength{\branchwidth}
\newlength{\parentwidth}
\newlength{\treewidth}
\newcommand{\ontop}[1]{\begin{tabular}{c}#1\end{tabular}}
\newcommand{\poptree}{%
\ifnum\value{treecount}=0\typeout{QobiTeX warning---Tree stack underflow}\fi%
\addtocounter{treecount}{-1}%
\setlength{\treeoffsettwo}{\treeoffsetthree}%
\setlength{\treeoffsetthree}{\treeoffsetfour}%
\setlength{\treeoffsetfour}{\treeoffsetfive}%
\setlength{\treeoffsetfive}{\treeoffsetsix}%
\setlength{\treeoffsetsix}{\treeoffsetseven}%
\setlength{\treeoffsetseven}{\treeoffseteight}%
\setlength{\treeoffseteight}{\treeoffsetnine}%
\setlength{\treeoffsetnine}{\treeoffsetten}%
\setlength{\treeoffsetten}{\treeoffseteleven}%
\setlength{\treeoffseteleven}{\treeoffsettwelve}%
\setlength{\treeoffsettwelve}{\treeoffsetthirteen}%
\setlength{\treeoffsetthirteen}{\treeoffsetfourteen}%
\setlength{\treeoffsetfourteen}{\treeoffsetfifteen}%
\setlength{\treeoffsetfifteen}{\treeoffsetsixteen}%
\setlength{\treeoffsetsixteen}{\treeoffsetseventeen}%
\setlength{\treeoffsetseventeen}{\treeoffseteighteen}%
\setlength{\treeoffseteighteen}{\treeoffsetnineteen}%
\setlength{\treeoffsetnineteen}{\treeoffsettwenty}%
\setlength{\treeshifttwo}{\treeshiftthree}%
\setlength{\treeshiftthree}{\treeshiftfour}%
\setlength{\treeshiftfour}{\treeshiftfive}%
\setlength{\treeshiftfive}{\treeshiftsix}%
\setlength{\treeshiftsix}{\treeshiftseven}%
\setlength{\treeshiftseven}{\treeshifteight}%
\setlength{\treeshifteight}{\treeshiftnine}%
\setlength{\treeshiftnine}{\treeshiftten}%
\setlength{\treeshiftten}{\treeshifteleven}%
\setlength{\treeshifteleven}{\treeshifttwelve}%
\setlength{\treeshifttwelve}{\treeshiftthirteen}%
\setlength{\treeshiftthirteen}{\treeshiftfourteen}%
\setlength{\treeshiftfourteen}{\treeshiftfifteen}%
\setlength{\treeshiftfifteen}{\treeshiftsixteen}%
\setlength{\treeshiftsixteen}{\treeshiftseventeen}%
\setlength{\treeshiftseventeen}{\treeshifteighteen}%
\setlength{\treeshifteighteen}{\treeshiftnineteen}%
\setlength{\treeshiftnineteen}{\treeshifttwenty}%
\setlength{\treewidthtwo}{\treewidththree}%
\setlength{\treewidththree}{\treewidthfour}%
\setlength{\treewidthfour}{\treewidthfive}%
\setlength{\treewidthfive}{\treewidthsix}%
\setlength{\treewidthsix}{\treewidthseven}%
\setlength{\treewidthseven}{\treewidtheight}%
\setlength{\treewidtheight}{\treewidthnine}%
\setlength{\treewidthnine}{\treewidthten}%
\setlength{\treewidthten}{\treewidtheleven}%
\setlength{\treewidtheleven}{\treewidthtwelve}%
\setlength{\treewidthtwelve}{\treewidththirteen}%
\setlength{\treewidththirteen}{\treewidthfourteen}%
\setlength{\treewidthfourteen}{\treewidthfifteen}%
\setlength{\treewidthfifteen}{\treewidthsixteen}%
\setlength{\treewidthsixteen}{\treewidthseventeen}%
\setlength{\treewidthseventeen}{\treewidtheighteen}%
\setlength{\treewidtheighteen}{\treewidthnineteen}%
\setlength{\treewidthnineteen}{\treewidthtwenty}%
\sbox{\treeboxtwo}{\usebox{\treeboxthree}}%
\sbox{\treeboxthree}{\usebox{\treeboxfour}}%
\sbox{\treeboxfour}{\usebox{\treeboxfive}}%
\sbox{\treeboxfive}{\usebox{\treeboxsix}}%
\sbox{\treeboxsix}{\usebox{\treeboxseven}}%
\sbox{\treeboxseven}{\usebox{\treeboxeight}}%
\sbox{\treeboxeight}{\usebox{\treeboxnine}}%
\sbox{\treeboxnine}{\usebox{\treeboxten}}%
\sbox{\treeboxten}{\usebox{\treeboxeleven}}%
\sbox{\treeboxeleven}{\usebox{\treeboxtwelve}}%
\sbox{\treeboxtwelve}{\usebox{\treeboxthirteen}}%
\sbox{\treeboxthirteen}{\usebox{\treeboxfourteen}}%
\sbox{\treeboxfourteen}{\usebox{\treeboxfifteen}}%
\sbox{\treeboxfifteen}{\usebox{\treeboxsixteen}}%
\sbox{\treeboxsixteen}{\usebox{\treeboxseventeen}}%
\sbox{\treeboxseventeen}{\usebox{\treeboxeighteen}}%
\sbox{\treeboxeighteen}{\usebox{\treeboxnineteen}}%
\sbox{\treeboxnineteen}{\usebox{\treeboxtwenty}}}
\newcommand{\leaf}[1]{%
\ifnum\value{treecount}=20\typeout{QobiTeX warning---Tree stack overflow}\fi%
\addtocounter{treecount}{1}%
\sbox{\treeboxtwenty}{\usebox{\treeboxnineteen}}%
\sbox{\treeboxnineteen}{\usebox{\treeboxeighteen}}%
\sbox{\treeboxeighteen}{\usebox{\treeboxseventeen}}%
\sbox{\treeboxseventeen}{\usebox{\treeboxsixteen}}%
\sbox{\treeboxsixteen}{\usebox{\treeboxfifteen}}%
\sbox{\treeboxfifteen}{\usebox{\treeboxfourteen}}%
\sbox{\treeboxfourteen}{\usebox{\treeboxthirteen}}%
\sbox{\treeboxthirteen}{\usebox{\treeboxtwelve}}%
\sbox{\treeboxtwelve}{\usebox{\treeboxeleven}}%
\sbox{\treeboxeleven}{\usebox{\treeboxten}}%
\sbox{\treeboxten}{\usebox{\treeboxnine}}%
\sbox{\treeboxnine}{\usebox{\treeboxeight}}%
\sbox{\treeboxeight}{\usebox{\treeboxseven}}%
\sbox{\treeboxseven}{\usebox{\treeboxsix}}%
\sbox{\treeboxsix}{\usebox{\treeboxfive}}%
\sbox{\treeboxfive}{\usebox{\treeboxfour}}%
\sbox{\treeboxfour}{\usebox{\treeboxthree}}%
\sbox{\treeboxthree}{\usebox{\treeboxtwo}}%
\sbox{\treeboxtwo}{\usebox{\treeboxone}}%
\sbox{\treeboxone}{\ontop{#1}}%
\sbox{\treeboxone}{\raisebox{-\ht\treeboxone}{\usebox{\treeboxone}}}%
\setlength{\treeoffsettwenty}{\treeoffsetnineteen}%
\setlength{\treeoffsetnineteen}{\treeoffseteighteen}%
\setlength{\treeoffseteighteen}{\treeoffsetseventeen}%
\setlength{\treeoffsetseventeen}{\treeoffsetsixteen}%
\setlength{\treeoffsetsixteen}{\treeoffsetfifteen}%
\setlength{\treeoffsetfifteen}{\treeoffsetfourteen}%
\setlength{\treeoffsetfourteen}{\treeoffsetthirteen}%
\setlength{\treeoffsetthirteen}{\treeoffsettwelve}%
\setlength{\treeoffsettwelve}{\treeoffseteleven}%
\setlength{\treeoffseteleven}{\treeoffsetten}%
\setlength{\treeoffsetten}{\treeoffsetnine}%
\setlength{\treeoffsetnine}{\treeoffseteight}%
\setlength{\treeoffseteight}{\treeoffsetseven}%
\setlength{\treeoffsetseven}{\treeoffsetsix}%
\setlength{\treeoffsetsix}{\treeoffsetfive}%
\setlength{\treeoffsetfive}{\treeoffsetfour}%
\setlength{\treeoffsetfour}{\treeoffsetthree}%
\setlength{\treeoffsetthree}{\treeoffsettwo}%
\setlength{\treeoffsettwo}{\treeoffsetone}%
\setlength{\treeoffsetone}{0.5\wd\treeboxone}%
\setlength{\treeshifttwenty}{\treeshiftnineteen}%
\setlength{\treeshiftnineteen}{\treeshifteighteen}%
\setlength{\treeshifteighteen}{\treeshiftseventeen}%
\setlength{\treeshiftseventeen}{\treeshiftsixteen}%
\setlength{\treeshiftsixteen}{\treeshiftfifteen}%
\setlength{\treeshiftfifteen}{\treeshiftfourteen}%
\setlength{\treeshiftfourteen}{\treeshiftthirteen}%
\setlength{\treeshiftthirteen}{\treeshifttwelve}%
\setlength{\treeshifttwelve}{\treeshifteleven}%
\setlength{\treeshifteleven}{\treeshiftten}%
\setlength{\treeshiftten}{\treeshiftnine}%
\setlength{\treeshiftnine}{\treeshifteight}%
\setlength{\treeshifteight}{\treeshiftseven}%
\setlength{\treeshiftseven}{\treeshiftsix}%
\setlength{\treeshiftsix}{\treeshiftfive}%
\setlength{\treeshiftfive}{\treeshiftfour}%
\setlength{\treeshiftfour}{\treeshiftthree}%
\setlength{\treeshiftthree}{\treeshifttwo}%
\setlength{\treeshifttwo}{\treeshiftone}%
\setlength{\treeshiftone}{0pt}%
\setlength{\treewidthtwenty}{\treewidthnineteen}%
\setlength{\treewidthnineteen}{\treewidtheighteen}%
\setlength{\treewidtheighteen}{\treewidthseventeen}%
\setlength{\treewidthseventeen}{\treewidthsixteen}%
\setlength{\treewidthsixteen}{\treewidthfifteen}%
\setlength{\treewidthfifteen}{\treewidthfourteen}%
\setlength{\treewidthfourteen}{\treewidththirteen}%
\setlength{\treewidththirteen}{\treewidthtwelve}%
\setlength{\treewidthtwelve}{\treewidtheleven}%
\setlength{\treewidtheleven}{\treewidthten}%
\setlength{\treewidthten}{\treewidthnine}%
\setlength{\treewidthnine}{\treewidtheight}%
\setlength{\treewidtheight}{\treewidthseven}%
\setlength{\treewidthseven}{\treewidthsix}%
\setlength{\treewidthsix}{\treewidthfive}%
\setlength{\treewidthfive}{\treewidthfour}%
\setlength{\treewidthfour}{\treewidththree}%
\setlength{\treewidththree}{\treewidthtwo}%
\setlength{\treewidthtwo}{\treewidthone}%
\setlength{\treewidthone}{\wd\treeboxone}}
\newcommand{\branch}[2]{%
\setcounter{branchcount}{#1}%
\ifnum\value{branchcount}=1\sbox{\parentbox}{\ontop{#2}}%
\setlength{\parentoffset}{\treeoffsetone}%
\addtolength{\parentoffset}{-0.5\wd\parentbox}%
\setlength{\daughteroffset}{0in}%
\ifdim\parentoffset<0in%
\setlength{\daughteroffset}{-\parentoffset}%
\setlength{\parentoffset}{0in}\fi%
\setlength{\parentwidth}{\parentoffset}%
\addtolength{\parentwidth}{\wd\parentbox}%
\setlength{\treeoffset}{\daughteroffset}%
\addtolength{\treeoffset}{\treeoffsetone}%
\setlength{\treewidth}{\wd\treeboxone}%
\addtolength{\treewidth}{\daughteroffset}%
\ifdim\treewidth<\parentwidth\setlength{\treewidth}{\parentwidth}\fi%
\sbox{\treebox}{\begin{minipage}{\treewidth}%
\begin{flushleft}%
\hspace*{\parentoffset}\usebox{\parentbox}\\
{\setlength{\unitlength}{2ex}%
\hspace*{\treeoffset}\begin{picture}(0,1)%
\put(0,0){\line(0,1){1}}%
\end{picture}}\\
\vspace{-\baselineskip}
\hspace*{\daughteroffset}%
\raisebox{-\ht\treeboxone}{\usebox{\treeboxone}}%
\end{flushleft}%
\end{minipage}}%
\setlength{\treeoffsetone}{\parentoffset}%
\addtolength{\treeoffsetone}{0.5\wd\parentbox}%
\setlength{\treeshiftone}{0pt}%
\setlength{\treewidthone}{\treewidth}%
\sbox{\treeboxone}{\usebox{\treebox}}%
\else\ifnum\value{branchcount}=2\sbox{\parentbox}{\ontop{#2}}%
\setlength{\branchwidthone}{\treewidthtwo}%
\addtolength{\branchwidthone}{\treeoffsetone}%
\addtolength{\branchwidthone}{-\treeshiftone}%
\addtolength{\branchwidthone}{-\treeoffsettwo}%
\setlength{\branchwidth}{\branchwidthone}%
\setlength{\daughteroffsetone}{\branchwidth}%
\addtolength{\daughteroffsetone}{-\branchwidthone}%
\addtolength{\daughteroffsetone}{-\treeshiftone}%
\setlength{\parentoffset}{-0.5\wd\parentbox}%
\addtolength{\parentoffset}{\treeoffsettwo}%
\addtolength{\parentoffset}{0.5\branchwidth}%
\setlength{\daughteroffset}{0in}%
\ifdim\parentoffset<0in%
\setlength{\daughteroffset}{-\parentoffset}%
\setlength{\parentoffset}{0in}\fi%
\setlength{\parentwidth}{\parentoffset}%
\addtolength{\parentwidth}{\wd\parentbox}%
\setlength{\treeoffset}{\daughteroffset}%
\addtolength{\treeoffset}{\treeoffsettwo}%
\setlength{\treewidth}{\wd\treeboxone}%
\addtolength{\treewidth}{\daughteroffsetone}%
\addtolength{\treewidth}{\treewidthtwo}%
\addtolength{\treewidth}{\daughteroffset}%
\ifdim\treewidth<\parentwidth\setlength{\treewidth}{\parentwidth}\fi%
\sbox{\treebox}{\begin{minipage}{\treewidth}%
\begin{flushleft}%
\hspace*{\parentoffset}\usebox{\parentbox}\\
{\setlength{\unitlength}{0.5\branchwidth}%
\hspace*{\treeoffset}\begin{picture}(2,0.5)%
\put(0,0){\line(2,1){1}}%
\put(2,0){\line(-2,1){1}}%
\end{picture}}\\
\vspace{-\baselineskip}
\hspace*{\daughteroffset}%
\makebox[\treewidthtwo][l]%
{\raisebox{-\ht\treeboxtwo}{\usebox{\treeboxtwo}}}%
\hspace*{\daughteroffsetone}%
\raisebox{-\ht\treeboxone}{\usebox{\treeboxone}}%
\end{flushleft}%
\end{minipage}}%
\setlength{\treeoffsetone}{\parentoffset}%
\addtolength{\treeoffsetone}{0.5\wd\parentbox}%
\setlength{\treeshiftone}{0pt}%
\setlength{\treewidthone}{\treewidth}%
\sbox{\treeboxone}{\usebox{\treebox}}\poptree%
\else\ifnum\value{branchcount}=3\sbox{\parentbox}{\ontop{#2}}%
\setlength{\branchwidthone}{\treewidthtwo}%
\addtolength{\branchwidthone}{\treeoffsetone}%
\addtolength{\branchwidthone}{-\treeshiftone}%
\addtolength{\branchwidthone}{-\treeoffsettwo}%
\setlength{\branchwidthtwo}{\treewidththree}%
\addtolength{\branchwidthtwo}{\treeoffsettwo}%
\addtolength{\branchwidthtwo}{-\treeshifttwo}%
\addtolength{\branchwidthtwo}{-\treeoffsetthree}%
\setlength{\branchwidth}{\branchwidthone}%
\ifdim\branchwidthtwo>\branchwidth%
\setlength{\branchwidth}{\branchwidthtwo}\fi%
\setlength{\daughteroffsetone}{\branchwidth}%
\addtolength{\daughteroffsetone}{-\branchwidthone}%
\addtolength{\daughteroffsetone}{-\treeshiftone}%
\setlength{\daughteroffsettwo}{\branchwidth}%
\addtolength{\daughteroffsettwo}{-\branchwidthtwo}%
\addtolength{\daughteroffsettwo}{-\treeshifttwo}%
\setlength{\parentoffset}{-0.5\wd\parentbox}%
\addtolength{\parentoffset}{\treeoffsetthree}%
\addtolength{\parentoffset}{\branchwidth}%
\setlength{\daughteroffset}{0in}%
\ifdim\parentoffset<0in%
\setlength{\daughteroffset}{-\parentoffset}%
\setlength{\parentoffset}{0in}\fi%
\setlength{\parentwidth}{\parentoffset}%
\addtolength{\parentwidth}{\wd\parentbox}%
\setlength{\treeoffset}{\daughteroffset}%
\addtolength{\treeoffset}{\treeoffsetthree}%
\setlength{\treewidth}{\wd\treeboxone}%
\addtolength{\treewidth}{\daughteroffsetone}%
\addtolength{\treewidth}{\treewidthtwo}%
\addtolength{\treewidth}{\daughteroffsettwo}%
\addtolength{\treewidth}{\treewidththree}%
\addtolength{\treewidth}{\daughteroffset}%
\ifdim\treewidth<\parentwidth\setlength{\treewidth}{\parentwidth}\fi%
\sbox{\treebox}{\begin{minipage}{\treewidth}%
\begin{flushleft}%
\hspace*{\parentoffset}\usebox{\parentbox}\\
{\setlength{\unitlength}{0.5\branchwidth}%
\hspace*{\treeoffset}\begin{picture}(4,1)%
\put(0,0){\line(2,1){2}}%
\put(2,0){\line(0,1){1}}%
\put(4,0){\line(-2,1){2}}%
\end{picture}}\\
\vspace{-\baselineskip}
\hspace*{\daughteroffset}%
\makebox[\treewidththree][l]%
{\raisebox{-\ht\treeboxthree}{\usebox{\treeboxthree}}}%
\hspace*{\daughteroffsettwo}%
\makebox[\treewidthtwo][l]%
{\raisebox{-\ht\treeboxtwo}{\usebox{\treeboxtwo}}}%
\hspace*{\daughteroffsetone}%
\raisebox{-\ht\treeboxone}{\usebox{\treeboxone}}%
\end{flushleft}%
\end{minipage}}%
\setlength{\treeoffsetone}{\parentoffset}%
\addtolength{\treeoffsetone}{0.5\wd\parentbox}%
\setlength{\treeshiftone}{0pt}%
\setlength{\treewidthone}{\treewidth}%
\sbox{\treeboxone}{\usebox{\treebox}}\poptree\poptree%
\else\ifnum\value{branchcount}=4\sbox{\parentbox}{\ontop{#2}}%
\setlength{\branchwidthone}{\treewidthtwo}%
\addtolength{\branchwidthone}{\treeoffsetone}%
\addtolength{\branchwidthone}{-\treeshiftone}%
\addtolength{\branchwidthone}{-\treeoffsettwo}%
\setlength{\branchwidthtwo}{\treewidththree}%
\addtolength{\branchwidthtwo}{\treeoffsettwo}%
\addtolength{\branchwidthtwo}{-\treeshifttwo}%
\addtolength{\branchwidthtwo}{-\treeoffsetthree}%
\setlength{\branchwidththree}{\treewidthfour}%
\addtolength{\branchwidththree}{\treeoffsetthree}%
\addtolength{\branchwidththree}{-\treeshiftthree}%
\addtolength{\branchwidththree}{-\treeoffsetfour}%
\setlength{\branchwidth}{\branchwidthone}%
\ifdim\branchwidthtwo>\branchwidth%
\setlength{\branchwidth}{\branchwidthtwo}\fi%
\ifdim\branchwidththree>\branchwidth%
\setlength{\branchwidth}{\branchwidththree}\fi%
\setlength{\daughteroffsetone}{\branchwidth}%
\addtolength{\daughteroffsetone}{-\branchwidthone}%
\addtolength{\daughteroffsetone}{-\treeshiftone}%
\setlength{\daughteroffsettwo}{\branchwidth}%
\addtolength{\daughteroffsettwo}{-\branchwidthtwo}%
\addtolength{\daughteroffsettwo}{-\treeshifttwo}%
\setlength{\daughteroffsetthree}{\branchwidth}%
\addtolength{\daughteroffsetthree}{-\branchwidththree}%
\addtolength{\daughteroffsetthree}{-\treeshiftthree}%
\setlength{\parentoffset}{-0.5\wd\parentbox}%
\addtolength{\parentoffset}{\treeoffsetfour}%
\addtolength{\parentoffset}{1.5\branchwidth}%
\setlength{\daughteroffset}{0in}%
\ifdim\parentoffset<0in%
\setlength{\daughteroffset}{-\parentoffset}%
\setlength{\parentoffset}{0in}\fi%
\setlength{\parentwidth}{\parentoffset}%
\addtolength{\parentwidth}{\wd\parentbox}%
\setlength{\treeoffset}{\daughteroffset}%
\addtolength{\treeoffset}{\treeoffsetfour}%
\setlength{\treewidth}{\wd\treeboxone}%
\addtolength{\treewidth}{\daughteroffsetone}%
\addtolength{\treewidth}{\treewidthtwo}%
\addtolength{\treewidth}{\daughteroffsettwo}%
\addtolength{\treewidth}{\treewidththree}%
\addtolength{\treewidth}{\daughteroffsetthree}%
\addtolength{\treewidth}{\treewidthfour}%
\addtolength{\treewidth}{\daughteroffset}%
\ifdim\treewidth<\parentwidth\setlength{\treewidth}{\parentwidth}\fi%
\sbox{\treebox}{\begin{minipage}{\treewidth}%
\begin{flushleft}%
\hspace*{\parentoffset}\usebox{\parentbox}\\
{\setlength{\unitlength}{0.5\branchwidth}%
\hspace*{\treeoffset}\begin{picture}(6,1)%
\put(0,0){\line(3,1){3}}%
\put(2,0){\line(1,1){1}}%
\put(4,0){\line(-1,1){1}}%
\put(6,0){\line(-3,1){3}}%
\end{picture}}\\
\vspace{-\baselineskip}
\hspace*{\daughteroffset}%
\makebox[\treewidthfour][l]%
{\raisebox{-\ht\treeboxfour}{\usebox{\treeboxfour}}}%
\hspace*{\daughteroffsetthree}%
\makebox[\treewidththree][l]%
{\raisebox{-\ht\treeboxthree}{\usebox{\treeboxthree}}}%
\hspace*{\daughteroffsettwo}%
\makebox[\treewidthtwo][l]%
{\raisebox{-\ht\treeboxtwo}{\usebox{\treeboxtwo}}}%
\hspace*{\daughteroffsetone}%
\raisebox{-\ht\treeboxone}{\usebox{\treeboxone}}%
\end{flushleft}%
\end{minipage}}%
\setlength{\treeoffsetone}{\parentoffset}%
\addtolength{\treeoffsetone}{0.5\wd\parentbox}%
\setlength{\treeshiftone}{0pt}%
\setlength{\treewidthone}{\treewidth}%
\sbox{\treeboxone}{\usebox{\treebox}}\poptree\poptree\poptree%
\else\ifnum\value{branchcount}=5\sbox{\parentbox}{\ontop{#2}}%
\setlength{\branchwidthone}{\treewidthtwo}%
\addtolength{\branchwidthone}{\treeoffsetone}%
\addtolength{\branchwidthone}{-\treeshiftone}%
\addtolength{\branchwidthone}{-\treeoffsettwo}%
\setlength{\branchwidthtwo}{\treewidththree}%
\addtolength{\branchwidthtwo}{\treeoffsettwo}%
\addtolength{\branchwidthtwo}{-\treeshifttwo}%
\addtolength{\branchwidthtwo}{-\treeoffsetthree}%
\setlength{\branchwidththree}{\treewidthfour}%
\addtolength{\branchwidththree}{\treeoffsetthree}%
\addtolength{\branchwidththree}{-\treeshiftthree}%
\addtolength{\branchwidththree}{-\treeoffsetfour}%
\setlength{\branchwidthfour}{\treewidthfive}%
\addtolength{\branchwidthfour}{\treeoffsetfour}%
\addtolength{\branchwidthfour}{-\treeshiftfour}%
\addtolength{\branchwidthfour}{-\treeoffsetfive}%
\setlength{\branchwidth}{\branchwidthone}%
\ifdim\branchwidthtwo>\branchwidth%
\setlength{\branchwidth}{\branchwidthtwo}\fi%
\ifdim\branchwidththree>\branchwidth%
\setlength{\branchwidth}{\branchwidththree}\fi%
\ifdim\branchwidthfour>\branchwidth%
\setlength{\branchwidth}{\branchwidthfour}\fi%
\setlength{\daughteroffsetone}{\branchwidth}%
\addtolength{\daughteroffsetone}{-\branchwidthone}%
\addtolength{\daughteroffsetone}{-\treeshiftone}%
\setlength{\daughteroffsettwo}{\branchwidth}%
\addtolength{\daughteroffsettwo}{-\branchwidthtwo}%
\addtolength{\daughteroffsettwo}{-\treeshifttwo}%
\setlength{\daughteroffsetthree}{\branchwidth}%
\addtolength{\daughteroffsetthree}{-\branchwidththree}%
\addtolength{\daughteroffsetthree}{-\treeshiftthree}%
\setlength{\daughteroffsetfour}{\branchwidth}%
\addtolength{\daughteroffsetfour}{-\branchwidthfour}%
\addtolength{\daughteroffsetfour}{-\treeshiftfour}%
\setlength{\parentoffset}{-0.5\wd\parentbox}%
\addtolength{\parentoffset}{\treeoffsetfive}%
\addtolength{\parentoffset}{2\branchwidth}%
\setlength{\daughteroffset}{0in}%
\ifdim\parentoffset<0in%
\setlength{\daughteroffset}{-\parentoffset}%
\setlength{\parentoffset}{0in}\fi%
\setlength{\parentwidth}{\parentoffset}%
\addtolength{\parentwidth}{\wd\parentbox}%
\setlength{\treeoffset}{\daughteroffset}%
\addtolength{\treeoffset}{\treeoffsetfive}%
\setlength{\treewidth}{\wd\treeboxone}%
\addtolength{\treewidth}{\daughteroffsetone}%
\addtolength{\treewidth}{\treewidthtwo}%
\addtolength{\treewidth}{\daughteroffsettwo}%
\addtolength{\treewidth}{\treewidththree}%
\addtolength{\treewidth}{\daughteroffsetthree}%
\addtolength{\treewidth}{\treewidthfour}%
\addtolength{\treewidth}{\daughteroffsetfour}%
\addtolength{\treewidth}{\treewidthfive}%
\addtolength{\treewidth}{\daughteroffset}%
\ifdim\treewidth<\parentwidth\setlength{\treewidth}{\parentwidth}\fi%
\sbox{\treebox}{\begin{minipage}{\treewidth}%
\begin{flushleft}%
\hspace*{\parentoffset}\usebox{\parentbox}\\
{\setlength{\unitlength}{0.5\branchwidth}%
\hspace*{\treeoffset}\begin{picture}(8,1)%
\put(0,0){\line(4,1){4}}%
\put(2,0){\line(2,1){2}}%
\put(4,0){\line(0,1){1}}%
\put(6,0){\line(-2,1){2}}%
\put(8,0){\line(-4,1){4}}%
\end{picture}}\\
\vspace{-\baselineskip}
\hspace*{\daughteroffset}%
\makebox[\treewidthfive][l]%
{\raisebox{-\ht\treeboxfour}{\usebox{\treeboxfive}}}%
\hspace*{\daughteroffsetfour}%
\makebox[\treewidthfour][l]%
{\raisebox{-\ht\treeboxfour}{\usebox{\treeboxfour}}}%
\hspace*{\daughteroffsetthree}%
\makebox[\treewidththree][l]%
{\raisebox{-\ht\treeboxthree}{\usebox{\treeboxthree}}}%
\hspace*{\daughteroffsettwo}%
\makebox[\treewidthtwo][l]%
{\raisebox{-\ht\treeboxtwo}{\usebox{\treeboxtwo}}}%
\hspace*{\daughteroffsetone}%
\raisebox{-\ht\treeboxone}{\usebox{\treeboxone}}%
\end{flushleft}%
\end{minipage}}%
\setlength{\treeoffsetone}{\parentoffset}%
\addtolength{\treeoffsetone}{0.5\wd\parentbox}%
\setlength{\treeshiftone}{0pt}%
\setlength{\treewidthone}{\treewidth}%
\sbox{\treeboxone}{\usebox{\treebox}}\poptree\poptree\poptree\poptree%
\else\typeout{QobiTeX warning--- Can't handle #1 branching}\fi\fi\fi\fi\fi}
\newcommand{\faketreewidth}[1]{%
\sbox{\parentbox}{\ontop{#1}}%
\setlength{\treewidthone}{0.5\wd\parentbox}%
\addtolength{\treewidthone}{\treeoffsetone}%
\setlength{\treeshiftone}{\treeoffsetone}%
\addtolength{\treeshiftone}{-0.5\wd\parentbox}}
\newcommand{\tree}{%
\usebox{\treeboxone}
\setlength{\treeoffsetone}{\treeoffsettwo}%
\sbox{\treeboxone}{\usebox{\treeboxtwo}}%
\poptree}

\input{psfig}
\input{amssym.def}
\renewcommand{\Box}{\diamond}
\newcommand{\startproof}{\noindent {\em Proof \hspace{3em}}}

\def\understackrel#1#2{\mathrel{\mathop{#2}\limits_{#1}}}
\newcommand{\derives}{\stackrel{\ast}{\Rightarrow}}
\newcommand{\Dderives}{\stackrel{D}{\Rightarrow}}
\newcommand{\Ederives}{\stackrel{E}{\Rightarrow}}

\newcounter{list:metrics}

\newcommand{\constituent}[3]{\langle #1, #2, #3\rangle}
\newcommand{\constituenttwo}[2]{\langle #1, #2 \rangle}

\newcommand{\etal}{{\em et al.\ }}
\newcommand{\smalletal}{{\em et al.}}

\newcommand{\rspace}{\hspace{2em}}
\newcommand{\smallrspace}{\hspace{1.2em}}
\newcommand{\mi}[1]{\mbox{\it #1}}
\newcommand{\topn}{\mathit{topn}}
\newcommand{\smi}[1]{\mbox{\scriptsize\it #1}}
\newcommand{\longsum}[2]
{\begin{array}{l}{\scriptstyle #1}\\{\scriptstyle #2}\end{array}}
\newcommand{\longersum}[3]
{\begin{array}{l}{\scriptstyle #1}\\{\scriptstyle #2}\\{\scriptstyle #3}\end{array}}
\newcommand{\sumindent}{\hspace{1em}}
\newcommand{\alist}[1]{\langle #1 \rangle}

\newcommand{\aldo}{{\bf do }}
\newcommand{\alelse}{{\bf else }}
\newcommand{\alFunction}{{\bf Function }}
\newcommand{\alif}{{\bf if }}
\newcommand{\alfloat}{{\bf float }}
\newcommand{\alfor}{{\bf for }}
\newcommand{\alforeach}{{\bf for each }}

\newcommand{\alnot}{{\bf not }}

\newcommand{\alreturn}{{\bf return }}
\newcommand{\alswitch}{{\bf switch }}
\newcommand{\alto}{{\bf to }}
\newcommand{\alwhile}{{\bf while }}
\newcommand{\shortinfer}[3]{{\mbox{$\begin{array}{c}#1 \\ \hline {#2}\end{array}\;#3$}}}
\newcommand{\smallinfer}[3]{\shortinfer{\scriptstyle #1}{\scriptstyle
#2}{\scriptstyle #3}}
\newcommand{\mediuminfer}[3]{{\rule{0em}{2.5em}\shortinfer{#1}{#2}{#3}}}
\newcommand{\infer}[3]{{\rule{0em}{3em}\shortinfer{#1}{#2}{#3}}}

\newcounter{mytheoremctr}[chapter]
\renewcommand{\themytheoremctr}{\arabic{chapter}.\arabic{mytheoremctr}}
\newenvironment{mytheoremlike}[1]{

\medskip
\noindent
{\bf #1} \refstepcounter{mytheoremctr} {\bf  \themytheoremctr}\\
}
{
\medskip

}

\newenvironment{mytheorem}{\begin{mytheoremlike}{Theorem}}{\end{mytheoremlike}}
\newenvironment{mylemma}{\begin{mytheoremlike}{Lemma}}{\end{mytheoremlike}}
\newenvironment{mycorollary}{\begin{mytheoremlike}{Corollary}}{\end{mytheoremlike}}

\newenvironment{myoldtheorem}[1]{

\medskip
\noindent
{\bf Theorem \ref{#1}}\\
}
{
\medskip

}

\newcommand\beginappendix{\par
  \setcounter{section}{0}%
  \renewcommand\thesection{{\thechapter--\Alph{section}}}%
}
\newcommand{\appendixsection}[1]{\clearpage{\noindent\Large\bfseries Appendix}\vspace{-0.3in}\section{#1}}

\newcommand{\bigotimesbar}[1]{\overline{\bigotimes_{#1}}}
\newcommand{\smallst}{\;\mbox{\scriptsize s.t.}\;}

\newcommand{\bara}{\overline{a}}
\newcommand{\barb}{\overline{b}}
\newcommand{\barc}{\overline{c}}
\newcommand{\bard}{\overline{d}}
\newcommand{\bare}{\overline{e}}

\newcommand{\PA}{\PAend\ }
\newcommand{\PAend}{$\mathcal{P}(\Bbb{A})$}

\newcommand{\BbbA}{\Bbb{A}}  
\newcommand{\power}[1]{#1\langle \langle\Sigma^*\rangle \rangle}
\newcommand{\powerb}[1]{\power{\Bbb{#1}} }
\newcommand{\powera}{\powerb{A} }

\newcommand{\prederives}{\stackrel{\smi{pre}}{\Rightarrow}}

\newcommand{\derivesby}[2]{{\stackrel{#1\rightarrow #2}{\Rightarrow}}}

\newcommand{\maxvit}{\understackrel{\smi{Vit}}{\mbox{max}}}
\newcommand{\maxvitn}{\understackrel{\smi{Vit-n}}{\mbox{max}}}
\newcommand{\timesvit}{\understackrel{\smi{Vit}}{\times}}
\newcommand{\timesvitn}{\understackrel{\smi{Vit-n}}{\times}}

\def\understackrel#1#2{\mathrel{\mathop{#2}\limits_{#1}}}

\newcommand{\cdotsj}{\stackrel{-j}{\cdots}}
\newcommand{\ldotsj}{\stackrel{-j}{\ldots}}
\newcommand{\VZVin}{{\scriptstyle \frac{VZ}{V}_{\smi{in}}}}

\begin{document}

\bibliographystyle{thesis}
\pagenumbering{roman}

% -*- mode: latex; -*-

%
%	$Id: title.tex,v 1.4 1998/05/19 15:46:05 goodman Exp $
%

\thispagestyle{empty}

\mbox{}
\vfill

{\large
\begin{center}
%
%
%
%{\Large DRAFT: \today} \\
%\bigskip \bigskip
%
%
%
{\bf \LARGE Parsing Inside-Out} \\
\mbox{} \\
A thesis presented \\
by \\
\bigskip
{\bf \Large Joshua T. Goodman} \\
\bigskip
to \\
The Division of Engineering and Applied Sciences \\
in partial fulfillment of the requirements \\
for the degree of \\
Doctor of Philosophy \\
in the subject of \\
Computer Science \\
\mbox{} \\
Harvard University \\
Cambridge, Massachusetts \\
\mbox{} \\
May 1998
\end{center}
}

\vfill

%
%	$Id: copyright.tex,v 1.1 1997/10/03 15:54:16 goodman Exp $
%

\mbox{}
\vfill

\begin{center}
\copyright 1998 by Joshua T. Goodman\\
All rights reserved.
\end{center}

\vfill

%
%	$Id: abstract.tex,v 1.2 1998/05/19 15:41:31 goodman Exp $
%

%%%%%%%%%%%%%%%%%%%%%%%%%%%%%%%%%%%%%%%%%%%%%%%%%%%%%%%%%%%%%%%%%%%%%%%%
%
\chapter*{Abstract}
%
%%%%%%%%%%%%%%%%%%%%%%%%%%%%%%%%%%%%%%%%%%%%%%%%%%%%%%%%%%%%%%%%%%%%%%%%

Probabilistic Context-Free Grammars (PCFGs) and variations on them
have recently become some of the most common formalisms for parsing.
It is common with PCFGs to compute the inside and outside
probabilities.  When these probabilities are multiplied together and
normalized, they produce the probability that any given non-terminal
covers any piece of the input sentence.  The traditional use of these
probabilities is to improve the probabilities of grammar rules.  In
this thesis we show that these values are useful for solving many
other problems in Statistical Natural Language Processing.

We give a framework for describing parsers.  The framework generalizes
the inside and outside values to semirings.  It makes it easy to
describe parsers that compute a wide variety of interesting
quantities, including the inside and outside probabilities, as well as
related quantities such as Viterbi probabilities and n-best lists.  We
also present three novel uses for the inside and outside
probabilities.  The first novel use is an algorithm that gets improved
performance by optimizing metrics other than the exact match rate.
The next novel use is a similar algorithm that, in combination with
other techniques, speeds Data-Oriented Parsing, by a factor of 500.
The third use is to speed parsing for PCFGs using thresholding
techniques that approximate the inside-outside product; the
thresholding techniques lead to a 30 times speedup at the same
accuracy level as conventional methods.  At the time this research was
done, no state of the art grammar formalism could be used to compute
inside and outside probabilities.  We present the Probabilistic
Feature Grammar formalism, which achieves state of the art accuracy,
and can compute these probabilities.

{ \singlespace
% -*- mode: latex; -*-

%
%	$Id: acknowledgements.tex,v 1.2 1998/05/19 15:42:39 goodman Exp $
%

%%%%%%%%%%%%%%%%%%%%%%%%%%%%%%%%%%%%%%%%%%%%%%%%%%%%%%%%%%%%%%%%%%%%%%%%
%
\chapter*{Acknowledgements}
%
%%%%%%%%%%%%%%%%%%%%%%%%%%%%%%%%%%%%%%%%%%%%%%%%%%%%%%%%%%%%%%%%%%%%%%%%

\setlength{\parskip}{\smallskipamount}

%	Stan's hack to make fit in two pages
%\vskip-\baselineskip

%\begin{center}
{\em To Erica \\
my love, my help}
\vspace{0.5in}
%\end{center}

There are too many people for me to thank\footnote{\tiny In the AI
Research Group, acknowledgments are traditionally funny.  This
encourages people to read the acknowledgements, and skip the thesis.
I have therefore written rather dry acknowledgments, and put one joke
in this thesis, to encourage the reading of this work, in its
entirety.  The joke is not very funny, so you will have to read the
whole thing to be sure you have found it.  I will send \$5 to the
first person each year to find the joke without help.  Some
restrictions apply.}, and for too many things.  Most of all, this
thesis is for Erica, who I met a month after I started graduate
school, and who I will marry a month after I finish.

Next, I want to thank my committee.  Stuart Shieber is an amazing
advisor, who gave me wonderful support, and just the right amount of
guidance.  Barbara Grosz has been a great acting advisor in my last
year, putting her foot down on important things, but allowing me to
occassionally split infinitives when it did not matter.  Fernando
Pereira has given me very good advice and is a terrific source of
knowledge.  Thanks also to Leslie Valiant for the almost thankless
task of serving on my committee.

For my first year or two, I shared an office with Stan Chen and Andy
Kehler.  I'd rather share a small office with them than have a large
office to myself -- they taught me as much as anyone here and made
grad school fun.  They, and the other members of the AI-Research group
read endless drafts of my papers, attended no end of nearly-identical
practice talks, and listened to hundreds of bad ideas.  Wheeler Ruml
in particular has done countless small favors.  Lillian Lee has been a
source of guidance.  I'm happy to call Rebecca Hwa my friend.  Others
-- Luke Hunsberger, Ellie Baker, Kathy Ryall, Jon Christenson,
Christine Nakatani, Nadia Shalaby, Greg Galperin, and David Magerman
(for poker, and rejecting my ACL paper) -- have all helped me through
and made my time here better.

A year after getting here, I developed tendonitis in my hands.  I
needed more help than usual to graduate, and I got it, both from the
Division of Engineering and Applied Sciences, and from many of the
people I have named so far.

A Ph.D. is built on a deep foundation.  My parents and family have
always pushed me and supported me, and continue to do so.  My father
encouraged me from a ridiculously early age (what kind of person gives
a 5 year old Scientific American?) while my mother, through what she
called gentle teasing, gave me her own special kind of encouragement.
I want to thank Edward Siegfried, my mentor for nine years, who taught
me so much about computers.  As an undergraduate, I had many helpful
professors, among whom Harry Lewis stands out.  After college, I
worked at Dragon Systems, where Dean Sturtevant, Larry Gillick, and
Bob Roth initiated me into the black art of speech recognition, and
taught me how to be a good programmer.

I'd also like to thank the National Science Foundation for providing
most of my funding with Grant IRI-9350192, Grant IRI 9712068, and an
NSF Graduate Student Fellowship.

% Thanks from DOP
%\thanks{\hspace{1em}I would like to acknowledge support from National
%Science Foundation Grant IRI-9350192 and a National Science Foundation
%Graduate Student Fellowship.  I would also like to thank Rens Bod,
%Stan Chen, Andrew Kehler, David Magerman, Wheeler Ruml, Stuart
%Shieber, and Khalil Sima'an for helpful discussions, and comments on
%earlier drafts, and the comments of the anonymous reviewers.} }

%thanks from PFGs
%\title{Probabilistic Feature Grammars \thanks{\hspace{.3em} This
%material is based upon work supported by the National Science
%Foundation under Grant No. IRI-9350192 and a National Science
%Foundation Graduate Student Fellowship.  I would like to thank Stanley
%Chen, Lillian Lee, David Magerman, Wheeler Ruml, Stuart Shieber, and
%Fernando Pereira for helpful comments and discussions, as well as the
%anonymous reviewers for their useful comments and suggestions.  }}

%Thanks from thresholding
%\title{Global Thresholding and Multiple-Pass
%Parsing\thanks{\hspace{.3em}This material is based in part upon work
%supported by the National Science Foundation under Grant
%No. IRI-9350192 and a National Science Foundation Graduate Student
%Fellowship.  I would also like to thank Michael Collins, Rebecca Hwa,
%Lillian Lee, Wheeler Ruml, and Stuart Shieber for helpful discussions,
%and comments on earlier drafts, and the anonymous reviewers for their
%extensive comments.} }

%
%	$Id: s-toc.tex,v 1.1 1996/02/01 21:40:19 sfc Exp $
%

\tableofcontents
\listoffigures
\listoftables

}

\pagenumbering{arabic}
% -*- mode: latex; -*-

\chapter{Introduction} \label{ch:intro}

\section{Background}

Consider the sentence ``Stuart loves his thesis, and Barbara does
too.''  If a human being, or a computer, attempts to understand this
sentence, what will the steps be?  One might imagine that the first
step would be a gross analysis of the sentence, a syntactic parsing to
determine its structure, breaking the sentence into a conjunction and
two sentential clauses, breaking each sentential clause into a noun
phrase ($\mi{NP}$) and a verb phrase ($\mi{VP}$), and so on,
recursively to the words.
\begin{center}
\leaf{$\mathit{Stuart}$}
\branch{1}{$\mathit{PN}$}
\branch{1}{$\mathit{NP}$}
\leaf{$\mathit{loves}$}
\branch{1}{$\mathit{verb}$}
\leaf{$\mathit{his}$}
\branch{1}{$\mathit{pos}$}
\leaf{$\mathit{thesis}$}
\branch{1}{$\mathit{N}$}
\branch{2}{$\mathit{NP}$}
\branch{2}{$\mathit{VP}$}
\branch{2}{$\mathit{S}$}
\leaf{$\mathit{and}$}
\branch{1}{$\mathit{conj}$}
\leaf{$\mathit{Barbara}$}
\branch{1}{$\mathit{PN}$}
\branch{1}{$\mathit{NP}$}
\leaf{$\mathit{does}$}
\branch{1}{$\mathit{verb}$}
\leaf{$\mathit{too}$}
\branch{1}{$\mathit{adv}$}
\branch{2}{$\mathit{VP}$}
\branch{2}{$\mathit{S}$}
\branch{3}{$\mathit{S}$}
\tree
\end{center}
After this first syntactic step, a semantic step would follow.  The
sentence would be converted into a logical representation, ruling out
incorrect interpretations such as ``Stuart loves someone's thesis, and
Barbara loves someone else's thesis,'' and allowing only
interpretations such as ``Stuart loves his own thesis, and Barbara
loves Stuart's thesis too'', and ``Stuart loves someone's thesis, and
Barbara loves that thesis too.''  After this semantic step, a
pragmatic step would be necessary to disambiguate between these
interpretations.  Given that Stuart is notoriously hard to please, we
would probably decide that it is Stuart's own thesis which is the
intended antecedent, rather than some more recent one.

The eventual goal of Natural Language Processing (NLP) research is to
understand the meaning of sentences of human language.  This is a
difficult goal, and one we are still a long way from achieving.
Rather than attacking the full problem all at once, it is prudent to
attack subproblems separately, such as syntax, semantics, or
pragmatics.  We will be solely concerned with the first problem:
syntax.  There are two broad approaches to syntax: the rule-based and
the statistical.  The statistical approach has become possible only
recently with the advent of bodies of text (corpora) which have had
their structures hand annotated, called tree banks.  Each sentence in
the corpus has been assigned a parse tree -- a representation of its
structure -- by a human being. The majority of these sentences can be
used to train a statistical model, and another portion can be used to
test the accuracy of the model.

In this thesis, we will be particularly concerned with two special
quantities in statistical NLP, the inside and outside probabilities.
For each span of terminal symbols (words), and for each nonterminal
symbol (i.e., a phrase type, such as a noun phrase), the inside
probability is the probability that that nonterminal would consist of
exactly those terminals.  For instance, if one out of every ten thousand
noun phrases is ``his thesis,'' then
\begin{figure}
\begin{center}
\begin{tabular}{c}
\psfig{figure=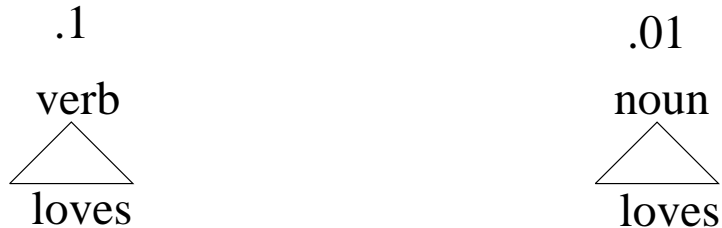}
\end{tabular}
\end{center}
\caption{Inside probability example}\label{fig:lovesinside}
\end{figure}
$$
\mathit{inside}(\mbox{``his thesis''}, \mathit{NP}) =.0001
$$
While a phrase like ``his thesis'' can be interpreted only as a noun
phrase, natural language is replete with ambiguities.  For instance,
the word ``loves'' can be either a verb or a noun.  Let us assume that
there is a one in ten chance that a given verb will be loves (as seems
true in the literature of computational linguistics examples), and a
one in one hundred chance that a given noun will be loves.  Then
$$
\mathit{inside}(\mbox{``loves''}, \mathit{verb}) =.1
$$
$$
\mathit{inside}(\mbox{``loves''}, \mathit{noun}) =.01
$$
We might denote this graphically as in Figure \ref{fig:lovesinside}.

\begin{figure}
\begin{center}
\begin{tabular}{c}
\psfig{figure=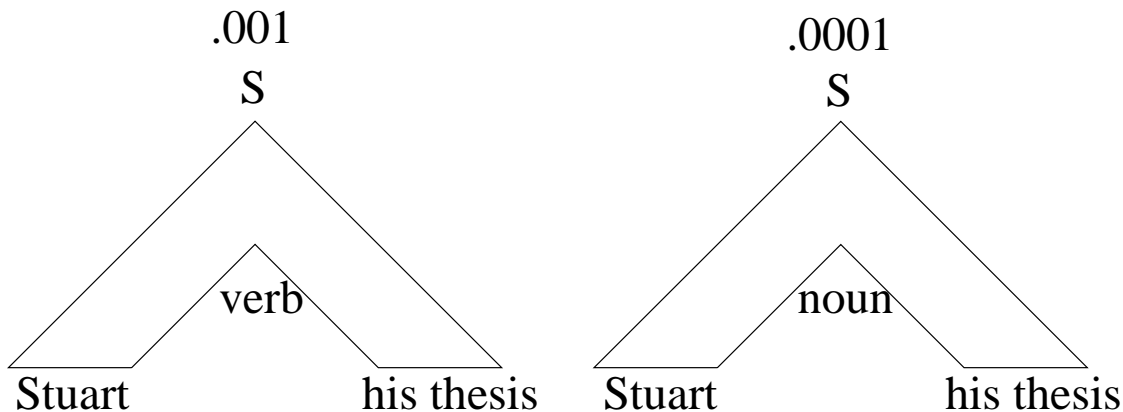}
\end{tabular}
\end{center}
\caption{Outside probability example}\label{fig:lovesoutside}
\end{figure}
The outside probabilities can be thought of as the probability of
everything surrounding a phrase.  For instance, if the probability of
a sentence of the form ``Stuart {\it verb} his thesis'' is say one in
one thousand, and one of the form ``Stuart {\it noun} his thesis'' is
one in ten thousand, then
$$
\mathit{outside}(\mbox{``Stuart \rule{3em}{.03em} his thesis''},
\mathit{verb}) = .001
$$
$$
\mathit{outside}(\mbox{``Stuart \rule{3em}{.03em} his thesis''},
\mathit{noun}) = .0001
$$
This is illustrated in Figure \ref{fig:lovesoutside}.

\begin{figure}
\begin{center}
\begin{tabular}{c}
\psfig{figure=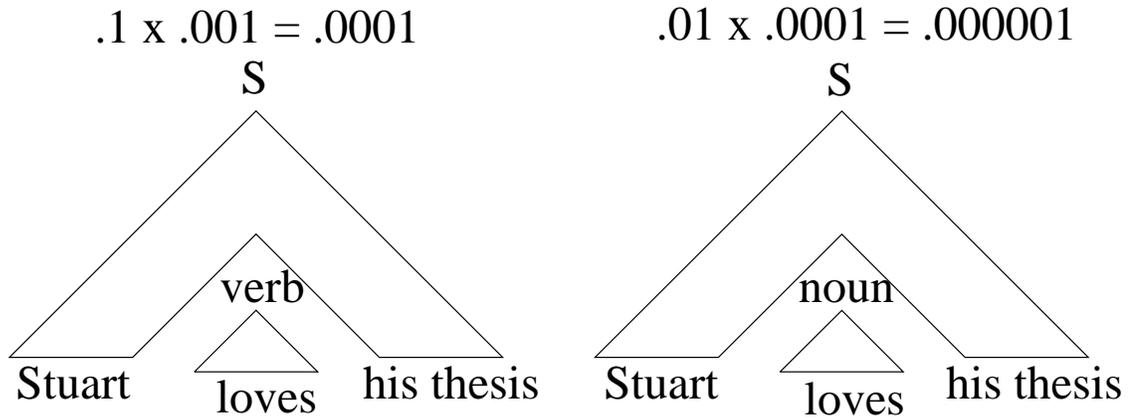}
\end{tabular}
\end{center}
\caption{Inside-Outside probability example}\label{fig:lovesinsideoutside}
\end{figure}
Now, we can multiply the inside probabilities by the outside
probabilities, as illustrated in Figure \ref{fig:lovesinsideoutside}.
The product of the inside and outside probabilities gives us the
overall probability of the sentence having the indicated structure.
For instance, there is a $.1 \times .001 = .0001$ chance of the
sentence ``Stuart loves his thesis'', with ``loves'' being a verb, and
a $.01 \times .0001 = .000001$ chance of the sentence with ``loves''
being a noun.  Since these are the only two possible parts of speech
for ``loves,'' the overall probability of the sentence is $.0001 +
.000001 = .000101$.  We can divide by this overall probability to get
the conditional probabilities.  That is, $\frac{.0001}{.000101}
\approx 0.99$ is the conditional probability that ``loves'' is a verb,
given the sentence as a whole, while $\frac{.000001}{.000101} \approx
0.01$ is the conditional probability that ``loves'' is a noun.

Traditionally, the inside-outside probabilities have been used as a
tool to learn the parameters of a statistical model.  In particular,
the inside-outside probabilities of a training set using one model can
be used to find the parameters of another model, with typically
improved performance. This procedure can be iterated, and is
guaranteed to converge to a local optimum.

While learning the probabilities of a statistical grammar is the
traditional use for the inside and outside probabilities, the goal of
this thesis is different.  Our goal is to show that the inside-outside
values are useful for solving many other problems in statistical
parsing, and to provide useful tools for finding these values.  

\section{Introduction to Statistical NLP}
\label{sec:background}
In this section, we will give a very brief introduction to statistical
NLP, describing context-free grammars (CFGs), probabilistic
context-free grammars (PCFGs), and algorithms for parsing CFGs and
PCFGs.

\subsection{Context-Free Grammars}
\label{sec:CFG}

We begin by quickly reviewing Context-Free Grammars (CFGs), less with
the intention of aiding the novice reader, than of clarifying the
relationship to PCFGs, in the next subsection.

A CFG $G$ is a 4-tuple $\langle N,\Sigma, R, S\rangle$ where $N$ is
the set of nonterminals including the start symbol $S$, $\Sigma$ is
the set of terminal symbols, and $R$ is the set of rules (we use $R$
rather than the more conventional $P$ to avoid confusion with
probabilities, which will be introduced later).  Let $V$, the
vocabulary of the grammar, be the set $N\cup\Sigma$.  We will use
lowercase letters $a, b, c,...$ to represent terminal symbols,
uppercase letters $A, B, C,...$ for nonterminals, and Greek symbols,
$\alpha$, $\beta$, $\gamma$,... to represent a string of zero or more
terminals and nonterminals.  We will use the special symbol $\epsilon$
to represent a string of zero symbols.  Rules in the grammar are all
of the form $A \rightarrow \alpha$.

\begin{figure}
\begin{tabbing}
\verb|    |\=\verb|    |\=\verb|    |\=\verb|    |\=\verb|    |\=\verb|    |\=\verb|    |\=\verb|    |\=\verb|    |\=\verb|    |\= \kill
$\mi{boolean chart}[1..n, 1..|N|, 1..n\!+\!1] := \mi{FALSE};$ \\
\alforeach start $s$ \\
 \> \alforeach rule $A \rightarrow w_s$ \\
 \>  \> $\mi{chart}[s, A, s\!+\!1] := \mi{TRUE}$;  \\
\alforeach length $l$, shortest to longest \\
 \> \alforeach start $s$ \\
 \>  \> \alforeach split length $t$ \\
 \>  \>  \> \alforeach rule $A \rightarrow BC\;\in R$\\
 \>  \>  \>  \> {\em /* extra TRUE for expository purposes */} \\
 \>  \>  \>  \> $\mi{chart}[s, A, s\!+\!l] := \mi{chart}[s, A, s\!+\!l] \;\vee$ \\
 \>  \>  \>  \>  \>  \> $\mi{chart}[s, B, s\!+\!t] \wedge \mi{chart}[s\!+\!t,
C, s\!+\!l] \wedge \mi{TRUE}$;\\
 \>  \>  \>  \>  \> \\
\alreturn $\mi{chart}[1, S, n\!+\!1]$;
\end{tabbing}
\caption{CKY algorithm}\label{fig:CKYalgorithm}
\end{figure}

Given a string $\alpha A\beta$ and a grammar rule $A
\rightarrow\gamma\in R$, we write
$$
\alpha A\beta \Rightarrow \alpha\gamma\beta
$$
to indicate that the first string produces the second string by
substituting $\gamma$ for $A$.  A sequence of zero or more such
substitutions, called a derivation, is indicated by $\derives$.  For
instance, if we have an input sentence $w_1... w_n$, and a sequence of
substitutions starting with the start symbol $S$ derives the sentence,
then we write $S\derives w_1... w_n$.

There are several well known algorithms for determining whether for a
given input sentence $w_1... w_n$ such a sequence of substitutions
exists.  The two best known algorithms are the CKY algorithm
\cite{Kasami:65a,Younger:67a} and Earley's algorithm \cite{Earley:70a}.
The CKY algorithm makes the simplifying assumption that the grammar is
in a special form, Chomsky Normal Form (CNF), in which all productions
are of the form $A \rightarrow BC$ or $A \rightarrow a$.  The CKY
algorithm is given in Figure \ref{fig:CKYalgorithm}.

Most of the parsing algorithms we use in this thesis will strongly
resemble the CKY algorithm, so it is important that the reader
understand this algorithm.  Briefly, the algorithm can be described as
follows.  The central data structure in the CKY algorithm is a boolean
three dimensional array, the chart.  An entry $\mi{chart}[i, A, j]$
contains $\mi{TRUE}$ if $A \derives w_i... w_{j-1}$, $\mi{FALSE}$
otherwise.  The key line in the algorithm,
\begin{tabbing}
\verb|    |\=\verb|    |\=\verb|    |\=\verb|    |\=\verb|    |\=\verb|    |\=\verb|    |\=\verb|    |\=\verb|    |\=\verb|    |\= \kill
 \>  \>  \>  \> {\em /* extra TRUE for expository purposes */} \\
 \>  \>  \>  \> $\mi{chart}[s, A, s\!+\!l] := \mi{chart}[s, A, s\!+\!l] \;\vee$ \\
 \>  \>  \>  \>  \>  \> $\mi{chart}[s, B, s\!+\!t] \wedge \mi{chart}[s\!+\!t,
C, s\!+\!l] \wedge \mi{TRUE}$; 
\end{tabbing}
says that if $A \rightarrow BC$ and $B\derives w_s...w_{s+t-1}$ and $C
\derives w_{s+t}...w_{s+l-1}$, then $A \derives w_s...w_{s+l-1}$.
Notice that once all spans of length one have been examined, which
occurs in the first double set of loops, we can then proceed to
examine all spans of length two, and from there all spans of length
three (which must be formed only from shorter spans), and so on.  Thus
the outermost loop of the main set of loops is a loop over lengths,
from shortest to longest.  The next three loops examine all possible
combinations of start positions, split lengths, and rules, so that the
main inner statement examines all possibilities.  Because array
elements covering shorter spans are filled in first, this style of
parser is also called a bottom-up chart parser.

\subsection{Probabilistic Context-Free Grammars}
\label{sec:PCFGintroduction}

In this thesis, we will primarily be concerned with a variation on
context-free grammars, probabilistic context-free grammars (PCFGs).  A
PCFG is simply a CFG augmented with probabilities.  We denote the
probability of a rule $A \rightarrow \alpha$ by $P(A \rightarrow
\alpha)$.  We can also discuss the probability of a particular
derivation or of all possible derivations of one string from another.
In order to make sure that equivalent derivations are not counted
twice, we need the concept of a {\em leftmost derivation}.  A leftmost
derivation is one in which the leftmost nonterminal symbol is the one
that is substituted for.  We will write $\alpha \derivesby{A}{\beta}
\gamma$ to indicate a leftmost derivation using a substitution of
$\beta$ for $A$.  Now, we define the probability of a one step
derivation to be
$$
P(\alpha \derivesby{A}{\beta} \gamma) = P(A \rightarrow\beta)
$$
We can define the probability of a string of substitutions
$$
P(\alpha \derivesby{A_1}{\beta_1} \gamma_1
 \derivesby{A_2}{\beta_2} \gamma_2 \derivesby{A_3}{\beta_3}...
\derivesby{A_k}{\beta_k} \gamma_k) = \prod_{i=1}^k P(A_i \rightarrow
\beta_i)
$$
Finally, we can define the probability of all of the leftmost
derivations of some string $\delta$ from some initial string $\alpha$.
In particular, we define
$$
P(\alpha \derives \delta) = \sum_{k, A_1, \beta_1,\gamma_1,...,A_k,
\beta_k, \gamma_k \;\mbox{\scriptsize s.t. }\;\alpha \derivesby{A_1}{\beta_1} \gamma_1
... \derivesby{A_k}{\beta_k} \gamma_k \wedge \gamma_k = \delta} 
P(\alpha \derivesby{A_1}{\beta_1} \gamma_1 \derivesby{A_2}{\beta_2} ... \derivesby{A_k}{\beta_k} \gamma_k)
$$

Several probabilities of this form are of special interest.  In
particular, we say that the probability of a sentence $w_1... w_n$ is
$P(S\derives w_1... w_n)$, the sum of the probabilities of all
(leftmost) derivations of the sentence.  In general, we will be
interested in probabilities of nonterminals deriving sections of the
sentence, $P(A\derives w_i... w_{j-1})$.  We will call this
probability the {\em inside probability} of $A$ over the span $i$ to
$j$, and will write it as $\mi{inside}(i, A, j)$.

\begin{figure}
\begin{center}
\begin{tabular}{c}
\psfig{figure=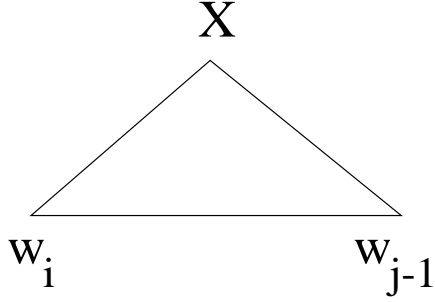}
\end{tabular}
\end{center}
\caption{Inside probabilities}\label{fig:insideprobabilities}
\end{figure}

\begin{figure}
\begin{tabbing}
\verb|    |\=\verb|    |\=\verb|    |\=\verb|    |\=\verb|    |\=\verb|    |\=\verb|    |\=\verb|    |\=\verb|    |\=\verb|    |\= \kill
$\mi{float chart}[1..n, 1..|N|, 1..n\!+\!1] := 0;$ \\
\alforeach start $s$ \\
 \> \alforeach rule $A \rightarrow w_s$ \\
 \>  \> $\mi{inside}[s, A,s\!+\!1]:= P(A\rightarrow w_s)$; \\
\alforeach length $l$, shortest to longest \\
 \> \alforeach start $s$ \\
 \>  \> \alforeach split length $t$ \\
 \>  \>  \> \alforeach rule $A \rightarrow BC\;\in R$\\
 \>  \>  \>  \>  $\mi{inside}[s, A, s\!+\!l] := \mi{inside}[s, A, s\!+\!l] \;+$ \\
 \>  \>  \>  \>  \>  \> $\mi{inside}[s, B, s\!+\!t] \times \mi{inside}[s\!+\!t, C, s\!+\!l] \times P(A\rightarrow BC)$; \\
\alreturn $\mi{inside}[1, S, n\!+\!1]$;
\end{tabbing}
\caption{Inside algorithm}\label{fig:insidealgorithm}
\end{figure}

Note that in general, a derivation of the form $A \derives \alpha$ can
be drawn as a parse tree, in which the internal branches of the parse
tree represent the nonterminals where substitutions occured; for each
substitution $A \rightarrow B_1...B_k$ there will be a node with parent
$A$ and children $B_1...B_k$.  The leaves of the tree when
concatenated form the string $\alpha$.  There is a one-to-one
correspondence between leftmost derivations and parse trees, so we
will use the concepts interchangeably.

When we consider inside probabilities, we are summing over the
probabilities of all possible parse trees with a given root node
covering a given span.  Since the internal structure is summed over
for an inside probability, we graphically represent the inside
probability of a nonterminal $X$ covering words $w_i... w_{j-1}$ as
shown in Figure \ref{fig:insideprobabilities}.  The inside algorithm,
shown in Figure \ref{fig:insidealgorithm}, computes these
probabilities.  Notice that the inside algorithm is extremely similar
to the CKY algorithm of Figure \ref{fig:CKYalgorithm}.  The inside
algorithm was created by \newcite{Baker:79b}, and \newcite{Lari:90a}
have written a good tutorial explaining it.

In many practical applications, we are not interested only in the sum
of probabilities of all derivations, but also in the most probable
derivation.  The inside algorithm of Figure \ref{fig:insidealgorithm}
can be easily modified to return the probability of the most probable
derivation (which corresponds uniquely to a most probable parse tree)
instead of the sum of the probabilities of all derivations.  We simply
change the inner loop of the inside algorithm to read:
$$
\mi{Viterbi}[s, A, s\!+\!l] := \max(\mi{Viterbi}[s, A, s\!+\!l], \mi{Viterbi}[s, B, s\!+\!t] \times \mi{Viterbi}[s\!+\!t, C, s\!+\!l] \times P(A\!\rightarrow\! BC)); \\
$$
These probabilities are known as the Viterbi probabilities, by analogy
to the Viterbi probabilities for Hidden Markov Models (HMMs)
\cite{Rabiner:89a,Viterbi:67a}.  With slightly larger changes, the
algorithm can record the actual productions used to create this most
probable parse tree.

\begin{figure}
\begin{center}
\begin{tabular}{c}
\psfig{figure=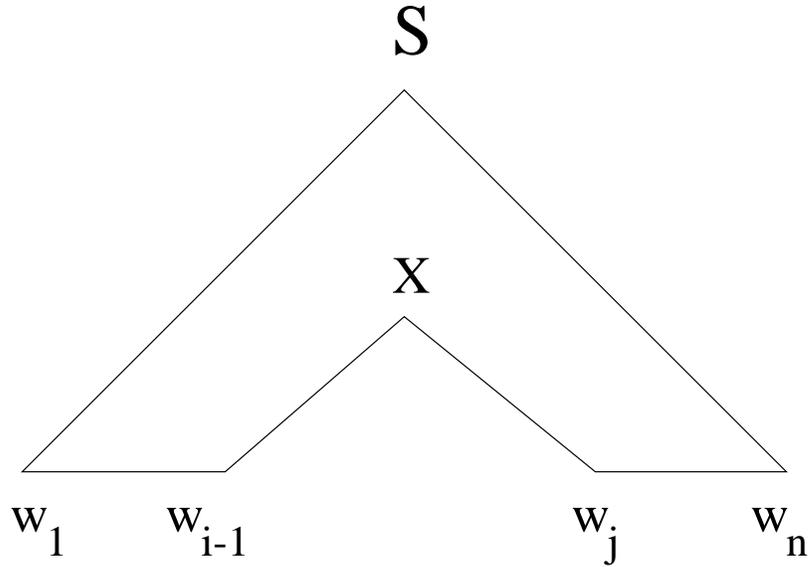}
\end{tabular}
\end{center}
\caption{Outside probabilities}\label{fig:outsideprobabilities}
\end{figure}

Another useful probability is the {\em outside probability}.  The outside
probability of a nonterminal $X$ covering $w_i$ to $w_{j-1}$ is
$$
P(S \derives w_1...w_{i-1} X w_j...w_n)
$$
This probability is illustrated in Figure
\ref{fig:outsideprobabilities}.  The outside probability can be
computed using the outside algorithm, as given in Figure
\ref{fig:outsidealgorithm} \cite{Baker:79b,Lari:90a}.

\begin{figure}
\begin{tabbing}
\verb|    |\=\verb|    |\=\verb|    |\=\verb|    |\=\verb|    |\=\verb|    |\=\verb|    |\=\verb|    |\=\verb|    |\=\verb|    |\= \kill
\alforeach length $l$, longest {\bf downto} shortest \\
 \> \alforeach start $s$ \\
 \>  \> \alforeach split length $t$ \\
 \>  \>  \> \alforeach rule $A \rightarrow BC\;\in R$\\
 \>  \>  \>  \>  $\mi{outside}[s, B, s\!+\!t] := \mi{outside}[s, B, s\!+\!t] \;+$ \\
 \>  \>  \>  \>  \>  \> $\mi{outside}[s, A, s\!+\!l] \times
\mi{inside}[s\!+\!t, C, s\!+\!l] \times P(A\rightarrow BC)$; \\
 \>  \>  \>  \>  $\mi{outside}[s\!+\!t, C, s\!+\!l] := \mi{outside}[s\!+\!t, C, s\!+\!l] \;+$ \\
 \>  \>  \>  \>  \>  \> $\mi{outside}[s, A, s\!+\!l] \times \mi{inside}[s, B, s\!+\!t] \times P(A\rightarrow BC)$; \\
\end{tabbing}
\caption{Outside algorithm}\label{fig:outsidealgorithm}
\end{figure}

\begin{figure}
\begin{center}
\begin{tabular}{c}
\psfig{figure=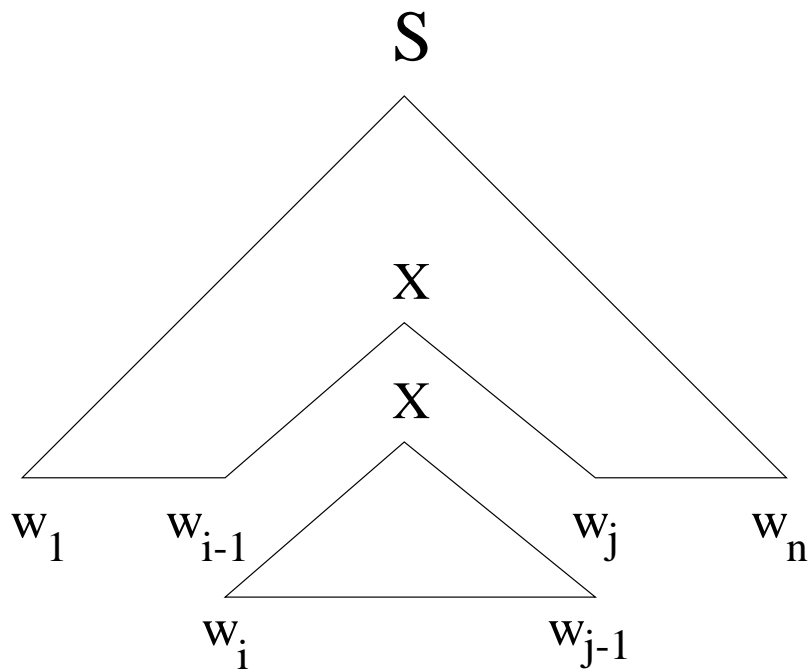}
\end{tabular}
\end{center}
\caption{Inside-outside probabilities}\label{fig:triangleprobabilities}
\end{figure}

Notice that if we multiply an inside probability by the corresponding
outside probability, we get
\begin{eqnarray*}
\mi{inside}(i,X, j) \times \mi{outside}(i, X, j) & =&
P(X \derives w_i... w_{j-1}) \times P(S \derives w_1...w_{i-1} X
w_j...w_n) \\
& = & P(S \derives w_1...w_{i-1} X w_j...w_n \derives w_1...w_n)
\end{eqnarray*}
which is the probability that a derivation of the whole sentence uses
a constituent $X$ covering $w_i$ to $w_{j-1}$.  This combination is
illustrated in Figure \ref{fig:triangleprobabilities}.  Now, the
probability of the sentence as a whole is just 
$$
\mi{inside}(1,S,n\!+\!1) = P(S \derives w_1...w_n)
$$
If we normalize by dividing by the probability of the sentence, we get
the conditional probability that $X$ covers $w_i...w_{j-1}$ given the
sentence:
\begin{eqnarray*}
\frac{\mi{inside}(i,X, j) \times \mi{outside}(i, X, j)}{\mi{inside}(1,
S, n\!+\!1)} & =&
\frac{P(X \derives w_i...w_{j-1}) \times P(S \derives w_1...w_{i-1} X
w_j...w_n)}{P(S\derives w_1...w_n)} \\
&=& P(S \derives w_1...w_{i-1} X
w_j...w_n \derives w_1...w_n | S \derives w_1...w_n) 
\end{eqnarray*}
We might also want the probability that a particular rule was used to
cover a particular span, given the sentence
\begin{eqnarray*}
\lefteqn{\frac{\sum_{k}
\mi{inside}(i, B, k) \times \mi{inside}(k, C, j) \times
\mi{outside}(i, A, j) \times P(A \rightarrow \alpha)}{\mi{inside}(1, S,
n\!+\!1)} } \\
&= & P(S \derives w_1...w_{i-1} A w_j...w_n \derivesby{A}{\alpha} w_1...w_{i-1} \alpha w_j...w_n \derives
w_1...w_n | S \derives w_1...w_n) \\
\end{eqnarray*}

\begin{figure}
\begin{tabbing}
\verb|    |\=\verb|    |\=\verb|    |\=\verb|    |\=\verb|    |\=\verb|    |\=\verb|    |\=\verb|    |\=\verb|    |\=\verb|    |\= \kill
\alforeach iteration $i$ until the probabilities have converged \\
 \> \alforeach rule $A \rightarrow \alpha$ \\
 \>  \> $C[A \rightarrow \alpha] := 0$; \\
 \> \alfor $j:= 1$ \alto number of sentences\\
 \>  \> compute inside probabilities of sentence $j$ using $P_i$; \\
 \>  \> compute outside probabilities of sentence $j$ using $P_i$; \\
 \>  \> \alforeach length $l$, shortest to longest \\
 \>  \>  \> \alforeach start $s$ \\
 \>  \>  \>  \> \alforeach rule $A \rightarrow \alpha$\\
 \>  \>  \>  \>  \> $C[A \rightarrow \alpha] := C[A \rightarrow \alpha] + $ \\
% \>  \>  \>  \>  \>  \>  \> $P(
%\begin{array}{rcl}
%S & \derives & w_1...w_{i-1} A w_j...w_n \\
%  & \derivesby{A}{\alpha} & w_1...w_{i-1} \alpha w_j...w_n \\
%  & \derives & w_1...w_n | S \derives w_1...w_n)
%\end{array}$ \\
 \>  \>  \>  \>  $P_i(S \derives  w_1...w_{i-1} A w_j...w_n \derivesby{A}{\alpha}  w_1...w_{i-1} \alpha w_j...w_n \derives  w_1...w_n | S \derives w_1...w_n)$; \\
 \> \alforeach rule $A \rightarrow \alpha$\\
 \>  \> $P_{i+1} (A \rightarrow \alpha) := \frac{C[A \rightarrow
\alpha]}{\sum_\beta C[A \rightarrow \beta]}$; \\
\end{tabbing}
\caption{Inside-outside algorithm}\label{fig:insideoutsidealgorithm}
\end{figure}

This conditional probability is extremely useful.  Traditionally, it
has been used for estimating the probabilities of a PCFG from training
sentences, using the {\em inside-outside algorithm}, shown in Figure
\ref{fig:insideoutsidealgorithm}.  In this algorithm, we start with
some initial probability estimate, $P_1(A \rightarrow \alpha)$.  Then,
for each sentence of training data, we determine the inside and
outside probabilities to compute, for each production, how likely it
is that that production was used as part of the derivation of that
sentence.  This gives us a number of counts for each production for
each sentence.  Summing these counts across sentences gives us an
estimate of the total number of times each production was used to
produce the sentences in the training corpus.  Dividing by the total
counts of productions used for each nonterminal $A$ gives us a new
estimate of the probability of the production.  It is an important
theorem that this new estimate will assign a higher probability to the
training data than the old estimate, and that when this algorithm is
run repeatedly, the rule probabilities converge towards locally
optimum values \cite{Baker:79b}, in terms of maximizing the
probability of the training data.

\section{Overview}

While the traditional use for the inside-outside probabilities is to
estimate the parameters of a PCFG, our goal is different: our goal is
to demonstrate that the inside-outside probabilities are useful for
solving many other problems in statistical parsing, and to provide
useful tools for finding these values.  In the remaining chapters of
this thesis, we will first provide a general framework for specifying
parsers that makes it easy to compute inside and outside
probabilities.  Next, we will show three novel uses: improved parser
performance on specific criteria; faster parsing for the Data-Oriented
Parsing (DOP) model; and faster parsing more generally by using
thresholding.  Finally, we describe a state-of-the-art parsing
formalism that can compute inside and outside probabilities.

In Chapter \ref{ch:semi}, we develop a novel framework for specifying
parsers that makes it easy to compute the inside and outside
probabilities, and others.  The CKY algorithm of Figure
\ref{fig:CKYalgorithm} is very similar to the inside algorithm of
Figure \ref{fig:insidealgorithm}, and the inside algorithm in some
ways resembles the outside algorithm of Figure
\ref{fig:outsidealgorithm}.  For a simple parsing technique, such as
CKY parsing, it is not too much work to derive each of these
algorithms separately, ignoring their commonalities, but for more
complicated algorithms and formalisms, the duplicated effort is
significant.  In particular, for sophisticated formalisms or
techniques, the outside formula can be especially complicated to
derive.  Also, for parsing algorithms that can handle loops, like
those that result from a grammar rule like $A \rightarrow A$, the
inside and outside algorithms may become yet more complicated, because
of the infinite summations that result.  We develop a framework that
allows a single parsing description to be used to compute recognition,
inside, outside, and Viterbi values, among others.  The framework uses
a description language that is independent of the values being
derived, and thus allows the complicated manipulations required to
handle infinite summations to be separated out from the construction
of the parsing algorithms.  Using this framework, we show how to
easily compute many interesting values, including the set of all
parses of a grammar, the top $n$ parses of a sentence, the most
probable completion of a sentence, and many others.  With this format,
it will be simple to specify the thresholding and parsing algorithms
of the following chapters, although we will also use traditional
pseudocode as well, in an effort to keep the chapters self-contained.

In Chapter \ref{ch:max}, we present our first novel use for the
inside-outside probabilities, tailoring parsing algorithms to various
metrics.  Most probabilistic parsing algorithms are similar to the
Viterbi algorithm.  They attempt to maximize a single metric, the
probability that the guessed parse tree is exactly correct.  If the
score that the parser receives is given by the number of exactly
correct guessed trees, then this approach is correct.  However, in
practice, many other metrics are typically used, such as precision and
recall, or crossing brackets.  These metrics measure, in one way or
another, how many pieces of the sentence are correct, rather than
whether the whole sentence is exactly correct.  Because the
inside-outside product is proportional to the probability that a given
constituent is correct, we can use it maximize correct pieces rather
than the whole.  We give various algorithms using the inside-outside
probabilities for maximizing performance on these piecewise metrics.
We also show that surprisingly, a similar problem, maximizing
performance on the well known zero crossing brackets rate, is
NP-Complete.  Finally, we give an algorithm that, using the inside and
outside probabilities, allows the tradeoff of precision versus recall.
These algorithms can be easily specified in the format of Chapter
\ref{ch:semi}.

Next, we show another use for the inside-outside probabilities:
Data-Oriented Parsing.  When we began this research, DOP was one of
the most promising techniques available, with reported results an
order of magnitude better than other parsers.  However, the only
algorithms available for parsing using the DOP model required
exponentially large grammars.  Furthermore, these algorithms were
randomized algorithms with some chance of failure.  We show how to use
the inside-outside techniques of Chapter \ref{ch:max} to parse the DOP
model in $O(n^3)$ time, deterministically, without sampling at all.
We also show a grammar construction technique that is linear in
sentence length, rather than exponential.  Using these techniques, we
are able to parse 500 times faster than previous algorithms.  Our
results are not as good as the previously published results, and we
give an analysis of the data that shows that these previous results
are probably due to a fortuitous split of the data into test and
training sections, or to easy data.

The third novel use we give for the inside-outside probabilities is to
speed parsing, which we discuss in Chapter \ref{ch:thresh}.  Parsers
can be sped up using {\em thresholding}, a technique in which some low
probability hypotheses are discarded, speeding later parsing.  Since
the normalized inside-outside probability gives the probability that
any constituent is correct, it is the mathematically ideal probability
to use to determine which hypotheses to discard.  However, since the
outside probability cannot be determined until after parsing is
complete, we instead use three approximations to the inside-outside
probabilities.  These include a variation on beam search in which,
rather than thresholding based only on the inside probability, we also
include a very simple approximation to the outside probability;
another thresholding technique that uses a more complicated
approximation to the inside-outside probability which takes into
account the whole sentence; and a multiple pass technique that uses
the inside-outside probability from one pass to threshold later
passes.  All three of these algorithms lead to significantly improved
speed.  In order to maximize performance using all of these algorithms
at once, it is necessary to maximize many parameters simultaneously.
We give a novel algorithm for maximizing the parameters of multiple
thresholding algorithms.  Rather than directly maximizing accuracy,
this algorithm maximizes the inside probability, which turns out to be
much more efficient.  Combining all of the thresholding algorithms
together leads to about a factor of 30 speedup over traditional
thresholding algorithms at the same error rates.  All of the
thresholding algorithms can be succinctly described using the
item-based descriptions of Chapter \ref{ch:semi}.

Despite the usefulness of the inside and outside probabilities, as
shown in the previous chapters, most state of the art parsing
formalisms cannot be used to compute either the inside or the outside
scores.  In Chapter \ref{ch:PFG}, we introduce a grammar formalism,
Probabilistic Feature Grammar (PFG), that combines the best properties
of most of the previous existing formalisms, but more elegantly.  PFGs
can be used to compute both inside and outside probabilities, meaning
that they can be used with the previously introduced algorithms.  PFGs
achieve state of the art performance on parsing tasks.

This thesis shows that using the inside-outside probabilities is a
powerful, general technique.  We have simplified the process of
computing inside and outside probabilities for new parsing algorithms.
We have shown how to use these probabilities to improve performance by
matching parsing algorithms to metrics; to quickly parse DOP grammars;
and to quickly parse Probabilistic Context-Free Grammars (PCFGs) and
PFGs with novel thresholding techniques.  Finally, we have introduced
a novel formalism, PFG, for which the inside and outside probabilities
can be easily computed, and that achieves state of the art
performance.

In writing this thesis, we have taken into account that it will be a
rare person who wishes to read the thesis in its entirety.  Thus,
whenever it is reasonable to make a chapter self-contained, or to
isolate interdependencies, we have sought to do so.

\begin{figure}
\begin{center}
\begin{tabular}{c}
\psfig{figure=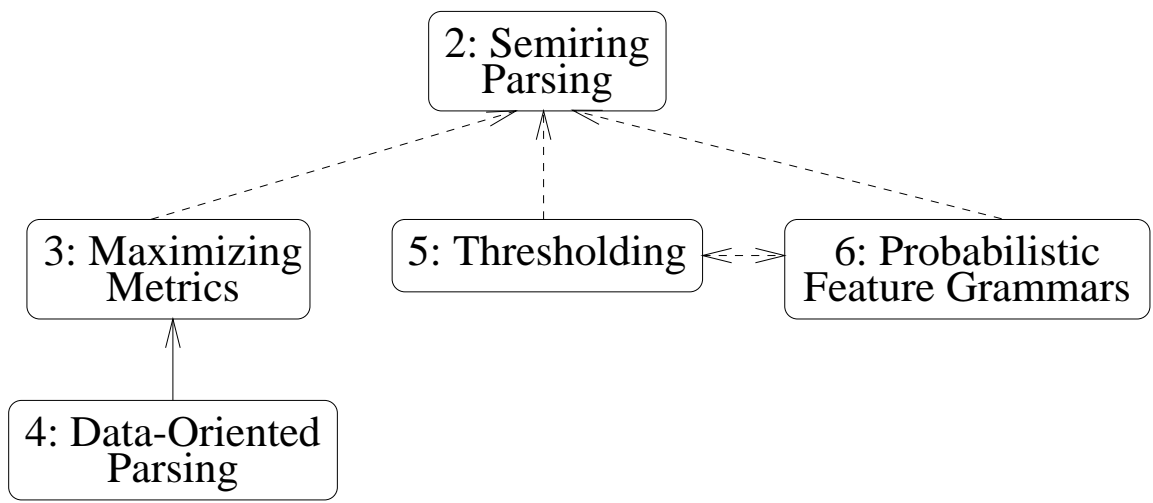,width=6in}
\end{tabular}
\end{center}
\caption{Dependencies in the thesis}\label{fig:organization}
\end{figure}

Figure \ref{fig:organization} shows the organization and dependencies
of the content chapters of the thesis.  The chapters on maximizing
metrics, thresholding, and probabilistic feature grammars all depend
somewhat on the semiring parsing chapter, in that algorithms in these
chapters are given using the format and theory of semiring parsing.
However, to keep the chapters self-contained, all algorithms in these
chapters are also presented in traditional pseudocode.  The most
important dependency is that of the Data-Oriented Parsing chapter,
which uses algorithms and ideas from the chapter on maximizing
metrics, and should probably not be read on its own.  The thresholding
and probabilistic feature grammar chapters each depend somewhat on
the other, and can best be appreciated as a pair, although each can
be read separately.

% -*- mode: latex; -*-

\chapter{Semiring Parsing} \label{ch:semi}

\newcommand{\thesis}[1]{#1}
\newcommand{\paper}[1]{}
\newcommand{\chapterpaper}{chapter}

\paper{
\begin{abstract}
We present a system for describing parsers that allows a single,
simple representation to be used for describing parsers that compute
recognition, derivation forests, Viterbi, inside, and outside values,
and other values.  We generalize the outside computation to semirings.
We show how to use the same representation to describe grammar
transformations, which can be used to transform grammars in ways that
preserve the values in commutative semirings, including the inside and
Viterbi semirings.
\end{abstract}
}%paper

\thesis{In this chapter, we present a system for describing parsers
that allows a single simple representation to be used for describing
parsers that compute inside and outside probabilities, as well as many
other values, including recognition, derivation forests, and Viterbi
values.  This representation will be used throughout the thesis to
describe the parsing algorithms we develop.}%thesis

\section{Introduction}  

For a given grammar and string, there are many interesting quantities
we can compute.  We can determine whether the string is generated by
the grammar; we can enumerate all of the derivations of the string; if
the grammar is probabilistic, we can compute the inside and outside
probabilities of components of the string.  Traditionally, a different
parser description has been needed to compute each of these values.
For some parsers, such as CKY parsers, all of these algorithms (except
for the outside parser) strongly resemble each other.  For other
parsers, such as Earley parsers, the algorithms for computing each
value are somewhat different, and a fair amount of work can be
required to construct each one.  We present a formalism for describing
parsers such that a single simple description can be used to generate
parsers that compute all of these quantities and others.  This will be
especially useful for finding parsers for outside values, and for
parsers that can handle general grammars, like Earley-style parsers.

\paper{Although our description format is not limited to Context-Free
Grammars (CFGs), we will begin by considering parsers for this common
formalism.  The input string will be denoted $w_1 w_2... w_n$.  We
will refer to the complete string as the sentence.  A CFG $G$ is a
4-tuple $\langle N,\Sigma, R, S\rangle$ where $N$ is the set of
nonterminals including the start symbol $S$, $\Sigma$ is the set of
terminal symbols, and $R$ is the set of rules, each of the form $A
\rightarrow \alpha$ for $A\in N$ and $\alpha\in(N\cup\Sigma)^*$.  (We
use $R$ rather than the more common $P$ to avoid confusion with
probabilities, which will be introduced later.)  Let $V=N\cup\Sigma$.
We will use the symbol $\Rightarrow$ for immediate derivation and
$\derives$ for its reflexive, transitive closure.} %paper
%this text mostly stolen from Shieber:93a

\begin{figure}
\begin{tabbing}
\verb|    |\=\verb|    |\=\verb|    |\=\verb|    |\=\verb|    |\=\verb|    |\=\verb|    |\=\verb|    |\=\verb|    |\=\verb|    |\= \kill
$\mi{boolean chart}[1..n, 1..|N|, 1..n\!+\!1] := \mi{FALSE};$ \\
\alforeach start $s$ \\
 \> \alforeach rule $A \rightarrow w_s$ \\
 \>  \> $\mi{chart}[s, A, s\!+\!1] := \mi{TRUE}$;  \\
\alforeach length $l$, shortest to longest \\
 \> \alforeach start $s$ \\
 \>  \> \alforeach split length $t$ \\
 \>  \>  \> \alforeach rule $A \rightarrow BC\;\in R$\\
 \>  \>  \>  \> {\em /* extra TRUE for expository purposes */} \\
 \>  \>  \>  \> $\mi{chart}[s, A, s\!+\!l] := \mi{chart}[s, A, s\!+\!l] \;\vee$ \\
 \>  \>  \>  \>  \>  \> $\mi{chart}[s, B, s\!+\!t] \wedge \mi{chart}[s\!+\!t,
C, s\!+\!l] \wedge \mi{TRUE}$;\\
 \>  \>  \>  \>  \> \\
\alreturn $\mi{chart}[1, S, n\!+\!1]$;
\end{tabbing}
\caption{CKY Recognition Algorithm}\label{fig:CKYiterative}
\end{figure}

\paper{We will illustrate the similarity of parsers for computing
different values using the CKY algorithm 
\cite{Kasami:65a,Younger:67a} as an example.  We can write this
algorithm in its iterative form as shown in Figure
\ref{fig:CKYiterative}.  Here, we explicitly construct a boolean
chart, $\mi{chart}[1 .. n, 1..|N|, 1 .. n\!+\!1]$.  Element
$\mi{chart}[i, A, j]$ contains $\mi{TRUE}$ if and only if $A\derives
w_i... w_{j-1}$.  The algorithm consists of a first set of loops to
handle the singleton productions; a second set of loops to handle the
binary productions; and a return of the start symbol's chart
entry.}%paper

\begin{figure}
\begin{tabbing}
\verb|    |\=\verb|    |\=\verb|    |\=\verb|    |\=\verb|    |\=\verb|    |\=\verb|    |\=\verb|    |\=\verb|    |\=\verb|    |\= \kill
$\mi{float chart}[1..n, 1..|N|, 1..n\!+\!1] := 0;$ \\
\alforeach start $s$ \\
 \> \alforeach rule $A \rightarrow w_s$ \\
 \>  \> $\mi{chart}[s, A,s\!+\!1]:= P(A\rightarrow w_s)$; \\
\alforeach length $l$, shortest to longest \\
 \> \alforeach start $s$ \\
 \>  \> \alforeach split length $t$ \\
 \>  \>  \> \alforeach rule $A \rightarrow BC\;\in R$\\
 \>  \>  \>  \>  $\mi{chart}[s, A, s\!+\!l] := \mi{chart}[s, A, s\!+\!l] \;+$ \\
 \>  \>  \>  \>  \>  \> $\mi{chart}[s, B, s\!+\!t] \times \mi{chart}[s\!+\!t, C, s\!+\!l] \times P(A\rightarrow BC)$; \\
\alreturn $\mi{chart}[1, S, n\!+\!1]$;
\end{tabbing}
\caption{CKY Inside Algorithm}\label{fig:CKYinside}
\end{figure}

\paper{Next, we consider probabilistic grammars, in which we associate a
probability with every rule, $P(A \rightarrow \alpha)$.  These
probabilities can be used to associate a probability with a particular
derivation, equal to the product of the rule probabilities used in the
derivation, or to associate a probability with a set of derivations,
$A\derives w_i... w_{j-1}$ equal to the sum of the probabilities of
the individual derivations.  We call this latter probability the {\it
inside} probability of $i, A, j$.  We can rewrite the CKY algorithm to
compute the inside probabilities, as shown in Figure
\ref{fig:CKYinside} \cite{Baker:79b,Lari:90a}.}%paper

\thesis{We will compare the CKY algorithm
\cite{Kasami:65a,Younger:67a} to the inside algorithm
\cite{Baker:79b,Lari:90a} to illustrate the similarity of parsers for
computing different values.  Both of these parsers were described in
in Section \ref{sec:CFG}; we repeat the code here in Figures
\ref{fig:CKYiterative} and \ref{fig:CKYinside}.}%thesis
{} Notice how similar the inside algorithm is to the recognition
algorithm: essentially, all that has been done is to substitute $+$
for $\vee$, $\times$ for $\wedge$, and $P(A \rightarrow w_s)$ and $P(A
\rightarrow BC)$ for $\mi{TRUE}$.  For many parsing algorithms, this,
or a similarly simple modification, is all that is needed to create a
probabilistic version of the algorithm.  On the other hand, a simple
substitution is not always sufficient.  To give a trivial example, if
in the CKY recognition algorithm we had written
$$
\mi{chart}[s, A, s\!+\!l] := \mi{chart}[s, A, s\!+\!l] \;\vee \;
 \mi{chart}[s, B, s\!+\!t] \wedge \mi{chart}[s\!+\!t,C, s\!+\!l];
$$
instead of the less natural
$$
\mi{chart}[s, A, s\!+\!l] := \mi{chart}[s, A, s\!+\!l] \;\vee \;
 \mi{chart}[s, B, s\!+\!t] \wedge \mi{chart}[s\!+\!t,C, s\!+\!l] \wedge \mi{TRUE};
$$
larger changes would be necessary to create the inside algorithm.

Besides recognition, there are four other quantities that are commonly
computed by parsing algorithms: derivation forests, Viterbi scores,
number of parses, and outside probabilities.  The first quantity, a
derivation forest, is a data structure that allows one to efficiently
compute the set of legal derivations of the input string.  The
derivation forest is typically found by modifying the recognition
algorithm to keep track of ``back pointers'' for each cell of how it
was produced.  The second quantity often computed is the Viterbi
score, the probability of the most probable derivation of the
sentence.  This can typically be computed by substituting $\times$ for
$\wedge$ and $\max$ for $\vee$.  Less commonly computed is the total
number of parses of the sentence, which like the inside values, can be
computed using multiplication and addition; unlike for the inside
values, the probabilities of the rules are not multiplied into the
scores.  \paper{There is one last commonly computed quantity, the
outside probabilities, which we will describe later, in Section
\ref{sec:reverse}.}\thesis{One last commonly computed quantity, the
outside probability, cannot be found with modifications as simple as
the others.  We will discuss how to compute outside quantities later,
in Section \ref{sec:reverse}.}

One of the key ideas of this \chapterpaper{} is that all five of these
commonly computed quantities can be described as elements of {\em
complete semirings} \cite{Kuich:97a}.  A complete semiring is a set of
values over which a multiplicative operator and a commutative additive
operator have been defined, and for which infinite summations are
defined.  For parsing algorithms satisfying certain conditions, the
multiplicative and additive operations of any complete semiring can be
used in place of $\wedge$ and $\vee$, and correct values will be
returned.  We will give a simple normal form for describing parsers,
then precisely define complete semirings, the conditions for
correctness, and a simple normal form for describing parsers.

We now describe our normal form for parsers, which is very similar to
that used by \newcite{Shieber:93a} and by \newcite{Sikkel:93a}.  In
most parsers, there is at least one chart of some form.  In our normal
form, we will use a corresponding concept, $\mi{items}$.  Rather than,
for instance, a chart element $\mi{chart}[i, A, j]$, we will use an
item $[i, A, j]$.  Conceptually, chart elements and items are
equivalent.  Furthermore, rather than use explicit, procedural
descriptions, such as
$$
\mi{chart}[s, A, s\!+\!l] := \mi{chart}[s, A, s\!+\!l] \;\vee \;
 \mi{chart}[s, B, s\!+\!t] \wedge \mi{chart}[s\!+\!t,C, s\!+\!l] \wedge \mi{TRUE}
$$
we will use {\em inference rules} such as
$$
\shortinfer{R(A \rightarrow BC)\rspace [i, B, k]\rspace [k, C, j]}{[i, A,j]}{}
$$
The meaning of an inference rule is that if the top line is all true,
then we can conclude the bottom line.  For instance, this example
inference rule can be read as saying that if $A \rightarrow BC$ and $B
\derives w_i...w_{k-1}$ and $C \derives w_k...w_{j-1}$, then $A
\derives w_1...w_{j-1}$.

The general form for an inference rule will be
$$
\shortinfer{A_1 \cdots A_k}{B}{}
$$
where if the conditions $A_1... A_k$ are all true, then we infer that
$B$ is also true.  The $A_i$ can be either items, or (in an extension
to the usual convention for inference rules), can be rules, such as
$R(A \rightarrow BC)$.  We write $R(A \rightarrow BC)$ rather than $A
\rightarrow BC$ to indicate that we could be interested in a value
associated with the rule, such as the probability of the rule if we
were computing inside probabilities.  If an $A_i$ is in the form
$R(...)$, we call it a {\em rule}.  All of the $A_i$ must be rules or items;
when we wish to refer to both rules and items, we use the word {\em
terms}.

\begin{figure}
$$
\begin{array}{cl}
\mbox{\bf Item form:} \\
{[i, A, j]} \\ \\
\mbox{\bf Goal:} \\
{[1, S, n\!+\!1]} \\\\
\mbox{\bf Rules:} \\
\infer{R(A \rightarrow w_i)}{[i, A, i\!+\!1] }{} & \mbox{Unary} \\
\infer{R(A \rightarrow BC)\rspace [i, B, k]\rspace [k, C, j]}{[i, A,
j]}{}
& \mbox {Binary}
\end{array}
$$
\caption{Item-based description of a CKY parser}\label{fig:CKYitem}
\end{figure}

We now give an example of an item-based description, and its
semantics.  Figure \ref{fig:CKYitem} gives a description of a CKY
style parser.  For this example, we will use the inside semiring,
whose additive operator is addition and whose multiplicative operator
is multiplication.  We use the input string $xxx$ to the following
grammar:
\begin{equation}
\begin{array}{rcll}
S & \rightarrow & XX &  1.0 \\
X & \rightarrow & XX &  0.2\\
X & \rightarrow & x &  0.8
\end{array}
\label{eqn:firstgrammar}
\end{equation}
Our first step is to use the unary rule,
$$
\shortinfer{R(A \rightarrow w_i)}{[i, A, i\!+\!1] }{}
$$
The effect of the unary rule will exactly parallel the first set of
loops in the CKY inside algorithm.  We will instantiate the free
variables of the unary rule in every possible way.  For instance, we
instantiate the free variable $i$ with the value 1, and the free
variable $A$ with the nonterminal $X$.  Since $w_1 = x$, the
instantiated rule is then
$$
\shortinfer{R(X \rightarrow x)}{[1, X, 2]}{}
$$
Because the value of the top line of the instantiated unary rule, $R(X
\rightarrow x)$, has value 0.8, we deduce that the bottom line, $[1,
X, 2]$, has value $0.8$.  We instantiate the rule in two other ways,
and compute the following chart values:
$$
\begin{array}{rcl}
{[1, X,2]} & = &0.8 \\
{[2, X,3]} & = &0.8 \\
{[3, X,4]} & = &0.8 
\end{array}
$$
Next, we will use the binary rule,
$$
\shortinfer{R(A \rightarrow BC)\rspace [i, B, k]\rspace [k, C, j]}{[i, A,j]}{}
$$
The effect of the binary rule will parallel the second set of loops
for the CKY inside algorithm.  Consider the instantiation $i=1$,
$k=2$, $j=3$, $A=X$, $B=X$, $C=X$,
$$
\shortinfer{R(X \rightarrow XX)\rspace [1, X, 2]\rspace [2, X, 3]}{[1, X, 3]}{}
$$
We use the multiplicative operator of the semiring of interest to
multiply together the values of the top line.  In the inside semiring,
the multiplicative operator is just multiplication, so we get:
$0.2\times 0.8\times 0.8 = 0.128$, and deduce that $[1, X, 3]$ has
value 0.128.  We can do the same thing for the instantiation $i=2$,
$k=3$, $j=4$, $A=X$, $B=X$, $C=X$, getting the following item values:
$$
\begin{array}{rcl}
{[1, X,3]} & = &0.128 \\
{[2, X,4]} & = &0.128 
\end{array}
$$
We can also deduce that 
$$
\begin{array}{rcl}
{[1, S,3]} & = &0.128 \\
{[2, S,4]} & = &0.128 
\end{array}
$$
There are two more ways to instantiate the conditions of the binary
rule: 
$$
\begin{array}{c}
\shortinfer{R(S \rightarrow XX)\rspace [1, X, 2]\rspace [2, X, 4]}{[1, S, 4]}{} \\
\infer{R(S \rightarrow XX)\rspace [1, X, 3]\rspace [3, X, 4]}{[1, S, 4]}{}
\end{array}
$$
The first has the value $1 \times 0.8 \times 0.128 = 0.1024$, and the
second also has the value 0.1024.  When there is more than one way to
derive a value for an item, we use the additive operator of the
semiring to sum them up.  In the inside semiring, the additive operator is just
addition, so the value of this item is $[1, S, 4] = 0.2048$.  Notice
that the goal item for the CKY parser is $[1, S, 4]$.  Thus we now
know that the inside value for $xxx$ is $0.2048$.  The goal item
exactly parallels the return statement of the CKY inside algorithm.

Besides the fact that this item-based description is simpler than the
explicit looping description, there will be other reasons we wish to
use it.  Unlike the other quantities we wish to compute, the outside
probabilities\paper{, values related to the inside % probabilities
that we will define later,}{} cannot be computed by simply
substituting a different semiring into either an iterative or
item-based description.  Instead, we will show how to compute the
outside probabilities using a modified interpreter of the same
item-based description used for computing the inside probabilities.
%Furthermore, we will show how different interpreters
%of the same item-based description can lead to more efficient parsers
%for some grammars.

\subsection{Earley Parsing}

Many parsers are much more complicated than the CKY parser.  This will
make our description of semiring parsing a bit longer, but will also
explain why our format is so useful: these complexities occur in many
different parsers, and the ability of semiring parsing to handle them
automatically will prove to be its main attraction.

Most of the interesting complexities we wish to discuss are exhibited
by Earley parsing \cite{Earley:70a}.  Earley's parser is often
described as a bottom-up parser with top-down filtering.  In a
probabilistic framework, the bottom-up and top-down aspects are very
different; the bottom-up sections compute probabilities, while the
top-down filtering non-probabilistically removes items that cannot be
derived.  In order to capture these differences, we expand our
notation for deduction rules, to the following form:
$$
\shortinfer{A_1 \cdots A_k}{B}{C_1 \cdots C_l}
$$
$C_1 \cdots C_l$ are {\em side conditions}, interpreted
non-probabilistically, while $A_1 \cdots A_k$ are {\em main
conditions} with values in whichever semiring we are using.  While the
values of all main conditions are multiplied together to yield the
value for the item under the line, the side conditions are interpreted
in a boolean manner: either they all have non-zero value or not.  The rule
can only be used if all of the side conditions have non-zero value,
but other than that, their values are ignored.

\begin{figure}
$$
\begin{array}{cl}
\mbox{\bf Item form:} \\
{[i, A\rightarrow \alpha\bullet\beta, j]} \\ \\
\mbox{\bf Goal:} \\
{[1, S' \rightarrow S \bullet {}, n\!+\!1]} \\& \\
\mbox{\bf Rules:} \\
\infer{}{[1, S'\rightarrow {} \bullet S, 1]}{} &\mbox{Initialization} \\
\infer{[i, A \rightarrow \alpha\bullet w_j \beta, j]}
{[i, A \rightarrow \alpha w_j\bullet \beta, j\!+\!1]}{}
 &\mbox{Scanning}\\
\infer{R(B \rightarrow\gamma)}{[j, B \rightarrow {} \bullet\gamma, j]}
{[i, A \rightarrow \alpha\bullet B\beta, j]} &\mbox{Prediction} \\
\infer{[i, A \rightarrow \alpha\bullet B\beta, k]\rspace
[k, B \rightarrow\gamma\bullet {}, j]}
{[i, A \rightarrow \alpha B\bullet\beta, j]}{}  &\mbox{Completion}
\end{array}
$$
\caption{Earley Parsing}\label{fig:Earley}
\end{figure}

Figure \ref{fig:Earley} gives an item-based description of Earley's
parser.  We assume the addition of a distinguished nonterminal $S'$
with a single rule $S' \rightarrow S$.  An item of the form $[i, A
\rightarrow \alpha\bullet\beta, j]$ asserts that $A \Rightarrow
\alpha\beta\derives w_i... w_{j-1}\beta$.  

The prediction rule includes a side condition, making it a good
example.  The rule is:
$$
\infer{R(B \rightarrow\gamma)}{[j, B \rightarrow {} \bullet\gamma, j]}
{[i, A \rightarrow \alpha\bullet B\beta, j]}
$$
Through the prediction rule, Earley's algorithm guarantees that an
item of the form $[j, B \rightarrow {} \bullet\gamma, j]$ can only be
produced if $S \derives w_1...w_{j-1} B \delta$ for some $\delta$;
this top down filtering leads to significantly more efficient parsing
for some grammars than the CKY algorithm.  The prediction rule
combines side and main conditions.  The side condition
$$
[i, A \rightarrow \alpha \bullet B\beta, j]
$$
provides the top-down filtering, ensuring that only items that might
be used later by the completion rule can be predicted, while the main
condition,
$$
R(B \rightarrow \gamma)
$$
provides the probability of the relevant rule.  The side condition is
interpreted in a boolean fashion, while the main condition's actual
probability is used.

Unlike the CKY algorithm, Earley's algorithm can handle grammars with
epsilon ($\epsilon$), unary, and n-ary branching rules.  In some
cases, this can significantly complicate parsing.  For instance, given
unary rules $A \rightarrow B$ and $B \rightarrow A$, a cycle exists.
This kind of cycle may allow an infinite number of different
derivations, requiring an infinite summation to compute the inside
probabilities.  The ability of item-based parsers to handle these
infinite loops with ease is a major attraction.

\subsection{Overview}

This \chapterpaper{} will simplify the development of new parsers in several
ways.  First, it will simplify specification of parsers: the
item-based description is simpler than a procedural description.
Second, it will make it easier to generalize parsers to other tasks: a
single item-based description can be used to compute values in a
variety of semirings, and outside values as well.  This will be
especially advantageous for parsers that can handle loops resulting
from rules like $A \rightarrow A$ and computations resulting from
$\epsilon$ productions, both of which typically lead to infinite sums.
In these cases, the procedure for computing an infinite sum differs
from semiring to semiring, and the fact that we can specify that a
parser computes an infinite sum separately from its method of
computing that sum will be very helpful.

In the next section, we describe the basics of semiring parsing.  In
the following sections, we derive formulae for computing the values of
items in semiring parsers, and then describe an algorithm for
interpreting an item-based description.  Next, we discuss using this
same formalism for performing grammar transformations.  At the end of
the \chapterpaper{}, we give examples of using semiring parsers to solve a
variety of problems.

\section{Semiring Parsing}
\label{sec:semiringparsing}

In this section we first describe the inputs to a semiring parser: a
semiring, an item-based description, and a grammar.  Next, we give the
conditions under which a semiring parser gives correct results.  At
the end of this section we discuss three especially complicated and
interesting semirings.

\subsection{Semiring}

\label{sec:semiring}

In this subsection, we define and discuss semirings.  The best
introduction to semirings that we know of, and the one we follow here,
is that of \newcite{Kuich:97a}, who also gives more formal definitions
than those given in this \chapterpaper.

A semiring has two operations, $\oplus$ and $\otimes$, that
intuitively have most (but not necessarily all) of the properties of
the conventional $+$ and $\times$ operations on the positive integers.
In particular, we require the following properties: $\oplus$ is
associative and commutative; $\otimes$ is associative and distributes
over $\oplus$.  If $\otimes$ is commutative, we will say that the
semiring is commutative.  We assume an additive identity element,
which we write as 0, and a multiplicative identity element, which we
write as 1.  Both addition and multiplication can be defined over
finite sets of elements; if the set is empty, then the value is the
respective identity element, 0 or 1.  We also assume that $x \otimes 0
= 0 \otimes x = 0$ for all $x$.  In other words, a semiring is just
like a ring, except that the additive operator need not have an
inverse.  We will write
$$
\langle \Bbb{A}, \oplus, \otimes, 0, 1 \rangle
$$
to indicate a semiring over the set $\Bbb{A}$ with additive operator
$\oplus$, multiplicative operator $\otimes$, additive identity 0, and
multiplicative identity 1.
% Don't forget to distinguish left from right distributivity, if write
% out formulas..

For parsers with loops, i.e. those in which an item can be used to
derive itself, we will also require that sums of an infinite number of
elements be well defined.  In particular, we will require that the
semirings be {\em complete} \cite[p. 611]{Kuich:97a}.  This means that
sums of an infinite number of elements should be associative,
commutative, and distributive just like finite sums.  All of the
semirings we will deal with in this \chapterpaper{} are complete.  Completeness
is a somewhat stronger condition than we really need; we could,
instead, require that limits be appropriately defined for those infinite
sums that occur while parsing, but this weaker condition is more
complicated to describe precisely.

Certain semirings are {\em naturally ordered}, meaning that we can
define a partial ordering, $\sqsubseteq$, such that $x \sqsubseteq y$
if and only if there exists $z$ such that $x + z = y$.  We will call a
naturally ordered complete semiring $\omega$-continuous
\cite[p. 612]{Kuich:97a} if for any sequence $x_1, x_2, ...$ and for
any constant $y$, if for all $n$, $\bigoplus_{0\leq i\leq n} x_i
\sqsubseteq y$, then $\bigoplus_{i} x_i \sqsubseteq y$.  That is, if
every partial sum is less than or equal to $y$, then the infinite sum
is also less than or equal to $y$.  This important property makes it
easy to compute, or at least approximate, infinite sums.  All of the semirings we discuss
here will be both naturally ordered and $\omega$-continuous.

There will be several especially useful semirings in this
\chapterpaper{}, which are defined in Figure \ref{fig:semirings}.  We
will write $\Bbb{R}_a^b$ to indicate the set of real numbers from $a$
to $b$ inclusive, with similar notation for the natural numbers,
$\Bbb{N}$.  We will write $\Bbb{E}$ to indicate the set of all
derivations, where a derivation is an ordered list of grammar rules.
We will write $2^{\Bbb{E}}$ to indicate the set of all sets of
derivations.  There are three derivation semirings: the derivation
forest semiring, the Viterbi-derivation semiring, and the
Viterbi-n-best semiring.  The operators used in the derivation
semirings $(\cdot, \maxvit, \timesvit, \maxvitn,$ and $\timesvitn$)
will be described later, in Section \ref{sec:parseforest}.

\begin{figure}
\begin{tabular}{lp{4in}}
{\bf boolean} & $\langle \{\mi{TRUE}, \mi{FALSE}\,\}, \vee, \wedge,
\mi{FALSE}, \mi{TRUE} \rangle$ \\

{\bf inside} & $\langle \Bbb{R}_0^{\infty}, +, \times, 0, 1 \rangle$
\\

{\bf Viterbi} & $\langle \Bbb{R}_0^1, \max, \times, 0, 1 \rangle$ \\

{\bf counting} & $\langle \Bbb{N}_0^{\infty}, +, \times, 0, 1 \rangle$ \\

{\bf tropical semiring} & $\langle \Bbb{R}_0^{\infty}, \min, +,
\infty, 0 \rangle$ \\

{\bf arctic semiring} & $\langle \Bbb{R} \cup \{ -\infty \}, \max,
+, -\infty, 0 \rangle$ \\

{\bf derivation forest} & $ \langle 2^{\Bbb{E}}, \cup, \cdot, \emptyset,
\{\langle\rangle\} \rangle$ \\

{\bf Viterbi-derivation} &  $\langle \Bbb{R}_0^1 \times 2^{\Bbb{E}},
\maxvit, \timesvit, \langle 0, \emptyset \rangle,
\langle 1, \{ \langle\rangle \} \rangle \rangle $ \\

{\bf Viterbi-n-best} & $\langle \{ \mi{topn}(X) | X\in 2^{\Bbb{R}_0^1
\times \Bbb{E}}\}, \maxvitn, \timesvitn, \emptyset , \{ \langle 1, \{
\langle\rangle \} \rangle \} \rangle$

%\\

%{\bf DCG} & $\langle 2^\mi{terms}, \cup, \sqcup, \emptyset, \{X\} \rangle

%\\

\end{tabular}
\caption{Semirings Used: $\langle A, \oplus, \otimes, 0, 1 \rangle$} \label{fig:semirings}
\end{figure}

The inside semiring includes all non-negative real numbers, to be
closed under addition, and includes infinity to be closed under
infinite sums, while the Viterbi semiring contains only numbers up to
1, since that is all that is required to be closed under $\max$.  

There are two additional semirings, the tropical semiring (which
usually is restricted to natural numbers, but is extended to real
numbers here), and another semiring, which we have named the arctic
semiring, since it is the opposite of the tropical semiring, taking
maxima rather than minima.

The three derivation forest semirings can be used to find especially
important values: the derivation forest semiring computes all
derivations of a sentence; the Viterbi-derivation semiring computes
the most probable derivation; and the Viterbi-n-best semiring computes
the $n$ best derivations.  A derivation is simply a list of rules from
the grammar.  From a derivation, a parse tree can be derived, so the
derivation forest semiring is analogous to conventional parse forests.
Unlike the other semirings, all three of these semirings are
non-commutative.  The additive operation of these semirings is
essentially union or maximum, while the multiplicative operation is
essentially concatenation.  These semirings are relatively
complicated, and are described in more detail in Section
\ref{sec:parseforest}.

%The last semiring is the DCG semiring, inspired by
%\newcite{Tendeau:97a}.  Notice that we will be describing DCGs as a
%semiring, not as a grammar formalism.  The value of an item in the DCG
%semiring will be the set of term values it could have.  For instance,
%if a DCG had rules that constructed the semantic representation of a
%sentence, then the value of $[1, S, n\!+\!1]$ in the CKY semiring would be
%the set of all terms giving possible semantic representations, and the
%empty set if the sentence is unparsable.  The multiplicative operation
%is unification (actually, the cross product of possible unifications)
%while the additive operation is union.  The DCG semiring is described
%in more detail in Seciont \ref{sec:DCG}.

\subsection{Item-Based Description}

\label{sec:itembased}

A semiring parser requires an item-based description, $\mathcal{D}$,
of the parsing algorithm, in the form given earlier.  So far, we have
skipped one important detail of semiring parsing.  In a simple
recognition system, as used in deduction systems, all that matters is
whether an item can be deduced or not.  Thus, in these simple systems,
the order of processing items is relatively unimportant, as long as
some simple constraints are met.  On the other hand, for a semiring
such as the inside semiring, there are important ordering constraints:
for instance, we cannot compute the inside value of a CKY-style chart
element until the inside values of all of its children have been
computed.

Thus, we need to impose an ordering on the items, in such a way that
no item precedes any item on which it depends.  We will associate with
each item $x$ a ``bucket'' $B$ and write $\mi{bucket}(x) = B$.  We
order the buckets in such a way that if item $y$ depends on item $x$,
then $\mi{bucket}(x)\leq \mi{bucket}(y)$.  We will write $\mi{first}$,
$\mi{last}$, $\mi{next}(B)$, and $\mi{previous}(B)$ for the first,
last, next and previous buckets respectively.  For some pairs of
items, it may be that both depend, directly or indirectly, on each
other; we associate these item with special ``looping'' buckets, whose
values may require infinite sums to compute.  We will also call a
bucket looping if an item in it depends on itself.  The predicate
$\mi{loop}(B)$ will be true for looping buckets $B$.

One way to achieve a bucketing with the required ordering constraints
is to create a graph of the dependencies, with a node for each item,
and an edge from each item $x$ to each item $b$ that depends on it.
We then separate the graph into its strongly connected components, and
perform a topological sort.  Items forming singleton strongly
connected components are in their own buckets; items forming
non-singleton strongly connected components are together in looping
buckets.  

An actual example may help here.  Consider an example grammar, such as
$$
\begin{array}{rcl}
S & \rightarrow & CAC \\
C& \rightarrow &c \\
B& \rightarrow &A \\
A& \rightarrow &B \\
A& \rightarrow &a
\end{array}
$$
and an input sentence $cac$, parsed with the Earley parser of Figure
\ref {fig:Earley}.
%There are about 14 productions used to parse the sentence; too many
%to give them all.
It will be possible to derive items such as $[1, C \rightarrow \bullet
c, 1]$ through prediction, $[1, C\rightarrow c\bullet {},2]$ through
scanning, and $[1, S \rightarrow C\bullet AC, 2]$ through completion.
Each of these items would form a singleton strongly connected
component, and could be put into its own bucket.  Now, using
completion, we see that
$$
\shortinfer{[2, B \rightarrow {} \bullet A, 2] \rspace[2, A
\rightarrow a \bullet {}, 3] }{[2, B\rightarrow A \bullet {}, 3]}{}
$$
Next comes the looping part.  Notice that
$$
\shortinfer{[2, A \rightarrow {} \bullet B, 2] \rspace[2, B
\rightarrow A \bullet {}, 3] }{[2, A\rightarrow B \bullet {}, 3]}{}
$$
and
$$
\shortinfer{[2, B \rightarrow {} \bullet A, 2] \rspace[2, A
\rightarrow B \bullet {}, 3] }{[2, B\rightarrow A \bullet {}, 3]}{}
$$
Thus, the items $[2, B\rightarrow A \bullet {}, 3]$ and $[2, A\rightarrow
B \bullet {}, 3]$ can each be derived from the other, and since they
depend on each other, will be placed together in a looping bucket.

%For a CKY style parser with buckets formed from the strongly connected
%components, each item will be in its own bucket; there are many
%possible orderings, but one simple possibility is to sort buckets from
%shortest to longest, sub-sorted from left to right.  This bucket
%ordering yields the usual order of computation for a CKY style parser.
%Next, consider an Earley style parser applied to a grammar with a
%large number of unary and epsilon productions.  Each item would
%typically be in a looping bucket together with all other items
%covering the same span.
%%and not containing terminal symbols, but ignore that for now
%The items covering a given span would need to be in a looping bucket,
%since they may all depend on each other, because with a large enough
%number of unary and epsilon productions, each item covering a span can
%be used to produce the others.  As in the CKY case, the buckets would
%perhaps be ordered from shortest to longest.

A topological sort is not the only way to bucket the items.  In
particular, for items such that neither depends on the other, it is
possible to place them into a bucket together with no loss of
efficiency.  For some descriptions, this could be used to produce
faster, simpler parsing algorithms.  For instance, in a CKY style
parser, we could simply place all items of the same length in the same
bucket, ordering buckets from shortest to longest, avoiding the need
to perform a topological sort.

Later, when we discuss algorithms for interpreting an item-based
description, we will need another concept.  Of all the items
associated with a bucket $B$, we will be able to find derivations for
only a subset.  If we can derive an item $x$ associated with bucket
$B$, we write $x \in B$, and say that item $x$ is in bucket $B$.  For
example, the goal item of a parser will almost always be {\em
associated} with the last bucket; if the sentence is grammatical, the
goal item will be {\em in} the last bucket, and if it is not
grammatical, it won't be.

It will be useful to assume that there is a single, variable free goal
item, and that this goal item does not occur as a condition for any
rules.  We can always add a new goal item $[\mi{goal}]$ and a rule
$\shortinfer{[\mi{old-goal}]}{[\mi{goal}]}{}$ where $[\mi{old-goal}]$ is
the goal in the original description.  We will assume in general that this
transformation has been made, or is not necessary.

\subsection{The Grammar}
A semiring parser also requires a grammar as input.  We will need a
list of rules in the grammar, and a function that gives the value for
each rule in the grammar.  This latter function will be semiring
specific.  For instance, for computing the inside and Viterbi
probabilities, the value of a grammar rule is just the conditional
probability of that rule, or 0 if it is not in the grammar.  For the
boolean semiring, the value is $\mi{TRUE}$ if the rule is in the grammar,
$\mi{FALSE}$ otherwise.  For the counting semiring, the value is 1 if the
rule is in the grammar, 0 otherwise.  We call this function
$R(\mi{rule})$.  This function replaces the set of rules $R$ of a
conventional grammar description; a $\mi{rule}$ is in the grammar if
$R(\mi{rule})$ is not the zero element of the semiring.

%The second function we might wish to calculate is the list of rules
%consistent with a rule template.  For instance, in a CKY style parser,
%given a nonterminal in the left cell, and a nonterminal in the right
%cell, we might wish to know all the possible parent nonterminals.  We
%call this function $\mi{rules}(\mi{template})$.

\subsection{Conditions for Correct Processing}
\label{sec:correctprocessing}

We will say that a semiring parser works correctly if for any grammar,
input and semiring, the value of the input according to the grammar equals
the value of the input using the parser.  In this subsection, we will
define the value of an input according to the grammar; the value of an input
using the parser; and give a sufficient condition for a semiring
parser to work correctly.

From this point onwards, unless we specifically mention otherwise, we
will assume that some fixed semiring, item-based description, and
grammar have been given, without specifically mentioning which ones.

\subsubsection{Value according to grammar}

Under certain conditions, a semiring parser will work correctly for any
grammar, input, and semiring.  First, we must define what we mean by
working correctly.  Essentially, we mean that the value of a sentence
according to the grammar equals the value of the sentence using the parser.

Consider a derivation $E$, consisting of grammar rules $e_1, e_2,...,
e_l$.  We define the value of the derivation to be simply the product
(in the semiring) of the values of the rules used in $E$:
$$
V_G(E) = \bigotimes_{i=1}^l R(e_i)
$$
Then we can define the value of a sentence that can be derived using
grammar derivations $E^1, E^2,..., E^k$ to be:
$$
V_G = \bigoplus_{j=1}^k V_G(E^j)
$$
where $k$ is potentially infinite.  In other words, the value of the
sentence according to the grammar is the sum of the values of all
derivations.  We will assume that in each grammar formalism there is
some way to define derivations uniquely; for instance, in CFGs, one
way would be using left-most derivations.  For simplicity, we will
simply refer to derivations, rather than e.g. left-most derivations,
since we are never interested in non-unique derivations.

\subsubsection{Example of value according to grammar}

A short example will help clarify.  We consider the following grammar:
\begin{equation}
\begin{array}{rcll}
S & \rightarrow & AA & R(S \rightarrow AA) \\
A & \rightarrow & AA & R(A \rightarrow AA) \\
A & \rightarrow & a & R(A \rightarrow a) 
\end{array}
\label{eqn:simplegrammar}
\end{equation}
and the input string $aaa$.  There are two grammar derivations, the
first of which is
$$
S \derivesby{S}{AA} AA \derivesby{A}{AA} AAA \derivesby{A}{a} aAA \derivesby{A}{a} aaA
\derivesby{A}{a} aaa
$$
which has value
$$
R(S \rightarrow AA) \otimes R(A \rightarrow AA) \otimes R(A
\rightarrow a) \otimes R(A \rightarrow a) \otimes R(A \rightarrow a)
$$
Notice that the rules in the value are the same rules in the same
order as in the derivation.  The other grammar derivation is
$$
S \derivesby{S}{AA} AA \derivesby{A}{a} aA \derivesby{A}{AA} aAA \derivesby{A}{a} aaA
\derivesby{A}{a} aaa
$$
which has value
$$
R(S \rightarrow AA) \otimes R(A \rightarrow a) \otimes R(A
\rightarrow AA) \otimes R(A \rightarrow a) \otimes R(A \rightarrow a)
$$
We note that for commutative semirings, the value of the two grammar derivations
are equal, but for non-commutative semirings, they differ.

The value of the sentence is the sum of the values of the two
derivations,
$$
\begin{array}{l}
R(S \rightarrow AA) \otimes R(A \rightarrow AA) \otimes R(A
\rightarrow a) \otimes R(A \rightarrow a) \otimes R(A \rightarrow a) \\
\oplus \\
R(S \rightarrow AA) \otimes R(A \rightarrow a) \otimes R(A
\rightarrow AA) \otimes R(A \rightarrow a) \otimes R(A \rightarrow a)
\end{array}
$$

\subsubsection{Item derivations}

Next, we must define item derivations, i.e. derivations using the
item-based description of the parser.  We will define item derivation
in such a way that for a correct parser description, there will be
exactly one item derivation for each grammar derivation.  The value of
a sentence using the parser is the sum of the value of all item
derivations of the goal item.

We say that $\shortinfer{a_1...a_k}{b}{c_1...c_l}$ is an {\em
instantiation} of deduction rule
$\shortinfer{A_1... A_k}{B}{C_1... C_l}$ whenever the first expression
is a variable-free instance of the second; that is, the first
expression is the result of consistently substituting constant terms
for each variable in the second.  Now, we can define an {\em item
derivation tree}.  Intuitively, an item derivation tree for $x$ just
gives a way of deducing $x$ from {\em ground} items (items that don't
depend on other items, i.e. items that can be deduced using rules that
have no items in the $A_i$.)  We define an item derivation tree
recursively.  The base case is rules of the grammar: $\langle r
\rangle$ is an item derivation tree, where $r$ is a rule of the
grammar.  Also, if $D_{a_1},...,D_{a_k}, D_{c_1},...,D_{c_l}$ are
derivation trees headed by $a_1...a_k, c_1...c_l$ respectively, and if
$\shortinfer{a_1...a_k}{b}{c_1...c_l}$ is the instantiation of a
deduction rule, then $\langle b : D_{a_1}, ..., D_{a_k} \rangle$ is
also a derivation tree.  Notice that the $D_{c_1}...D_{c_l}$ do not
occur in this tree: they are side conditions, and although their
existence is required to prove that $c_1...c_l$ could be derived, they
do not contribute to the value of the tree.  We will write
$\shortinfer{a_1...a_k}{b}{}$ to indicate that there is an item
derivation tree of the form $\langle b: D_{a_1}, ..., D_{a_k}
\rangle$.  As mentioned in Section \ref{sec:itembased}, we will write
$x \in B$ if $\mi{bucket}(x) = B$ and there is an item derivation tree
for $x$.

\subsubsection{Example of item derivation}

We can continue the example of parsing $aaa$, now using the item
based CKY parser of Figure \ref{fig:CKYitem}.  There are two item
derivation trees for the goal item; we give the first as an example,
displaying it as a tree, rather than with angle bracket notation, for
simplicity.  Figure \ref{fig:itemexample} shows this tree and the
corresponding grammar derivation.

\begin{figure}
\begin{center}
\begin{tabular}{c}
$S \derivesby{S}{AA} AA \derivesby{A}{AA} AAA \derivesby{A}{a} aAA
\derivesby{A}{a} aaA \derivesby{A}{a} aaa$
\\
Grammar Derivation
\\
\\
\hspace{-5em}
\leaf{$a$}
\branch{1}{$R(A\rightarrow a)$}
\leaf{$a$}
\branch{1}{$R(A\rightarrow a)$}
\branch{2}{$R(A\rightarrow AA)$}
\leaf{a}
\branch{1}{$R(A\rightarrow a)$}
\branch{2}{$R(S\rightarrow AA)$}
\tree
\\
Grammar Derivation Tree
\\
\\
\hspace{-5em}
\leaf{$R(S \rightarrow AA)$}
\leaf{$R(A \rightarrow AA)$}
\leaf{$R(A \rightarrow a)$}
\branch{1}{$[1, A, 2]$}
\leaf{$R(A \rightarrow a)$}
\branch{1}{$[2, A, 3]$}
\branch{3}{$[1, A, 3]$}
\leaf{$R(A \rightarrow a)$}
\branch{1}{$[3, A, 4]$}
\branch{3}{$[1, S, 4]$}
\tree
\\
Item Derivation Tree
\\
\\
$R(S \rightarrow AA) \otimes R(A \rightarrow AA) \otimes R(A
\rightarrow a) \otimes R(A \rightarrow a) \otimes R(A \rightarrow a)$
\\
Derivation Value
\end{tabular}

%\leaf{$R(S \rightarrow AA)$}
%\branch{1}{$[1, A, 2]$}
%\leaf{$R(A \rightarrow a)$}
%\branch{1}{$[2, A, 3]$}
%\leaf{$R(A \rightarrow AA)$}
%\leaf{$R(A \rightarrow a)$}
%\branch{1}{$[3, A, 4]$}
%\branch{3}{$[2, A, 4]$}
%\branch{3}{$[1, S, 4]$}
%\leaf{$R(A \rightarrow a)$}
%\tree
\end{center}
%\caption{Two example item derivation trees}
\caption{Grammar derivation tree; item derivation tree; value}
\label{fig:itemexample}
\end{figure}

Notice that an item derivation is a tree, not a directed graph.  Thus,
an item sub-derivation could occur multiple times in a given item
derivation.  This means that we can have a one-to-one correspondence
between item derivations and grammar derivations; loops in the grammar
lead to an infinite number of grammar derivations, and an infinite
number of corresponding item derivations.

A grammar including rules such as
$$
\begin{array}{rcl}
S&\rightarrow &AAA \\
A& \rightarrow & B \\
A& \rightarrow &a \\
B& \rightarrow & A \\
B& \rightarrow &\epsilon
\end{array}
$$
would allow derivations such as $S \Rightarrow AAA \Rightarrow BAA
\Rightarrow AA \Rightarrow BA \Rightarrow A \Rightarrow B
\Rightarrow\epsilon$.  Depending on the parser, we might include the
exact same item derivation showing $A \Rightarrow B \Rightarrow \epsilon$ three times.
Similarly, for a derivation such as $A \Rightarrow B \Rightarrow A
\Rightarrow B \Rightarrow A \Rightarrow a$, we would have a
corresponding item derivation tree that included multiple uses of the
$A \rightarrow B$ and $B \rightarrow A$ rules.

\subsubsection{Value of item derivation}

The value of an item derivation $D$, $V(D)$, is the product of the
value of its rules, $R(r)$, in the same order that they appear in the
item derivation tree.  Since rules occur only in the leaves of item
derivation trees, there is no ambiguity as to order in this
definition.  For an item derivation tree $D$ with rule values $d_1,
d_2, ..., d_d$ as its leaves,
\begin{equation}
V(D) = \bigotimes_{i=1}^d R(d_i)
\label{eqn:ordered}
\end{equation}
Alternatively, we can write this equation recursively as
\begin{equation}
V(D) =
\left\{ \begin{array}{ll}
{R(D)}{\mbox{ \em if } D \mbox{ is a rule}} \\
{\bigotimes_{i = 1}^k V(D_i)}{\mbox{ \em if } D=\langle b:D_1,...,D_k \rangle}
\end{array}
\right.
\label{eqn:derivationvalue}
\end{equation}
Continuing our example, the value of the item derivation tree of
Figure \ref{fig:itemexample} is
$$
R(S \rightarrow AA) \otimes R(A \rightarrow a) \otimes R(A
\rightarrow AA) \otimes R(A \rightarrow a) \otimes R(A \rightarrow a)
$$
the same as the value of the first grammar derivation.

Notice that Equation \ref{eqn:derivationvalue} is just a recursive
expression for the product of the rule values appearing in the leaves
of the tree.  Thus, 

Let $\mi{inner}(x)$ represent the set of all item derivation trees
headed by an item $x$.  Then the value of $x$ is the sum of
all the values of all item derivation trees headed by $x$.  Formally,
$$
V(x) = \bigoplus_{D\in\smi{inner}(x)} V(D)
$$
The value of a sentence is just the value of the goal item,
$V(\mi{goal})$.

\subsubsection{Iso-valued derivations} In certain cases, a particular grammar
derivation and a particular item-derivation will have the same value
for any semiring and any rule value function $R$.  In particular, if
the same rules occur in the same order in both the grammar derivation
and the item derivation, then their values will be the same no matter
what.  If a grammar derivation and an item derivation meet this
condition, then we define a new term to describe them, {\em
iso-valued}.  In Figure \ref{fig:itemexample}, the grammar derivation
and item derivation both have the rules $R(S \rightarrow AA), R(A
\rightarrow AA), R(A \rightarrow a), R(A \rightarrow a), R(A
\rightarrow a)$, and so they are iso-valued.

In some cases, a grammar derivation and an item-derivation will have
the same value for any commutative semiring and any rule value
function.  If the same rules occur the same number of times in both
the grammar derivation and the item derivation, then they will have
the same value in any commutative semiring.  We say that a grammar
derivation and an item derivation meeting this condition are
{\em commutatively iso-valued}.

Iso-valued and commutatively iso-valued derivations will be important
when we discuss conditions for correctness.

\subsubsection{Example value of item derivation}

Finishing our example, the value of the goal item given our example
sentence is just the sum of the values of the two item-based
derivations,
$$
\begin{array}{l}
R(S \rightarrow AA) \otimes R(A \rightarrow AA) \otimes R(A
\rightarrow a) \otimes R(A \rightarrow a) \otimes R(A \rightarrow a) \\
\oplus \\
R(S \rightarrow AA) \otimes R(A \rightarrow a) \otimes R(A
\rightarrow AA) \otimes R(A \rightarrow a) \otimes R(A \rightarrow a)
\end{array}
$$
This value is the same as the value of the sentence according to the grammar.

\subsubsection{Conditions for correctness}

We can now specify the conditions for an item-based description to be
correct.

\begin{mytheorem}
Given \label{theorem:itemgrammar} an item-based description
$\mathcal{D}$, if for every grammar $G$, there exists a one-to-one
correspondence between the item derivations using $\mathcal{D}$ and
the grammar derivations, and the corresponding derivations are
iso-valued, then for every complete semiring, the value of a given
input $w_1... w_n$ is the same according to the grammar as the value
of the goal item.  If the semiring is commutative, then the
corresponding derivations need only be commutatively iso-valued.
\end{mytheorem}
{\em Proof \hspace{3em}} The proof is very simple; essentially, each
term in each sum occurs in the other.  We separate the proof into two
cases.  First is the non-commutative case.  In this case, by
hypothesis, for a given input there are grammar derivations
$E_1... E_k$ (for $0 \leq k \leq \infty$) and corresponding iso-valued
item derivation trees $D_1... D_k$ of the goal item.  Since
corresponding items are iso-valued, for all $i$, $V(E_i) = V(D_i)$.
Now, since the value of the string according to the grammar is just $\sum_i
V(E_i) = \sum_i V(D_i)$, and the value of the goal item is $\sum_i
V(D_i)$, the value of the string according to the grammar equals the value of
the goal item.

The second case, the commutative case, follows analogously. $\Box$

There is one additional condition for an item-based description to be
usable in practice, which is that there be only a finite number of
derivable items for a given input sentence; there may, however, be an
infinite number of derivations of any item.

\subsection{The derivation semirings}
\label{sec:parseforest}

All of the semirings we use should be familiar, except for the
derivation semirings, which we now describe.  These semirings, unlike
the other semirings described in Figure \ref{fig:semirings}, are not
commutative under their multiplicative operator, concatenation.

In many parsers, it is conventional to compute parse forests: compact
representations of the set of trees consistent with the input.  We
will use a related concept, derivation forests, a compact
representation of the set of derivations consistent with the input,
which corresponds to the parse forest for CNF grammars, but is easily
extended to other formalisms.  Although the terminology we use is
different, the representation of derivation forests is similar to that
used by \newcite{Billot:89a}.  

Often, we will not be interested in the set of all derivations, but
only in the most probable derivation.  The Viterbi-derivation semiring
computes this value.  Alternatively, we might want the $n$ best
derivations, which would be useful if the output of the parser were
passed to another stage, such as semantic disambiguation; this value
is computed by the Viterbi-n-best derivation semiring.

Notice that each of the derivation semirings can also be used to
create transducers.  That is, we simply associate strings rather than
grammar rules with each rule value.  Instead of grammar rule
concatenation, we perform string concatenation.  The derivation
semiring then corresponds to nondeterministic transductions; the
Viterbi semiring corresponds to a weighted or probabilistic
transducer; and the inside semiring could be used to, for instance,
perform re-estimation of probabilistic transducers.

\subsubsection{Derivation Forest}

The derivation forest semiring consists of sets of derivations, where
a derivation is a list of rules of the grammar.  In the CFG case,
these rules would form, for instance, a left-most derivation.  The
additive operator $\cup$ produces a union of derivations, and the
multiplicative operator $\cdot$ produces the concatenation, one
derivation concatenated with the next.  The concatenation operation
($\cdot$) is defined on both derivations and sets of derivations; when
applied to a set of derivations, it produces the set of pairwise
concatenations.  The simplest derivations are simply rules of the
grammar, such as $X \rightarrow YZ$ for a CFG.  Sets containing one
rule, such as $\{\langle X \rightarrow YZ\rangle\}$ constitute the
primitive elements of the semiring.\footnote{
%Thanks to Elizabeth Wilmer for pointing this out:
The derivation forest semiring is equivalent to a semiring well known
to mathematicians, the polynomials over non-commuting variables.
Such a polynomial is a sum of terms, each of which is an ordered
product of variables.  If these variables correspond to the basic
elements of this semiring, then each term in the polynomial
corresponds to a derivation.  
%In the polynomial version, each
%term may occur several times, i.e. $abc+abc=2abc$, corresponding to a
%bag (set with counts) of lists of trees, rather than sets of lists of
%trees.  However, the efficient implementation we use is actually for
%bags of trees, not sets, and in a correct parser, no repeated terms
%should occur anyway, since these would correspond to multiple item
%derivations for a single grammar derivation.
}

A few examples may help.  The additive identity, the zero element, is
simply the empty set, $\emptyset$: union with the empty set is an
identity operation.  The multiplicative identity is the set containing
the empty derivation, $\{\langle \rangle\}$: concatenation with the
empty derivation is an identity operation.  Derivations need not be
complete.  For instance, assuming we are using left-most derivations
with CFGs, $\{\langle X \rightarrow YZ,\;Y \rightarrow y\rangle\}$ is
a valid element, as is $\{\langle Y \rightarrow y,\; X \rightarrow x
\rangle\}$.  In fact, $\{\langle X \rightarrow A,\; B \rightarrow b
\rangle\}$ is a valid element of the semiring, even though it could
not occur in a valid grammar derivation; this value should never occur
in a correctly functioning parser.

The obvious implementation of derivation forests, as actual sets of
derivations, would be extremely inefficient.  In the worst case, when
we allow infinite unions, a case we will wish to consider, the obvious
implementation does not work at all.  However, in Section
\ref{sec:solvinginfinite}, we will show how to use pointers to
efficiently implement infinite unions of derivation forests, in a
manner analogous to the traditional implementation of parse forests.

\begin{figure}
\begin{tabbing}
\verb|    |\=\verb|    |\=\verb|    |\=\verb|    |\=\verb|    |\=\verb|    |\=\verb|    |\=\verb|    |\=\verb|    |\=\verb|    |\= \kill

\alFunction $\mi{Concatenate}(f,\; g)$ \\
 \> \alreturn $\langle \mbox{``Concatenate''}\; f\; g\rangle$; \\
 \\
\alFunction $\mi{Union}(f,\; g)$ \\
 \> \alreturn $\langle \mbox{``Union''}\; f\; g\rangle $; \\
 \\
\alFunction $\mi{Create}(\mi{rule})$ \\
 \> \alreturn $\langle \mbox{``Create''}\; \mi{rule}\rangle$; \\
 \\
\alFunction $\mi{Extract}(f) $ \\
 \> \alswitch $f_1$: \\
 \>  \> ``Concatenate'': \\
 \>  \>  \> \alreturn $\mi{Concat!}(\mi{Extract}(f_2),\; \mi{Extract}(f_3))$; \\
 \>  \> ``Union'': \\
 \>  \>  \> \alreturn $\mi{choose}(\mi{Extract}(f_2),\; \mi{Extract}(f_3))$; \\
 \>  \> ``Create'': \\
 \>  \>  \> \alreturn $\langle f_2 \rangle)$; \\
 \\
\end{tabbing}
\caption{Derivation Forest Implementation}\label{fig:parsecode}
\end{figure}

We can now describe a simple, efficient implementation of the
derivation forest semiring.  We will assume that four operations are
desired: concatenation, union, primitive creation, and extraction.
Primitive creation is used to create the basic elements of the
semiring, sets containing a single rule.  Extraction
non-deterministically extracts a single derivation from the derivation
forest.  Code for these four functions is given in Figure
\ref{fig:parsecode}.  All of the code is straightforward.  We assume a
function {\em choose}, needed to handle unions in the extract
function, that nondeterministically chooses between its inputs.

A straightforward implementation of this algorithm would work fine,
but slight variations are required for good efficiency.  The problem
comes from the concatenation operation.  Typically, in LISP-like
languages, concatenation is implemented as a copy operation.  If we
were to build up a derivation of length $n$ one rule at a time, then
the run time would be $O(n^2)$, since we would first copy one element,
then two, then three, etc., resulting in $O(n^2)$ rules being copied.
To get good efficiency, we need to implement concatenation
destructively; we assume that lists, indicated with angle brackets,
are implemented with linked lists, with a pointer to the last
element.  (For those skilled in Prolog, this implementation of linked
lists is essentially equivalent to difference lists.)  List creation
can be performed efficiently using non-destructive operations.
Destructive concatenation can operate in constant time.  Given this
destructive operation, any interpreter of this nondeterministic 
algorithm would need to keep a list of destructive changes to undo
during backtracking, as is done in many implementations of unification
grammars, or of unification in Prolog.

This modified implementation is efficient, in the following senses.
First, concatenation and union are constant time operations.  Second,
if we were to use $\mi{Extract}$ with a non-deterministic interpreter
to generate all derivations in the derivation forest, the time used
would be at worst proportional to the total size of all trees
generated.

\newcite{Billot:89a} show how to create grammars to represent parse
forests.  Readers familiar with their work will recognize the similarity between their
representation and ours.  Where we write
$$
e := \mi{Concatenate}(f, g)
$$
Billot \etal write
$$
e \rightarrow fg
$$
Where we write
$$
e := \mi{Union}(f, g)
$$
they write
$$
\begin{array}{c}
e \rightarrow f \\
e \rightarrow g
\end{array}
$$

\subsubsection{Viterbi-derivation Semiring}

The Viterbi-derivation semiring computes the most probable derivation
of the sentence, given a probabilistic grammar.  Elements of this
semiring are a pair of a real number $v$ and a derivation forest $E$,
i.e. the set of derivations with score $v$.  We define $\maxvit$, the
additive operator, as
$$
\maxvit(\langle v, E \rangle, \langle w, D \rangle ) =
\left\{\begin{array}{ll}
\langle v, E \rangle & \mbox{\em if } v > w \\
\langle w, D \rangle & \mbox{\em if } v < w \\
\langle v, E \cup D \rangle & \mbox{\em if } v = w 
\end{array} \right.
$$
In typical practical Viterbi parsers, when two derivations have the
same value, one of the derivations is arbitrarily chosen.  In
practice, this is usually a fine solution, and one that could be used
in a real-world implementation of the ideas in this \chapterpaper{},
but from a theoretical viewpoint, the arbitrary choice destroys the
associative property of the additive operator, $\maxvit$.  To preserve
associativity, we keep derivation forests of all elements that tie for
best.  An alternate technique for preserving associativity would be to
choose between derivations using some ordering, but the derivation
forest solution simplifies the discussion in Section
\ref{sec:pairsemirings}

The definition for $\maxvitn$ is only defined for two elements.  Since
the operator is associative, it is clear how to define $\maxvitn$ for
any finite number of elements, but we also need infinite summations to
be defined.  We require an operator, $\sup$, the {\em supremum}, for
this definition.  The supremum of a set is the smallest value at least
as large as all elements of the set; that is, it is a maximum that is
defined in the infinite case.

We can now define $\maxvitn$ for the case of infinite sums.  First,
let
$$
w = \sup_{\langle v, E\rangle \in X} v
$$
Then, let 
$$
D = \{E \vert \langle w, E \rangle\in X\}
$$
Then $\maxvit X = \langle w, D\rangle$.  In the finite case, this is
equivalent to our original definition.  In the infinite case, $D$ is
potentially empty, but this causes us no problems in theory, and
infinite sums with an empty $D$ will not appear in practice.

We define the multiplicative operator, $\timesvit$, as
$$
\langle v, E\rangle\times \langle w, D\rangle = \langle v\times
w, E\cdot D \rangle
$$
where $E \cdot D$ represents the concatenation of the two
derivation forests.

\subsubsection{Viterbi-n-best semiring}
\label{sec:vitnbest}

The last kind of derivation semiring is the Viterbi-n-best semiring,
which is used for constructing n-best lists.  Intuitively, the value
of a string using this semiring will be the $n$ most likely
derivations of that string (unless there are fewer than $n$ total
derivations.)  Furthermore, in a practical implementation, this is
actually how a Viterbi-n-best semiring would typically be implemented.
From a theoretical viewpoint, however, this implementation is
inadequate, since we must also define infinite sums and be sure that
the distributive property holds.  Thus, we introduce two
complications.  First, when not only are there more than $n$ total
derivations, but there is a tie for the $n$'th most likely, there will
be more than $n$ entries, which we can represent efficiently with a
derivation forest.  Second, in order to make infinite sums well
defined, it will be useful to have an additional value, $\infty$,
counted as a legal derivation.  The value $\infty$ will arise due to
infinite sums of elements approaching a supremum.  Thus, we will want
to consider $\infty$ to represent an infinite number of values
approaching the derivation value.

The best way to define the Viterbi-n-best semiring is as a
homomorphism from a simpler semiring, the Viterbi-all semiring.  The
Viterbi-all semiring keeps all derivations and their values.  The
additive operator is set union, and the multiplicative operator is
$\star$, defined as
$$
X\star Y =\{\alist{vw, d\cdot e}\vert \alist{v, d}\in X \wedge
\alist{w, e}\in Y\}
$$
Then, the Viterbi-all semiring is
$$
\langle  2^{\Bbb{R}_0^1 \times
\Bbb{E}}, \cup, \star, \emptyset
, \{ \langle 1, \langle\rangle  \rangle \} \rangle
$$

Now, we can define a helper function we will need for the homomorphism
to the Viterbi-n-best semiring.  We define $\mi{simpletopn}$, which
returns the $n$ highest valued elements.  Ties for last are kept; this
property will make the additive operator commutative and associative.
$$
\mi{simpletopn}(X) = \{\alist{v, d}\in X\vert \mbox{there are at most }
n-1 \mbox{ items }\alist{w, e}\in X \mbox{ s.t. } w < v\}
$$

We can now define a function, $\topn$, which will provide the
homomorphism.  Like $\mi{simpletopn}$, $\topn$ returns the $n$ highest
valued elements, keeping ties.  In addition, if there is an infinite
number of elements approaching a supremum, $\topn$ returns a special
element whose value is the supremum, and whose derivation is the
symbol $\infty$.
$$
\topn(X) = \mi{simpletopn}(X)\cup \left\{\begin{array}{ll}
\emptyset & \mbox{\em if }\vert \mi{simpletopn}(X)\vert \geq n \\
\emptyset & \mbox{\em if } X = \mi{simpletopn}(X) \\
\{\alist{\sup_{v\vert\alist{v, d}\in X-\smi{simpletopn}(X)} v, \infty}\}
& \mbox{\em otherwise}
\end{array}\right.
$$

We can now define the Viterbi-n-best semiring as a homomorphism from
the Viterbi-all semiring.  In particular, we define the elements of
the semiring to be $\{\topn(X) \vert X \in 2^{\Bbb{R}_0^1 \times
\Bbb{E}}\}$.  Because of this definition, for every $A, B$ in the
Viterbi-n-best semiring, there is some $X, Y$ such that $\topn(X) = A$
and $\topn(Y) = B$.  We can then define
$$
\maxvitn A, B = C
$$
where $C = \topn(X \cup Y)$, for some, $X, Y$ such that $\topn(X) = A$
and $\topn(Y) = B$.  In appendix \ref{sec:vitnbestproof}, we prove that $C$
is uniquely defined by this relationship.  Similarly, we define the
multiplicative operator $\timesvitn$ to be
$$
A \timesvitn B = C
$$
where $C = \topn(X \star Y)$, for some, $X, Y$ such that $\topn(X) = A$
and $\topn(Y) = B$, and again prove the uniqueness of the relationship
in the appendix.  Also in the appendix, we prove that these operators
do indeed form an $\omega$-continuous semiring.

\section{Efficient Computation of Item Values}

Recall that the value of an item $x$ is just $V(x) =
\bigoplus_{D\in\smi{inner}(x)} V(D)$.  This definition may require
summing over exponentially many or even infinitely many terms.  In
this section, we give relatively efficient formulas for computing the
values of items.  There are three cases that must be handled.  First
is the base case: if $x$ is a rule, then
$$
V(x) = \bigoplus_{D\in\smi{inner}(x)} V(D) = \bigoplus_{D\in\{\langle
x \rangle\}} V(D) = V(\langle x \rangle) = R(x)
$$

The second and third cases occur when $x$ is an item.  Recall that
each item is associated with a bucket, and that the buckets are
ordered.  Each item $x$ is either associated with a non-looping
bucket, in which case its value depends only on the values of items in
earlier buckets; or with a looping bucket, in which case its value
depends potentially on the values of other items in the same bucket.
In the second case, when the item is associated with a non-looping
bucket, and if we compute items in the same order as their buckets, we
can assume that the values of items $a_1... a_k$ contributing to the
value of item $b$ are known.  We give a formula for computing the
value of item $b$ that depends only on the values of items in earlier
buckets.

For the third case, in which $x$ is associated with a looping bucket,
infinite loops may occur, when the value of two items in the same
bucket are mutually dependent, or an item depends on its own value.
These infinite loops may require computation of infinite sums.  Still,
we can express these infinite sums in a relatively simple form,
allowing them to be efficiently computed or approximated.

\subsection{Item Value Formula}

\begin{mytheorem}
If an item $x$ is not in a looping bucket, then
\label{theorem:itemvalue}
\begin{equation}
\label{eqn:itemvalue}
V(x) = \bigoplus_{a_1...a_k \;\mbox{\scriptsize
s.t.}\;\smallinfer{a_1... a_k}{x}{}} \bigotimes_{i = 1}^k
V(a_i)
\end{equation}
\end{mytheorem}
{\em Proof \hspace{3em}}
Let us expand our notion of inner to include
deduction rules: $\mi{inner}(\shortinfer{a_1...a_k}{b}{})$ is the set of
all derivation trees of the form $\langle b: \langle a_1 ... \rangle
\langle a_2... \rangle... \langle a_k... \rangle \rangle$.  For any
item derivation tree that is not a simple rule, there is some
$a_1... a_k, b$ such that $D\in
\mi{inner}(\shortinfer{a_1...a_k}{b}{})$.  Thus, for any item $x$,
\begin{eqnarray}
V(x) & = & \bigoplus_{D\in\smi{inner}(x)} V(D) \nonumber \\
\label{eqn:intermediateV}
& = & \bigoplus_{a_1... a_k}
\bigoplus_{D\in\smi{inner}(\smallinfer{a_1... a_k}{x}{})} V(D) 
\end{eqnarray}
Consider item derivation trees $D_{a_1}...D_{a_k}$ headed by items
$a_1...a_k$ such that \smallinfer{a_1...a_k}{x}{}.  Recall that
$\langle x: D_{a_1},...,D_{a_k} \rangle$ is the item derivation tree
formed by combining each of these trees into a full tree, and notice
that
$$
\bigcup_{\longsum{D_{a_1}\in\smi{inner}(a_1),...,}{D_{a_k}\in\smi{inner}(a_k)}}
\langle x : D_{a_1},...,D_{a_k} \rangle
=
\mi{inner}(\shortinfer{a_1...a_k}{x}{})
$$
Therefore
$$
\bigoplus_{D\in\smi{inner}(\smallinfer{a_1... a_k}{x}{})} V(D)  =
\bigoplus_{\longsum{D_{a_1}\in\smi{inner}(a_1),...,}{D_{a_k}\in\smi{inner}(a_k)}}
V(\langle x: D_{a_1}, ..., D_{a_k}\rangle)
$$
Also notice that $V(\langle x : D_{a_1},...,D_{a_k} \rangle) =
\bigotimes_{i=1}^k V(D_{a_i})$.  Thus,
$$
\bigoplus_{\longsum{D_{a_1}\in\smi{inner}(a_1),...,}{D_{a_k}\in\smi{inner}(a_k)}}
V(\langle x: D_{a_1}, ..., D_{a_k}\rangle)
=
\bigoplus_{\longsum{D_{a_1}\in\smi{inner}(a_1),...,}{D_{a_k}\in\smi{inner}(a_k)}}
\bigotimes_{i=1}^k V(D_{a_i}) 
$$
Since for all semirings, both operations are associative, and multiplication
distributes over addition, we can rearrange summations and
products:
\begin{eqnarray*}
\bigoplus_{\longsum{D_{a_1}\in\smi{inner}(a_1),...,}{D_{a_k}\in\smi{inner}(a_k)}}
\bigotimes_{i=1}^k V(D_{a_i}) 
& = &
\bigotimes_{i=1}^k \bigoplus_{D_{a_i}\in\smi{inner}(a_i)} V(D_{a_i}) \\
&=&
\bigotimes_{i=1}^k V(a_i) 
\end{eqnarray*}
Substituting this back into Equation \ref{eqn:intermediateV}, we get
$$
V(x) = \bigoplus_{a_1...a_k \;\mbox{\scriptsize s.t.}\;
\smallinfer{a_1... a_k}{x}{}} \bigotimes_{i = 1}^k
V(a_i)
$$
completing the proof. $\Box$

Now, we address the case in which $x$ is an item in a looping bucket.
This case requires computation of an infinite sum.  We will write out
this infinite sum, and discuss how to compute it exactly in all cases,
except for one, where we approximate it.

Consider the derivable items $x_1...x_m$ in some looping bucket $B$.
If we build up derivation trees incrementally, when we begin
processing bucket $B$, only those trees with no items from bucket $B$
will be available, what we will call 0th generation derivation trees.
We can put these 0th generation trees together to form first
generation trees, headed by elements in $B$.  We can combine these
first generation trees with each other and with 0th generation trees
to form second generation trees, and so on.  Formally, we define the
{\em generation} of a derivation tree headed by $x$ in bucket $B$ to
be the largest number of items in $B$ we can encounter on a path from
the root to a leaf.  It will be convenient to define the generation of
the derivation of any item $x$ in a bucket preceding $B$ to be
generation 0; generation 0 will not contain derivations of any items
in bucket $B$.  

Consider the set of all trees of generation at most $g$ headed by $x$.
Call this set $\mi{inner}_{\leq g}(x, B)$.  We can define the $\leq g$
{\em generation value} of an item $x$ in bucket $B$, $V_{\leq g}(x,
B)$:
$$
V_{\leq g}(x, B) = \bigoplus_{D \in \smi{inner}_{\leq g}(x, B)} V(x)
$$

Intuitively, as $g$ increases, for $x\in B$, $\mi{inner}_{\leq g}(x,
B)$ becomes closer and closer to $\mi{inner}(x)$.  Thus, intuitively,
the finite sum of values in the latter approaches the infinite sum of
values in the former.  For $\omega$-continuous semirings (which
includes all of the semirings considered in this \chapterpaper{}), an
infinite sum is equal to the supremum of the partial sums
\cite[p. 613]{Kuich:97a}.  Thus,
$$
V(x) = \bigoplus_{D \in \smi{inner}(x, B)} V(x) = \sup_{g} V_{\leq g}(x, B)
$$
It will be easier to compute the supremum if we find a simple formula
for $V_{\leq g}(x, B)$.

Notice that for items $x$ in buckets preceding $B$, since these items
are all included in generation 0,
\begin{equation}
\label{eqn:notinBinner}
V_{\leq g}(x, B) = V(x)
\end{equation}
Also notice that for items $x \in B$, there will be no generation 0
derivations, so
$$
V_{\leq 0}(x, B) = 0
$$
Thus, generation 0 makes a trivial base for a recursive formula.  Now,
we can consider the general case:

\begin{mytheorem}
For $x$ an item in a looping bucket $B$, and for $g \geq 1$,
\label{theorem:inloop}
\begin{equation}
V_{\leq g}(x, B) =
\bigoplus_{a_1...a_k \;\mbox{\scriptsize s.t.}\; \smallinfer{a_1... a_k}{x}{}}
\;\;\bigotimes_{i=1..k} 
\left\{
\begin{array}{ll}
{V(a_i)}{\mbox{ \em if }a_i\notin B} \\
{V_{\leq g\!-\!1}(a_i, B)}{\mbox{ \em if }a_i\in B}
\end{array}
\right.
\label{eqn:inloop}
\end{equation}
\end{mytheorem}

The proof parallels that of Theorem \ref{theorem:itemvalue}, and is
given in Appendix \ref{sec:additionalproofs}.

\subsection{Solving the Infinite Summation}
\label{sec:solvinginfinite}

A formula for $V_{\leq g}(x, B)$ is useful, but what we really need is
specific techniques for computing the supremum, $V(x) = \sup_g V_{\leq
g}(x,B)$.

For all $\omega$-continuous semirings, the supremum of iteratively
approximating the value of a set of polynomial equations, as we are
essentially doing in Equation \ref{eqn:inloop}, is equal to the
smallest solution to the equations \cite[p. 622]{Kuich:97a}.  In
particular, consider the equations:
\begin{equation}
V_{\leq \infty}(x, B) =
\bigoplus_{a_1...a_k \;\mbox{\scriptsize s.t.}\; \smallinfer{a_1... a_k}{x}{}}
\;\;\bigotimes_{i=1..k} 
\left\{
\begin{array}{ll}
{V(a_i)}{\mbox{ \em if }a_i\notin B} \\
{V_{\leq \infty}(a_i, B)}{\mbox{ \em if }a_i\in B}
\end{array}
\right.
\label{eqn:Vinfinity}
\end{equation}
where $V_{\leq \infty}(x, B)$ can be thought of as indicating $\vert B
\vert$ different variables, one for each item $x$ in the looping
bucket $B$.  Equation \ref{eqn:inloop} represents the iterative
approximation of this equation, and therefore the smallest solution to
this equation represents the supremum of that one.

One fact will be useful for several semirings: whenever the values of all
items $x\in B$ at generation $g\!+\!1$ are the same as the values of all
items in the preceding generation, $g$, they will be the same at all
succeeding generations, as well.  Thus, the value at generation $g$
will be the value of the supremum.  The proof of this fact is trivial: we
substitute the value of $V_{\leq g}(x, B)$ for $V_{\leq g\!+\!1}(x, B)$ to
show that $V_{\leq g+2}(x, B) = V_{\leq g\!+\!1}(x, B)$.
\begin{eqnarray*}
V_{\leq g+2}(x, B)
& = &
\bigoplus_{a_1...a_k \;\mbox{\scriptsize s.t.}\; \smallinfer{a_1... a_k}{x}{}}
\;\;\bigotimes_{i=1..k} 
\left\{
\begin{array}{ll}
{V(a_i)}{\mbox{ \em if }a_i\notin B} \\
{V_{\leq g\!+\!1}(a_i, B)}{\mbox{ \em if }a_i\in B} \\
\end{array}
\right. \\
& = &
\bigoplus_{a_1...a_k \;\mbox{\scriptsize s.t.}\; \smallinfer{a_1... a_k}{x}{}}
\;\;\bigotimes_{i=1..k} 
\left\{
\begin{array}{ll}
{V(a_i)}{\mbox{ \em if }a_i\notin B} \\
{V_{\leq g}(a_i, B)}{\mbox{ \em if }a_i\in B} \\
\end{array}
\right. \\
& = & V_{\leq g\!+\!1}(x, B)
\end{eqnarray*}

Now, we can consider various semiring specific algorithms for
computing the supremum.  We first examine the simplest case, the
boolean semiring (booleans under $\vee$ and $\wedge$).  Notice that
whenever a particular item has value $\mi{TRUE}$ at generation $g$, it must
also have value $\mi{TRUE}$ at generation $g\!+\!1$, since if the item can be
derived in at most $g$ generations then it can certainly be derived in
at most $g\!+\!1$ generations.  Thus, since the number of $\mi{TRUE}$ valued items
is non-decreasing, and is at most $\vert B\vert$, eventually the
values of all items must not change from one generation to the next.
Therefore, for the boolean semiring, a simple algorithm suffices:
keep computing successive generations, until no change is detected in
some generation; the result is the supremum.  We can perform this
computation efficiently if we keep track of items that change value
in generation $g$ and only examine items that depend on them in
generation $g\!+\!1$.  This algorithm is then similar to the algorithm of
\newcite{Shieber:93a}.

For the next three semirings -- the counting semiring, the Viterbi
semiring, and the derivation forest semiring -- we need the concept of
a {\em derivation subgraph}.  In Section \ref{sec:itembased} we
considered the strongly connected components of the dependency graph,
consisting of items that for some sentence could possibly depend on
each other, and we put these possibly interdependent items together in
looping buckets.  For a given sentence and grammar, not all items will
have derivations.  We will find the subgraph of the dependency graph
of items with derivations, and compute the strongly connected
components of this subgraph.  The strongly connected components of
this subgraph correspond to loops that actually occur given the
sentence and the grammar, as opposed to loops that might occur for
some sentence and grammar, given the parser alone.  We call this
subgraph the derivation subgraph, and we will say that items in a
strongly connected component of the derivation subgraph are part of a
loop.

Now, we can discuss the counting semiring (integers under + and
$\times$.)  In the counting semiring, for each item, there are three
cases: the item can be in a loop; the item can depend (directly or
indirectly) on an item in a loop; or the item does not depend on
loops.  If the item is in a loop or depends on a loop, its value is
infinite.  If the item does not depend on a loop in the current
bucket, then its value becomes fixed after some generation.  We can
now give the algorithm: first, compute successive generations until
the set of items in $B$ does not change from one generation to the
next.  Next, compute the derivation subgraph, and its strongly
connected components.  Items in a strongly connected component (a
loop) have an infinite number of derivations, and thus an infinite
value.  Compute items that depend directly or indirectly on items in
loops: these items also have infinite value.  Any other items can only
be derived in finitely many ways using items in the current bucket, so
compute successive generations until the values of these items do not
change.

The arctic semiring,$\langle \Bbb{R} \cup \{ -\infty \}, \max, +,
-\infty, 0 \rangle$, is analogous to the counting semiring: loops lead
to infinite values.  Thus, we can use the same algorithm.

The method for solving the infinite summation for the derivation
forest semiring depends on the implementation of derivation forests.
Essentially, that representation will use pointers to efficiently
represent derivation forests.  Pointers, in various forms, allow one
to efficiently represent infinite circular references, either directly
\paper{\cite{Goodman:98b}, or indirectly \cite{Goodman:98a}.}
\thesis{\cite{Goodman:98a}, or indirectly \cite{Goodman:98b}.}

The algorithm we use is to compute the derivation subgraph, and then
create pointers analogous to the directed edges in the derivation
subgraph, including pointers in loops whenever there is a loop in the
derivation subgraph (corresponding to an infinite number of
derivations).  For the representation described in Section
\ref{sec:parseforest} we perform two steps.  First, for each item $x
\in B$, we will set
$$
V(x) = \langle \mbox{``Union''} \; x_1 \; \langle
\mbox{``Union''} \; x_2 \; \langle \mbox{``Union''} \; x_3 \; \langle \cdots
\rangle\rangle\rangle\rangle
$$
where there is one $x_i$ for each instantiation of a rule
$\smallinfer{a_1...a_k}{x}{}$.  Next, for each derivation of $x$,
$\smallinfer{a_1...a_k}{x}{}$, we will create a value
$$
\langle \mbox{``Concatenate''}\; V(a_1)\; \langle
\mbox{``Concatenate''}\; V(a_2)\; \langle \cdots 
\langle \mbox{``Concatenate''}\; V(a_{k-1})\; V(a_k) \rangle 
\cdots
\rangle\rangle\rangle
$$
and destructively change an appropriate $x_i$ to this value.  In a
LISP-like language, this will create the appropriate circular
pointers.  As in the finite case, this representation is equivalent to
that of \newcite{Billot:89a}.

For the Viterbi semiring, the algorithm is analogous to the boolean
case.  Derivations using loops in these semi-rings will always have
values lower than derivations not using loops, since the value with
the loop will be the same as some value without the loop, multiplied
by some set of rule probabilities that are at most 1.  Thus, loops
do not change values.  Therefore, we can simply compute successive
generations until values fail to change from one iteration to the
next.  The tropical semiring is analogous: again, loops do not change
values, so we can compute successive generations until values don't
change.

Now, consider implementations of the Viterbi-derivation semiring in
practice, in which we keep only a representative derivation, rather
than the whole derivation forest.  In this case, loops do not change
values, and we use the same algorithm as for the Viterbi semiring.  On
the other hand, for a theoretically correct implementation, loops with
value 1 can lead to an infinite number of derivations, in the same way
they did for the derivation forest semiring.  Thus, for a
theoretically correct implementation, we must use the same techniques
we used for the derivation semiring.  In an implementation of the
Viterbi-n-best semiring, in practice, loops can change values, but at
most $n$ times, so the same algorithm used for the Viterbi semiring
still works.  Again, in a theoretical implementation, we need to use
the same mechanism as in the derivation forest semiring.

The last semiring we consider is the inside semiring.  This semiring
is the most difficult.  There are two cases of interest, one of which
we can solve exactly, and the other of which requires approximations.
In many cases involving looping buckets, all deduction rules will be
of the form $\shortinfer{a_1 x}{b}{}$, where $a_1$ and $b$ are items
in the looping bucket, and $x$ is either a rule, or an item in a
previously computed bucket.  This case corresponds to the items used for
deducing singleton productions, such as Earley's algorithm uses for
rules such as $A \rightarrow B$ and $B \rightarrow A$.  In this case,
Equation \ref{eqn:Vinfinity} forms a set of linear equations that can
be solved by matrix inversion.  In the more general case, as is likely
to happen with epsilon rules, we get a set of nonlinear equations, and
must solve them by approximation techniques, such as simply computing
successive generations for many iterations.\footnote{Note that even in
the case where we can only use approximation techniques, this
algorithm is relatively efficient.  By assumption, in this case, there
is at least one deduction rule with two items in the current
generation; thus, the number of deduction trees over which we are
summing grows exponentially with the number of generations: a linear
amount of computation yields the sum of the values of exponentially
many trees.}  \newcite{Stolcke:93a} provides an excellent discussion
of these cases, including a discussion of sparse matrix inversion,
useful for speeding up some computations.

\section{Reverse Values}
\label{sec:reverse}

The previous section showed how to compute several of the most
commonly used values for parsers, including boolean, inside, Viterbi,
counting, and derivation forest values, among others.  Noticeably
absent from the list are the outside probabilities\paper{, which we
define below}.  In general, computing outside probabilities is
significantly more complicated than computing inside probabilities.

\begin{figure}
\begin{tabbing}
\verb|    |\=\verb|    |\=\verb|    |\=\verb|    |\=\verb|    |\=\verb|    |\=\verb|    |\=\verb|    |\=\verb|    |\=\verb|    |\= \kill
\alforeach length $l$, longest {\bf downto} shortest \\
 \> \alforeach start $s$ \\
 \>  \> \alforeach split length $t$ \\
 \>  \>  \> \alforeach rule $A \rightarrow BC\;\in R$\\
 \>  \>  \>  \>  $\mi{outside}[s, B, s\!+\!t] := \mi{outside}[s, B, s\!+\!t] \;+$ \\
 \>  \>  \>  \>  \>  \> $\mi{outside}[s, A, s\!+\!l] \times
\mi{inside}[s\!+\!t, C, s\!+\!l] \times P(A\rightarrow BC)$; \\
 \>  \>  \>  \>  $\mi{outside}[s\!+\!t, C, s\!+\!l] := \mi{outside}[s\!+\!t, C, s\!+\!l] \;+$ \\
 \>  \>  \>  \>  \>  \> $\mi{outside}[s, A, s\!+\!l] \times \mi{inside}[s, B, s\!+\!t] \times P(A\rightarrow BC)$; \\
\end{tabbing}
\caption{Outside algorithm}\label{fig:CKYoutside}
\end{figure}

In this section, we show how to compute outside probabilities from the
same item-based descriptions used for computing inside values.
Outside probabilities have many uses, including for reestimating
grammar probabilities \cite{Baker:79b}, for improving parser
performance on some criteria
\paper{\cite{Goodman:96a}}\thesis{(Chapter \ref{ch:max})}, for
speeding parsing in some formalisms, such as Data-Oriented Parsing
\paper{\cite{Goodman:96b}}\thesis{(Chapter \ref{ch:DOP})}, and for
good thresholding algorithms
\paper{\cite{Goodman:97b}}\thesis{(Chapter \ref{ch:thresh})}.
%\footnote{The
%author confesses that he likes outside probabilities.}

We will show that by substituting other semirings, we can get
values analogous to the outside probabilities for any commutative
semiring, and in Sections \ref{sec:reversenonsmall} and
\ref{sec:reversenon} that we can get similar values for many
non-commutative semirings as well.  We will refer to these analogous
quantities as {\em reverse} values.  For instance, the quantity
analogous to the outside value for the Viterbi semiring will be called
the reverse Viterbi value.  Notice that the inside semiring values of
a Hidden Markov Model (HMM) correspond to the forward values of HMMs,
and the reverse inside values of an HMM correspond to the backwards
values.

Compare the outside algorithm\cite{Baker:79b,Lari:90a,Lari:91a}, given
in Figure \ref{fig:CKYoutside}, to the inside algorithm of Figure
\ref{fig:CKYinside}.  Notice that while the inside and recognition
algorithms were very similar, the outside algorithm is quite a bit
different.  In particular, while the inside and recognition algorithms
looped over items from shortest to longest, the outside algorithm
loops over items in the reverse order, from longest to shortest.
Also, compare the inside algorithm's main loop formula to the outside
algorithm's main loop formula.  While there is clearly a relationship
between the two equations, the exact pattern of the relationship is
not obvious.  Notice that the outside formula is about twice as
complicated as the inside formula.  This doubled complexity is typical
of outside formulas, and partially explains why the item-based
description format is so useful: descriptions for the simpler inside
values can be developed with relative ease, and then automatically
transformed used to compute the twice as complicated outside values.

For a context-free grammar, using the CKY parser of Figure
\ref{fig:CKYitem}, recall that the inside probability for an item $[i,
A, j]$ is $P(A \rightarrow w_i... w_{j-1})$.  The outside probability
for the same item is $P(S \derives w_1... w_{i-1} Aw_j... w_n)$.
Thus, the outside probability has the property that when multiplied by
the inside probability, it gives the probability that the start symbol
generates the sentence using the given item, $P(S\derives
w_1... w_{i-1} Aw_j...w_n\derives w_1... w_n)$.  This probability
equals the sum of the probabilities of all derivations using the given
item.  Formally, letting $P(D)$ represent the probability of a
particular derivation, and $C(D, [i, X, j])$ represent the number of
occurrences of item $[i, X, j]$ in derivation $D$ (which for some
parsers could be more than one if $X$ were part of a loop),
$$
\mi{inside}(i, X, j) \times \mi{outside}(i, X, j) = 
\sum_{D \mbox{ \footnotesize a derivation}} P(D)\; C(D, [i, X, j])
$$

The reverse values in general have an analogous meaning.  Let
$C(D,x)$ represent the number of occurrences (the count) of item $x$
in item derivation tree $D$.  Then, for an item $x$, the reverse value $Z(x)$
should have the property
\begin{equation}
\label{eqn:forwardsreverse}
V(x) \otimes Z(x) =\bigoplus_{D \mbox{ \footnotesize a derivation}} V(D) C(D, x)
\end{equation}
Notice that we have multiplied an element of the semiring, $V(D)$, by an
integer, $C(D, x)$.  This multiplication is meant to indicate
repeated addition, using the additive operator of the semiring.  Thus, for
instance, in the Viterbi semiring, multiplying by a count other than 0 has
no effect, since $x \oplus x = \max(x,x)=x$, while in the inside semiring,
it corresponds to actual multiplication.  This value represents the
sum of the values of all derivation trees that the item $x$ occurs
in; if an item $x$ occurs more than once in a derivation tree $D$,
then the value of $D$ is counted more than once.

\begin{figure}
\begin{tabular}{cc}
\psfig{figure=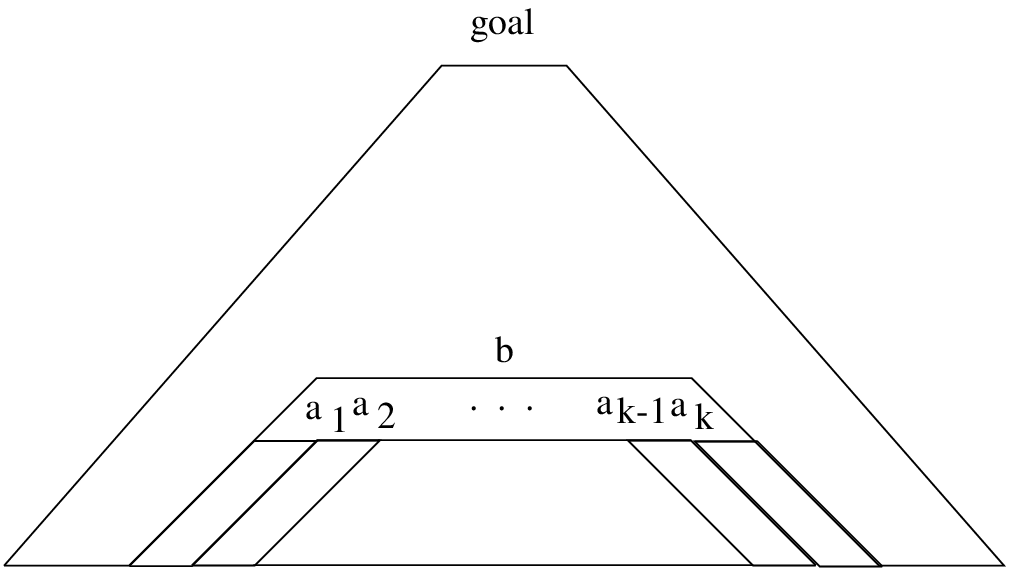,width=2.5in} & \psfig{figure=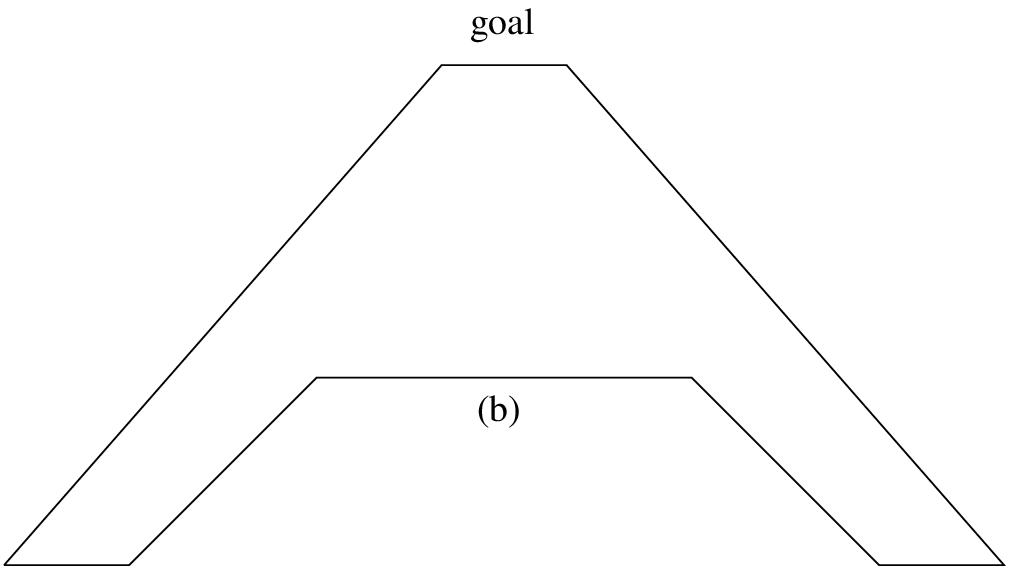,width=2.5in} \\
Derivation of [goal] & Outer tree of $[b]$ 
\end{tabular}
\caption{Goal tree, outer tree}\label{fig:rtriangle}
\end{figure}

To formally define the reverse value of an item $x$, we must first
define the {\em outer} trees $\mi{outer}(x)$.  Consider an item
derivation tree of the goal item, containing one or more instances of
item $x$.  Remove one of these instances of $x$, and its children too,
leaving a gap in its place.  This tree is an outer tree of $x$.
Figure \ref{fig:rtriangle} shows an item derivation tree of the goal
item, including a subderivation of an item $b$, derived from terms
$a_1,...,a_k$.  It also shows an outer tree of $b$, with $b$ and its
children removed; the spot $b$ was removed from is labelled by $(b)$.

For an outer tree $D\in\mi{outer}(x)$, we define its value, $Z(D)$, to
be the product of the value of all rules in $D$, $\bigotimes_{r\in D}
R(r)$.  Then, the reverse value of an item can be formally defined as
\begin{equation}
\label{eqn:reverse}
Z(x) = \bigoplus_{D\in \smi{outer}(x)} Z(D)
\end{equation}
That is, the reverse value of $x$ is the sum of the values of each
outer tree of $x$.

Now, we show that this definition of reverse values has the
property described by Equation \ref{eqn:forwardsreverse}.\footnote{We
note that satisfying Equation \ref{eqn:forwardsreverse} is a useful
but not sufficient condition for using reverse inside values for
grammar re-estimation.  While this definition will typically provide
the necessary values for the E step of an E-M algorithm, additional
work will typically be required to prove this fact; Equation
\ref{eqn:forwardsreverse} should be useful in such a proof.}
\begin{mytheorem}
\label{theorem:reverseproperty}
$$
V(x) \otimes Z(x) =\bigoplus_{D \mbox{ \footnotesize a derivation}} V(D) C(D, x)
$$
\end{mytheorem}
{\em Proof \hspace{3em}}
\begin{eqnarray}
V(x) \otimes Z(x) &= & V(x) \otimes \bigoplus_{O\in
\smi{outer}(x)} Z(O) \nonumber \\
& = & \left( \bigoplus_{I\in\smi{inner}(x)} V(I) \right) \otimes
\bigoplus_{O \in \smi{outer}(x)} Z(O) \nonumber \\
& = & \bigoplus_{I\in\smi{inner}(x)} \bigoplus_{O\in
\smi{outer}(x)} V(I) \otimes Z(O) \label{eqn:innerouterone}
\end{eqnarray}
Next, we argue that this last expression equals the expression on
the right hand side of Equation \ref{eqn:forwardsreverse},
$\bigoplus_D V(D) C(D, x)$.  For an item $x$, any outer part of an
item derivation tree for $x$ can be combined with any inner part to
form a complete item derivation tree.  That is, any $O\in
\mi{outer}(x)$ and any $I\in \mi{inner}(x)$ can be combined to form an
item derivation tree $D$ containing $x$, and any item derivation tree
$D$ containing $x$ can be decomposed into such outer and inner trees.
Thus, the list of all combinations of outer and inner trees
corresponds exactly to the list of all item derivation trees
containing $x$.  In fact, for an item derivation tree $D$ containing
$C(D, x)$ instances of $x$, there are $C(D, x)$ ways to form $D$ from
combinations of outer and inner trees.  Also, notice that for $D$
combined from $O$ and $I$
$$
V(I) \otimes Z(O) = \bigotimes_{r\in I} R(r) \otimes \bigotimes_{r\in O} R(r) = 
\bigotimes_{r\in D} R(r) =
V(D)
$$
Thus,
\begin{equation}
\bigoplus_{I\in\smi{inner}(x)} \bigoplus_{O\in \smi{outer}(x)}
V(I) \otimes Z(O) = \bigoplus_D V(D) C(D, x) \label{eqn:inneroutertwo}
\end{equation}
Combining Equation \ref{eqn:innerouterone} with Equation
\ref{eqn:inneroutertwo}, we see that
$$
V(x) \otimes Z(x) =\bigoplus_{D \mbox{ \footnotesize a derivation}} V(D) C(D, x)
$$
completing the proof. $\Box$

There is a simple, recursive formula for efficiently computing
reverse values.  Recall that the basic equation for computing
forward values not involved in loops was
$$
V(x) = \bigoplus_{a_1... a_k
\;\mbox{\scriptsize s.t.}\;\smallinfer{a_1... a_k}{x}{}} \bigotimes_{i =
1}^k V(a_i)
$$

At this point, for conciseness, we introduce a nonstandard notation.
We will soon be using many sequences of the form
$1,2,...,j-2,j-1,j+1,j+2,...,k-1,k$.  We indicate such sequences by
$1 ,\ldotsj, k$, significantly simplifying some expressions.  By
extension, we will also write $f(1), \ldotsj, f(k)$ to indicate a
sequence of the form
$f(1),f(2),...,f(j-2),f(j-1),f(j+1),f(j+2),...,f(k-1),f(k)$.

Now, we can give a simple formula for computing reverse values $Z(x)$
not involved in loops:

\begin{mytheorem}
For items $x \in B$ where $B$ is non-looping, 
\label{theorem:simplereverse}
\begin{equation}
Z(x) = \bigoplus_{j, a_1... a_k, b
\;\mbox{\scriptsize s.t.}\;\smallinfer{a_1... a_k}{b}{}\wedge x=a_j} Z(b)
\bigotimes_{i = 1 ,\ldotsj, k} V(a_i)
\label{eqn:noloopformula}
\end{equation}
unless $x$ is the goal item, in which case $Z(x) = 1$, the
multiplicative identity of the semiring.
\end{mytheorem}

%We show this by induction on bucket order.  Recall that each item
%belongs to some bucket, and all of the buckets are ordered.
%Furthermore, for a given sentence, only a finite number of items can
%be generated (this was one of our conditions of correctness for an
%item-based description).  Thus, we can induce on the number of
%non-empty buckets following the bucket for $x$.

%
\begin{figure}
\psfig{figure=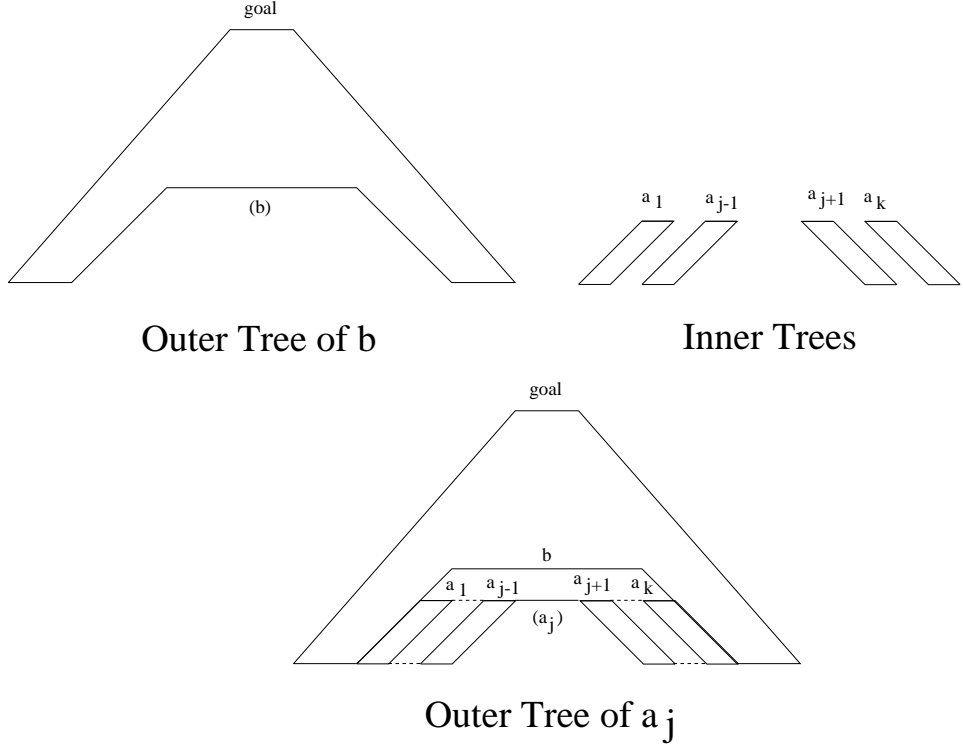,width=5in}
\caption{Combining an outer tree with inner trees to form an outer
tree}
\label{fig:makeouter}
\end{figure}

{\em Proof \hspace{3em}} The simple case is when $x$ is the goal item.
Since an outer tree of the goal item is a derivation of the goal item,
with the goal item and its children removed, and since we assumed in
Section \ref{sec:itembased} that the goal item can only appear in the
root of a derivation tree, the outer trees of the goal item are all
empty.  Thus,
$$
Z(\mi{goal}) =  \bigoplus_{D\in\smi{outer}(\smi{goal})} Z(D) = 
Z(\{\langle\rangle\}) = \bigotimes_{r\in\{\langle\rangle\}} R(r) = 1
$$
As mentioned in Section \ref{sec:semiring}, the product of zero
elements is the multiplicative identity.

Now, we consider the general case.  We need to expand our concept of
{\em outer} to include deduction rules, where $\mi{outer}\left(j,
\shortinfer{a_1... a_k}{b}{}\right)$ is an item derivation tree of the
goal item with one sub-tree removed, a sub-tree headed by $a_j$ whose
parent is $b$ and whose siblings are headed by $a_1 ,\ldotsj, a_k$.
Notice that for every outer tree $D\in \mi{outer}(x)$, there is
exactly one $j, a_1,..., a_k,$ and $b$ such that $x = a_j$ and
$D\in\mi{outer}\left(j, \shortinfer{a_1... a_k}{b}{}\right)$: this
corresponds to the deduction rule used at the spot in the tree where
the sub-tree headed by $x$ was deleted.  Figure \ref{fig:makeouter}
illustrates the idea of putting together an outer tree of $b$ with
inner trees for $a_1,\ldotsj,a_k$ to form an outer tree of $x=a_j$.  Using
this observation,
\begin{eqnarray}
Z(x) & = & \bigoplus_{D\in \smi{outer}(x)} Z(D) \nonumber \\
& = & \bigoplus_{j, a_1... a_k, b
\;\mbox{\scriptsize s.t.}\;\smallinfer{a_1... a_k}{b}{}\wedge x=a_j} 
\;\bigoplus_{D\in
\smi{outer}\left(j,\smallinfer{a_1... a_k}{b}{}\right)} Z(D)
\label{eqn:sumallouter}
\end{eqnarray}
Now, consider all of the outer trees $\mi{outer}\left(j,
\shortinfer{a_1... a_k}{b}{}\right)$.  For each item derivation tree
$D_{a_1}\in \mi{inner}(a_1),\ldotsj,D_{a_k}\in \mi{inner}(a_k)$ and
for each outer tree $D_b\in\mi{outer}(b)$, there will be one outer
tree in the set $\mi{outer}\left(j,
\shortinfer{a_1... a_k}{b}{}\right)$.  Similarly, each tree in
$\mi{outer}\left(j, \shortinfer{a_1... a_k}{b}{}\right)$ can be
decomposed into an outer tree in $\mi{outer}(b)$ and derivation trees
for $a_1 ,\ldotsj, a_k$.  Then,
{
\begin{eqnarray}
\lefteqn{\bigoplus_{D\in \smi{outer}\left(j,\smallinfer{a_1... a_k}{b}{}\right)} Z(D)} 
&&\nonumber \\
&=&\bigoplus_{\longersum{D_b\in \smi{outer}(b),}{D_{a_1}\in
\smi{inner}(a_1) ,\ldotsj,}{\sumindent D_{a_k}\in \smi{inner}(a_k)}} Z(D_b) \otimes V(D_{a_1}) \otimes \cdotsj \otimes V(D_{a_k}) 
\nonumber \\
&=&\left( \bigoplus_{D_{b}\in \smi{outer}(b)} Z(D_b) \right) \otimes
\left( \bigoplus_{D_{a_1}\in \smi{inner}(a_1)} V(D_{a_1}) \right) \otimes 
\cdotsj \otimes
\left( \bigoplus_{D_{a_k}\in \smi{inner}(a_k)} V(D_{a_k}) \right) 
\nonumber \\
&=& Z(b) \otimes
V(a_1) \otimes 
\cdotsj \otimes
V(a_k)
\nonumber \\
&=& Z(b) \otimes \bigotimes_{i=1 ,\ldotsj, k}V(a_i) 
\label{eqn:ZbVaonek}
\end{eqnarray}
Substituting equation \ref{eqn:ZbVaonek} into equation
\ref{eqn:sumallouter}, we conclude that 
$$
Z(x) = \bigoplus_{j, a_1... a_k, b
\;\mbox{\scriptsize s.t.}\;\smallinfer{a_1... a_k}{b}{}\wedge x=a_j} Z(b)
\;\otimes \bigotimes_{i = 1 ,\ldotsj, k} V(a_i)
$$
completing the general case.  $\Box$

Computing the reverse values for loops is somewhat more complicated.
As in the forward case, it requires an infinite sum.  Computation of
this infinite sum will be semiring specific.  Also as in the forward
case, we use the concept of generation.  Let us define the generation
$g$ of an outer tree $D$ of item $x$ in bucket $B$ to be the number of
items in bucket $B$ on the path between the root and the removal
point, inclusive.  Thus, outer trees of items in buckets following $B$
will be in generation 0.  Let $\mi{outer}_{\leq g}(x, B)$ represent
the set of outer trees of $x$ with generation at most $g$.  It should
be clear that as $g$ approaches $\infty$, $\mi{outer}_{\leq g}(x, B)$
approaches $\mi{outer}(x)$.  Now, we can define the $\leq g$
generation reverse value of an item $x$ in bucket $B$, $Z_{\leq g}(x,
B)$:
$$
Z_{\leq g}(x, B) = \bigoplus_{D\in\smi{outer}_{\leq g}(x, B)} Z(D)
$$

For $\omega$-continuous semirings, an infinite sum is equal to the
supremum of the partial sums:
$$
\bigoplus_{D\in\smi{outer}(x, B)} Z(D) = Z_{\leq \infty}(x, B) = \sup_{g} Z_{\leq g}(x, B)
$$
Thus, we wish to find a simple formula for $Z_{\leq g}(x, B)$.

Notice that  for $x \in C$, where $C$ is a bucket following $B$, by
the inclusion of derivations of these items in generation 0,
\begin{equation}
\label{eqn:notinBouter}
Z_{\leq g}(x, B) = Z(x)
\end{equation}
Also notice that for $g=0$ and $x \in B$,
$$
Z_{\leq 0}(x, B) = \bigoplus_{D\in\smi{outer}_{\leq g}(x, B)} Z(D) =
\bigoplus_{D\in\emptyset} Z(D) = 0
$$
Thus, generation 0 makes a simple base for a recursive formula for
outer values.  Now, we can consider the general case, for $g \geq 1$,

\begin{mytheorem}
For items $x \in B$ and $g \geq 1$,
\label{theorem:reverseloop}
\begin{equation}
Z_{\leq g}(x, B) = 
\bigoplus_{{j, a_1...a_k, b}\;\mbox{\scriptsize s.t.}\;
\smallinfer{a_1... a_k}{b}{}\wedge x=a_j}
\;\; \left( \bigotimes_{i =1 ,\ldotsj, k} V(a_i) \right)
\;\;\otimes
\left\{\begin{array}{ll}
{Z_{\leq g\!-\!1}(b, B)}&{\mbox{ \em if } b\in B} \\
{Z(b)}&{\mbox{ \em if } b\notin B}
\end{array} \right.
\label{eqn:reverseloop}
\end{equation}
\end{mytheorem}

The proof parallels that of Theorem \ref{theorem:simplereverse}, and
is given in Appendix \ref{sec:additionalproofs}.

\subsection{Reverse Values in Non-commutative Semirings}
\label{sec:reversenonsmall}

Equations \ref{eqn:noloopformula} and \ref{eqn:reverseloop} apply only
to commutative semirings, since their derivation makes use of the
commutativity of the multiplicative operator.  We might wish to
compute reverse values in non-commutative semirings as well.  For
instance, the reverse values in the derivation semiring could be used
to compute the set of all parses that include a particular
constituent.  It turns out that there is no equation in the
non-commutative semirings corresponding directly to Equations
\ref{eqn:noloopformula} and \ref{eqn:reverseloop}; in fact, in
general, there are no values in non-commutative semirings
corresponding directly to the reverse values.

When we compute reverse values, we need them to be such that the
product of the forward and the reverse values gives the sum of the
values over all derivation trees using the item.  Consider a case in
which the forward value of the item is $b$ and there are derivation
trees with the values $abc$ and $dbe$.  The product of the forward and
reverse values should thus be $abc+dbe$, but in a non-commutative
semiring there will not be any reverse value $x$ such that $xb$ = $abc
+ dbe$.  Instead, one must create what we call Pair semirings,
corresponding to multi-sets of pairs of values in the base semiring.
In this example, the reverse value would be the multiset $\{\langle a,
c\rangle, \langle d, e \rangle\}$ and a special operator would combine
$b$ with this set to yield $abc+dbe$.  Using Pair semirings, one can
find equations directly analogous to Equations \ref{eqn:noloopformula}
and \ref{eqn:reverseloop}.  A complete explication of reverse values
for non-commutative semirings and the derivation of the equations are
given in Appendix \ref{sec:reversenon}.

\section{Semiring Parser Execution}
\subsection{Bucketing}

Executing a semiring parser is fairly simple.  There is, however, one
issue that must be dealt with before we can actually begin parsing.  A
semiring parser computes the values of items in the order of the
buckets they fall into.  Thus, before we can begin parsing, we need to
know which items fall into which buckets, and the ordering of those
buckets.  There are three approaches to determining the buckets and
ordering that we will discuss in this section.  The first approach is
a simple, brute force enumeration of all items, derivable or not,
followed by a topological sort. This approach will have suboptimal
time and space complexity for some item-based descriptions.  The
second approach is to use an agenda parser in the boolean semiring to
determine the derivable items and their dependencies, and to then
perform a topological sort.  This approach has optimal time
complexity, but typically suboptimal space complexity.  The final
approach is to use bucketing code specific to the item-based
interpreter.  This achieves optimal performance for additional
programming effort.

The simplest way to determine the bucketing is to simply enumerate all
possible items for the given item-based description, grammar and input
sentence.  Then, we compute the strongly connected components and a
partial ordering; both steps can be done in time proportional to the
number of items plus the number of dependencies
\cite[ch. 23]{Cormen:90a}.  For some parsers, this technique has
optimal time complexity, although poor space complexity.  In
particular, for the CKY algorithm, the time complexity is optimal, but
since it requires computing and storing all possible $O(n^3)$
dependencies between the items, it takes significantly more space than
the $O(n^2)$ space required in the best implementation.  In general,
the brute force technique raises the space complexity to be the same
as the time complexity.  Furthermore, for some algorithms, such as
Earley's algorithm, there could be a significant time complexity added
as well.  In particular, Earley's algorithm may not need to examine
all possible items.  For certain grammars, Earley's algorithm examines
only a linear number of items and a linear number of dependencies,
even though there are $O(n^2)$ possible items, and $O(n^3)$ possible
dependencies.  Thus the brute force approach would require $O(n^3)$
time and space instead of $O(n)$ time and space.

The next approach to finding the bucketing solves the time complexity
problem.  In this approach, we first parse in the boolean semiring,
using the agenda parser described by \newcite{Shieber:93a}, and then
we perform a topological sort.  Shieber \etal use an interpreter
which, after an item is derived, determines all items that could {\em
trigger} off of that item.  For instance, in an Earley-style parser,
such as that of Figure \ref{fig:Earley}, if an item $[i, A \rightarrow
\alpha\bullet B \beta, j]$ is processed, and there is a rule $B
\rightarrow \gamma$, then, by the prediction rule, $[j, B \rightarrow
{} \bullet \gamma, j]$ can be derived.  Thus, immediately after $[i, A
\rightarrow \alpha\bullet B \beta, j]$ is processed, $[j, B
\rightarrow {} \bullet \gamma, j]$ is added to an agenda; new items to
be processed are taken off of the agenda.  This approach works fine
for the boolean semiring, where items only have value $\mi{TRUE}$ or
$\mi{FALSE}$, but cannot be used directly for other semirings.  For
other semirings, we need to make sure that the values of items are not
computed until after the values of all items they depend on are
computed.  However, we can use the algorithm of Shieber \etal to
compute all of the items that are derivable, and to store all of the
dependencies between the items.  Then we perform a topological sort on
the items.  The time complexity of both the agenda parser and the
topological sort will be proportional to the number of dependencies,
which will be proportional to the optimal time complexity.
Unfortunately, we still have the space complexity problem, since
again, the space used will be proportional to the number of
dependencies, rather than to the number of items.

The third approach to bucketing is to create algorithm-specific
bucketing code; this results in parsers with both optimal time and
optimal space complexity.  For instance, in a CKY style parser, we can
simply create one bucket for each length, and place each item into the
bucket for its length.  For some algorithms, such as Earley's
algorithm, special-purpose code for bucketing might have to be
combined with code for triggering, just as in the algorithm of Shieber
\etal, in order to achieve optimal performance.  

\subsection{Interpreter}
\begin{figure}
\begin{tabbing}
\verb|    |\=\verb|    |\=\verb|    |\=\verb|    |\=\verb|    |\=\verb|    |\=\verb|    |\=\verb|    |\=\verb|    |\=\verb|    |\= \kill
$\mi{current} := \mi{first};$ \\
\aldo \\
 \> \alif $\mi{loop}(\mi{current})$\\
 \>  \> {\it /* replace with semiring specific code */} \\
 \>  \> \alfor $x \in \mi{current}$ \\
 \>  \>  \> $V[x,0] = 0$; \\
 \>  \> \alfor $g := 1\;$ \alto $\infty$ \\
 \>  \>  \> \alforeach $x\in \mi{current}, a_1...a_k$ s.t. $\shortinfer{a_1...a_k}{x}{}$\\
%\>  \>  \> \alforeach instantiation $\shortinfer{a_1...a_k}{x}{}$ s.t. $x \in \mi{current}$ \\
 \>  \>  \>  \> $V[x,g] := V[x, g] \oplus \bigotimes_{i=1}^k 
\left\{ \begin{array}{ll} 
V[a_i] & a_i \notin \mi{current} \\
V[a_i, g\!-\!1] & a_i \in \mi{current} \\
\end{array}
\right.$ \\
 \>  \> \alforeach  $x \in \mi{current}$ \\
 \>  \>  \> $V[x] := V[x, \infty]$; \\
 \> \alelse \\
 \>  \> \alforeach $x\in \mi{current}, a_1...a_k$ s.t. $\shortinfer{a_1...a_k}{x}{}$\\
%\>  \> \alforeach instantiation $\shortinfer{a_1...a_k}{x}{}$ s.t. $x \in \mi{current}$ \\
 \>  \>  \> $V[x] := V[x] \oplus \bigotimes_{i=1}^k V[a_i]$; \\ 
 \> $\mi{oldCurrrent} := \mi{current}$;\\ 
 \> $\mi{current} := \mi{next}(\mi{current})$; \\
\alwhile $\mi{oldCurrent} \neq \mi{last}$ \\
\alreturn $V[\mi{goal}]$; \\
\end{tabbing}
\caption{Forward Semiring Parser Interpreter}\label{fig:ringparser}
\end{figure}

Once we have the bucketing, the parsing step is fairly simple.  The
basic algorithm appears in Figure \ref{fig:ringparser}.  We simply
loop over each item in each bucket.  There are two types of buckets:
looping buckets, and non-looping buckets.  If the current bucket is a
looping bucket, we compute the infinite sum needed to determine the
bucket's values; in a working system, we substitute semiring specific
code for this section, as described in Section
\ref{sec:solvinginfinite}.  If the bucket is not a looping bucket, we
simply compute all of the possible instantiations that could
contribute to the values of items in that bucket.  Finally, we return
the value of the goal item.

\begin{figure}
\begin{tabbing}
\verb|    |\=\verb|    |\=\verb|    |\=\verb|    |\=\verb|    |\=\verb|    |\=\verb|    |\=\verb|    |\=\verb|    |\=\verb|    |\= \kill
$\mi{current} := \mi{last}$; \\
\aldo \\
 \> \alif $\mi{loop}(\mi{current})$\\
 \>  \> {\it /* replace with semiring specific code */} \\
 \>  \> \alfor $x \in \mi{current}$ \\
 \>  \>  \> $Z[x,0] = 0$; \\
 \>  \> \alfor $g := 1\;$ \alto $\infty$ \\
 \>  \>  \> \alforeach $j, a_1... a_k, x$ s.t. $\shortinfer{a_1...a_k}{x}{}$ and $a_j \in \mi{current}$ \\
 \>  \>  \>  \> $Z[a_j,g] := Z[a_j, g] \oplus \bigotimes_{i=1,\ldotsj, k}  V[a_i] \otimes
\left\{ \begin{array}{ll} 
Z[x] & x \notin \mi{current} \\
Z[x, g\!-\!1] & x \in \mi{current} \\
\end{array}
\right.$ \\
 \>  \> \alforeach  $x \in \mi{current}$ \\
 \>  \>  \> $Z[x] := Z[x, \infty]$; \\
 \> \alelse \\
 \>  \> \alforeach $j, a_1... a_k, x$ s.t. $\shortinfer{a_1...a_k}{x}{}$ and $a_j \in \mi{current}$ \\
%\>  \> \alforeach $j$, instantiation $\shortinfer{a_1,...,a_k}{x}{}$ s.t. $a_j \in \mi{current}$ \\
 \>  \>  \> $Z[a_j] := Z[a_j] \oplus Z[x] \otimes \bigotimes_{j=1,\ldotsj, k} V[a_j]$; \\
 \> $\mi{oldCurrrent} := \mi{current}$;\\ 
 \> $\mi{current} := \mi{previous}(\mi{current})$; \\
\alwhile $\mi{oldCurrent} \neq \mi{first}$ \\
\end{tabbing}
\caption{Reverse Semiring Parser Interpreter}\label{fig:reverseparser}
\end{figure}

The reverse semiring parser interpreter is very similar to the forward
semiring parser interpreter.  The differences are that in the reverse
semiring parser interpreter, we traverse the buckets in reverse order, and
we use the formula for the reverse values, rather than the forward
values.

Both interpreters are closely based on formulas derived earlier.  The
forward semiring parser interpreter uses the code
\begin{tabbing}
\verb|    |\=\verb|    |\=\verb|    |\=\verb|    |\=\verb|    |\=\verb|    |\=\verb|    |\=\verb|    |\=\verb|    |\=\verb|    |\= \kill
 \> \alforeach $x\in \mi{current}, a_1...a_k$ s.t. $\shortinfer{a_1...a_k}{x}{}$\\
 \>  \> $V[x,g] := V[x, g] \oplus \bigotimes_{i=1}^k 
\left\{ \begin{array}{ll} 
V[a_i] & a_i \notin \mi{current} \\
V[a_i, g\!-\!1] & a_i \in \mi{current} \\
\end{array}
\right.$
\end{tabbing}
to implement Equation \ref{eqn:inloop} for computing the values of looping buckets.
$$
V_{\leq g}(x, B) =
\bigoplus_{a_1...a_k \;\mbox{\scriptsize s.t.}\; \smallinfer{a_1... a_k}{x}{}}
\;\;\bigotimes_{i=1..k} 
\left\{
\begin{array}{ll}
{V(a_i)}{\mbox{ \em if }a_i\notin B} \\
{V_{\leq g\!-\!1}(a_i, B)}{\mbox{ \em if }a_i\in B}
\end{array}
\right.
$$
For computing the values of non-looping buckets, the interpreter uses
the code
\begin{tabbing}
\verb|    |\=\verb|    |\=\verb|    |\=\verb|    |\=\verb|    |\=\verb|    |\=\verb|    |\=\verb|    |\=\verb|    |\=\verb|    |\= \kill
 \>  \> \alforeach $x\in \mi{current}, a_1...a_k$ s.t. $\shortinfer{a_1...a_k}{x}{}$\\
 \>  \>  \> $V[x] := V[x] \oplus \bigotimes_{i=1}^k V[a_i]$ ;
\end{tabbing}
which is simply an implementation of Equation \ref{eqn:itemvalue}
$$
V(x) = \bigoplus_{a_1...a_k \;\mbox{\scriptsize
s.t.}\;\smallinfer{a_1... a_k}{x}{}} \bigotimes_{i = 1}^k
V(a_i)
$$
The corresponding lines in the reverse interpreter correspond to
Equations \ref{eqn:reverseloop} and \ref{eqn:noloopformula},
respectively.  

Using these equations, a simple inductive proof shows that the
semiring parser interpreter is correct, and an analogous theorem holds
for the reverse semiring parser interpreter.

\begin{mytheorem}\label{theorem:correctinterpret}%
The forward semiring parser interpreter correctly computes the value
of all items.
\end{mytheorem}
A sketch of the proof is given in Appendix \ref{sec:additionalproofs}.

There are two other implementation issues.  First, for some parsers,
it will be possible to discard some items.  That is, some items serve
the role of temporary variables, and can be discarded after they are
no longer needed, especially if only the forward values are going to
be computed.  Also, some items do not depend on the input string, but
only on the rule value function of the grammar.  The values of these
items can be precomputed, using the forward semiring parser
interpreter.

\section{Grammar Transformations}

\label{sec:grammar}

We can apply the same techniques to grammar transformations that we
have so far applied to parsing.  Consider a grammar transformation,
such as the Chomsky Normal Form (CNF) grammar transformation, which
takes a grammar with epsilon, unary, and n-ary branching productions,
and converts it into one in which all productions are of the form $A
\rightarrow BC$ or $A \rightarrow a$.  For any sentence $w_1... w_n$
its value under the original grammar in the boolean semiring
($\mi{TRUE}$ if the sentence can be generated by the grammar, $\mi{FALSE}$
otherwise) is the same as its value under a transformed grammar.
Therefore, we say that this grammar transformation is {\em value
preserving} under the boolean semiring.  We can generalize this
concept of value preserving to other semirings.  For instance, if
properly specified, the CNF transformation also preserves value under
any complete commutative semiring, so that the value of any sentence
in the transformed grammar is the same as the value of the sentence in
the original grammar.  Thus, for instance, we could start with a
grammar with rule probabilities, transform it using the CNF
transformation, and find Viterbi values using a CKY parser and the
transformed grammar; the Viterbi values for any sentence would be the
values using the original grammar.

We now show how to specify grammar transformations using almost the
same item-based descriptions we used for parsing.  We give a value
preserving transformation to CNF in this section, and in Appendix
\ref{sec:GNFappendix}, we give a value preserving transformation to
Greibach Normal Form (GNF).  While item-based descriptions have been
used to specify parsers by \newcite{Shieber:93a} and
\newcite{Sikkel:93a}, we do not know of previous uses for specifying
grammar transformations.

The concept of value preserving grammar transformation is known in the
intersection of formal language theory and algebra.  Kuich
\shortcite{Kuich:97a,Kuich:86a} shows how to perform transformations
to both CNF and GNF, with a value-preserving formula.
\newcite{Teitelbaum:73a} shows how to convert to CNF for a subclass
of $\omega$-continuous semirings.  The contribution, then, of this
section is to show that these value preserving transformations can be
fairly simply given as item-based descriptions, allowing the same
computational machinery to be used for grammar transformations as is
used for parsing, and to some extent showing the relationship between
certain grammar transformations and certain parsers, such as that of
\newcite{Graham:80a}, discussed in Section \ref{sec:GHR} and Appendix
\ref{sec:GHRappendix}.  While the relationship between grammar
transformations and parsers is already known in the literature on
covering grammars \cite{Nijholt:80a,Leermakers:89a}, our treatment is
clearer, because we use the same machinery for specifying both the
transformations and the parsers, allowing commonalities to be
expressed in the same way in both cases.

%\subsection{Unary Removal Transformation Example}
\begin{figure}
$$
\begin{array}{cl}
\mbox{\bf Item form:} & \\
{[A \rightarrow \alpha  \beta]} & \\\\
\mbox{\bf Rule Goal}& \\
R_2(A \rightarrow \alpha) \\
\\
\mbox{\bf Rules:}& \\
\infer{R_1(A \rightarrow B C \alpha)}{[A \rightarrow  B C \alpha]}{} &
 \mbox{N-ary}\\
\infer{R_1(A \rightarrow a)}{[A \rightarrow a]}{} & 
 \mbox{Unary}\\
\infer{R_1(A \rightarrow B)\rspace[B \rightarrow \alpha]}{[A \rightarrow \alpha]}{} &
 \mbox{Extension}\\
\infer{[A \rightarrow \alpha]}{R_2(A \rightarrow \alpha)}{}& \mbox{Output}
\end{array}
$$
\caption{Removal of Unary Productions}\label{fig:nounary}
\end{figure}

There are three steps to the CNF transformation: removal of epsilon
productions; removal of unary productions; and, finally, splitting of
n-ary productions.  Of these three, the best one for expository
purposes is the removal of unary productions, shown in Figure
\ref{fig:nounary}, so we will explicate this transformation first,
even though logically it belongs second.  This transformation assumes
that the grammar contains no epsilon rules.  It will be convenient to
allow variables $A$, $B$, $C$ to represent either nonterminals or
terminals throughout this section.

To distinguish rules in the old grammar from rules in the new grammar,
we have numbered the rule functions, $R_1$ in this transformation for
the original grammar, and $R_2$ for the new grammar.  The only
difference between an item-based parser description and an item-based
grammar transformation description is the goals section; instead of a
single goal item, there is a rule goal with variables, such as $R_2(A
\rightarrow \alpha)$, giving the value of items in the new grammar.

An item of the form $[A \rightarrow B C \alpha]$ can derived if and
only there is a derivation of the form
$$
A \Rightarrow D \Rightarrow E \Rightarrow \cdots \Rightarrow F
\Rightarrow B C \alpha
$$
An item in this form is simply derived using the extension rule
several times, combined with the N-ary rule.  Similarly, an item
of the form $[A \rightarrow a]$ can be derived if and only if there is
a derivation of the form
$$
A \Rightarrow D \Rightarrow E \Rightarrow \cdots \Rightarrow F
\Rightarrow a
$$
Items of the form $[A \rightarrow B]$ where $B$ is a nonterminal
cannot be derived. 

A short example will help illustrate how this grammar transformation
works.  Consider the following grammar, with values in the inside
semiring:
\begin{equation}
\label{eqn:unarygrammar}
\begin{array}{rcll}
S & \rightarrow & A a & (1.0)\\
A & \rightarrow & a & (0.5)\\
A & \rightarrow & A & (0.5)\\
\end{array}
\end{equation}
There are an infinite number of derivations of the item $[A
\rightarrow a]$.  It can be derived using just the unary rule, with
value 0.5; it can be derived using the unary rule and the extension
rule, with value 0.25; it can be derived using the unary rule and
using the extension rule twice, with value 0.125; and so on.  The
total of all derivations is 1.0.  There is just one derivation for the
item $[S \rightarrow A a]$, which has value 1.0 also.  Thus, the
resulting grammar is:
\begin{equation}
\label{eqn:insidegrammar}
\begin{array}{rcll}
S & \rightarrow & A a & (1.0)\\
A & \rightarrow & a & (1.0)\\
\end{array}
\end{equation}
which has no nonterminal unary rules.

%\subsection{Conditions for Correctness of Grammar Transformations}

Now, with an example finished, we can discuss conditions for
correctness for a grammar transformation.

Consider a grammar derivation $D$ (as always, left-most) in the
original grammar, and a grammar derivation $E$ in the new grammar,
using item derivations $I$ to derive the rules used in $E$.  Roughly,
if there is a one-to-one pairing between old-grammar derivations $D$
and pairs $(E, I)$, then, the transformation is value preserving.
Formally,

\begin{mytheorem}
Consider a derivation $D$ in the original grammar, using rules
$D_1... D_d$.  Consider also a derivation in the new grammar $E$ using
rules $E_1... E_e$.  For each new grammar rule $E_i$, there is some
set of item-based derivations of that rule, $I_i^1...I_i^{j_i}$.  We
can consider sequences of such rule derivations, $I_1^{k_1},
I_2^{k_2},...,I_e^{k_e}$, selecting one rule derivation $I_i^{k_i}$
for each new grammar rule $E_i$.  A grammar transformation will be
value preserving for a commutative semiring if there is a one-to-one
pairing between derivations $D_1...D_d$ and pairs $(E_1...E_e,
I_1^{k_1}...I_e^{k_e})$ and if all rule values (from the original
grammar) occur the same number of times in $I_1^{k_1}...I_e^{k_e}$ as
they do in $D_1... D_d$.
\end{mytheorem}
{\em Proof \hspace{3em}} The proof is obvious, since each term in the
sum over derivations in the original grammar has a term in the sum
over derivations in the new grammar.  The details essentially follow
Theorem \ref{theorem:itemgrammar}.  $\Box$

There are two important caveats to note about this form of grammar
transformation.  The first is that the grammar transformation is
semiring specific.  Consider the probabilistic grammar example,
\ref{eqn:unarygrammar}, transformed with the unary productions removal
transformation.  If we transform it using the inside semiring, we get
Grammar \ref{eqn:insidegrammar}.  On the other hand, if we transform
the grammar using the Viterbi semiring, we get
$$
\begin{array}{rcll}
S & \rightarrow & A a & (1.0) \\
A & \rightarrow & a & (0.5)
\end{array}
$$
which is also correct: the Viterbi semiring value of the string $a a$
in the original grammar is 0.5, just as it is in the transformed
grammar.  Notice that the values differ; it is important to remember
that grammar transformations are semiring specific.  Furthermore, notice
that the probabilities do not sum to 1 in the transformed grammar in
the Viterbi semiring.  While for this example, and the unary productions
removal transformation in general, transformations using the inside
ring do preserve summation to 1, this is not always true, as we will
show during our discussion of the epsilon removal transformation.

\begin{figure}
$$
\begin{array}{cl}
\mbox{\bf Item form:} \\
{[A \rightarrow \alpha \bullet \beta]} \\
\\
\mbox{\bf Rule Goal}\\
R_1(A \rightarrow \alpha)\\
\\
\mbox{\bf Rules:} \\

\infer{R_0(A\rightarrow \alpha)}{[A\rightarrow {} \bullet\alpha]}{}
& \mbox{Prediction} \\

\infer{[A\rightarrow \alpha \bullet B \beta ]\rspace[B \rightarrow {} \bullet {} ]}{[A \rightarrow \alpha\bullet \beta]}{}
& \mbox{Epsilon Completion} \\

\infer{[A\rightarrow \alpha \bullet B \beta ]}{[A \rightarrow \alpha B
\bullet \beta]}{[B \rightarrow C \gamma \bullet{}]}
& \mbox{Non-epsilon Completion} \\

\infer{[A\rightarrow \alpha \bullet a \beta ]}{[A \rightarrow \alpha a
\bullet \beta]}{}
& \mbox{Scanning} \\

\infer{[A \rightarrow B\alpha \bullet {} ]}{R_1(A \rightarrow B\alpha)}{}
& \mbox{N-ary Output}\\

\infer{[S \rightarrow {} \bullet {} ]}{R_1(S\rightarrow \epsilon)}{}
& \mbox{Epsilon Output}

\end{array}
$$
\caption{Removal of Epsilon Productions}\label{fig:noepsilon}
\end{figure}

Next, we consider the epsilon removal transformation.  This transformation is fairly simple; it is derived
from Earley's algorithm.  This transformation assumes $S$ does not
occur on the right hand side of any productions.  There is one item
form, $[A \rightarrow \alpha \bullet \beta]$ that can be derived if
and only if there is a derivation of the form $A
\rightarrow\gamma\beta\derives\alpha\beta$, using only substitutions
of the form $B\derives\epsilon$, and where each symbol $C$ in $\alpha$
can derive some string of terminals.

There are six rules.  The first, prediction, simply makes sure there
is one initial item, $[A \rightarrow {} \bullet \alpha]$ for each rule of
the form $A \rightarrow \alpha$.  The next rule, epsilon completion,
allows deletion of symbols $B$ that derive epsilon from rules of the
form $A \rightarrow \alpha B\beta$, while the following rule,
non-epsilon completion, simply moves the dot over symbols $B$ that
can derive terminal strings.  The fourth rule, scanning, moves the dot
over terminals.  The last two rules, $n$-ary output and epsilon
output, derive the output values.  

For simplicity, the removal of unary productions
transformation and the removal of $n$-ary productions transformations
do not handle the rule $S \rightarrow \epsilon$ produced by the
epsilon output rule, but could easily be modified to do so.

While the epsilon removal transformation preserves the value of any
string using the inside semiring, it does not in general produce a grammar
with probabilities that sum to 1.  Consider the probabilistic grammar
$$
\begin{array}{rcll}
S & \rightarrow & a B & (1.0) \\
B & \rightarrow &\epsilon & (0.7) \\
B & \rightarrow & b & (0.3) \\
\end{array}
$$
The grammar with epsilons removed using the inside semiring will be
$$
\begin{array}{rcll}
S & \rightarrow & a B & (1.0) \\
S & \rightarrow & a & (0.7) \\
B & \rightarrow & b & (0.3) \\
\end{array}
$$
which generates the same strings with the same probabilities: value is
preserved; notice however that probabilities of individual
nonterminals sum to both more and less
than one and that simply normalizing probabilities by dividing through
by the total for each left hand side leads to a grammar with different
string probabilities.  The lack of summation to one could potentially
make this grammar less useful in a system that needed, for instance, 
intermediate probabilities for thresholding.

\begin{figure}
$$
\begin{array}{cl}
\mbox{\bf Item form:} &\\
{[A\rightarrow \alpha\bullet\beta]} &\\ &\\
\mbox{\bf Goal:} &\\
{[S' \rightarrow S \bullet {} ]} &\\& \\
\mbox{\bf Rules:} \\
\infer{}{[S'\rightarrow {} \bullet S]}{} &\mbox{Initialization} \\

\infer{[A \rightarrow \alpha\bullet a \beta]}
{[A \rightarrow \alpha a\bullet \beta]}
{} &\mbox{Scanning}\\

\infer{R(B \rightarrow\gamma)}{[B \rightarrow {} \bullet\gamma]}
{[A \rightarrow \alpha\bullet B\beta]} &\mbox{Prediction} \\

\infer{[A \rightarrow \alpha\bullet B\beta]\rspace
[B \rightarrow\gamma\bullet {} ]}
{[A \rightarrow \alpha B\bullet\beta]}{}  &\mbox{Completion}

\end{array}
$$
\caption{Renormalization Parsing}\label{fig:renormalize}
\end{figure}

We can correctly renormalize using a modified item-based description
of Earley's algorithm, in Figure \ref{fig:renormalize}.  This parser
is just like Earley's parser, except that the indices have been
removed, and the scanning rule does not make reference to the words of
the sentence.  There is one valid derivation in this parser for each
derivation in the grammar.  Computing 
$$
\frac{V([A \rightarrow \alpha\bullet {} ])\times Z([A \rightarrow \alpha\bullet {} ])}
{\sum_{\beta} V([A \rightarrow \beta\bullet {} ])\times Z([A \rightarrow \beta\bullet {} ])}
$$
gives the normalized probability $P(A \rightarrow \alpha)$.  The
intuition behind this formula is simply that it is the usual formula
for inside-outside re-estimation, and inside-outside re-estimation,
when in a local minimum, stays in the same place.  Since a grammar is
a local minimum of itself, we should get essentially the same grammar,
but with normalized rule probabilities.  A stronger argument is given
in Appendix \ref{sec:additionalproofs}, Theorem
\ref{theorem:renormalize}.

The renormalization parser has some other useful properties.  For
instance, in the counting semiring, $V([S' \rightarrow S \bullet {} ])$ gives
the total number of parses in the language; in the Viterbi-n-best semiring,
it gives the probabilities and parses of the $n$ most probable parses
in the language; and in the parse-forest semiring, it gives a derivation
forest for the entire language.  In the inside semiring, it should always
be 1.0, assuming a proper grammar as input.

One last interesting property of the renormalization parser is that it can be
used to remove useless rules.  The forward boolean value times the
reverse boolean value of an item $[A \rightarrow \alpha
\bullet {} ]$ will be $\mi{TRUE}$ if and only if the rule $A \rightarrow \alpha$
is useful; that is, if it can appear in the derivation of some string.
Useless rules can be eliminated.

\begin{figure}
$$
\begin{array}{c} 
\mbox{\bf Item form:} \\
\\
\mbox{\bf Rule Goal}\\
R_3(\alpha \rightarrow \beta) \\
\\
\mbox{\bf Rules:} \\
\infer{R_2(A\rightarrow B\;C)}{R_3(A \rightarrow B\;C)}{}\\
\infer{R_2(A\rightarrow B\;C\;D\;\alpha)}{R_3(A \rightarrow B\; \langle
CD\alpha \rangle)}{}\\
\infer{}{R_3(\langle C D \rangle \rightarrow C\; D)}{R_2(A \rightarrow B \alpha C
D)}\\ 
\infer{}{R_3(\langle C D  E \alpha \rangle \rightarrow C \;  \langle
DE\alpha \rangle)}{R_2(A \rightarrow B \beta C D E \alpha)} \\
\\
\end{array}
$$
\caption{Removal of n-ary Productions}\label{fig:nonary}
\end{figure}

For completeness, there are two more steps in the CNF transformation:
conversion from n-ary branching rules ($n \geq 2$) to binary branching
rules, and conversion from binary branching rules with terminal
symbols to those with nonterminals.  We show how to convert from n-ary to
binary branching rules in Figure \ref{fig:nonary}.  This
transformation assumes that all unary and $\epsilon$ rules have been
removed.  We construct many new nonterminals in this transformation,
each of the form $\langle C D \alpha \rangle$.  This step is
fairly simple, so we won't explicate it.  The remaining step, removal
of terminal symbols from binary branching rules, is trivial and we do
not present it.

We should note that not all transformations have a value-preserving
version.  For instance, in the general case, the transformation of a
non-deterministic finite state automaton (NFA) into a deterministic
finite state automaton (DFA) cannot be made value preserving: the
problem is that in the DFA, there is exactly one derivation for a
given input string, so it is not possible to get a one-to-one
correspondence between derivations in the NFA and the DFA, meaning
that these techniques cannot be used.  (\newcite{Mohri:97a} discusses
the conditions under which some NFAs in certain semirings can be made
deterministic.)

\section{Examples}

In this section, we give examples of several parsing algorithms,
expressed as item-based descriptions.  

\subsection{Finite State Automata and Hidden Markov Models}
NFAs and HMMs can both be expressed using a single item-based
description, shown in Figure \ref{fig:NFA}.  We will express
transitions from state $A$ to state $B$ emitting symbol $a$ as $A
\rightarrow a, B$.  As usual, we will let $R(A \rightarrow a, B)$ have
different values depending on the semiring used.

\begin{center}
\paper{\begin{tabular}{lp{4.5in}}}
\thesis{\begin{tabular}{lp{5.0in}}}
Boolean & $\mi{TRUE}$ if there is a transition from $A$ to $B$ emitting
$a$, $\mi{FALSE}$ otherwise \\
Counting & 1 if there is a transition from $A$ to $B$ emitting $a$, 0 otherwise \\
Derivation & $\{\langle A \rightarrow a, B\rangle\}$ if there
is a transition from $A$ to $B$ emitting $a$, $\emptyset$ otherwise\\
Viterbi &  Probability of a transition from $A$ to $B$ emitting $a$ \\
Inside & Probability of a transition from $A$ to $B$ emitting $a$ 
\end{tabular}
\end{center}

We assume there is a single start state $S$ and a single final state
$F$, and we allow $\epsilon$ transitions.

For HMMs, notice that the forward algorithm is obtained simply by
using the inside semiring; the backwards algorithm is obtained using the
reverse values of the inside semiring; and the Viterbi algorithm is
obtained using the Viterbi semiring.  For NFAs, we can use the
boolean semiring to determine whether a string is in the language of
an NFA; we can use the counting semiring to determine how many state
sequences there are in the NFA for a given string; and we can use the
derivation forest semiring to get a compact representation of all state
sequences in an NFA for an input string.

\begin{figure}
$$
\begin{array}{cl}
\mbox{\bf Item form:} \\
{[A, i]} \\
\\
\mbox{\bf Goal}\\
{[F, n\!+\!1]} \\
\\
\mbox{\bf Rules:} \\
\infer{}{[S,1]}{} & \mbox{Start Axiom}\\

\infer{[A, i]\rspace R(A \rightarrow w_i, B)}{[B, i\!+\!1]}{} &
\mbox{Scanning} \\

\infer{[A, i]\rspace R(A \rightarrow\epsilon, B)}{[B, i]}{} &
\mbox{Epsilon Scanning}

\end{array}
$$
\caption{NFA/HMM parser}\label{fig:NFA}
\end{figure}

\subsection{Prefix Values}

For language modeling, it may be useful to compute the prefix
probability of a string.  That is, given a string $w_1... w_n$, we may
wish to know the total probability of all sentences beginning with
that string,
$$
\sum_{k \geq 0,x_1, ...x_k} P(S \rightarrow w_1... w_nx_1...x_k)
$$
\newcite{Jelinek:91b} and \newcite{Stolcke:93a} both give algorithms
for computing the prefix probabilities.  However, the derivations are
somewhat complex, requiring five pages of Jelinek and Lafferty's nine
page paper.

\begin{figure}
$$
\begin{array}{cl}
\mbox{\bf Item form:} \\
{[i, A, j]} \\
{[i, A]} \\
{[A]} \\
\\
\mbox{\bf Goal}\\
{[1, S]} \\
\\
\mbox{\bf Rules:} \\
\infer{R(A \rightarrow w_i)}{[i, A, i\!+\!1] }{} & \mbox{Unary In}\\
\infer{ R(A \rightarrow BC)\rspace[i, B, k]\rspace [k, C, j] }{[i, A, j]}{}
& \mbox {In In} \\
\infer{R(A \rightarrow a)}{[A]}{} & \mbox{Unary Out} \\
\infer{R(A \rightarrow BC)\rspace [B]\rspace [C]}{[A]}{}
& \mbox {Out Out} \\
\infer{R(A \rightarrow w_{n})}{[n,A]}{}{}
& \mbox{Unary Between} \\
\infer{R(A\rightarrow BC)\rspace[i,B,j]\rspace[j,C]}{[i, A]}{}
& \mbox{In Between} \\
\infer{R(A\rightarrow BC)\rspace[i,B]\rspace[C]}{[i, A]}{}
& \mbox{Between Out}
\end{array}
$$
\caption{Prefix Derivation Rules}\label{fig:prefix}
\end{figure}

In contrast, we give a fairly simple item-based description in Figure
\ref{fig:prefix}, which we will explicate in detail below.  We will
call a {\em prefix derivation} $X \prederives w_i... w_j$ a derivation
in which $X \derives w_i... w_jx_1x_2... x_k$.

The prefix value of a string is the sum of the products of the values
used in the prefix derivation.  A brief example will help: consider
the prefix $a$ and the grammar
$$
\begin{array}{rcll}
S & \rightarrow & s A & (1) \\
A & \rightarrow & a & (0.3) \\
A & \rightarrow & b  & (0.7)
\end{array}
$$
There are two prefix derivations of $s$ of the form: $S \derives sA
\derives sa$ (0.3) and $S \derives sA \derives sb$ (0.7).  The inside
score is the sum, 1, while the Viterbi score is the $\max$, 0.7.

Figure \ref{fig:prefix} gives an item-based description for finding
prefix values for CNF grammars; it would be straightforward to modify
this algorithm for an Earley-style parser, allowing it to handle any
CFG, as shown by \newcite{Stolcke:93a}.

There are three item types in this description.  The first item type,
$[i, A, j]$, called {\em In}, can be derived only if $A \derives
w_i... w_{j-1}$.  The second type, $[i, A]$, called {\em Between}, can
be derived only if $A \derives w_i... w_nx_1... x_k$.  The final type,
$[A]$, called {\em Out}, can be derived only if $A \derives
x_1... x_k$.  The Unary In and In In rules correspond to the usual
unary and binary rules.  The Unary Out and Out Out rules correspond to
the usual unary and binary rules, but for symbols after the prefix.
For instance, the Out Out rule says that if $A \rightarrow BC$ and $B
\derives x_1...x_k$ and $C \derives y_1...y_l$ then $A \derives
x_1...x_ky_1...y_l$.  The last three rules deal with Between items.  Unary
Between says that if $A \rightarrow w_n$ then $A \derives w_n$.  In
Between says that if $A \rightarrow BC$ and $B \derives w_i...w_{j-1}$
and $C \derives w_j...w_nx_1...x_k$ then $A \derives
w_i...w_nx_1...x_k$.  Finally, Between Out says that if $A \rightarrow
BC$ and $B \derives w_i...w_nx_1...x_k$ and $C \derives y_1...y_l$
then $A \derives w_i...w_nx_1...y_l$.  Figure \ref{fig:prefixdrawing}
illustrates the seven different rules.  Words in the prefix are
indicated by solid triangles, and words that could follow the prefix
are indicated by dashed ones.

\begin{figure}
\psfig{figure=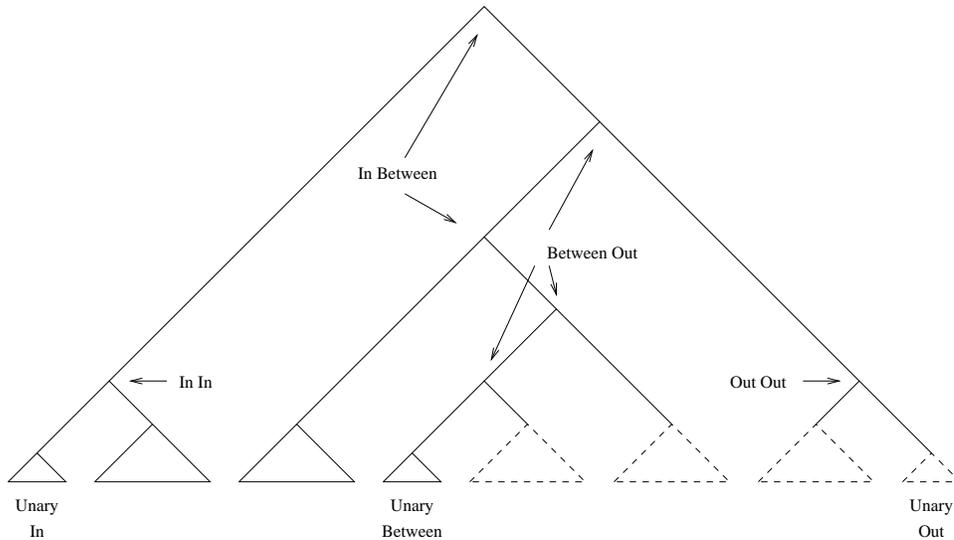,width=5in}
\caption{Prederivation Illustration} \label{fig:prefixdrawing}
\end{figure}

There is one problem with the item-based description of Figure
\ref{fig:prefix}.  Both the Out and the Between items are all
associated with looping buckets.  Since the Between items can only be
computed on line (i.e. only once we know the input sentence), this
means that we must perform time consuming infinite sums on line.  It
turns out that with a few modifications, the values of all items
associated with looping buckets can be computed off-line, once per
grammar, significantly speeding up the on-line part of the computation.

\begin{figure}
$$
\begin{array}{cl}
\mbox{\bf Item form:} \\
{[i, A, j]} \\
{[i, A]} \\
{[A]} \\
{[A\prederives B]} \\
\\
\mbox{\bf Goal}\\
{[1, S]} \\
\\
\mbox{\bf Rules:} \\
\infer{R(A \rightarrow w_i)}{[i, A, i\!+\!1] }{} & \mbox{Unary In}\\
\infer{ R(A \rightarrow BC)\rspace [k, C, j]\rspace[i, B, k] }{[i, A, j]}{}
& \mbox {In In} \\
\infer{R(A \rightarrow a)}{[A]}{} & \mbox{Unary Out} \\
\infer{R(A \rightarrow BC)\rspace [C]\rspace [B]}{[A]}{}
& \mbox {Out Out} \\
\infer{}{[A\prederives A]}{} 
& \mbox{Prederivation Axiom}\\
\infer{R(A \rightarrow BC) \rspace [C]\rspace [B\prederives D]}{[A\prederives
D]}{} 
& \mbox{Prederivation Completion} \\
\infer{[A \prederives B]\rspace \rspace R(B \rightarrow w_{n})}{[n, A]}{}
& \mbox{Between Initialization} \\
\infer{[A \prederives B]\rspace R(B\rightarrow C D)\rspace[j,D]\rspace[i,C,j]}{[i, A]}{}
& \mbox{Between Continuation}
\end{array}
$$
\caption{Fast Prefix Derivation Rules}\label{fig:fastprefix}
\end{figure}

Figure \ref{fig:fastprefix} gives a faster item-based description.
The fast version contains a new item type, $[A \prederives B]$ that
can be derived only if there is a derivation of the form $A \derives B
x_1...x_k$.  It turns out that for the fast description, we need to
compute right-most derivations rather than left-most derivations.
There are eight deduction rules, the first four of which are the same
as before, modified for right-most derivations.  The next two rules,
the Prederivation Axiom, and Prederivation Completion, compute all
items of the form $[A \prederives B]$.  The last rule, Between
Continuation is the most complicated.  The idea behind this rule is
the following.  In our previous implementation, there would be an In
Between rule followed by a string of Between Out rules.  It was
because of the Between Out rules that the Between items were
associated with looping buckets.  In the fast version, we collapse the
In Between followed by a string of many Between Outs into a single
Between Continuation rule.  Between Continuation is the same as In
Between, except with $[A \prederives B]$ prepended; the $[A
\prederives B]$ is essentially equivalent to a string of zero or more
Between Out rules.  It is prepended rather than appended because we
are now finding right-most derivations.  We use a similar trick for
Unary Between; in the original description, there could be a Unary
Between rule followed by a string of Between Out rules.  Again, we
collapse the string of Between Out rules using a single item of the
form $[A \prederives B]$.  Figure \ref{fig:fastprefixdraw} shows a
schematic tree with examples of the rules.

\begin{figure}
\psfig{figure=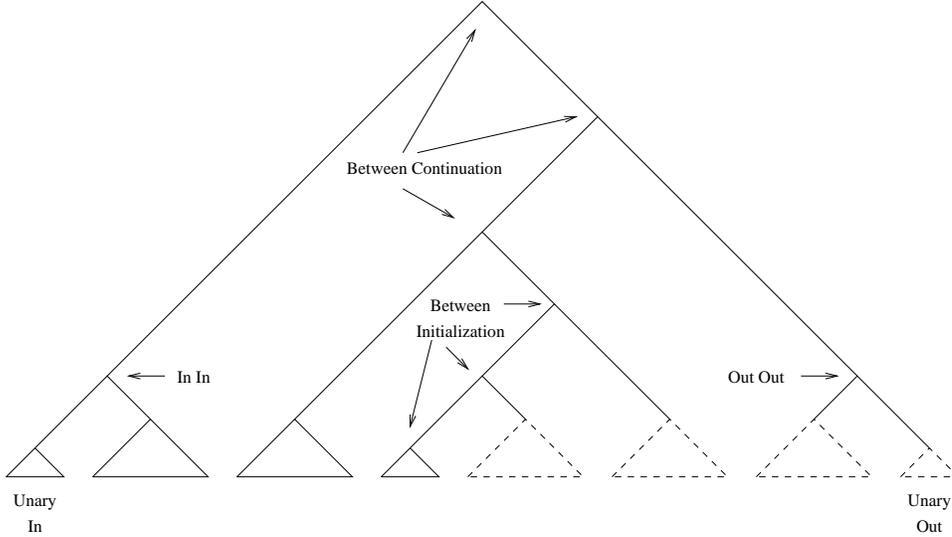,width=5in}
\caption{Fast Prederivation Illustration} \label{fig:fastprefixdraw}
\end{figure}

We can give a more formal justification for each of the last four
rules.  The Prederivation Axiom says that for all $A$, $A \derives A$.
The Prederivation Completion rule says that if $A \rightarrow BC$ and
$B \derives D x_1...x_k$ and $C \derives y_1...y_l$ then $A \derives D
x_1...x_ky_1...y_l$.  The Between Initialization rule says that
if $A \derives B x_1...x_k$ and $B \rightarrow w_n$ then $A \derives
w_nx_1... x_k$.  Finally, Between Continuation says that if $A
\derives B x_1...x_k$ and $B \rightarrow CD$ and and $C\derives
w_i... w_{j-1}$ and $D\derives w_j... w_ny_1... y_l$ then $A \derives
w_i... w_ny_1...y_lx_1...x_k$.

The careful reader can verify that there is a one-to-one
correspondence between item derivations in this system, and prefix
derivations of the input string.  Notice that the only items
associated with a looping bucket are of the form $[A \prederives B]$
and $[A]$; these can all be precomputed, independent of the input
string.  This algorithm is essentially the same as that of
\newcite{Jelinek:91b}.
% Here is the correspondence to the Jelinek algorithm:
% H<i, j> is just like our [i, H, j+1]
% Jelinek uses k while we use n, and includes an extraneous variable
% in the next item:
% H<<i, k is just like our [i, H]
% Q_L(H \Rightarrow G) is just like our [H \prederives G] except
% Jelinek doesn't include [H \prederives H] by default
% Jelinek has an added Q_L(H \Rightarrow G_1 G_2) equivalent to our
% [H \prederives A] R(A \rightarrow G_1 G_2) -- his is slightly more
% efficient because he precomputes this product.
% In the inside ring, for proper grammars, [H] is always 1, so Jelinek
% ignores this term.

% This algorithm is a bit complicated for a few reasons.
% The  main reason is because we wanted to get rid of looping buckets,
% which meant that we needed to create the extra item $[A \prederives
% B]$.  Also, in Jelinek's version, $[A]$ has value one in the inside
% semiring for all proper grammars, and so can be implicitly
% neglected.  The need for right-most derivations is a side effect of
% the $[A \prederives B]$ items which use a parent rule, and a
% derivation to the right.  So, with left-most derivations, there is no
% way to get the left child derivation between the parent and right
% child derivation.

As usual, there are advantages to the item based description, besides
its simplicity.  For instance, we can use the same item-based
description with the Viterbi semiring to find the Viterbi prefix
probabilities: the probability of the most likely prefix derivation;
with the boolean semiring to compute the valid prefix property:
whether $S \prederives w_1 ... w_n$, etc.

Prefix derivation values are potentially useful for starting
interpretation of sentences before they are completed.  For instance,
a travel agent program, on hearing ``Show me flights on April 21 so I
can'' could compute the Viterbi-derivation prefix parse so that it could begin
processing the transaction before the user was finished speaking.
Similar values, such as the prefix derivation-forest value, would
yield the set of all possible derivations that could complete the
sentence.

We note that in the inside semiring,
\begin{eqnarray*}
\lefteqn{\frac{V([i, A, j])\times Z([i, A, j])}{V([1, S])} =} \\
&&\sum_{k, x_1,..,x_k}
P(S \derives w_1...w_{i-1}A w_j...w_nx_1...x_k \derives
w_1...w_nx_1...x_k | w_1...w_n )
\end{eqnarray*}
which is the probability that symbol $A$ covers terminals $i$ to $j-1$
given all input symbols seen so far.  From an information content
point of view, this formula is the optimal one to use for
thresholding, although the resulting algorithm would be $O(n^4)$,
since the outside values, which require time $O(n^3)$ to compute,
would need to be recomputed after each new input symbol.  Thus,
although thresholding with this formula probably would not provide a
speedup, it could be useful for incremental interpretation, or for
analyzing garden path sentences.

\subsection{Beyond Context-Free}

There has been quite a bit of previous work on the intersection of
formal language theory and algebra, as described by
\newcite{Kuich:97a}, among others.  This previous work has made heavy
use of the fact that there is a strong correspondence between
algebraic equations in certain non-commutative semirings, and CFGs.
This correspondence has made it possible to manipulate algebraic
systems, rather than grammar systems, simplifying many operations.

On the other hand, there is an inherent limit to such an approach,
namely a limit to context-free systems.  It is then perhaps slightly
surprising that we can avoid these limitations, and create item-based
descriptions of parsers for weakly context-sensitive grammars, such as
Tree Adjoining Grammars (TAGs).  We avoid the limitations of previous
approaches using two techniques.  One technique is, rather than
computing parse trees for TAGs, we compute derivation trees.
Computing derivation trees for TAGs is significantly easier than
computing parse trees, since the derivation trees are context-free.
The other trick we use is that while earlier formulations created one
set of equations for each grammar, our parsing approach can be thought
of as creating a set of equations for each grammar and string length.
Because the number of equations grows with the string length, we can
recognize strings in weakly context-sensitive languages.

A further explication of this subject, including an item-based
description for a simple TAG parser is given in the appendix, in
Section \ref{sec:appbeyond}.

\subsection{Tomita Parsing}

Our goal in this section has been to show that item-based descriptions
can be used to simply describe almost all parsers of interest.  One
parsing algorithm that would seem particularly difficult to describe
is Tomita's graph-structured-stack LR parsing algorithm.  This
algorithm at first glance bears little resemblance to other parsing
algorithms, and worse, uses pointers extensively.  Since there is no
obvious way to emulate pointers in an item-based description, it would
appear that this parser has no simple item-based description.
However, \newcite{Sikkel:93a} gives an item-based description for a
Tomita-style parser for the boolean semiring, which is also more
efficient than Tomita's algorithm.  Sikkel's format is similar enough
to ours that his description can be easily converted to our format,
where it can be used for $\omega$-continuous semirings in general.

\subsection{Graham Harrison Ruzzo Parsing}
\label{sec:GHR}

\newcite{Graham:80a} describe a parser similar to Earley's, but with
several speedups that lead to significant improvements.  Essentially,
there are three improvements in the GHR parser.  First, epsilon
productions are precomputed.  Second, unary productions are
precomputed; and, finally, completion is separated into two steps,
allowing better dynamic programming.  

In Appendix \ref{sec:GHRappendix}, we give a full item-based
description of a GHR parser.  The forward values of many of the items
in our parser related to unary and epsilon productions can be computed
off-line, once per grammar.  This idea of precomputing values off-line
in a probabilistic GHR-style parser is due to \newcite{Stolcke:93a}.
Since reverse values require entire strings, the reverse values of
these items cannot be computed until the input string is known.
Because we use a single item-based description for precomputed items
and non-precomputed items, and for forward and reverse values, this
combination of off-line and on-line computation is easily and
compactly specified.

\section{Previous Work}

The previous work in this area is extensive, including work in
deductive parsing, work in statistical parsing, and work in the
combination of formal language theory and algebra.  This \chapterpaper{} can
be thought of as synthetic, combining the work in all three areas,
although in the course of synthesis, several general formulas have
been found, most notably the general formula for reverse values.  A
comprehensive examination of all three areas is beyond the scope of
this \chapterpaper{}, but we can touch on a few significant areas of each.

First, there is the work in deductive parsing.  This work in some
sense dates back to \newcite{Earley:70a}, in which the use of items
in parsers is introduced.  More recent work
\cite{Pereira:83a,Pereira:87a} demonstrates how to use deduction
engines for parsing.  Finally, both
\newcite{Shieber:93a} and \newcite{Sikkel:93a} have shown how to specify
parsers in a simple, interpretable, item-based format.  This format is
roughly the format we have used here, although there are differences
due to the fact that their work was strictly in the boolean semiring.

Work in statistical parsing has also greatly influenced this work.  We
can trace this work back to research in HMMs by Baum and his
colleagues \cite{Baum:67a,Baum:72a}.  A good introduction to this work
(and its practical application to speech recognition) was written by
\newcite{Rabiner:89a}.  In particular, the work of Baum developed the
concept of backward probabilities (in the inside semiring), as well as
many of the techniques for computing in the inside semiring.
\newcite{Viterbi:67a} developed corresponding algorithms for
computing in the Viterbi semiring.
% I haven't read that paper, but Rabiner cites it as the source for
% the algorithm.
\newcite{Baker:79b} extended the work of Baum \etal to PCFGs,
including to computation of the outside values (or reverse inside
values in our terminology.)  Baker's work is described by Lari and
Young \shortcite{Lari:90a,Lari:91a}.  Baker's work was only for PCFGs
in Chomsky Normal Form, avoiding the need to compute infinite
summations.  \newcite{Jelinek:91b} showed how to compute some of the
infinite summations in the inside semiring, those needed to compute the
prefix probabilities of PCFGs in CNF.  \newcite{Stolcke:93a} showed how to use
the same techniques to compute inside probabilities for Earley
parsing, dealing with the difficult problems of unary transitions, and
the more difficult problems of epsilon transitions.  He thus solved
all of the important problems encountered in using an item-based
parser to compute the inside and outside values (forward and reverse
inside values); he also showed how to compute the forward Viterbi
values.

The final area of work is in formal language theory and algebra.
Although it is not widely known, there has been quite a bit of work
showing how to use formal power series to elegantly derive results in
formal language theory.  In particular, the major classic results can
be derived in this framework, but with the added benefit that they
typically apply to all $\omega$-continuous semirings, or at least to
all commutative $\omega$-continuous semirings.  The most accessible
introduction to this literature we have found is by
\newcite{Kuich:97a}.  There are also books by \newcite{Salomaa:78a}
and \newcite{Kuich:86a}, as well as a book concentrating primarily on
the finite state case by \newcite{Berstel:88a}.  This work dates back
to to \newcite{Chomsky:63a}.  \newcite{Kuich:97a} gives a much more
complete bibliography than we give here.

One piece of work deserves special mention.  \newcite{Teitelbaum:73a}
showed that any semiring could be used in the CKY algorithm.  He
further showed that a subset of complete semirings could be used in
value-preserving transformations to CNF.  Thus, he laid the foundation
for much of the work that followed.

In summary, this \chapterpaper{} synthesizes work from several different related
fields, including deductive parsing, statistical parsing, and formal
language theory; we emulate and expand on the earlier synthesis of
Teitelbaum.  The synthesis here is powerful: by generalizing and
integrating many results, we make the computation of a much wider
variety of values possible.

\subsection{Recent similar work}
There has also been recent similar work by Tendeau
\shortcite{Tendeau:97a,Tendeau:97b}. \newcite{Tendeau:97a} gives an
Earley-like algorithm that can be adapted to work with complete
semirings satisfying certain conditions.  Unlike our version of
Earley's algorithm, Tendeau's version requires time $O(n^{L+1})$ where
$L$ is
% page 204, Tendeau himself gives this run time
the length of the longest right hand side, as opposed to $O(n^3)$ for
the classic version, and for our description.  There is also not much
detail about how the algorithm handles looping productions.

\newcite{Tendeau:97b} also shows how to compute values in an abstract
semiring with a CKY style algorithm, in a manner similar to
\newcite{Teitelbaum:73a}.  More importantly, he gives a generic
description for dynamic programming algorithms.  His description is
very similar to our item-based descriptions with two caveats.  First,
he includes a single rule term, on the left, rather than a set of rule
values intermixed with item values.  For commutative semirings this
limitation is fine, but for non-commutative semirings, it may lead to
inelegancies.  Second, he does not have an equivalent to our side
conditions.  This means that algorithms such as Earley's algorithm
which rely on side conditions for efficiency cannot be described in
this formalism in a way that captures those efficiency considerations.

Tendeau \shortcite{Tendeau:97a,Tendeau:97b} introduces a parse forest
semiring, similar to our derivation forest semiring, in that it
encodes a parse forest succinctly.  Tendeau's parse forest has
an advantage over ours in that it is commutative.  However, it has two
disadvantages.  First, it is partially because of the encoding of this
parse forest that Tendeau's version of Earley's algorithm has its poor
time complexity.  Second, to implement this semiring, Tendeau's
version of rule value functions take as their input not only a
nonterminal, but also the span that it covers; this is somewhat less
elegant than our version.  \newcite{Tendeau:97a} also shows that Definite
Clause Grammars (DCGs) can be described as semirings, with some
caveats, including the same problems as the parse forest semiring.

%\newcite{Tendeau:97a} also shows that Definite Clause Grammars (DCGs)
%can be described as semirings, with some caveats, and is the
%inspiration for our description of the DCG semiring.  The problems
%with Tendeau's DCG semiring are similar to the problems with his parse
%forest semiring, and we attempt to rectify them in similar ways.

\section{Conclusion}

In this \chapterpaper{}, we have shown that a simple item-based description
format can be used to describe a very wide variety of parsers.  These
parsers include the CKY algorithm, Earley's algorithm, prefix
probability computation, a TAG parsing algorithm, Graham, Harrison,
Ruzzo parsing, and HMM computations.  We have shown that this
description format makes it easy to find parsers that compute values
in any $\omega$-continuous semiring.  The same description can be used
to find reverse values in commutative $\omega$-continuous semirings,
and in many non-commutative ones as well.  We have also shown that
this description format can be used to describe grammar
transformations, including transformations to CNF and GNF, which
preserve values in any commutative $\omega$-continuous semiring.

While theoretical in nature, this \chapterpaper{} is of some practical value.
There are three reasons the results of this \chapterpaper{} would be used in
practice: first, these techniques make computation of the outside
values simple and mechanical; second, these techniques make it easy to
show that a parser will work in any $\omega$-continuous semiring; and
third, these techniques isolate computation of infinite sums in a
given semiring from the parser specification process.

Probably the way in which these results will be used most is to find
formulas for outside values.  For parsers such as CKY parsers, finding
outside formulas is not particularly burdensome, but for complicated
parsers such as TAG parsers, Graham, Harrison, Ruzzo parsers, and
others, it can require a fair amount of thought to find these
equations through conventional reasoning.  With these techniques, the
formulas can be found in a simple mechanical way.

The second advantage comes from clarifying the conditions under which
a parser can be converted from computing values in the boolean
semiring (a recognizer) to computing values in any $\omega$-continuous
semiring.  We should note that because in the boolean semiring,
infinite summations can be computed trivially ($\mi{TRUE}$ if any
element is $\mi{TRUE}$) and because repeatedly adding a term does not
change results, it is not uncommon for parsers that work in the
boolean semiring to require significant modification for other
semirings.  For parsers like CKY parsers, verifying that the parser
will work in any semiring is trivial, but for other parsers the
conditions are more complex.  With the techniques in this
\chapterpaper{}, all that is necessary is to show that there is a
one-to-one correspondence between item derivations and grammar
derivations.  Once that correspondence has been shown, Theorem
\ref{theorem:itemgrammar} states that any $\omega$-continuous semiring
can be used.

The third use of this \chapterpaper{} is to separate the computation of infinite
sums from the main parsing process.  Infinite sums can come from
several different phenomena, such as loops from productions like $A
\rightarrow A$; productions involving $\epsilon$; and sometimes left
recursion.  In traditional procedural specifications, the solution to
these difficult problems is intermixed with the parser specification,
and makes the parser specific to semirings using the same techniques
for solving the summations.

It is important to notice that the algorithms for solving these
infinite summations vary fairly widely, depending on the semiring.  On
the one hand, boolean infinite summations are nearly trivial to
compute.  For other semirings, such as the counting semiring, or
derivation forest semiring, more complicated computations are
required, including the detection of loops.  Finally, for the inside
semiring, in most cases only approximate techniques can be used,
although in some cases, matrix inversion can be used.  Thus, the
actual parsing algorithm, if specified procedurally, can vary quite a
bit depending on the semiring.

On the other hand, using our techniques makes infinite sums easier to
deal with in two ways.  First, these difficult problems are separated
out, relegated conceptually to the parser interpreter, where they can
be ignored by the constructor of parsing algorithms.  Second, because
they are separated out, they can be solved once, rather than again and
again.  Both of these advantages make it significantly easier to
construct parsers.

In summary, the techniques of this \chapterpaper{} will make it easier to
compute outside values, easier to construct parsers that work for any
$\omega$-continuous semiring, and easier to compute infinite sums in
those semirings.  In 1973, Teitelbaum wrote:
\begin{quote}
We have pointed out the relevance of the theory of algebraic power
series in non-commuting variables in order to minimize further
piecemeal rediscovery.
\end{quote}
Many of the techniques needed to parse in specific semirings continue
to be rediscovered, and outside formulas are derived without
observation of the basic formula given here.  We hope this \chapterpaper{} will
bring about Teitelbaum's wish.

{\beginappendix

\appendixsection{Additional Proofs}
\label{sec:additionalproofs}

In this appendix, we prove theorems given earlier.  It will be helpful
to refer back to the original statements of the theorems for context.

\begin{myoldtheorem}{theorem:inloop}
For $x$ an item in a looping bucket $B$, and for $g \geq 1$,
%\label{theorem:inloop}
%
$$
V_{\leq g}(x, B) =
\bigoplus_{a_1...a_k \;\mbox{\scriptsize s.t.}\; \smallinfer{a_1... a_k}{x}{}}
\;\;\bigotimes_{i=1..k} 
\left\{
\begin{array}{ll}
{V(a_i)}{\mbox{ \em if }a_i\notin B} \\
{V_{\leq g\!-\!1}(a_i, B)}{\mbox{ \em if }a_i\in B}
\end{array}
\right.
%\label{eqn:inloop}
$$
\end{myoldtheorem}

{\em Proof \hspace{3em}} Observe that for any item $x$ in a bucket preceding $B$,
$\mi{inner}_{\leq g}(x, B) = \mi{inner}(x)$.  Then
\begin{eqnarray}
V_{\leq g}(x, B) &=&
\bigoplus_{D \in \smi{inner}_{\leq g}(x,B)} V(x) \nonumber \\
&=& \bigoplus_{a_1...a_k \;\mbox{\scriptsize s.t.}\;
\smallinfer{a_1... a_k}{x}{}} 
\bigoplus_{\longsum{D_{a_1}\in\smi{inner}_{\leq g\!-\!1}(a_1,B),...,}{D_{a_k}\in\smi{inner}_{\leq g\!-\!1}(a_k,B)}}
V(\langle x: (D_{a_1},...,D_{a_k} \rangle) \nonumber \\
& = & 
\bigoplus_{a_1...a_k \;\mbox{\scriptsize s.t.}\; \smallinfer{a_1... a_k}{x}{}}
\;\;\bigoplus_{\longsum{D_{a_1}\in\smi{inner}_{\leq g\!-\!1}(a_1,B),...,}{D_{a_k}\in\smi{inner}_{\leq g\!-\!1}(a_k,B)}}
\;\;\bigotimes_{i=1..k} V(D_{a_i}) \nonumber \\
& = &
\bigoplus_{a_1...a_k \;\mbox{\scriptsize s.t.}\; \smallinfer{a_1... a_k}{x}{}}
\;\;\bigotimes_{i=1..k} \bigoplus_{D_{a_i}\in\smi{inner}_{\leq g\!-\!1}(a_i,B)} V(D_{a_i}) \nonumber \\
& = &
\bigoplus_{a_1...a_k \;\mbox{\scriptsize s.t.}\; \smallinfer{a_1... a_k}{x}{}}
\;\;\bigotimes_{i=1..k} 
V_{\leq g\!-\!1}(a_i, B) \label{eqn:lastinner}
\end{eqnarray}
Now, for elements $x \notin B$, we can substitute Equation
\ref{eqn:notinBinner} into Equation \ref{eqn:lastinner} to yield
$$
V_{\leq g}(x, B) =
\bigoplus_{a_1...a_k \;\mbox{\scriptsize s.t.}\; \smallinfer{a_1... a_k}{x}{}}
\;\;\bigotimes_{i=1..k} 
\left\{
\begin{array}{ll}
{V(a_i)}{\mbox{ \em if }a_i\notin B} \\
{V_{\leq g\!-\!1}(a_i, B)}{\mbox{ \em if }a_i\in B}
\end{array}
\right.
$$
completing the proof.  $\Box$

\begin{myoldtheorem}{theorem:reverseloop}
For items $x \in B$ and $g \geq 1$,
%\label{theorem:reverseloop}
%
$$
Z_{\leq g}(x, B) = 
\bigoplus_{{j, a_1...a_k, b}\;\mbox{\scriptsize s.t.}\;
\smallinfer{a_1... a_k}{b}{}\wedge x=a_j}
\;\; \left( \bigotimes_{i =1 ,\ldotsj, k} V(a_i) \right)
\;\;\otimes
\left\{\begin{array}{ll}
{Z_{\leq g\!-\!1}(b, B)}&{\mbox{ \em if } b\in B} \\
{Z(b)}&{\mbox{ \em if } b\notin B}
\end{array} \right.
%\label{eqn:reverseloop}
$$
\end{myoldtheorem}

{\em Proof \hspace{3em}} Define $\mi{MakeOuter}(j, D_{a_1},\ldotsj, D_{a_k}, D_b)$
to be a function that puts together the specified trees to form an
outer tree for $a_j$.  Then,
\begin{eqnarray}
\lefteqn{Z_{\leq g}(x, B)}
\nonumber \\ 
&=&\bigoplus_{D\in\smi{outer}_{\leq g\!-\!1}(x, B)} Z(D) 
\nonumber \\
&=&\bigoplus_{{j, a_1...a_k, b}\;\mbox{\scriptsize s.t.}\;
\smallinfer{a_1... a_k}{b}{\wedge x=a_j}}
\;\;\bigoplus_{\longersum
{D_{a_1}\in\smi{inner}(a_1),\ldotsj,}
{\sumindent D_{a_k}\in\smi{inner}(a_k),}
{D_{b}\in\smi{outer}_{\leq g\!-\!1}(b, B)}}
Z(\mi{MakeOuter}(j, D_{a_1},\ldotsj, D_{a_k}, D_b)) 
\nonumber \\
&=&\bigoplus_{{j, a_1...a_k, b}\;\mbox{\scriptsize s.t.}\;
\smallinfer{a_1... a_k}{b}{\wedge x=a_j}}
\;\;\bigoplus_{\longersum
{D_{a_1}\in\smi{inner}(a_1),\ldotsj,}
{\sumindent D_{a_k}\in\smi{inner}(a_k),}
{D_{b}\in\smi{outer}_{\leq g\!-\!1}(b, B)}}
\;\; Z(D_b) \otimes \bigotimes_{i = 1,\ldotsj, k} V(D_{a_i}) 
\nonumber \\
&=&\bigoplus_{{j, a_1...a_k, b}\;\mbox{\scriptsize s.t.} % THIS IS TOO LONG
\smallinfer{a_1... a_k}{b}{\wedge x=a_j}}
\bigoplus_{\longsum
{D_{a_1}\in\smi{inner}(a_1),\ldotsj,}
{D_{a_k}\in\smi{inner}(a_k)}}
\; \bigoplus_{D_{b}\in\smi{outer}_{\leq g\!-\!1}(b, B)}
\!\!\! Z(D_b) \otimes \! \bigotimes_{i = 1,\ldotsj, k} \! \! \! V(D_{a_i})
\nonumber \\
&=&\bigoplus_{{j, a_1...a_k, b}\;\mbox{\scriptsize s.t.} % THIS IS TOO LONG...
\smallinfer{a_1... a_k}{b}{\wedge x=a_j}}\!
\left( \! \bigoplus_{\longsum
{D_{a_1}\in\smi{inner}(a_1),\ldotsj,}
{D_{a_k}\in\smi{inner}(a_k)}}
\bigotimes_{i = 1,\ldotsj, k} \!\! V(D_{a_i}) \! \right) \!
\otimes
\!\!\!\bigoplus_{D_{b}\in\smi{outer}_{\leq g\!-\!1}(b, B)} \!\!\!\!\! Z(D_b) 
\nonumber \\
&=&\bigoplus_{{j, a_1...a_k, b}\;\mbox{\scriptsize s.t.}\;
\smallinfer{a_1... a_k}{b}{\wedge x=a_j}}
\;\; \left( \bigotimes_{i =1 ,\ldotsj, k} \;\;\bigoplus_{D_{a_i}\in\smi{inner}(a_i,B)} V(D_{a_i})\right)
\;\; \otimes
\;\; Z_{\leq g\!-\!1}(b, B) 
 \nonumber \\
&=&\bigoplus_{{j, a_1...a_k, b}\;\mbox{\scriptsize s.t.}\;
\smallinfer{a_1... a_k}{b}{\wedge x=a_j}}
\;\; \left( \bigotimes_{i =1 ,\ldotsj, k} V(a_i) \right)
\;\;\otimes
\;\; Z_{\leq g\!-\!1}(b, B) \label{eqn:lastouter}
\end{eqnarray}
Now, substituting Equation \ref{eqn:notinBouter} into Equation
\ref{eqn:lastouter}, we get
$$
Z_{\leq g}(x, B)  =
\bigoplus_{{j, a_1...a_k, b}\;\mbox{\scriptsize s.t.}\;
\smallinfer{a_1... a_k}{b}{\wedge x=a_j}}
\;\; \left( \bigotimes_{i =1 ,\ldotsj, k} V(a_i) \right)
\;\;\otimes
\left\{\begin{array}{ll}
{Z_{\leq g\!-\!1}(b, B)}&{\mbox{ \em if } b\in B} \\
{Z(b)}&{\mbox{ \em if } b\notin B}
\end{array}
\right.
$$
$\Box$

\begin{myoldtheorem}{theorem:correctinterpret}
The forward semiring parser interpreter correctly computes the value
of all items.
\end{myoldtheorem}
{\em Sketch of proof \hspace{3em}} We show that every item in each
bucket has its value correctly computed.  The proof is by induction on
the buckets.  The base case is the first bucket.  Since items in each
bucket depend only on the values of items in preceding buckets, items
in the first bucket depend on no previous buckets.  Now, if the first
bucket is a looping bucket, the interpreter uses an implementation of
Equation \ref{eqn:inloop}, previously shown correct in Theorem
\ref{theorem:inloop}.  If the first bucket is a non-looping bucket,
the interpreter uses an implementation of Equation
\ref{eqn:itemvalue}, previously shown correct in Theorem
\ref{eqn:itemvalue}.  Since these equations refer to items in the
first bucket, and such items do not depend on values outside the first
bucket, other than rule values, the first bucket is correctly
computed.

For the inductive step, consider the current bucket of the loop.
Assume that the values of all items in all previous buckets have been
correctly computed.  Then, depending on whether the current bucket is
a looping bucket or not, either an implementation of Equation
\ref{eqn:inloop} or \ref{eqn:itemvalue} is used.  These equations only
depend on other values in the current bucket and on values in
previously correctly computed buckets, so the values in the current
bucket are correctly computed.

Thus, by induction, the values of all items in all buckets are
correctly computed.  $\Box$

We now discuss the renormalization parser of Figure
\ref{fig:renormalize}.  We claim that using the equation
\begin{equation}
P_{\mi{new}}(A \rightarrow \alpha) = \frac{V([A \rightarrow \alpha\bullet {} ])\times Z([A \rightarrow \alpha\bullet {} ])}
{\sum_{\beta} V([A \rightarrow \beta\bullet {} ])\times Z([A
\rightarrow \beta\bullet {} ])}
\label{eqn:renormalize}
\end{equation}
yields new probabilities that produce the same trees with the same
probabilities as the original grammar and that clearly, these
probabilities sum to 1 for each nonterminal (which they may not have
done in the original grammar).  The intuition behind this statement is
that Equation \ref{eqn:renormalize} is simply the inside-outside
reestimation formula.  We have applied it here to the probability
distribution of all trees produced by the original grammar, but this
should be very similar to applying the inside-outside formula to a
very large (approaching infinite) sample of trees from that
distribution.  Since the inside-outside probabilities approach a local
optimum, and since a grammar producing the same trees with the same
probabilities will be optimal, we expect to achieve that grammar.

We will now give a formal proof that the resulting trees have the same
probabilities as the original trees, subject to one caveat.  In
particular, we do not know how to show that the resulting probability
distribution is tight, that is, that the sum of the probabilities of
all of the trees equals 1.  (Not all grammars are tight.  For
instance, $S \rightarrow SS (.9), S \rightarrow a (.1)$ is not tight.)
\newcite{Chi:98a} showed that CNF grammars produced by the
inside-outside reestimation formula are tight, but it is not clear
whether is has ever been shown for the case of grammars with unary
productions.

\begin{mytheorem}
Assuming that the inside-outside reestimation formula produces tight
grammars, and that the original grammar is tight, the grammar produced
using the renormalization parser of Figure
\label{theorem:renormalize}
\ref{fig:renormalize}, and Equation \ref{eqn:renormalize} produces a
grammar that produces an identical probability distribution over trees
as the original grammar, but with each rule having a probability
between 0 and 1.
\end{mytheorem}
{\em Proof \hspace{3em}} We will show by induction that for each
left-most derivation $D$ of length at most $k$, the value of all
derivations starting with $D$ is the same in the new grammar as the
old.

The base case follows from our assumption that both the new grammar
and the old grammar are tight, since the sum of all derivations
starting with the empty derivation (k=0) is the sum of the probability
of all trees, or, by assumption, 1 in both cases.

%if the old grammar were not proper, the new one might be proper (and tight):
%S->SS (.9), S->a (.1)
%V(S) = .11111
%New S->SS: .1, S->a, .9, which is proper.

Next, for the inductive step, assume the theorem for $k-1$.  Consider
a left-most derivation $D$ of length $k$.  We can split $D$ into a
left-most derivation $E$ of length $k-1$, followed by the single step
$A \rightarrow \alpha$: $S \Ederives w_1...w_jA\gamma \Rightarrow
w_1...w_j\alpha\gamma$.  We show that the sum of the values of all
derivations that begin this way is the same in both old and new
grammars.  Since the new grammar is, by assumption, tight, and has
rule probabilities that sum to one, the probability of all derivations
starting with $D$ is just $P(D)$:
\begin{eqnarray}
P_{\mi{new}}( S \Dderives w_1...w_j \alpha\gamma) & = &
P_{\mi{new}}( S \Ederives w_1...w_jA \gamma \Rightarrow w_1...w_j \alpha\gamma)
\nonumber \\
&= & P_{\mi{new}}( S \Ederives w_1...w_jA \gamma) \times P_{\mi{new}}(A\rightarrow \alpha) 
\nonumber \\
&= & P_{\mi{new}}( S \Ederives w_1...w_jA \gamma) \times \frac{V([A
\rightarrow \alpha \bullet])}{\sum_{\beta}V([A \rightarrow \beta
\bullet])} \label{eqn:endnewvalue}
\end{eqnarray}

Now, we consider values in the original grammar.  Let the value of
$A$, $V(A)$, be $\sum_{\beta} V([A \rightarrow\beta\bullet{}])$.  Let the nonterminal
symbols in $\beta$ be denoted by $\beta_1...\beta_{|\beta |}$ and let
$V(\beta) = \prod_{i = 1}^{|\beta |} V(\beta_i)$.  Notice that given a
derivation of the form $S \Dderives w_1...w_j \alpha \gamma$, the sum
of the values of all derivations starting with $D$ is the value of $D$
times the product of the value of all of the nonterminals in
$\alpha\gamma$:
\begin{eqnarray}
\lefteqn{V( S \Dderives w_1...w_j \alpha\gamma) \times V(\alpha)
\times V(\gamma)} \nonumber \\ 
& = & V( S \Ederives w_1...w_jA \gamma \Rightarrow w_1...w_j\alpha\gamma) \times V(\alpha)
\times V(\gamma)
\nonumber \\
& = &
V( S \Ederives w_1...w_jA \gamma)\times P_{\mi{orig}}(A \rightarrow\alpha) \times V(\alpha)
\times V(\gamma)
\nonumber \\
& = &
V( S \Ederives w_1...w_jA \gamma) \times V([A \rightarrow \alpha\bullet{}])
\times V(\gamma) 
\label{eqn:middlerenormalize} 
\end{eqnarray}
Next, by the inductive assumption, 
$$
V( S \Ederives w_1...w_jA \gamma) \times V(A) \times V(\gamma) = 
P_{\mi{new}}( S \Ederives w_1...w_jA \gamma)
$$
Dividing both sides by $V(A)$,
$$
V( S \Ederives w_1...w_jA \gamma)  \times V(\gamma) = 
\frac{
P_{\mi{new}}( S \Ederives w_1...w_jA \gamma)
}
{V(A)}
$$
Substituting into Equation \ref{eqn:middlerenormalize}, we get
\begin{eqnarray}
V( S \Dderives w_1...w_j \alpha\gamma) \times V(\alpha)
\times V(\gamma) & = & 
P_{\mi{new}}( S \Ederives w_1...w_jA \gamma)
\times 
\frac{V([A \rightarrow \alpha\bullet{}])}{V(A)} \nonumber \\
& =&
P_{\mi{new}}( S \Ederives w_1...w_jA \gamma)
\times 
\frac
{V([A \rightarrow \alpha\bullet{}])} 
{\sum_{\beta}V([A \rightarrow \beta \bullet{}])}
\nonumber 
\end{eqnarray}
which equals Expression \ref{eqn:endnewvalue}, completing the inductive
step.  $\Box$

\subsection{Viterbi-n-best is a semiring}
\label{sec:vitnbestproof}

In this section we show that the Viterbi-n-best semiring, as described
in Section \ref{sec:vitnbest}, has all the required properties of a
semiring, and is $\omega$-continuous.  Recall that the Viterbi-n-best
semiring is a homomorphism from the Viterbi-all semiring.  We first
show that the Viterbi-all semiring is $\omega$-continuous.

We begin by arguing that the Viterbi-all semiring has all of the
properties of an $\omega$-continuous semiring, saving the most
complicated property, the distributive property, for last.  It should
be clear that $\cup$ and $\star$ are associative.  To show that the
semiring is $\omega$-continuous, we first note that the proper
convergence of infinite sums follows directly from the properties of
union, as does the associativity of infinite sums (unions).  The only
complicated property is that $\star$ distributes over $\cup$, even in
the infinite case.  We sketch this proof quickly.  We need to show
that for any set of $Y_i$, and any $I$,
$$
X \star (\bigcup_{i \in I} Y_i) = \bigcup_{i \in I} (X \star Y_i)
$$
Recall the definition of $\star$:
$$
X\star Y =\{\alist{vw, d\cdot e}\vert \alist{v, d}\in X \wedge
\alist{w, e}\in Y\}
$$
First, consider an element $\alist{vw, d\cdot e} \in X \star (\bigcup_{i
\in I} Y_i)$.  It must be the case that $v \in X$ and $w \in Y_i$ for
some $v,w,i$.  So then, $\alist{vw, d\cdot e} \in \bigcup_{i \in I} X
\star Y_i$.  A reverse argument also holds, proving equality.
Technically, we also need to show distributivity in the opposite
direction, that $(\bigcup_{i \in I} Y_i) \star X = \bigcup_{i \in I} (Y_i
\star X)$, but this follows from an exactly analogous argument.

Now, we will show that our definitions of operations in the
Viterbi-n-best semiring are well defined, and then show that $\topn$
really does define a homomorphism.  Recall our definitions: $\maxvitn
A, B = C$ if and only if there is some $X, Y$ in the Viterbi-all
semiring such that such that $\topn(X) = A$, $\topn(Y)= B$, and
$\topn(X \cup Y) = C$.  We need to show that this defines a one to one
relationship -- that there is exactly one $C$ which satisfies this
relationship for each $A$ and $B$.  The existence of at least one such
$C$ follows from the definition of the Viterbi-n-best semiring.  There
must be at least one $X, Y$ such that $\topn(X)=A$ and $\topn(Y)=B$
and thus at least one $C = \topn(X \cup Y)$ .

Now, consider any $X, X'$ such that $\topn(X)=\topn(X')$ and $Y, Y'$
such that $\topn(Y) = \topn(Y')$.  To show that $C$ is unique, we need
to show that $\topn(X \cup Y) = \topn(X' \cup Y')$.  We will say that
an element of a set $Z$ is {\em simple} if there are at most a finite
number of larger elements in $Z$, and {\em complex} otherwise.  We
first show equality of the simple elements in the two sets.  Consider
a simple element $z$ in $\topn(X \cup Y)$.  There are at most $n-1$
larger elements in $X \cup Y$.  $z$ must have been an element of
either $X$ or $Y$ or both.  Assume without loss of generality that it
was in $X$.  Then, since there are at most $n-1$ larger elements in $X
\cup Y$, there are at most $n-1$ larger elements in $X$, and thus $z
\in \topn(X)$.  Since $\topn(X) = \topn(X'),$ $z \in \topn(X')$, and therefore $z \in X'$, and
$z \in X' \cup Y'$.  By analogous reasoning, for any simple element
$z' \in \topn(X' \cup Y')$, $z' \in X \cup Y$.  Now, it must be the
case that $z \in \topn(X' \cup Y')$, since each element of $\topn(X'
\cup Y')$ is in $X \cup Y$ and if there were $n$ elements larger than
$z$ in $X' \cup Y'$, then each of these elements would be in $\topn(X'
\cup Y')$ and thus in $X \cup Y$, which would prevent $z$ from being
in $\topn(X \cup Y)$.  By analogous reasoning, every simple element
$z' \in \topn(X' \cup Y')$ is also in $\topn(X \cup Y)$.  Thus, the
simple elements of the two sets are equal.

Now, let us consider the infinite elements, $\alist{v, \infty}$, of
$\topn(X \cup Y)$.  There are several cases to consider, involving the
possibilities where zero, one, or both of $X, Y$ have an infinite
element.  We only consider the most complicated case, in which both
do.  Notice that if there are $n$ or more simple elements in $X$ and
$Y$, then $\topn(X \cup Y)$ will not have an infinite element, so we
need only consider the case where there are at most $n-1$ simple
elements in $X\cup Y$.  Then the infinite element of $\topn(X \cup Y)$
will just be the supremum of all of the complex elements of $X \cup
Y$, which will equal the supremum of the complex elements of $X$ union
the complex elements of $Y$, which will equal the maximum of the
supremums of the complex elements of $X$ and the complex elements of
$Y$, which will equal the maximum of the infinite elements of
$\topn(X)$ and $\topn(Y)$.  The same reasoning applies to any $X',
Y'$, and, since if $\topn(X) = \topn(X')$, and $\topn(Y) = \topn(Y')$,
their infinite elements must be the same, the maximum of their
infinite elements must be the same, and thus the infinite elements of
$\topn(X \cup Y)$ and $\topn(X' \cup Y')$ must be the same.  Since we
have already shown equality of the finite elements, this shows
equality overall, and thus shows the uniqueness of $C$ in our
definition of additive operation $\maxvitn$.  Therefore, $\maxvitn$ is
well defined.

Next, we must go through very similar reasoning for $\star$, showing
that the multiplicative operator, $\timesvitn$, is well defined.
Recall that we defined $A \timesvitn B = C$ if and only if there is
some $X, Y$ in the Viterbi-all semiring such that such that $\topn(X)
= A$, $\topn(Y)= B$, and $\topn(X \star Y) = C$.  There is at least
one such $C$ (equal to $\topn(X \star Y)$) and now we need to prove
its uniqueness.  If $x = \alist{v, d}$ and $y = \alist{w, e}$, then we
will write $xy$ to indicate the product $\alist{vw, de}$.  We examine
first the simple elements of $X$ and $Y$.  Consider a simple element
$z \in \topn(X\star Y)$.  $z = xy,$ for some $x \in X, y \in Y$.  Now,
$x \in \topn(X)$: otherwise, the $n$ larger elements of $X$ multiplied
by $y$ would result in $n$ larger elements in $X \star Y$, meaning
that $z \notin \topn(X \star Y)$. Similarly, $y \in \topn(Y)$.  Thus,
$z \in X' \star Y'$.  Similarly, each $z' \in \topn(X' \star Y')$ is
in $X \star Y$.  Following the same reasoning as before, the simple
elements of the two sets are equal.

Now, consider the infinite elements.  Again, we will consider only the
case where both $\topn(X)$ and $\topn(Y)$ have an infinite element.
If there is an infinite element in $\topn(X \star Y)$, it must be
formed in one of three ways: by taking the top element of $x$ and
multiplying it by the elements approaching a supremum in $Y$; by
taking the top element of $y$ multiplied by the elements approaching
a supremum in $X$; or if there is no top element in $X$ or $Y$, by
taking the elements approaching the supremum in $X$ and multiplying
them by the elements approaching the supremum in $Y$.  Thus, the
infinite element of $\topn(X \star Y)$ is the maximum of these three
quantities, which will be same for any $X'$ such that $\topn(X') =
\topn(X)$ and $Y'$ such that $\topn(Y') = \topn(Y)$.

Now that we have shown that both $\maxvitn$ and $\timesvitn$ are
well-defined, the associative and distributive properties in the
Viterbi-n-best semiring follow automatically.  We simply map the
expression on the left side backwards from the Viterbi-n-best semiring
into expressions in the Viterbi-all semiring; map the expression on
the right side into the Viterbi-all semiring; and then use the
appropriate property in the Viterbi-all semiring to show equality.  We
show how to use this technique for the associativity of $\maxvitn$; the
other properties follow analogously.  We simply note that since our
homomorphism is onto, for any A, B, C in the Viterbi-n-best semiring,
there must be some $X, Y, Z$ in the Viterbi-all semiring such that $A
= \topn(X)$, $B = \topn(Y)$, and $C = \topn(Z)$.  Consider
$$
\maxvitn (\maxvitn A, B), C
$$
There must be some $D = \;\maxvitn A, B = \topn(X \cup Y)$.  Then
$\maxvitn (\maxvitn A, B), C = \;\maxvitn D, C = \topn((X \cup Y) \cup
Z)$.  By analogous reasoning, $\maxvitn A, (\maxvitn B, C) = \topn(X
\cup (Y \cup Z))$, and thus by the associativity of $\cup$, the
associativity of $\maxvitn$ is proved.  Using the same reasoning, we
can show associativity of $\timesvitn$ and distributivity.

Now, we must show that the Viterbi-n-best semiring is complete (has the
associative and distributive property for infinite sums) and is
$\omega$-continuous.  To do this, we need to show that our
homomorphism works even for infinite sums.  Then, by the same mapping
argument we just made, we can show associativity, distributivity, and
$\omega$-continuity for infinite sums as well.

Consider $X_1...X_\infty$ and $X'...X'_\infty$ such that $\topn(X_i) =
\topn(X'_i)$.  We need to show that $\topn(\bigcup_i X_i) = \topn(\bigcup_i
X'_i)$.  Let $X = \bigcup_i X_i$ and let $X' = \bigcup_i X_i'$.  Consider a
simple element $z$ in $\topn(X)$.  Our reasoning will be nearly
identical to the finite case, and is included here for completeness.
There are at most $n-1$ larger elements in $X$.  For some $i$, $z$
must have been an element of $X_i$.  Then, since there are at most
$n-1$ larger elements in $X$, there are at most $n-1$ larger elements
in $\topn(X_i)$ and thus $z \in \topn(X_i)$.  Thus, $z \in
\topn(X_i')$, and therefore $z \in X_i'$ and $z \in X'$.  By analogous
reasoning, for any simple element $z' \in \topn(X')$, $z' \in X$.
Now, it must be the case that $z \in \topn(X')$, since each element of
$\topn(X')$ is in $X$ and if there were $n$ elements larger than $z$
in $X'$, then each of these elements would be in $\topn(X')$ and thus
in $X$, which would prevent $z$ from being in $\topn(X)$.  By
analogous reasoning, every simple element $z' \in \topn(X')$ is also in
$\topn(X)$.  Thus, the simple elements of the two sets are equal.

Now consider the case where there is an infinite element in
$\topn(X)$; this case is somewhat more complex.  Let
$$
w = \sup_{v\vert\alist{v, d}\in X-\smi{simpletopn}(X)}
$$
and
$$
w' = \sup_{v\vert\alist{v, d}\in X'-\smi{simpletopn}(X')}
$$
Take an infinite sequence of elements in $X-\mi{simpletopn}(X)$
approaching $w$.  Consider an element $x$ in this sequence.  $x$ must
have come from some $X_i-\mi{simpletopn}(X)$.  We will show that there
is an element of $X_i'-\mi{simpletopn}(X')$ which is at least as close
to $w$ as $x$ is, and thus that $w'$ is at least as large as $w$.

Here are the cases to consider.
\begin{itemize}
\item[a] $X_i$ has at most $n-1$ elements.  
\item[b] $X_i$ has at most $n-1$ simple elements and an infinite number
of complex elements.
\item[c] $X_i$ has at least $n$ simple elements.
\end{itemize}
In case a, since $\topn(X_i)=\topn(X_i')$, we deduce $X_i=X_i'$ and
thus that $x \in X_i'$.  In case b, either $x$ is a simple element, in
which case this reduces to case a, or $x$ is a complex element.  Now,
since $\topn(X_i) = \topn(X_i')$, the supremum of the complex elements
is the same in both cases, and thus there is an element at least as
large as $x$ in $X_i'$.  In case c, we notice that not all $n$ simple
elements can be in $\mi{simpletopn}(X)$, since otherwise there would
be $n$ simple elements in $X$, and we would not be concerned with the
supremum of the complex elements.  In particular, at least the $n$th
largest element cannot be in $\mi{simpletopn}(X)$.  Now, if $x$ is one
of the $n$ largest elements of $X_i$, then $x \in X_i'$, since
$\topn(X_i) = \topn(X_i')$.  And if $x$ is not as large as the $n$th
largest element, then the $n$th largest element of $X_i'$ is an
element of $X_i'$ which is larger than $x$ and in $X_i' -
\mi{simpletopn}(X')$.

This shows that even in the case of infinite sums, the homomorphism
works. Following the same reasoning as before, it should be clear how
to prove associativity and distributivity for infinite sums, and
$\omega$-continuity.

\appendixsection{Additional Examples}
\label{sec:additionalexamples}

\subsection{Graham, Harrison, and Ruzzo (GHR) Parsing}
\label{sec:GHRappendix}

In this section, we give a detailed description of a parser similar to
that of \newcite{Graham:80a}, as introduced in Section \ref{sec:GHR}.
The GHR parser is in many ways similar to Earley's parser, but with
several improvements, including that epsilon and unary chains are both
precomputed, and that completion is separated into two steps, allowing
better dynamic programming.

\newcite[p. 122]{Sikkel:93a} gives a deduction system for GHR
parsing.  Note, however, that Sikkel's parser appears to have more
than one derivation for a given parse; in particular, chains of unary
productions can be broken down in several ways.  In a boolean parser,
such as Sikkel's, this repetition leads to no problems, but in a general,
semiring parser, this is not acceptable. Another issue is derivation
rules using precomputed chains, such as a rule using a condition like
$A\derives\epsilon$.  In a boolean parser, we can simply have a side
condition, $A \derives\epsilon$.  However, in a semiring parser, we
will need to multiply in the value of the derivation, as well.  Thus,
we will need to explicitly compute items such as $[A\derives\epsilon]
$, recording the value of the derivations.  We will need similar items
for chain rules.

%A problem, from our point of view, with Sikkel's GHR: (p. 122)
%We require that each derivation be possible only once.
%But, given

%A -> B
%B -> C
%C -> D
%D -> E
%E -> e

%By D^scan, we get E -> e.
%By D^C2 and E->e. we get A -> B.; B->C.; C->D.; D->E.
%By e.g. D^C2 and C->D. we get A ->B.
%Thus, there are two different item derivations of  A -> B.
%which we can't tolerate (and which is inefficient.)

\begin{figure}
$$
\begin{array}{cl}
\mbox{\bf Item form:} \\
{[A \derives \alpha \circ \beta]} \\
{[A \derives B]} \\
{[i, A\rightarrow \alpha\bullet\beta, j, u]} \\
{[i, \mi{finished}(A), j]} \\
{[i, \mi{extendedFinished}(A), j]} \\
%{[i, \mi{predicting}(A)]} \\
\\
\mbox{\bf Goal:} &\\
{[1, \mi{extendedFinished}(S'), n\!+\!1]} \\
\\
\mbox{\bf Rules:} \\
\mediuminfer{R(A \rightarrow \alpha)}{[A \derives\circ\alpha]}{} &
\mbox{Initial Axiom} \\

\mediuminfer{[A \derives \alpha\circ B \beta]\rspace[B \derives
\circ]}{[A \derives \alpha\circ\beta]}{} &
\mbox{Initial Epsilon Scanning} \\

\mediuminfer{[A \derives \circ B \beta]}{[A \derives B \circ\beta]}{} &
\mbox{Initial Unary Scanning} \\

\mediuminfer{}{[A \derives A]}{} &
\mbox{Unary Axiom} \\

\mediuminfer{[A \derives B]\rspace[B \derives C\circ]}{[A \derives C]}{} &
\mbox{Unary Completion} \\

\mediuminfer{R(S' \rightarrow \alpha)}{[1,S' \rightarrow {} \bullet \alpha,1,0]}{} &
\mbox{Initialization} \\
% what if S' goes to epsilon?
% what if S' goes to A and A goes to B -- check prediction carefully
% in this and following cases

\mediuminfer{[i, A \rightarrow \alpha\bullet B \beta, j, u] \rspace [B
\derives \circ]}
{[i, A \rightarrow \alpha \bullet \beta, j\!+\!1, u]}{}
&\mbox{Epsilon Scanning}\\

\mediuminfer{R(B \rightarrow\gamma)}{[j, B \rightarrow {} \bullet \gamma, j, 0]}
{[i, A \rightarrow \alpha\bullet B\beta, j, u] } &\mbox{Prediction} \\

\mediuminfer{[i, A \rightarrow \alpha\bullet B\beta, k, u]\rspace 
[k, \mi{extendedFinished}(B), j]} {[i, A \rightarrow \alpha B\bullet\beta, j,
\min(u\!+\!1, 2)]}{i<k}&
\mbox{Completion} \\

\mediuminfer{}{[i, \mi{finished}(a), i\!+\!1]}{w_i = a} &
\mbox{Terminal Finishing} \\

\mediuminfer{[i, A \rightarrow \alpha \bullet {}, j, u]}{[i, \mi{finished}(A),
j]}{u=2 \vee A = S' \wedge u = 0} &
\mbox{Finishing} \\

\mediuminfer{[i, \mi{finished}(A), j] \rspace [B \derives A]}
{[i, \mi{extendedFinished}(B), j]}{} &
\mbox{Extended Finishing}\\

\end{array}
$$
\caption{Graham Harrison Ruzzo}\label{fig:GHR}
\end{figure}

Figure \ref{fig:GHR} gives an item-based GHR parser description.  We
assume that the start symbol, $S'$, does not occur on the right hand
side of any rule.  We note that this parser does not work for
non-commutative semirings.  However, since there will still be a
one-to-one correspondence between item derivations and grammar
derivations, with simple modifications, it would be possible to map
from the item values derived here for the non-commutative derivation
semirings to the grammar values, as a post-processing step.

There are five different item types for this parser.  The first item
type, $[A \derives \alpha \circ \beta]$ is used for two purposes:
determining which elements have derivations of the form $A \derives
\epsilon$ and for determining which items have derivations of the form
$A \rightarrow \alpha B \beta$ such that $\alpha \derives \epsilon$
and $\beta \derives \epsilon $; these are the rules which form a
single step in a chain of unary productions.  We can actually only
derive two sub-types of this item: $[A \derives \circ \beta]$, which
can be derived only if there is a derivation of the form $A
\Rightarrow \alpha \beta \derives \beta$; and $[A \derives A
\circ\gamma]$ which can be derived only if there is a derivation of
the form $A \Rightarrow \alpha A \beta \gamma \derives A \gamma$.
Items of the type $[A \derives \alpha \circ \beta]$ are derived using
three different rules: the initial axiom, initial epsilon scanning,
and initial unary scanning.  The first rule, the initial axiom, simply
says that if $A \rightarrow \alpha$ then $A \derives \alpha$.  The
next rule, initial epsilon scanning, says that if $A \derives \alpha B
\beta$ and $B \derives \epsilon$, then $A \derives \alpha \beta$.  And
finally, initial unary scanning is a technical rule, which simply
advances the circle over a single nonterminal.  It is because of this
rule that we cannot use non-commutative semirings.  Consider a grammar
$$
\begin{array}{rcll}
A & \rightarrow & E B E  & R(A \rightarrow EBE) \\
E & \rightarrow & \epsilon & R(E \rightarrow\epsilon)
\end{array}
$$
Then the value of $[A\derives B\circ]$ will be $R(A \rightarrow EBE)
\otimes R(E \rightarrow\epsilon) \otimes R(E \rightarrow\epsilon)$.
For a commutative semiring, this grammar will work fine, but for a
non-commutative semiring, the value is incorrect, since we would need
to put the (yet to be determined) value of $B$'s derivation between
the two $R(E \rightarrow\epsilon)$'s.  Essentially, this is the same
reason that our grammar transformations only work for commutative
semirings.  In some sense, then, the GHR parser performs a grammar
transformation to CNF, and then parses the transformed grammar.  Of
course, for many non-commutative semirings such as the derivation
semirings, there will still be a one-to-one correspondence between
item derivations and grammar derivations; with slight modifications we
could easily map from the transposed item derivation to the
corresponding correct grammar derivation.

The next item type, $[A \derives B]$, can be derived only if there is
a derivation of the form $A \derives B$.  This item is derived with
two rules.  The unary axiom simply states that $A \derives A$.  Unary
completion states that if $A \derives B$ and $B \derives C$, then $A
\derives C$.  It is important that unary completion uses items of the
form $[B \derives C \circ]$, rather than items $[B \derives C]$, since
this makes sure that there is a one-to-one correspondence between item
derivations and grammar derivations.

The next item type is $[i, A\rightarrow \alpha\bullet\beta, j, u]$.
This is almost exactly the usual Earley style item: it can be derived
only if $S \rightarrow w_1...w_{i-1} A \gamma$ and $A \derives \alpha \beta
\derives w_i...w_{j-1} \beta$.  There is one additional condition
however.  We wish to precompute $\epsilon$ and unary derivations.
Therefore, in order to avoid duplicate derivations, we need to make
sure that we do not recompute $\epsilon$ or unary derivations during
the body of the computation.  Thus, we keep track of how many
non-$\epsilon$ derivations have been used to compute $[i, A\rightarrow
\alpha\bullet\beta, j, u]$, and encode this in $u$.  If $u$ is zero,
only $\epsilon$ rules have been used; if one, then one non-$\epsilon$
rule has been used, and if two, then at least two non-$\epsilon$ rules
have been used.  $u$ can only take the values 0,1, or 2.

We wish to collapse unary derivations.  We can do this using two more
item types.  Roughly, we can deduce $[i, \mi{finished}(A), j]$ if
there is an item of the form $[i, A \rightarrow \alpha\bullet {}, j,2]$,
i.e. when there is a derivation of the form $A \derives
w_i... w_{j-1}$ using two non-epsilon rules.  We require the last
element to be 2, to avoid recomputing unary chains or epsilon chains.
We will also derive such an item if $A = S'$ and the last element is
0; this allows us to recognize sentences of the form
$S'\derives\epsilon$.  The second item type, $[i,
\mi{extendedFinished}(A), j]$, can be derived only if there is a
derivation of the form $A \derives w_i...w_{j-1}$.  The difference
between this item type and the previous one is that there are now no
restrictions on unary branches and non-epsilon rules; this item type
captures unary branching extensions.

These items are derived using many rules.  The initialization rule is
just the initialization rule of Earley parsing; the epsilon scanning
rule is new: it allows us to skip over nonterminals $A$ such that $A
\derives \epsilon$.  The prediction rule is the same as the Earley
prediction rule.  We thus implement prediction incrementally, without
collapsing unary chains; \newcite[p. 436]{Graham:80a} suggest this as
one possible implementation of prediction.  With a few more deduction
rules, we could implement prediction more efficiently, in a manner
analogous to the finishing rules.

Completion is the same as Earley completion, with a few caveats.
First, we use the extended finishing items, rather than the items used
in Earley's algorithm.  This automatically takes into account unary
chains.  Also, we increment the non-epsilon count, not letting it go
above two.

We use the terminal finishing rule to handle terminal symbols, rather
than the scanning rule of Earley's algorithm.  This lets us take into
account unary chains using the terminal symbols.  The finishing rule
simply computes items of the form $[i, \mi{finished}(A)]$, as
previously described, and the extended finishing rule takes the
finished rules, and extends them with precomputed unary chains.

Notice that the value of items of the form $[A \derives \alpha \circ
\beta]$ and $[A \derives B]$, which are the only items in looping
buckets, can be computed without the input sentence; thus the value of
these items can be computed off-line, from the grammar alone.  Notice
also that the reverse values of these items require the input
sentence.  The item-based description format makes it easy to specify
parsers that use off-line computation for some values, and on-line
computation for other similar values.

\subsection{Beyond Context-Free}
\label{sec:appbeyond}

In this section, we consider the problem of formalisms that are more
powerful than CFGs.  We will show that these formalisms pose a slight
problem for the conventional algebraic treatment of formal language
theory, but that we can solve this problem without much trouble.  In
particular, in the conventional view of formal language theory
combined with algebra, a language is described as a formal power
series.  The terms of the formal power series are strings of the
language; the coefficients give the values of the strings.  These
formal power series are most readily described as sets of algebraic
equations; the formal power series represents a solution to the
equations.  Sets of algebraic equations used in this
way can only describe context-free languages.  On the other hand, it
is straightforward to describe a Tree Adjoining Grammar (TAG) parser,
which also can be transformed into a set of algebraic equations.
Since Tree Adjoining Grammars are more powerful than context-free
languages, this is a useful result.

\subsubsection{Formal Language Theory Background}

At this point, we need to discuss some results from algebra/formal
language theory.  One of the primary results is that there is a
one-to-one correspondence between CFGs and certain sets of algebraic
equations.

We must begin by defining a {\em formal power series}.  A formal power
series $\powera$ is a semiring using elements from a semiring $\BbbA$ and an
alphabet $\Sigma$.  Elements in $\powera$ map from strings
$\alpha$ in $\Sigma^*$ to elements of the semiring $\BbbA$.  If $s$ is an
element of $\powera$, we will write $(s, \alpha)$ to indicate the value
$s$ maps to $\alpha$.  Essentially, $s$ can be thought of as a
language; $\BbbA$ is the semiring used to assign values to strings of the
language.  Let $\Bbb{B}$ represent the booleans; then formal power
series $s \in \powerb{B}$ represent formal languages; strings $\alpha$ such
that $(s, \alpha) = \mi{TRUE}$ correspond to the elements in the
language; if $\BbbA$ were the inside semiring, then $(s, \alpha)$
would equal $P(\alpha)$, the probability of the string.  

We can define sums $s + t$ in $\powera$ as
$$
(u, \alpha) = (s, \alpha) \oplus (t, \alpha) 
$$
We can defines products $s \times t$ in $\powera$ as
$$
(u, \alpha) = \bigoplus_{\beta, \gamma \vert \alpha = \beta \gamma}
(s, \beta) \otimes (t, \gamma) 
$$
The product and sum definitions are analogous to the way products are
defined for polynomials in variables that do not commute, which is why
$\powerb{A}$ is called a formal power series; this is, however, probably
not the best way to think of $\powerb{A}$.  We can define multiplication
by a constant $t \in \BbbA$, $t \times s$ as
$$
(u, \alpha) = t \otimes (s, \alpha)
$$

Now, a formal power series is called {\em algebraic} if it is the
solution to a set of algebraic equations.  Consider the formal power
series 
$$
z + xzy + xxzyy + xxxzyyy + xxxxzyyyy + \cdots
$$
in $\powerb{B}$.  This formal
power series is a solution to the following algebraic equations in
$\Bbb{B}$:
\begin{eqnarray*}
S &=&xSy + Z \\
Z &=& z
\end{eqnarray*}
which is very similar to the CFG
$$
\begin{array}{rcl}
S & \rightarrow & xSy \\
S & \rightarrow & Z \\
Z & \rightarrow & z 
\end{array}
$$

The preceding example is in the boolean semiring, but could be
extended to any $\omega$-continuous semiring, by adding constants to
the equations.  In general, given a CFG and a rule value function R,
we can form a set of algebraic equations.  For instance, for a
nonterminal $A$ with rules such as
$$
\begin{array}{rcll}
A& \rightarrow & \alpha & R(A\rightarrow\alpha) \\
A& \rightarrow &\beta & R(A\rightarrow\beta) \\
A& \rightarrow &\gamma & R(A\rightarrow\gamma) \\
&\vdots & 
\end{array}
$$
there is a corresponding algebraic equation
$$
A = R(A\rightarrow\alpha)\alpha +R(A\rightarrow\beta)\beta +R(A\rightarrow\gamma)\gamma...
$$
with analogous equations for the other nonterminals in the grammar.
Each nonterminal symbol represents a variable; each terminal symbol is
a member of the alphabet, $\Sigma$.  

% we should use a double line B, not cal B, for booleans

It is an important theorem from the intersection of formal language
theory and algebra, that for each CFG there is a set of algebraic
equations in $\powerb{B}$, the solutions to which represent the
strings of the language; for each algebraic equation in $\powerb{B}$,
there is a CFG, whose language corresponds to the solutions of the
equations.

\subsubsection{Tree Adjoining Grammars}

In this section, we address TAGs.  TAGs pose an interesting challenge
for semiring parsers, since the tree adjoining languages are weakly
context sensitive.  We develop an item-based description for a TAG
parser, essentially using the description of \newcite{Shieber:93a},
modified for our format, and including a few extra rule values, to
ensure that there is a one-to-one correspondence between item and
grammar derivations which can easily be recovered.  The reader is
strongly encouraged to refer to that work for background on TAGs and
explication of the parser.

\begin{figure}
$$
\begin{array}{cl}
\mbox{\bf Item form:} \\
{[\nu^{\bullet}, i, j, k, l]} \\
{[\nu_{\bullet}, i, j, k, l]} \\
{[\mi{goal}]} \\
\\
\mbox{\bf Goal}\\
{[\mi{goal}]} \\
\\
\mbox{\bf Rules:} \\
\infer{R(\mi{start}(\alpha)) \rspace [\alpha@\epsilon^{\bullet},0,\_,\_,n]}{[\mi{goal}]}{} &
\mbox{Find Start} \\

\infer{}{[\nu^{\bullet}, i,\_,\_, i\!+\!1]}{\mi{label}(\nu) = w_{i\!+\!1}} &
\mbox{Terminal Axiom} \\

\infer{}{[\nu^{\bullet}, i,\_,\_, i]}{\mi{label}(\nu) =\epsilon} &
\mbox{Empty String Axiom} \\

\infer{}{[\beta@\mi{foot}(\beta)_{\bullet},p,p,q,q]}{\beta \in A} &
\mbox{Foot Axiom} \\

\infer{[\alpha@(p\cdot1)^{\bullet},i,j,k,l]}
{[\alpha@p_{\bullet},i,j,k,l]}{\alpha@(p \cdot 2) \mbox{ undefined}} &
\mbox{Complete Unary} \\

\infer{[\alpha@(p\cdot1)^{\bullet},i,j,k,l]\rspace [\alpha@(p\cdot 2)^{\bullet},l,j',k',m]}
{[\alpha@p_{\bullet},i,j \cup j', k \cup k', m]}{} &
\mbox{Complete Binary} \\

\infer{R(\mi{noadjoin}(\nu)) \rspace [\nu_{\bullet},i,j,k,l]} {[\nu^{\bullet},i,j,k,l]}{}&
\mbox{No Adjoin} \\

\infer{R(\mi{adjoin}(\beta, \nu)) \rspace [\beta@\epsilon^{\bullet},i,p,q,l]
\rspace [\nu_{\bullet},p,j,k,q] }
{[\nu^\bullet {},i,j,k,l]}{} &
\mbox{Adjoin}

\end{array}
$$
\caption{TAG parser item-based description}\label{fig:TAG}
\end{figure}

In the TAG formalism, there are two trees that correspond to a parse
of a sentence.  One tree is the parse tree, which is a conventional
parse of the sentence.  The other tree is the derivation tree; a
traversal of the derivation tree gives the rules that would actually
be used in a derivation to produce the parse tree, in the order they
would be used.  While the parse trees of TAGs cannot be produced by a
CFG, the derivation trees can.  It is important to note that the
derivations produced by the parser of Figure \ref{fig:TAG} correspond
to derivation trees, not parse trees.  This is, in part, what allows
us to parse with a formalism more powerful than CFGs.

The other reason that we are able to parse with a more powerful
formalism is that while the algebraic formulation of formal language
theory specifies a language as a single set of algebraic equations, we
specify parsers as an input-string specific set of algebraic
equations.  Recall that the way we find the values of items in a
looping bucket is to create a set of algebraic equations for the
bucket.  Recall also that we can always place all items into a single
large looping bucket.  Thus, we can solve the parsing problem for any
specific input string by solving a set of algebraic equations.
However the algebraic equations we solve are specific to the input
string, and the number of equations will almost always grow with the
length of the input string.  This variability of the equations is the
other factor that allows us to parse formalisms more powerful than
CFGs.

\subsection{Greibach Normal Form} 
\label{sec:GNFappendix}

As we discussed in Section \ref{sec:grammar}, item-based descriptions
can be used to specify grammar transformations.  In Section
\ref{sec:grammar}, we showed how to use item-based descriptions to
convert to Chomsky Normal Form.  In this section, we show how to use
these descriptions to convert to Greibach Normal Form (GNF).  In GNF,
every rule is of the form $A \rightarrow a \alpha$.  We give here a
value-preserving transformation to GNF, following
\newcite{Hopcroft:79a}.  While value preserving GNF transformations
have been given before \cite{Kuich:86a}, this is the first item-based
description of such a transformation. 

Figure \ref{fig:GNF} gives an item-based description for a
value-preserving GNF transformation.  There is a sequence of steps in
GNF transformation, so we will use items of the form $[A \rightarrow
\alpha]_j$, where $j$ indicates the step number.  The first step in a
GNF transformation is to put the grammar in Chomsky Normal Form, which
we have previously shown how to do, in Section \ref{sec:grammar}.  We
will assume for this subsection that our nonterminal symbols are $A_1$
to $A_m$.  The next step is conversion to ``ascending'' form,
$$
\begin{array}{ll}
A_i \rightarrow A_j \alpha & (j \geq i) \\
A_i \rightarrow a \alpha &
\end{array}
$$

\begin{figure}
$$
\begin{array}{cl}
\mbox{\bf Item form:} &\\
{[A_i \rightarrow \alpha]_j} &\\
{[B_i \rightarrow \alpha]_j} &\\
&\\
\mbox{\bf Rule Goal}&\\
{R_1(X \rightarrow \alpha)}& \\
&\\
\mbox{\bf Rules:} &\\
\infer{R_0(A_i \rightarrow A\alpha)}{[A_i \rightarrow 
\alpha]_0}{} & \mbox{Rule Axiom}\\

\infer{[A_i \rightarrow A_j \alpha]_0 \rspace [A_j \rightarrow \beta]_1 }
{[A_i \rightarrow \beta \alpha]_0}{j < i} & 
\mbox{Ascent Substitution} \\

\infer{[A_i \rightarrow X \alpha]_0}{[A_i \rightarrow X \alpha]_1}
{X = a \vee (X = A_j \wedge j\geq i)} & \mbox{Ascent Completion} \\

\infer{[A_i \rightarrow X \beta]_1}{[A_i \rightarrow X \beta
B_i]_2}{X = a \vee (X = A_j \wedge j> i)} & \mbox{Right Beginning} \\

\infer{[A_i \rightarrow X \beta]_1}
{[A_i \rightarrow X \beta]_2}{X = a \vee (X = A_j \wedge j > i)} & \mbox{Right Single Step} \\

\infer{[A_i \rightarrow A_i \alpha]_1}{[B_i \rightarrow \alpha]_2}{j > i} & \mbox{Right Termination} \\

\infer{[A_i \rightarrow A_i \alpha]_1}{[B_i \rightarrow \alpha B_i]_2}{} &
\mbox{Right Continuation} \\

\infer{[A_i \rightarrow A_j \alpha]_2 \rspace [A_j \rightarrow a \beta]_3}
{[A_i \rightarrow a \beta \alpha]_3}{j > i} & 
\mbox{Descent Substitution} \\

\infer{[A_i \rightarrow a \alpha]_2}{[A_i \rightarrow a \alpha]_3}{} &
\mbox{Descent Non-substitution} \\

\infer{[B_i \rightarrow A_j \alpha]_2\rspace [A_j \rightarrow a\beta]_3}
{[B_i \rightarrow a\beta \alpha]_3}{} & \mbox{Auxiliary Substitution}  \\

\infer{[B_i \rightarrow a \alpha]_2}
{[B_i \rightarrow a\alpha]_3}{} & \mbox{Auxiliary Non-substitution} \\

\infer{[X \rightarrow \alpha]_3}{R_1(X \rightarrow \alpha)}{} & 
\mbox{Output}

\end{array}
$$
\caption{Greibach Normal Form Transformation}\label{fig:GNF}
\end{figure}
This will be accomplished by first putting all rules with $A_1$ into
this form, then $A_2$, etc.  To transform rules into this form, we
substitute the right hand sides of $A_j$ for $A_j$ in any rule of the
form $A_i \rightarrow A_j \alpha\; (j < i)$, a process that must
terminate after at most $i-1$ steps.  This step is done using three
rules, the Rule Axiom, which creates one item of the form $[A_i
\rightarrow \alpha]_0$ for every rule of the form $A_i \rightarrow
\alpha$, i.e. for every rule in the grammar.  We use two different
item forms: $[A_i \rightarrow A_j \alpha]_0$, $j\leq i$, for rules not
yet in ascending form; and $[A_i \rightarrow A_j \alpha]_1$, $j\geq
i$, for rules in ascending form.  The Ascent Substitution rule
substitutes the right hand side of a rule in ascending form into the
first nonterminal of a rule not in ascending form.  The Ascent
Completion rule detects that an item is in ascending form, and
promotes it, creating an item with subscript 1.

The next step is removal of all left branching rules
of the form $A_i \rightarrow A_i \alpha$ by conversion to the form
$$
\begin{array}{ll}
A_i \rightarrow A_j \alpha & (j > i) \\
A_i \rightarrow a \alpha &\\
B_i \rightarrow \alpha & 
\end{array}
$$
There are four rules that perform this operation: right beginning,
right single step, right termination, and right continuation, using
items in four different forms.  The first item form is $[A_i
\rightarrow A_j\beta]_1$, which is essentially the input to this step.
The other three item forms are $[A_i \rightarrow A_j \alpha]_2$,
$j>i$, $[A_i \rightarrow a \alpha]_2$, and $[B_i \rightarrow \alpha]_2$,
which are essentially the outputs of this step.  Consider a
left-branching derivation of the form
$$
A_i \derivesby{A_i}{\alpha} A_i
\alpha \derivesby{A_i}{\beta} A_i \beta \alpha \derivesby{A_i}{A_j\gamma} A_j \gamma \beta
\alpha
$$
The right branching equivalent will be 
$$
A_i \derivesby{A_i}{A_j\gamma B_i} A_j \gamma B_i
\derivesby{B_i}{\beta B_i} A_j \gamma \beta B_i \derivesby{B_i}{\alpha} A_j \gamma
\beta \alpha
$$
The item generating the rule used in the first substitution in the
right branching form will be derived using Right Beginning; the item
for the second substitution will be derived using Right Continuation;
and the item for the final substitution will be derived using Right
Termination.  The rule $A_i \rightarrow A_j \gamma$ is valid in both
forms; Right Single Step derives the needed item.

Now, we are ready for the next and penultimate step.  Notice that
since all items of the form $[A_i \rightarrow A_j \alpha]_2$ have $j >
i$, the last nonterminal, $A_m$, must have rules only of the form $A_m
\rightarrow a \alpha$, the target form.  We can substitute $A_m$'s
productions into any production of the form $A_{m-1} \rightarrow A_m
\alpha$, putting all $A_{m-1}$ productions into the target form.  This
recursive substitution can be repeated all the way down to terminal
$A_1$.  Formally, we substitute $A_j$ into any rule of the form $A_i
\rightarrow A_j \alpha \; (j > i)$, yielding rules of the form
$$
\begin{array}{ll}
A_i \rightarrow a \alpha &\\
\end{array}
$$
We call this penultimate step Descent Substitution.  Descent
Substitution is done using two rules: Descent Substitution and Descent
Non-substitution.  The inputs to this step are items in the form $[A_i
\rightarrow A_j \alpha]_2$ and the outputs are in the form $[A_i
\rightarrow a \beta]_3$.  Descent Non-substitution simply detects when
an item already has a terminal as its first symbol, and promotes it to
the output form.

Finally, we substitute the $A_i$ into the $B_j$ to yield rules of the
form:
$$
\begin{array}{ll}
A_i \rightarrow a \alpha &\\
B_i \rightarrow a \alpha & 
\end{array}
$$
This step is accomplished with the Auxiliary Substitution and
Auxiliary Non-substitution rules.  \newcite{Hopcroft:79a} shows that
all items $[B_i \rightarrow \alpha]_2$ are actually in the form $[B_i
\rightarrow A_j \alpha]_2$ or $[B_i \rightarrow a \alpha]_2$.  The
Auxiliary Substitution rule performs substitution into items of the
first form, and the Auxiliary Non-substitution rule promotes items of
the second form.

The output rule trivially states that items of the form $[X
\rightarrow \alpha]_3$ correspond to the rules of the transformed
grammar. 

Proving that this transformation is value-preserving is somewhat
tedious, and essentially follows the proof that grammars can be
converted to GNF given by \newcite{Hopcroft:79a}.

\appendixsection{Reverse Value of Non-commutative Semirings}
\label{sec:reversenon}

In this appendix, we consider the problem of finding reverse values in
non-commutative semirings; the situation is somewhat more complex than
in the commutative case.  The problem is, for non-commutative
semirings, there is no obvious equivalent to the reverse values.
Consider the following grammar:
$$
\begin{array}{rcl}
S & \rightarrow & SA \\
S & \rightarrow & B \\
A & \rightarrow & \epsilon \\
B & \rightarrow & b
\end{array}
$$
Now, consider the derivation forest semiring.  Without making reference to a
specific parser, let us call the item deriving the terminal symbol
$[B]$.  We will denote derivations using just the nonterminals on the
left hand side, for conciseness.  So, for instance, a derivation 
$$
S \derivesby{S}{SA} SA \derivesby{S}{SA} SAA \derivesby{S}{B} BAA
\derivesby{B}{b} bAA \derivesby{A}{\epsilon} bA
\derivesby{A}{\epsilon} b
$$
will be written as simply $SSSBAA$.  The inside value of $[B]$ will
just be
$$
V([B]) = \{ B \}
$$
The value of the sentence is the union of all derivations, namely:
$$
\{SB, SSBA, SSSBAA, SSSSBAAA, SSSSSBAAAA,...\}
$$
Now, since all derivations use the item $[B]$, it should be the case
that the forward value times the reverse value of $[B]$ should just be
the value of the sentence:
$$
V([B]) \cdot Z([B]) = \{SB, SSBA, SSSBAA, SSSSBAAA, SSSSSBAAAA,...\}
$$
Since $V([B]) = \{B\}$, we get
$$
\{B\} \cdot Z([B]) = \{SB, SSBA, SSSBAA, SSSSBAAA, SSSSSBAAAA,...\}
$$
for some $Z([B])$.  But it should be clear that there is no such value
for $Z([B])$ in a non-commutative semiring -- the problem, intuitively, is that we need to get
$V([B])$ into the center of the product, and there is no way to do
that.  One might think that what we need is two reverse values, a left
outside value $Z_L([B])$, and a right outside value $Z_R([B])$, so
that we can find values such that
$$
Z_L([B]) \cdot V([B]) \cdot Z_R([B]) = \{SB, SSBA, SSSBAA, SSSSBAAA, SSSSSBAAAA,...\}
$$
but some thought will show that even this is not possible.

The solution, then, is to keep track of pairs of values.  That is, we let
$Z([B])$ be a set of pairs of values, such as
$$ 
Z([B]) = \{ \langle \{S\},\{\} \rangle, \langle \{SS\},\{A\} \rangle,
\langle \{SSS\},\{AA\} \rangle, \langle \{SSSS\},\{AAA\} \rangle, ...\}
$$
and then define an appropriate way to combine these pairs of values
with inside values.  This lets us define reverse values in the
non-commutative case.  We can use this same technique of using pairs
of values to find an analog to the reverse values for most
non-commutative semirings.

\subsection{Pair Semirings}
\label{sec:pairsemirings}

There are many technical details.  In particular, we
will want to compute infinite sums using these pairs of values, in
ways similar to what we have already done.  Thus, it will be
convenient to define things so that we can use the mathematical
machinery of semirings.  Therefore, given a semiring $\Bbb{A}$, we
will define a new semiring \PAend.  Later, we will describe a property,
preserving pair order, and show that if $\Bbb{A}$ is
$\omega$-continuous and preserves pair order, then \PA is
$\omega$-continuous, allowing us to compute infinite sums just as
before.  We will call \PA the pair semiring of $\Bbb{A}$.

Intuitively, we simply want pairs of values, but in practice it will
be more convenient to define addition if we allow pairs to occur multiple times.
Therefore, we will define elements of \PA as a mapping function from
pairs $ a, a'$ to $\Bbb{N}^\infty$.  Thus, an element
of the semiring will be $r: \Bbb{A} \times \Bbb{A} \rightarrow
\Bbb{N}^{\infty}$.\footnote{In retrospect, it might have been simpler
to consider the semiring of functions from $\Bbb{A}$ to $\Bbb{A}$,
where the multiplicative operator is composition.  On the other hand,
the Pair semiring will simplify the discussion in Section
\ref{sec:specificpair}, where the fact that all elements of the
semiring have the form of multi-sets of pairs will be useful.}

Continuing our example, if $r = Z([B])$, then
$$
r(E_1, E_2) =
\left\{\begin{array}{ll}
1& \mbox{ \em if } E_1 =\{S^k\}\wedge E_2 =\{A^{k-1}\} \\
0& \mbox{ \em otherwise}
\end{array}\right.
$$

Next, we need to define combinations of an element $r \in$ \PA with an
element $a \in \Bbb{A}$, which we will write as $r.a$, to evoke
function application.  We simply multiply $a$ between each of the
pairs.  Formally, 
$$
r.a = \bigoplus_{b, b' \in \Bbb{A}} r(b, b')b a b'
$$
where the premultiplication by the integral value $r(b, b')$ indicates
repeated addition in $\Bbb{A}$.  

Furthermore, we define two elements, $r$ and $s$ to be
equivalent if whenever we combine them with any element of $\Bbb{A}$,
they yield the same value. Formally,
$$
r = s \mbox { iff } \forall a\; r.a = s.a
$$

It will be convenient to denote certain single pairs of elements of
$\Bbb{A}$ concisely.  Let $\lfloor a, a'\rfloor$ indicate
$$
\lfloor a, a'\rfloor(b, b') = 
\left\{\begin{array}{ll}
1 & \mbox{ \em if } a = b\wedge a' = b' \\
0 & \mbox{ \em otherwise}
\end{array}
\right.
$$
Intuitively, this value is the set containing the single pair $a, a'$.

We will need to make frequent reference to pairs of values.  We will
therefore let $\bara$ indicate the pair $a, a'$, and interchange these
notations freely.

Addition can be simply defined pairwise: if $t = r\oplus s$, then
$$
t(\bara) = r(\bara) +s(\bara)
$$
Commutativity of addition follows trivially.  We denote the zero
element by $\lfloor 0, 0 \rfloor$; notice that for all $a$, $\lfloor
0,0 \rfloor.a = 0$

We now show that application distributes over addition.
\begin{eqnarray*}
(r+s).a 
& = & \sum_{b, b'} (r+s)(b, b') bab' \\
& = & \sum_{b, b'} (r(b, b')+s(b,b')) bab' \\
& = & \sum_{b, b'} r(b, b')bab'+s(b,b')bab' \\
& = & \sum_{b, b'} r(b, b')bab'+\sum_{b,b'}s(b,b')bab' \\
& = & r.a + s.a
\end{eqnarray*}
A similar argument shows that application also distributes over infinite sums.

Next, we show that for any elements $r, s, t$, our definition of
equality is consistent with our definition of addition, i.e. if $r=s$
then $r+t = s+t$.  We simply notice that, for all $a \in \Bbb{A}$,
$(r+t).a = r.a + t.a = s.a+t.a = (s+t).a$.

Now, when we define multiplication, we want the property that
$$
\lfloor a, a'\rfloor \otimes \lfloor b, b'\rfloor = \lfloor ab,
b'a'\rfloor
$$
multiplying the first element on the right, and the second element on
the left.

More generally, we define multiplication as, for $t =
r \otimes s$, 
$$
t(c, c') =\sum_{a, a'} \;\; \sum_{b, b'\smallst ab = c\wedge
b'a'= c'} r(a, a') s(b, b')
$$
This definition of multiplication has the
desired property.  We will also write $\bara \otimes \barb$ to
indicate the pair $\lfloor ab, b'a' \rfloor$; this allows us to write the
multiplicative formula as:
$$
t(\barc) =\sum_{\bara}\sum_{\barb \smallst \bara \otimes \barb = \barc} r(\bara) s(\barb)
$$
It should be clear that $\lfloor 1, 1 \rfloor$ is a multiplicative identity.

We now show that for any elements $r, s, t$, our definition of
equality is consistent with our definition of multiplication, i.e. if
$r=s$, then $r \otimes t = s \otimes t$ and $t \otimes r = t \otimes s$.  We 
first show that for all $r,s$, $(r \otimes s).a = r.(s.a)$:
\begin{eqnarray}
(r \otimes s).a 
& = &
\sum_{b, b'} (r \otimes s)(b, b') b a b' \nonumber \\
& = &
\sum_{b, b'} \sum_{c, c'} \sum_{d, d' \smallst cd = b \wedge d'c' =b'}
r(c, c') s(d, d') bab' \nonumber \\
& = &
\sum_{c, c'} \sum_{d, d'} r(c, c') s(d, d') cdad'c' \nonumber \\
& = &
\sum_{c, c'} r(c, c') c \left( \sum_{d, d'} s(d, d') dad' \right) c' \nonumber \\
& = &
\sum_{c, c'} r(c, c') c (s.a) c' \nonumber \\
& = &
r.(s.a)
\end{eqnarray}
Now, if $r=s$ then for all $a$, 
$$(r \otimes t).a = r.(t.a) = s.(t.a) = (s \otimes t).a$$
Also, if $r=s$ then 
$$(t \otimes r).a = t.(r.a) = t.(s.a) = (t \otimes s).a$$ 
This shows that our definition of equality is consistent with our
definition of multiplication.

Multiplication is not commutative, but is
associative, meaning that $ (r \otimes s) \otimes t= r \otimes (s
\otimes t)$.  We show this now, first computing a simple
form for $ (r \otimes s) \otimes t$.
\begin{eqnarray*}
( (r \otimes s) \otimes t)(\bare) & =&
\sum_{\barc} \;\; \sum_{\bard|\barc\otimes\bard=\bare} (r\otimes s)(\barc)t(\bard) \\
& = &
\sum_{\barc} \;\; \sum_{\bard|\barc\otimes\bard=\bare}
\sum_{\bara} \;\;\sum_{\barb|\bara\otimes\barb=\barc}
r(\bara) s(\barb) t(\bard) \\
& = &
\sum_{\barc} \;\; 
\sum_{\bara}\sum_{\barb|\bara\otimes\barb=\barc} \;\;
\sum_{\bard|\barc\otimes\bard=\bare}
r(\bara) s(\barb) t(\bard) \\
& = &
\sum_{\bara}\sum_{\barb}\;
\sum_{\barc|\bara\otimes\barb=\barc} \; \;
\sum_{\bard|\barc\otimes\bard=\bare}
r(\bara) s(\barb) t(\bard) \\
&=&\sum_{\bara}\sum_{\barb} \;\;
\sum_{\bard|\bara \otimes \barb \otimes\bard=\bare}
r(\bara) s(\barb) t(\bard) \\
\end{eqnarray*}
A similar rearrangement yields the same formula for $ r \otimes (s
\otimes t)$, showing
associativity.

Now, we need to show that addition distributes over multiplication,
both left and right.  We do only the right case, i.e. $r \otimes (s
\oplus t) = r \otimes s \oplus r \otimes t$.  The left case is symmetric.
\begin{eqnarray*}
(r \otimes (s \oplus t))(\bara) 
&= &
\sum_{\barb} \;\;\sum_{\barc \smallst \barb\barc = \bara} r(\barb) (s\oplus
t)(\barc) \\
&= &
\sum_{\barb} \;\;\sum_{\barc \smallst \barb\barc = \bara} r(\barb)
(s(\barc) + t(\barc)) \\
&= &
\sum_{\barb} \;\;\sum_{\barc \smallst \barb\barc = \bara} r(\barb)
s(\barc) +
\sum_{\barb} \;\;\sum_{\barc \smallst \barb\barc = \bara} r(\barb) t(\barc)\\
& = &
(r\otimes s \oplus r\otimes t)(\bara)
\end{eqnarray*}

We have now shown all of the non-trivial properties of a semiring (and
a few of the trivial ones, as well.)

There is one more property we must show: that \PA is
$\omega$-continuous, assuming that $\Bbb{A}$ is $\omega$-continuous,
and assuming an additional quality, preserving pair order, which we
define below.  This involves several steps.  The first is to show that
\PA is naturally ordered, meaning that we can define an ordering
$\sqsubseteq$, such that $r \sqsubseteq s$ if and only if there exists
$t$ such that $r \oplus t = s$.  To show that this ordering defines a
true partial ordering, we must show that if $r \sqsubseteq s$ and $s
\sqsubseteq r$ then $r = s$.  (The other properties of a partial
ordering, reflexivity and transitivity, follow immediately from the
properties of addition.)  From the fact that $r \sqsubseteq s$, we
know that for some $t$, $r + t = s$.  Therefore, we know that for all
$a \in \Bbb{A}, (r + t).a = s.a$.  Thus, $r.a + t. a = s.a$.  Now,
since $\Bbb{A}$ is also $\omega$-continuous, it is also naturally
ordered, implying that $r.a \sqsubseteq s.a$.  Similarly, from the
fact that $s \sqsubseteq r$, we conclude that $s.a \sqsubseteq r.a$.
Thus, $r.a = s.a$, which shows that $r = s$.

The next step in showing $\omega$-continuity is to show that \PA is
complete, meaning that we must show that infinite sums are commutative
and satisfy the distributive law.  Associativity of infinite sums
means that given an index set $J$, and disjoint index sets $I_j$ for
$j \in J$,
$$
\bigoplus_{j \in J} \bigoplus_{i \in I_j} r_i = \bigoplus_{i \in \bigcup_j
I_j} r_i
$$
This property follows directly from the fact that $\Bbb{N}^\infty$ is
complete.  We also must show that multiplication distributes (both
left and right) over infinite sums.  We show one case; the other is
symmetric.  We use  a simple rearrangement of sums, which again relies on the
fact that $\Bbb{N}^{\infty}$ is complete.
\begin{eqnarray*}
\left( \bigoplus_{i \in I} r \otimes s_i \right) (\barc) & = &
\sum_{i \in I} \sum_{\bara} \;\;\sum_{\barb|\bara \barb = \barc} r(\bara) s_i(\barb) \\
& = &
\sum_{\bara} \;\;\sum_{\barb|\bara \barb = \barc} r(\bara) \;\;\sum_{i \in I} s_i(\barb) \\
& = &
\left( r\otimes \bigoplus_{i \in I} s_i \right) (\barc) 
\end{eqnarray*}

The final step in showing $\omega$-continuity is to show that for all
$s$, for all sequences $r_i$, if $\bigoplus_{0 \leq i \leq n} r_i
\sqsubseteq s$ for all $n \in \Bbb{N}$, then $\bigoplus_{i \in
\Bbb{N}} r_i \sqsubseteq s$.  We do not know how to show this for
$\omega$-continuous semirings $\Bbb{A}$ in general, although
$\omega$-continuity of $\Bbb{A}$ is certainly a requirement.  We
therefore make an additional assumption which is true of all semirings
discussed in this \chapterpaper{}: we assume that if for all $a \in \Bbb{A}, r.a
\sqsubseteq s.a$ then $r \sqsubseteq s$.  If this assumption is true for a
semiring $\Bbb{A}$, we shall say that $\Bbb{A}$ {\em preserves pair
order}.  Now, assuming $\Bbb{A}$ preserves pair order, it is
straightforward to show that \PA is $\omega$-continuous.  Given a
sequence $r_i$, let $r = \bigoplus_{0 \leq i} r_i$, and let $r_{\leq
n} = \bigoplus_{0 \leq i \leq n} r_i$.  Now, for all $n$, $r_{\leq n}
\sqsubseteq r$, since $r_{\leq n} \oplus \bigoplus_{i > n} r_i = r$.
Notice that for any $t$, $t \sqsubseteq s$ implies that for all $a$,
$t.a \sqsubseteq s.a$.  Therefore, since for all $n$ and all $a$,
$r_{\leq n} \sqsubseteq s$, we conclude that
\begin{equation}
r_{\leq n}.a \sqsubseteq s.a
\label{eqn:ralesssa}
\end{equation}
Now, it is a property of $\omega$-continuous semirings that
\cite[p. 613]{Kuich:97a}
$$
\sup_n \bigoplus_{0 \leq i \leq n} a_i = \bigoplus_{0 \leq i} a_i
$$
Notice that 
$$
r_{\leq n}.a = \left(\bigoplus_{0 \leq i \leq n}
r_i\right).a = \bigoplus_{0 \leq i \leq n} (r_i.a)
$$ 
and
$$
r.a = \left(\bigoplus_{0 \leq i} r_i\right).a = \bigoplus_{0 \leq i} (r_i.a)
$$
Thus,
\begin{equation}
\sup_n r_{\leq n}.a = r.a
\label{eqn:supreqr}
\end{equation}
From Equation \ref{eqn:ralesssa} and Equation \ref{eqn:supreqr}, we
conclude that for all $a$, $r.a \sqsubseteq s.a$, and from this and
our assumption that $\Bbb{A}$ preserves pair order, we conclude that
$r \sqsubseteq s$, which shows that \PA is an $\omega$-continuous
semiring.

\subsection{Specific Pair Semirings} \label{sec:specificpair}
Now, we can discuss specific semirings, showing that they preserve
pair order, and showing how to efficiently implement \PAend.
We first note that for any commutative semiring $\Bbb{A}$, \PA is
isomorphic to $\Bbb{A}$; the equivalence classes of \PA are in
direct correspondence with the elements of $\Bbb{A}$; application
$r.a$ is equivalent to multiplication.  It is thus straightforward to
show that $\Bbb{A}$ preserves pair order.  Thus, the formulae we will
give in the sequel for paired semirings hold equally well for all
commutative semirings.

Next, we notice that for all of the non-commutative semirings
discussed in this \chapterpaper{}, the three derivation semirings, if
$a \sqsubseteq b$ then $a \oplus b=b$.  Now, for such a semiring, if
for all $a$, $r.a \sqsubseteq s.a$ then for all $a$, $r.a + s.a = s.a$
and thus, $r + s = s$ which implies that $r \sqsubseteq s$.  Thus all
of the non-commutative semirings discussed in this \chapterpaper{}
preserve pair order.

%Consider the following derivations:

%x = bbbbb ccccccccc
%y = bbbbbbbbb ccccc

%Then, x.a = bbbbbaccccccccc,
%      y.a = bbbbbbbbbaccccc

%But x.d = bbbbbdccccccccc,
%    y.d = bbbbbbbbbdccccc

%So, depending on what we apply, we get different orders.

Implementations of the paired versions of the three derivation
semirings are straightforward.  To implement the Pair-derivation
semiring, we simply keep sets of pairs of derivation forests.  Recall
that for the Viterbi-derivation semiring, in theory we keep a
derivation forest of all top scoring values, but in practical
implementations, typically keep just an arbitrary element of this
forest.  In a theoretically correct implementation of the
Pair-Viterbi-derivation semiring, we simply maintain a top scoring
element $v$, and the set of pairs of derivations with value $v$, using
the same implementation as for the Pair-derivation semiring.  If in
practice we are only interested in a representative top scoring
derivation, rather than the set of all top scoring derivations, then
we can simply maintain a pair, $d, e$, of a left and a right
derivation, along with the value $v$.  Similar considerations are true
for both theoretical and practical implementations of the
Pair-Viterbi-n-best semiring.

\subsection{Derivation of Non-Commutative Reverse Value Formulas}

Now, we can rederive the reverse formulas, but using non-commutative
semirings, and the corresponding pair semirings.  These derivations
almost exactly follow the corresponding derivations in commutative
semirings.  

For non-commutative semirings $\Bbb{A}$, $Z(x)$ will represent a value in
\PAend.  We will construct reverse values in such a way that
\begin{equation}
\label{eqn:forwardsreverse2}
Z(x).V(x) =\bigoplus_{D\mbox{ \footnotesize a derivation}} V(D) C(D, x)
\end{equation}
which is directly analogous to Equation \ref{eqn:forwardsreverse}.

Recall that an outer tree $O$ has a hole in it, from where some inner
tree headed by $x$ was deleted.  We now define the left and right
reverse values of an outer tree $O$, $Z_L(O)$ and $Z_R(O)$, to be the
product of the values of the rules to the left and to the right of the
hole, respectively.
$$
Z_L(O) = \bigotimes_{r\in D\mbox{ \footnotesize to left}} R(r)
$$
$$
Z_R(O) = \bigotimes_{r\in D\mbox{ \footnotesize to right}} R(r) 
$$
Notice that these are values in $\Bbb{A}$.  Now, we will define the
value of an outer tree.  We will use the same notation, $Z(O)$ as we
used for non-commutative semirings for consistency, which we hope will
not lead to confusion.  If these formulas were applied to
commutative semirings, the values would be the same as with the
original formulas, so this redefinition should not be problematic.

We define the value of $Z(O)$ to be the pair of values of the
rules to the left and right of the hole.
$$
Z(O) = \lfloor Z_L(O), Z_R(O) \rfloor
$$
which is a value in \PAend.  Then, the reverse value of an item can be
defined, just as before in Equation \ref{eqn:reverse}.
$$
Z(x) = \bigoplus_{D\in \smi{outer}(x)} Z(D)
$$
That is, the reverse value of $x$ is the sum of the values of each
outer tree of $x$, this time in the pair semiring.

Next, we show that this new definition of reverse values has the
property described by Equation \ref{eqn:forwardsreverse2}, following
almost exactly the proof of Theorem \ref{theorem:reverseproperty}.
\begin{mytheorem}
$$
Z(x).V(x) =\bigoplus_{D \mbox{ \footnotesize a derivation}} V(D) C(D, x)
$$
\end{mytheorem}
{\em Proof \hspace{3em}}
\begin{eqnarray*}
Z(x).V(x)  &= & \left( \bigoplus_{O\in \smi{outer}(x)} Z(O) \right).V(x)  \\
& = & \left( \bigoplus_{O\in \smi{outer}(x)} Z(O)\right).
\left( \bigoplus_{I\in\smi{inner}(x)} V(I) \right) \\
& = & 
\left( \bigoplus_{O\in \smi{outer}(x)} 
\lfloor Z_L(O), Z_R(O) \rfloor
\right).
\left( \bigoplus_{I\in\smi{inner}(x)} V(I) \right) \\
& = & 
\bigoplus_{O\in \smi{outer}(x)} 
\left(
\lfloor Z_L(O), Z_R(O) \rfloor . \bigoplus_{I\in\smi{inner}(x)} V(I) 
\right) 
\\
& = & 
\bigoplus_{O\in \smi{outer}(x)} 
Z_L(O) \left( \bigoplus_{I\in\smi{inner}(x)} V(I) \right) Z_R(O) 
\\
& = & 
\bigoplus_{I\in\smi{inner}(x)} \bigoplus_{O\in \smi{outer}(x)} 
Z_L(O)  V(I) Z_R(O) 
\end{eqnarray*}
Now, using the same reasoning as in Theorem
\ref{theorem:reverseproperty}, it should be clear that this last
expression equals the expression on the right hand side of Equation
\ref{eqn:forwardsreverse2}, $\bigoplus_D V(D) C(D, x)$,
completing the proof. $\Box$

There is a simple, recursive formula for efficiently computing reverse
values, analogous to that shown in Theorem
\ref{theorem:simplereverse}, and using an analogous proof.  Recall
that the basic equation for computing forward values not involved in
loops was
$$
V(x) = \bigoplus_{a_1... a_k
\;\mbox{\scriptsize s.t.}\;\smallinfer{a_1... a_k}{x}{}} \bigotimes_{i =
1}^k V(a_i)
$$

Earlier, we introduced the notation $1,\ldotsj, k$ to indicate the
sequence $1,..., j-1, j+1,..., k$.  Now, we introduce additional
notation for constructing elements of the pair semiring.  Let 
$$
\bigotimesbar{i = 1,\ldotsj, k} a_i= \lfloor \bigotimes_{i = 1}^{j-1}
a_i, \bigotimes_{i = j\!+\!1}^k a_i \rfloor
$$
We need to show
\begin{mylemma}
$\bigotimesbar{i=1,\ldotsj,k}$ distributes over $\bigoplus$.
\end{mylemma}
{\em Proof \hspace{3em}} Let $H_1,\ldotsj,H_k$ be index sets such that
for $h_i \in H_i$, $x_{h_i}$ is a value in $ \Bbb{A}$.  Then
\begin{eqnarray*}
\bigotimesbar{i=1,\ldotsj,k} \bigoplus_{h_i \in H_i}  x_{h_i} 
& = & \left\lfloor \bigotimes_{i=1,...,j-1} \;\bigoplus_{h_i \in H_i}  x_{h_i}, \bigotimes_{i=j\!+\!1,...,k} \; \bigoplus_{h_i \in H_i}  x_{h_i} \right\rfloor \\
& = & \left\lfloor \bigotimes_{i=1,...,j-1} \;\bigoplus_{h_i \in H_i}  x_{h_i}, \;\; 1 \right\rfloor \otimes \left\lfloor 1, \; \bigotimes_{i=j\!+\!1,...,k} \;\bigoplus_{h_i \in H_i}  x_{h_i} \right\rfloor \\ 
& = & \bigotimes_{i=1,...,j-1} \left\lfloor \bigoplus_{h_i \in H_i}  x_{h_i}, \;\; 1 \right\rfloor \otimes \bigotimes_{i=k,...,j\!+\!1} \left\lfloor 1, \; \bigoplus_{h_i \in H_i}  x_{h_i} \right\rfloor \\
& = & \bigotimes_{i=1,...,j-1} \;\bigoplus_{h_i \in H_i} \left\lfloor x_{h_i}, 1 \right\rfloor \otimes \bigotimes_{i=k,...,j\!+\!1} \;\bigoplus_{h_i \in H_i} \left\lfloor 1, x_{h_i} \right\rfloor \\
& = & \bigoplus_{h_1 \in H_1,...,h_{j-1} \in H_{j-1}}
\bigotimes_{i=1...j\!-\!1} \left\lfloor x_{h_i}, 1 \right\rfloor \otimes
\bigoplus_{h_{j\!+\!1}\in H_{j\!+\!1},...,h_k \in H_k} \;\bigotimes_{i=k,...,j\!+\!1} \left\lfloor 1, x_{h_i} \right\rfloor \\
& = & \bigoplus_{h_1 \in H_1,\ldotsj,h_k \in H_k} \left(\bigotimes_{i=1,...,j-1} \left\lfloor x_{h_i}, 1 \right\rfloor \otimes \bigotimes_{i=k,...,j\!+\!1} \left\lfloor 1, x_{h_i} \right\rfloor\right) \\
& = & \bigoplus_{h_1 \in H_1,\ldotsj,h_k \in H_k} \left\lfloor \bigotimes_{i=1,...,j-1} x_{h_i}, \bigotimes_{i=j\!+\!1,...,k} x_{h_i} \right\rfloor \\ 
& = & \bigoplus_{h_1 \in H_1,\ldotsj,h_k \in H_k} \; \bigotimesbar{i=1,\ldotsj,k} x_{h_i} 
\end{eqnarray*}
$\Box$

Now, we can give a simple formula for computing reverse values $Z(x)$
not involved in loops:

\begin{mytheorem}
For items $x \in B$ where $B$ is non-looping, 
$$
Z(x) = \bigoplus_{j, a_1... a_k, b
\;\mbox{\scriptsize s.t.}\;\smallinfer{a_1... a_k}{b}{}\wedge x=a_j} Z(b)
\otimes
\bigotimesbar{i = 1 ,\ldotsj, k} V(a_i)
$$
unless $x$ is the goal item, in which case $Z(x) = \lfloor 1, 1 \rfloor$, the
multiplicative identity of the pair semiring.
\end{mytheorem}

{\em Proof \hspace{3em}} We begin with the goal item.  As before, the
outer trees of the goal item are all empty. Thus,
\begin{eqnarray*}
Z(\mi{goal}) & = &
\bigoplus_{D\in\smi{outer}(\smi{goal})} \lfloor Z_L(D), Z_R(D) \rfloor \\
& = &
\lfloor Z_L(\{\langle\rangle\}), Z_R(\{\langle\rangle\}) \rfloor \\
& = &
\lfloor \bigotimes_{r\in\{\langle\rangle\}} R(r) ,
\bigotimes_{r\in\{\langle\rangle\}} R(r)  \rfloor \\
& = &
\lfloor 1,1 \rfloor
\end{eqnarray*}

Using the same observation as in Theorem \ref{theorem:simplereverse},
that every outer tree of $a_j$, $D_{a_j},$ can be described as a
combination of the surrounding outer tree of $b$, $D_b$ and inner
trees of $a_1,\ldotsj, a_k$ where $\shortinfer{a_1... a_k}{b}{}$, and
observing that the value of such an outer tree is $Z(D_b) \otimes
\bigotimesbar{i=1,\ldotsj,k} V(D_{a_i})$,
\begin{eqnarray*}
Z(x) & = & \bigoplus_{D\in \smi{outer}(x)} Z(D)\\
& = & \bigoplus_{j, a_1... a_k, b
\smallst \smallinfer{a_1... a_k}{b}{}\wedge x=a_j} 
\;\bigoplus_{D\in
\smi{outer}\left(j,\smallinfer{a_1... a_k}{b}{}\right)} Z(D) \\
& =&
\bigoplus_{j, a_1... a_k, b \smallst \smallinfer{a_1... a_k}{b}{}\wedge x=a_j}
\bigoplus_{\longersum{D_{a_1}\in \smi{inner}(a_1)
,\ldotsj,}{\sumindent D_{a_k}\in \smi{inner}(a_k),}
{D_b\in \smi{outer}(b)}} Z(D_b) 
\otimes
\bigotimesbar{i=1,\ldotsj,k} V(D_{a_i}) \\
& = &
\bigoplus_{j, a_1... a_k, b \smallst \smallinfer{a_1... a_k}{b}{}\wedge x=a_j} 
\bigoplus_{D_b\in \smi{outer}(b)}  Z(D_b)
\otimes 
\bigoplus_{\longsum{D_{a_1}\in \smi{inner}(a_1) ,\ldotsj,}{D_{a_k}\in
\smi{inner}(a_k)}} \;\;
\bigotimesbar{i=1,\ldotsj,k}V(D_{a_i}) \\
& = &
\bigoplus_{j, a_1... a_k, b \smallst \smallinfer{a_1... a_k}{b}{}\wedge x=a_j} 
Z(b) 
\otimes
\bigotimesbar{i=1,\ldotsj,k}  \;
\bigoplus_{D_{a_i}\in \smi{inner}(a_i)} V(D_{a_i}) \\
& = &
\bigoplus_{j, a_1... a_k, b \smallst \smallinfer{a_1... a_k}{b}{}\wedge x=a_j} 
Z(b) \otimes \bigotimesbar{i=1,\ldotsj,k} V(a_i) 
\end{eqnarray*}
completing the general case.  $\Box$

The case for looping buckets, is, as usual, somewhat more complicated,
but follows Theorem \ref{theorem:reverseloop} very closely.  As in the
commutative case, it requires an infinite sum.  This infinite sum is
why we so carefully made sure that the pair semiring is
$\omega$-continuous: we wanted to make sure that we could easily
handle the infinite case, by using the properties of
$\omega$-continuous semirings.

Recall that $\mi{outer}_{\leq g}(x, B)$ represents the set of outer
trees of $x$ with generation at most $g$.  Now, we can define the left
and right $\leq g$ generation reverse value of an item $x$ in bucket
$B$
$$
Z_{\leq g}(x, B) = \bigoplus_{D\in\smi{outer}_{\leq g}(x, B)} \lfloor
Z_L(D), Z_R(D) \rfloor \\
$$

Since \PA is an $\omega$-continuous semiring, an infinite sum is equal to the
supremum of the partial sums:
$$
\bigoplus_{D\in\smi{outer}(x, B)} Z(D) = Z_{\leq \infty}(x, B) = \sup_{g} Z_{\leq g}(x, B)
$$
Thus, as before, we wish to find a simple formula for $Z_{\leq
g}(x, B)$.

Recall that for $x$ in a bucket following $B$, $Z_{\leq g}(x, B) =
Z(x)$ and that for $x \in B$, $Z_{\leq 0}(x, B) = 0$, providing a base
case.  We can then address the general case, for $g \geq 1$:

\begin{mytheorem}
For $x \in B$ and $g \geq 1$,
$$
Z_{\leq g}(x, B) = 
\bigoplus_{{j, a_1...a_k, b}\smallst\smallinfer{a_1... a_k}{b}{\wedge x=a_j}}
\left(
\left\{\begin{array}{ll}
{Z_{\leq g\!-\!1}(b, B)}&{\mbox{ \em if } b\in B} \\
{Z(b)}&{\mbox{ \em if } b\notin B}
\end{array}
\right.
\right)
\otimes
\bigotimesbar{i =1 ,\ldotsj, k} V(a_i) 
$$
\end{mytheorem}

{\em Proof \hspace{3em}} Recall that $\mi{MakeOuter}(j, D_{a_1},\ldotsj, D_{a_k}, D_b)$
is a function that puts together the specified trees to form an
outer tree for $a_j$.  Then,
\begin{eqnarray}
\lefteqn{
Z_{\leq g}(x, B)}
\nonumber \\ &=& 
\bigoplus_{D\in\smi{outer}_{\leq g}(x, B)} Z(D) =
\nonumber \\ &=&
\!\!\!\!\!\!\bigoplus_{{j, a_1...a_k, b}\smallst
\smallinfer{a_1... a_k}{b}{\wedge x=a_j}}
\;\;\bigoplus_{\longersum
{D_{b}\in\smi{outer}_{\leq g\!-\!1}(b, B),}
{D_{a_1}\in\smi{inner}(a_1),\ldotsj,}
{\sumindent D_{a_k}\in\smi{inner}(a_k)}
}
\!\!\!\!\!\!\!\!\!\! Z(\mi{MakeOuter}(j, D_{a_1},\ldotsj, D_{a_k}, D_b)) =
\nonumber \\ &=&
\!\!\!\!\!\!\bigoplus_{{j, a_1...a_k, b}\smallst
\smallinfer{a_1... a_k}{b}{\wedge x=a_j}}
\;\;\bigoplus_{\longersum
{D_{b}\in\smi{outer}_{\leq g\!-\!1}(b, B),}
{D_{a_1}\in\smi{inner}(a_1),\ldotsj,}
{\sumindent D_{a_k}\in\smi{inner}(a_kB)}
}
\;\; Z(D_b) \otimes \bigotimesbar{i = 1,\ldotsj, k} V(D_{a_i}) =
\nonumber \\ &=&
\!\!\!\!\!\!\bigoplus_{{j, a_1...a_k, b}\smallst
\smallinfer{a_1... a_k}{b}{\wedge x=a_j}}
\bigoplus_{D_{b}\in\smi{outer}_{\leq g\!-\!1}(b, B)} \!\! \!\!\! Z(D_b) 
\otimes \!\!\!\! \!\!\!\!\!
\bigoplus_{\longsum
{D_{a_1}\in\smi{inner}(a_1),\ldotsj,}
{D_{a_k}\in\smi{inner}(a_k)}}
\bigotimesbar{i = 1,\ldotsj, k} \!\!V(D_{a_i}) 
=
\nonumber \\ &=&
\!\!\!\!\!\!\bigoplus_{{j, a_1...a_k, b}\smallst\smallinfer{a_1... a_k}{b}{\wedge x=a_j}}
Z_{\leq g\!-\!1}(b, B) \otimes
\bigotimesbar{i =1 ,\ldotsj, k} \; \bigoplus_{D_{a_i}\in\smi{inner}(a_i,B)} V(D_{a_i})
 =
\nonumber \\ &=&
\!\!\!\!\!\!\bigoplus_{{j, a_1...a_k, b}\smallst \smallinfer{a_1... a_k}{b}{\wedge x=a_j}}
Z_{\leq g\!-\!1}(b, B)  \otimes
\bigotimesbar{i =1 ,\ldotsj, k} V(a_i) \label{eqn:lastouternoncom}
\end{eqnarray}
Substituting Equation \ref{eqn:notinBouter} into Equation
\ref{eqn:lastouternoncom}, we get
$$
Z_{\leq g}(x, B)  =
\!\!\!\!\!\!\bigoplus_{{j, a_1...a_k, b}\smallst\smallinfer{a_1... a_k}{b}{\wedge x=a_j}}
\left(
\left\{\begin{array}{ll}
{Z_{\leq g\!-\!1}(b, B)}&{\mbox{ \em if } b\in B} \\
{Z(b)}&{\mbox{ \em if } b\notin B}
\end{array}
\right.
\right)
\otimes
\bigotimesbar{i =1 ,\ldotsj, k} V(a_i) 
$$
$\Box$

%\begin{figure}
%$$
%\begin{array}{c}
%\mbox{\bf Item form:} \\
%\\
%\mbox{\bf Goal}\\
%\\
%\mbox{\bf Rules:} \\
%\end{array}
%$$
%\caption{}\label{fig:}
%\end{figure}

} %end beginappendix

% -*- mode: latex; -*-

\chapter{Maximizing Metrics} \label{ch:max}

It is well known how to find the parse with the highest probability of
being exactly right.  In this chapter, we show how to use the
inside-outside probabilities to find parses that are the best in
other senses, such as maximizing the expected number of correct constituents
\cite{Goodman:96a}.  We will use these algorithms in the next chapter
to parse the DOP model efficiently.

\section{Introduction}

In corpus-based approaches to parsing, researchers are given a
treebank (a collection of text annotated with the ``correct'' parse
tree).  The researchers then attempt to find algorithms that, given
unlabelled text from the treebank, produce as similar a parse as
possible to the one in the treebank.

Various methods can be used for finding these parses.  Some of the
most common involve inducing Probabilistic Context-Free Grammars
(PCFGs), and then using an algorithm, such as the Labelled Tree
(Viterbi) algorithm, which maximizes the probability that the output
of the parser (the ``guessed'' tree) is the one that the PCFG
produced.

There are many different ways to evaluate the output parses.  In
Section \ref{sec:evalmet}, we will define them, using for consistency
our own terminology.  We will often include the conventional term in
parentheses for those measures with standard names.  The most
common evaluation metrics include the Labelled Tree rate (also called
the Viterbi Criterion or Exact Match rate), Consistent Brackets Recall
rate (also called the Crossing Brackets rate), Consistent Brackets
Tree rate (also called the Zero Crossing Brackets rate), and Precision
and Recall.  Despite the variety of evaluation metrics, nearly all
researchers use algorithms that maximize performance on the Labelled
Tree rate, even when they evaluate performance using other criteria.

We propose that by creating algorithms that optimize the evaluation
criterion, rather than some general criterion, improved performance
can be achieved.  

There are two commonly used kinds of parse trees.  In one kind, binary
branching parse trees, the trees are constrained to have exactly two
branches for each nonterminal node.  In the other kind, n-ary
branching parse trees, each nonterminal node may have any number of
branches; however, in this chapter we will only be considering n-ary
trees with at least two branches.

In the first part of this chapter, we discuss the binary branching
case.  In Section \ref{sec:evalmet}, we define most of the evaluation
metrics used in this chapter and discuss previous approaches.  Then,
in Section \ref{sec:labconspars}, we discuss the Labelled Recall
algorithm, a new algorithm that maximizes performance on the Labelled
Recall rate.  In Section \ref{sec:brackconspars}, we discuss another
new algorithm, the Bracketed Recall algorithm, that maximizes
performance on the Bracketed Recall rate, closely related to the
Consistent Brackets Recall (Crossing Brackets) rate.  In Section
\ref{sec:results1}, we present experimental results using these two
algorithms on appropriate tasks, and compare them to the Viterbi
algorithm.  Next, in Section \ref{sec:npcomplete}, we show that
optimizing a similar criterion, the Bracketed Tree rate, which
resembles the Consistent Brackets Tree rate, is NP-Complete.

In the second part of the chapter, we extend these results in two
ways.  First, in Section \ref{sec:generalalg}, we show that the
algorithms of the first part are both special cases of a more general
algorithm, and give some potential applications for this more general
algorithm.  Second, in Section \ref{sec:nary}, we generalize the
algorithm further, so that it can handle n-ary branching parses.  In
particular, we give definitions for Recall, Precision, and a new,
related measure, Mistakes.  We then show how to maximize the weighted
difference of Recall minus Mistakes, which we call the Combined rate.
Finally, we give experimental results for the n-ary branching case.

\section{Evaluation Metrics}
\label{sec:evalmet}

In this section, we first define some of the basic terms and symbols,
including {\em parse tree}, and then specify the different kinds of
errors parsing algorithms can make.  Next, we define the different
metrics used in evaluation.  Finally, we discuss the relationship of
these metrics to parsing algorithms.

In this chapter, we spend quite a bit of time on the topic of parsing
metrics, and the use of a consistent naming convention will simplify
the discussion.  However, this consistent naming convention is not
standard, and so will only be used in this chapter.  A glossary of
these terms is provided in Appendix \ref{sec:glossary}.

\subsection{Basic Definitions}

In this chapter, we are primarily concerned with scoring parse trees
and finding parse trees that optimize certain scores.  Parse trees are
typically scored by the number of their constituents that are
correct, according to some measure, so it will be convenient in this
chapter to define parse trees as sets of constituents.  As we have
done throughout this thesis, we will let $w_1...w_n$ denote the
sequence of terminals (words) in the sentence under discussion.  Then
we define a parse tree $T$ as a set of constituents $\constituent{i}{X}{j}$.  A
triple $\constituent{i}{X}{j}$ indicates
that $w_iw_{i+1}... w_{j-1}$ can be parsed as a terminal or nonterminal
$X$.  The triples must meet the following requirements, which
enforce binary branching constraints and consistency:
\begin{itemize}

\item The sentence was generated by the start symbol, $S$.  Formally,
$\constituent{1}{S}{n+1} \in T$ and for all $X \neq S$,
$\constituent{1}{X}{n+1} \notin T$.
%
%\item Every word in the sentence is in the parse tree.  Formally,
%for every $i$ between 1 and $n$ ther  triple $\constituent{i}{w_i}{i} \in T$.
%
\item The tree is binary branching and consistent.  Formally, for
every $\constituent{i}{X}{j}$ in $T$, $i \neq j-1$, there is exactly one $k, Y,$ and $Z$
such that $i < k < j$ and $\constituent{i}{Y}{k} \in T$ and $\constituent{k}{Z}{j} \in T$.
%
%\item The tree does not contain epsilons.  Formally, there is no $\constituent{i,
%i, X}$ in $T$.
\end{itemize}

Let $T_C$ denote the ``correct'' parse (the one in the tree bank) and
let $T_G$ denote the ``guessed'' parse (the one output by the parsing
algorithm).  Let $N_C$ denote $\vert T_C \vert$, the number of
vocabulary symbols in the correct parse tree, and let $N_G$ denote
$\vert T_G \vert$, the number of vocabulary symbols in the guessed
parse tree.

\subsection{Evaluation Metrics}
There are various levels of strictness for determining whether a
constituent (element of $T_G$) is ``correct.''  The strictest of these
is {\em Labelled Match}.  A constituent $\constituent{i}{X}{j} \in T_G$ is correct
according to Labelled Match if and only if $\constituent{i}{X}{j} \in T_C.$ In
other words, a constituent in the guessed parse tree is correct if and
only if it occurs in the correct parse tree.

The next level of strictness is {\em Bracketed Match}.  Bracketed match is
like Labelled Match, except that the nonterminal label is ignored.
Formally, a constituent $\constituent{i}{X}{j} \in T_G$ is correct according to
Bracketed Match if and only if there exists a $Y$ such that $\constituent{i}{Y}{j}
\in T_C.$

\begin{figure}
\begin{center}
\begin{tabular}{ccccc}
\psfig{figure=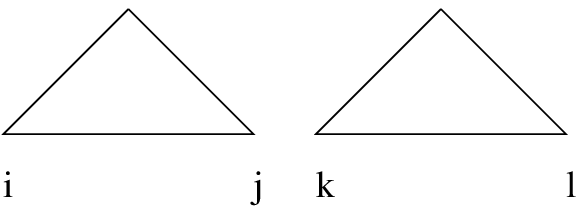} &
\hspace{0.15in} &
\psfig{figure=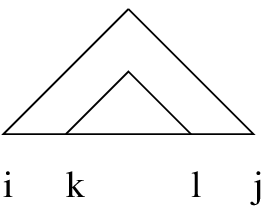} &
\hspace{0.15in} &
\psfig{figure=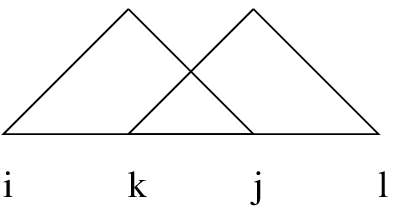} \\
Doesn't cross & &
Doesn't cross & &
Crosses
\end{tabular}
\end{center}
\caption{Non-crossing and crossing constituents}\label{fig:crosses}
\end{figure}

The least strict level is {\em Consistent Brackets} (traditionally
misnamed Crossing Brackets).  Consistent Brackets is like Bracketed
Match in that the label is ignored.  It is even less strict in that
the observed $\constituent{i}{X}{j}$ need not be in $T_C$---it must
simply not be ruled out by any $\constituent{k}{Y}{l} \in T_C$.  A
particular triple $\constituent{k}{Y}{l}$ rules out
$\constituent{i}{X}{j}$ if there is no way that
$\constituent{i}{X}{j}$ and $\constituent{k}{Y}{l}$ could both be in
the same parse tree.  Figure \ref{fig:crosses} shows examples of
non-crossing and crossing constituents.  In particular, if the
interval $\constituenttwo{i}{j}$ crosses the interval
$\constituenttwo{k}{l}$, then $\constituent{i}{X}{j}$ is ruled out and
counted as an error.  Formally, we say that $\constituenttwo{i}{j}$
crosses $\constituenttwo{k}{l}$ if and only if $i < k < j < l$ or $j <
i < k < l$.\footnote{This follows \newcite{Pereira:92a}, except that
in our notation spans include the first element but not the last.}

If $T_C$ is binary branching, then Consistent Brackets and Bracketed
Match are identical (a proof of this fact is given in Appendix
\ref{sec:crossproof}, immediately following this chapter).  The
following symbols denote the number of constituents that match
according to each of these criteria.

\begin{list}{}
\item $L = \vert \{ T_C \cap T_G\} \vert$: the number of constituents in $T_G$
that are correct according to Labelled Match.

\item $B = \vert \{\constituent{i}{X}{j} : \constituent{i}{X}{j} \in
T_G$ and for some $Y, \constituent{i}{Y}{j} \in T_C\}\vert $: the
number of constituents in $T_G$ that are correct according to
Bracketed Match.

\item $C = \vert \{\constituent{i}{X}{j} :\constituent{i}{X}{j} \in
T_G$  and there is no $\constituent{k}{Y}{l}
\in T_C$ crossing $\constituent{i}{X}{j}\} \vert$ : the number of constituents
in $T_G$ correct according to Consistent Brackets.

\end{list}

\smallskip

Following are the definitions of the six metrics used in this chapter for
evaluating binary branching trees:

\newcommand{\oneif}{\mbox{ 1 \it if }}
\newcommand{\bigoneif}[1]{\left\{\begin{array}{ll}1 & \mbox{\em if } #1 \\ 0 &
\mbox{\it otherwise} \end{array}\right.}

\begin{itemize}

\item {\em Labelled Recall Rate} = $L/N_C$

\item {\em Labelled Tree Rate} = $\bigoneif{L = N_C}$  \\This metric is also called 
the Viterbi criterion or the Exact Match rate.

\item {\em Bracketed Recall Rate} = $B/N_C$

\item {\em Bracketed Tree Rate} = $\bigoneif{B = N_C}$

\item {\em Consistent Brackets Recall Rate} = $C/N_G$  \\ This metric is often
called the Crossing Brackets rate.  In the case where the parses are
binary branching, this criterion is the same as the Bracketed Recall rate.

\item {\em Consistent Brackets Tree rate} = $\bigoneif{C = N_G}$ \\ This
metric is closely related to the Bracketed Tree rate.  In the case
where the parses are binary branching, the two metrics are the same.
This criterion is also called the Zero Crossing Brackets rate.

\end{itemize}

\noindent The preceding six metrics each correspond to cells in the following
table:

\smallskip

\begin{center}
\begin{tabular}{|l||r|r|} \hline
	          & Recall         & Tree                \\ \hline
                                                              \hline
Consistent Brackets & $C / N_G$            & $\bigoneif{C = N_G}$     \\
\hline
Bracketed          & $B / N_C$            & $\bigoneif{B = N_C}$     \\
\hline
Labelled            &  $L / N_C$           &  $\bigoneif{L = N_C}$      \\
\hline
\end{tabular}
\end{center}

\smallskip

We will define metrics for n-ary branching trees, including Precision
and Recall, in Section \ref{sec:nary}.

\subsection {Maximizing Metrics}
\label{sec:submaxmetrics}

Although several metrics are available for evaluating parsers, there is
only one metric most parsing algorithms attempt to maximize, namely
the Labelled Tree rate.  That is, most parsing algorithms attempt to
solve the following problem:
\begin{equation}
\label{eqn:labtremax}
\arg\max_{T_G} E\left(\left. \bigoneif{L = N_C}\right|\mathcal{M}\right)
\end{equation}
Here, $E$ is the expected value operator, and $\mathcal{M}$ is the
model under consideration, typically a probabilistic grammar.  In
words, this maximization finds that tree $T_G$ which maximizes the
expected Labelled Tree score, assuming that the sentence was generated
by the model.  Equation \ref{eqn:labtremax} is equivalent to
maximizing the probability that $T_G$ is exactly right.  The
assumption that the sentence was generated by the model, or at least
that the model is a good approximation, is key to any parsing
algorithm.  The Viterbi algorithm, the prefix probability estimate of
\newcite{Jelinek:91b}, language modeling using PCFGs, and n-best
parsing algorithms all implicitly make this approximation: without it,
very little can be determined.  Whether or not the approximation is a
good one is an empirical question.  Later, we will show experiments,
using grammars obtained in two different ways, that demonstrate that
the algorithms we derive using this approximation do indeed work well.
Since essentially every equation in this chapter requires this
approximation, we will implicitly assume the conditioning on
$\mathcal{M}$ from now on.

The maximization of Equation \ref{eqn:labtremax} is used by most
parsing algorithms, including the Labelled Tree (Viterbi) algorithm
and stochastic versions of Earley's algorithm \cite{Stolcke:93a}, and
variations such as those used in Picky parsing \cite{Magerman:92a},
and in current state-of-the-art systems, such as those of
\newcite{Charniak:97a} and \newcite{Collins:97a}.  Even in
probabilistic models not closely related to PCFGs, such as Spatter
parsing \cite{Magerman:94a}, Expression \ref{eqn:labtremax} is still
computed.  One notable exception is Brill's Transformation-Based Error
Driven system \cite{Brill:93a}, which induces a set of transformations
designed to maximize the Consistent Brackets Recall (Crossing
Brackets) rate.  However, Brill's system does not induce a PCFG, and
the techniques Brill introduced are not as powerful as modern
probabilistic techniques.  We will show that by matching the parsing
algorithm to the evaluation criteria, better performance can be
achieved for the more commonly used PCFG formalism, and 
variations.

Ideally, one might try to directly maximize the most commonly used
evaluation criteria, such as the Consistent Brackets Recall (Crossing
Brackets) rate or Consistent Brackets Tree (Zero Crossing Brackets)
rate.  However, these criteria are relatively difficult to
maximize, since it is time-consuming to compute the probability that a
particular constituent crosses some constituent in the correct parse.
On the other hand, the Bracketed Recall and Bracketed Tree rates are
easier to handle, since computing the probability that a bracket
matches one in the correct parse is not too difficult.  It is
plausible that algorithms which optimize these closely related
criteria will do well on the analogous Consistent Brackets criteria.

\subsection{Which Metrics to Use} 
When building a system, one should use the metric most appropriate for
the target problem.  For instance, if one were creating a database
query system, such as an automated travel agent, then the Labelled
Tree (Viterbi) metric would be most appropriate.  This is because a
single error in the syntactic representation of a query will likely
result in an error in the semantic representation, and therefore in an
incorrect database query, leading to an incorrect result.  For
instance, if the user request ``Find me all flights on Tuesday'' is
misparsed with the prepositional phrase attached to the verb, then the
system might wait until Tuesday before responding: a single error
leads to completely incorrect behavior.

On the other hand, for a machine-assisted translation system, in which
the system provides translations, and then a human fluent in the
target language manually edits them, Labelled Recall is more
appropriate.  If the system is given the foreign language equivalent
of ``His credentials are nothing which should be laughed at,'' and
makes the single mistake of attaching the relative clause at the
sentential level, it might translate the sentence as ``His credentials
are nothing, which should make you laugh.''  With this translation,
the human translator must make some changes, but certainly needs to do
less editing than if the sentence were completely misparsed.  The more
errors there are, the more editing the human translator needs to do.
Thus, a criterion such as Labelled Recall is appropriate for this
task, where the number of incorrect constituents correlates to
application performance.

\section{Labelled Recall Parsing}
\label{sec:labconspars}

The Labelled Recall parsing algorithm finds that tree $T_G$ that has the
highest expected value for the Labelled Recall rate, $L/N_C$ (where
$L$ is the number of correctly labelled constituents, and $N_C$ is the
number of nodes in the correct parse).  Formally, the algorithm finds
\begin{equation} \label{eqn:max1}
T_G=\arg \max_T E(L/N_C)
\end{equation}

\begin{figure}
\begin{center}
$$
\begin{array}{rcll}
S & \rightarrow  & A \; C  & 0.25 \\
S & \rightarrow  & A \; D  & 0.25 \\
S & \rightarrow  & E \; B  & 0.25 \\
S & \rightarrow  & F \; B  & 0.25 \\
A, B, C, D, E, F & \rightarrow & x x & 1.0  \\
\end{array}
$$
Sample grammar illustrating Labelled Recall

\bigskip

%
% Use faketreewidth to work around a bug
%
\hspace{-.5in}\begin{tabular}{cc}
\leaf{$x$} \faketreewidth{AA}
\leaf{$x$} \faketreewidth{AA}
\branch{2}{$A$}
\leaf{$x$} \faketreewidth{AA}
\leaf{$x$} \faketreewidth{AA}
\branch{2}{$C$}
\faketreewidth{AA}
\branch{2}{$S$}
\faketreewidth{SS}
\tree &
\leaf{$x$} \faketreewidth{AA}
\leaf{$x$} \faketreewidth{AA}
\branch{2}{$A$}
\leaf{$x$} \faketreewidth{AA}
\leaf{$x$} \faketreewidth{AA}
\branch{2}{$D$}
\faketreewidth{AA}
\branch{2}{$S$}
\faketreewidth{SS}
\tree \\
\leaf{$x$} \faketreewidth{AA}
\leaf{$x$} \faketreewidth{AA}
\branch{2}{$E$}
\leaf{$x$} \faketreewidth{AA}
\leaf{$x$} \faketreewidth{AA}
\branch{2}{$B$}
\faketreewidth{AA}
\branch{2}{$S$}
\faketreewidth{SS}
\tree &
\leaf{$x$} \faketreewidth{AA}
\leaf{$x$} \faketreewidth{AA}
\branch{2}{$F$}
\leaf{$x$} \faketreewidth{AA}
\leaf{$x$} \faketreewidth{AA}
\branch{2}{$B$}
\faketreewidth{AA}
\branch{2}{$S$}
\faketreewidth{SS}
\tree \\
\end{tabular} \\
Four trees in the sample grammar

\bigskip

\leaf{$x$} \faketreewidth{AA}
\leaf{$x$} \faketreewidth{AA}
\branch{2}{$A$}
\leaf{$x$} \faketreewidth{AA}
\leaf{$x$} \faketreewidth{AA}
\branch{2}{$B$}
\faketreewidth{SS}
\branch{2}{$S$}
\tree \\
Labelled Recall tree

\end{center}
\caption{Four trees of sample grammar, and Labelled Recall tree}
\label{fig:labrecallexample}
\end{figure}

The difference between the Labelled Recall maximization of Expression
\ref{eqn:max1} and the Labelled Tree maximization of Expression
\ref{eqn:labtremax} may be seen from the following example.  Figure
\ref{fig:labrecallexample} gives an example grammar that generates
four trees with equal probability.  For the top left tree in Figure
\ref{fig:labrecallexample}, the probabilities of being correct are
$S$: 100\%; $A$:50\%; and $C$: 25\%.  Similar counting holds for the
other three.  Thus, the expected value of $L$ for any of these trees
is 1.75.

In contrast, the optimal Labelled Recall parse is shown in the
bottom of Figure \ref{fig:labrecallexample}.  This tree has 0
probability according to the grammar, and thus is non-optimal
according to the Labelled Tree rate criterion.  However, for this tree
the probabilities of each node being correct are $S$: 100\%; $A$: 50\%;
and $B$: 50\%.  The expected value of $L$ is 2.0, the highest of any
tree.  This tree therefore optimizes the Labelled Recall rate.

\subsection{Formulas}
\label{sec:labelledalgorithm}
We now derive an algorithm for finding the parse that maximizes the
expected Labelled Recall rate.  We do this by expanding
Expression \ref{eqn:max1} out into a probabilistic
form, converting this into a recursive equation, and finally creating
an equivalent dynamic programming algorithm.

We begin by rewriting Expression \ref{eqn:max1}, expanding out the
expected value operator, and removing the $\frac{1}{N_C}$, which is
the same for all $T_G$, and so plays no role in the maximization.
\begin{equation} \label{eqn:max2}
\arg\max_{T_G} \sum_{T_C} P(T_C\mid w_1...w_n) \; \vert T_G \cap T_C \vert
\end{equation}
This can be further expanded to
\begin{equation} \label{eqn:max3}
\arg \max_{T_G} \sum_{T_C} P(T_C \mid w_1...w_n) \sum_{\constituent{i}{X}{j} \in T_G} 
\bigoneif{\constituent{i}{X}{j} \in T_C}
\end{equation}
Now, given a Probabilistic Context-Free Grammar $G$ with start symbol $S$,
the following equality holds:

$$
P(S \derives w_1 ... w_{i-1} X w_j ... w_n | w_1 ... w_n) =
\sum_{T_C} P(T_C|w_1...w_n) \times \bigoneif{\constituent{i}{X}{j} \in T_C}
$$

By rearranging the summation in Expression \ref{eqn:max3} and then  
substituting this equality, we get

$$
\arg \max_{T_G} \sum_{\constituent{i}{X}{j} \in T_G}
P(S \derives w_1 ... w_{i-1} X w_j ... w_n | w_1 ... w_n)
$$

At this point, it is useful to recall the inside and outside
probabilities, introduced in Section \ref{sec:background}.  Recall
that the inside probability is defined as $\mi{inside}(i, X, j) = P(X \derives w_i ... w_{j-1})$ and the outside probability is
$\mi{outside}(i,X,j) = P(S \derives w_1 ... w_{i-1} X
w_j ... w_n).$

Let us define a new symbol, $g(i, X, j)$.
\begin{eqnarray*}
g(i, X, j) &=&  P(S \derives w_1 ... w_{i-1} X w_{j} ... w_n | w_1
... w_n) \\
           &=& \frac{P(S \derives w_1...w_{i-1} X w_{j}...w_n ) \times
                     P(X \derives w_i...w_{j-1}) }
                    { P(S \derives w_1...w_n) } \\
           &=& \frac{\mi{outside}(i,X,j) \times
\mi{inside}(i,X,j)}{\mi{inside}(1, S, n+1)}
\end{eqnarray*}
%
%Recall from Section \ref{sec:PCFGintroduction} that $g$ can be written
%in the inside and outside probabilties:
%
Now, the definition of a Labelled Recall Parse can be rewritten as
$$
\arg \max_{T_G} \sum_{\constituent{i}{X}{j} \in T_G} g(i, X, j)
$$

\def\MAXC{\mathop{\rm MAXC}\nolimits}

\subsection{Pseudocode Algorithm}

Given the values of $g(i, X, j)$, it is a simple matter of dynamic
programming to determine the parse that maximizes the Labelled Recall
rate.  Define
\begin{equation}
\mi{MAXC}(i, j) = 
\max_{X} g(i, X, j) +  
\left\{ \begin{array}{ll} \max_{k \smallst i \leq k < j} \mi{MAXC}(i,
k) +\mi{MAXC}(k, j)&
\mbox{\em if } i \neq j-1 \\
0 & \mbox{\em if } i = j-1
\end{array} \right.
\label{eqn:maxc}
\end {equation}

We now show that this recursive equation is correct, with a simple
proof.  We will write $T_{i, j}$ to represent a subtree covering $w_i... w_{j-1}$
and we will write $L(T_{i, j})$ to represent a function giving the expected
number of correctly labelled constituents in $T_{i, j}$.
\begin{mytheorem}
$$
\mi{MAXC}(i, j) = \max_{T_{i, j}} L(T_{i, j})
$$
\end{mytheorem}
\startproof The proof is by induction over the size of the parse tree,
$j-i$.  The base case is parse trees of size 1, when $i=j-1$.  In this
case, there is a single constituent in the possible parse trees.
Thus,
\begin{eqnarray*}
\max_{T_{i, j}} L(T_{i, j}) 
& = & \max_{\{\constituent{i}{X}{j}\}} L(\{\constituent{i}{X}{j}\}) \\
& = & \max_{\{\constituent{i}{X}{j}\}} g(i, X, j) \\
& = & \max_{X} g(i, X, j) \\
& = & \mi{MAXC}(i, j)
\end{eqnarray*}
completing the base case.

In the inductive step, assume the theorem for lengths less than $j-i$.
Then, we notice that for trees of size greater than one, we can always
break down a maximal subtree
$$
T'_{i,j} = \arg \max_{T_{i, j}} L(T_{i, j})
$$
into a root node, $\constituent{i}{X}{j}$ and two smaller maximal
child trees:
$$T'_{i,k} = \arg \max_{T_{i, k}} L(T_{i, k})$$
$$T'_{k,j} = \arg \max_{T_{k, j}} L(T_{k, j})$$ 
Notice that if we were trying to maximize the Labelled Tree rate, the
most probable child trees would depend on the
parent nonterminal.  However,
when we maximize the Labelled Recall rate, the child trees do not
depend on the parent, as was illustrated in Figure
\ref{fig:labrecallexample}, where we showed that in fact, the joint
probability of the parent nonterminal and child trees could even be zero.
Thus,
\begin{eqnarray*}
\max_{T_{i, j}} L(T_{i, j}) & =& 
\max_X g(i, X, j) + 
\max_k \left( 
\max_{T_{i, k}} L(T_{i, k})  +
\max_{T_{k, j}} L(T_{k, j}) 
\right) \\
& = &
 \max_X g(i, X, j) + 
\max_k \left( \mi{MAXC(i, k)} + \mi{MAXC(k, j)} \right) \\
& = & \mi{MAXC}(i, j) \mbox{ \em if } i \neq j-1
\end{eqnarray*}
which completes the inductive step. $\Box$

Notice that $\mi{MAXC}(1, n+1)$ contains the score of the best parse
according to the Labelled Recall rate.

\begin{figure}
\begin{tabbing}
\verb|   |\=\verb|   |\=\verb|   |\=\verb|   |\=\verb|   |\=\verb|   |\=\verb|   |\=\verb|   |\=\verb|   |\=\verb|   |\= \kill
$\mi{float maxC}[1..n, 1..n\!+\!1] := 0;$ \\
\alfor $\mi{length} := 1$ \alto $n$ \\
 \> \alfor $i := 1$ \alto $n-\mi{length}+1$ \\
 \>  \> $j := i + \mi{length} ;$ \\
 \>  \> $\mi{maxG} := \max_X g(i,X,j)$ \\
 \>  \> \alif $\mi{length} \neq 1$  \\
 \>  \>  \> $\mi{bestSplit} := \max_{k\vert i < k < j} \mi{maxC}[i,k] + \mi{maxC}[k,j]$ \\
 \>  \> \alelse \\
 \>  \>  \> $\mi{bestSplit} := 0;$ \\
 \>  \> $\mi{maxC}[i, j] := \mi{maxG} + \mi{bestSplit};$ \\
\end{tabbing}
\caption{Labelled Recall Algorithm} \label{fig:maxcons}
\end{figure}

Equation \ref{eqn:maxc} can be converted into a dynamic programming
algorithm as shown in Figure \ref{fig:maxcons}.  For a grammar with
$r$ rules and $k$ nonterminals, the run time of this algorithm is
$O(n^3 + kn^2)$ since there are two layers of outer loops, each with
run time at most $n$, and an inner loop, over nonterminals and $n$.
However, the overall runtime is dominated by the computation of the
inside and outside probabilities, which takes time $O(rn^3)$.

By modifying the algorithm slightly to record the actual split used at each
node, we can recover the best parse.  The entry $\mi{maxC}[1, n\!+\!1]$
contains the expected number of correct constituents, given the model.

\subsection{Item-Based Description}

\label{sec:itemextend}

{\it For those not familiar with the notation of Chapter
\ref{ch:semi}, this section and succeeding references to item-based
descriptions may be skipped without loss of continuity.}

It is also possible to specify the Labelled Recall algorithm using an
item-based description, as in Chapter \ref{ch:semi}, although we will
need to slightly extend the notation of item-based descriptions to do
so.  These same extensions will be necessary in Chapter
\ref{ch:thresh} when we describe global thresholding and multiple-pass
parsing.  There are two extensions that will be required.  First, we
may have more than one goal item, leading to each item having a
separate outside value for each goal item.  Second, we will need to be
able to make reference to the inside-outside value of an item.  We
will denote the forwards value of an item $[x]$ in the inside semiring
by $V_{\smi{in}}([x])$ and its reverse value with goal item
$[\mi{goal}]$ by $Z_{\smi{in}}([x], [\mi{goal}])$.  We will abbreviate
the quantity $\frac{V_{\smi{in}}([x])Z_{\smi{in}}([x],
[\smi{goal}])}{V_{\smi{in}}[\smi{goal}]}$ by $\VZVin([x],
[\smi{goal}])$.

\begin{figure}
$$
\begin{array}{cl}
\mbox{\bf Item form:} \\
{[i, A, j]} & \mbox{inside semiring} \\
{[i, A, j]^*} & \mbox{arctic semiring, or similar} \\ \\
\mbox{\bf Primary Goal:} \\
{[1, S, n+1]^*} \\\\
\mbox{\bf Secondary Goal:} \\
{[1, S, n+1]} \\\\
\mbox{\bf Rules:} \\
\infer{R(A \rightarrow w_i)}{[i, A, i+1] }{} & \mbox{Unary} \\
\infer{R(A \rightarrow BC)\rspace [i, B, k]\rspace [k, C, j]}{[i, A,
j]}{}
& \mbox {Binary} \\
\infer{R(A, w_i, \VZVin([i, A, i+1], [1, S, n+1])}{[i, A,i+1]^*}{} &
\mbox{Unary Labels} \\
\infer{R(A, B, C, \VZVin([i, A, j], [1, S, n+1]))\rspace [i, B, k]^*\rspace [k, C, j]^*}
{[i, A, j]^*}{} xo
& \mbox {Binary Labels}

\end{array}
$$
\caption{Labelled Recall Description}\label{fig:labelledrecalldesc}
\end{figure}

Figure \ref{fig:labelledrecalldesc} gives the Labelled Recall
item-based description.  The description is similar to the procedural
version.  The first step is to compute the inside-outside values.  We
thus have the usual items $[i, A, j]$ in the inside semiring, and the
usual unary and binary rules for CKY parsing.  The inside-outside
values will simply be $\VZVin([i, A, j], [1, S, n+1])$, using the
notation we just defined.  Next, we need to find, for each $i, A, j$,
the inside-outside value  and the best sum
of splits for the span.  These will be stored in items of the form
$[i, A, j]^*$.  To get the inside-outside value for
$\constituent{i}{A}{j}$, we use a special rule value function,
$$
R(A, B, C, \VZVin([i, A, j], [1, S, n+1]))
$$
The value of this function is just the inside-outside value of the
constituent $\constituent{i}{A}{j}$, which is passed into the
function as the last argument.  We also need to find the best sum of
splits for the span, i.e., a maximum over a sum.  For this, we use a
special semiring, the arctic semiring, whose operations are $\max$, +
(as defined in Section \ref{sec:semiring}.)  Notice that our two
different item types use two different semirings: the inside semiring
for $[i, A, j]$, and the arctic semiring for $[i, A, j]^*$.  Using
the arctic semiring and our special rule value function, the unary
and binary labels rules compute the best nonterminal label and best
sum of splits for each span.  These rules are identical to the usual
unary and binary rules, except that they use our special rule value
function that gives the inside-outside value of the relevant constituent.

We must, of course, find the inside and outside values for the $[i, A,
j]$ before we can compute the $[i, A, j]^*$ values; this means that
the order of interpretation is changed as well.  Normally, the order
of interpretation is simply all of the forward values, in order,
followed by all of the reverse values, in the reverse order.  In this
new version, we first have all of the forward values of items $[i, A,
j]$, and then, in reverse order, all of the reverse values of these
items; next, we have all of the forward values of $[i, A, j]^*$, and
finally, optionally, in reverse order all of the reverse values of
$[i, A, j]^*$.

If we use the arctic semiring, we get only the maximum expected number
of correctly labelled constituents, but not the tree which gives this
number.  To get this tree, we would use the arctic-derivation
semiring.  As usual, there are advantages to describing the algorithm
with item-based descriptions.  For instance, we can easily compute
n-best derivations, just by using the arctic-top-n semiring.

A short example may help clarify the algorithm.  Consider the example
grammar of Figure \ref{fig:labrecallexample}.  For this grammar, we
have that the inside-outside values of the $[i, X, j]$, which equal
$\VZVin([i, X, j], [1, S, n+1]$, are as follows: \\
$$
\begin{array}{cc}
\hspace{-.5in}
\leaf{$x$} \faketreewidth{AA}
\leaf{$x$} \faketreewidth{AA}
%\branch{2}{$A$\makebox[0in][l]{:0.5, $E$:0.25, $F$:0.25}} \faketreewidth{AAAAAAAAAAAAAAAAAAAAAA}
\branch{2}{$\begin{array}{c}
A\makebox[0in][l]{ (0.5)}\\
E\makebox[0in][l]{ (0.25)}\\
F\makebox[0in][l]{ (0.25)}
\end{array}$}
\faketreewidth{AAAAAAAAAA}
\leaf{$x$} \faketreewidth{AA}
\leaf{$x$} \faketreewidth{AA}
%\branch{2}{$B$\makebox[0in][l]{:0.5, $C$:0.25, $D$:0.25}} \faketreewidth{AAAAAAAAAAAAAAAAAAAAAA}
\branch{2}{$\begin{array}{c}
B\makebox[0in][l]{ (0.5)}\\
C\makebox[0in][l]{ (0.25)}\\
D\makebox[0in][l]{ (0.25)}
\end{array}$}
\branch{2}{$S$\makebox[0in][l]{ (1.0)}}
\faketreewidth{SS}
\tree \hspace{1in}
&
\begin{array}{rcl}
{[1, S, 5]}  &=& 1.0 \\
{[1, A, 3]}  &=& 0.5 \\
{[1, E, 3]}  &=& 0.25 \\
{[1, F, 3]}  &=& 0.25 \\
{[3, B, 5]}  &=& 0.5 \\
{[3, C, 5]}  &=& 0.25 \\
{[3, D, 5]}  &=& 0.25 \\ 
\end{array}
\end{array}
$$
We have that the forward arctic values of the $[i, X, j]^*$ are: 
$$
\begin{array}{cc}
\hspace{-.5in}
\leaf{$x$} \faketreewidth{AA}
\leaf{$x$} \faketreewidth{AA}
%\branch{2}{$A$\makebox[0in][l]{:0.5, $E$:0.25, $F$:0.25}} \faketreewidth{AAAAAAAAAAAAAAAAAAAAAA}
\branch{2}{$\begin{array}{c}
A^*\makebox[0in][l]{ (0.5)}\\
E^*\makebox[0in][l]{ (0.25)}\\
F^*\makebox[0in][l]{ (0.25)}
\end{array}$}
\faketreewidth{AAAAAAAAAA}
\leaf{$x$} \faketreewidth{AA}
\leaf{$x$} \faketreewidth{AA}
%\branch{2}{$B$\makebox[0in][l]{:0.5, $C$:0.25, $D$:0.25}} \faketreewidth{AAAAAAAAAAAAAAAAAAAAAA}
\branch{2}{$\begin{array}{c}
B^*\makebox[0in][l]{ (0.5)}\\
C^*\makebox[0in][l]{ (0.25)}\\
D^*\makebox[0in][l]{ (0.25)}
\end{array}$}
\branch{2}{$S^*$\makebox[0in][l]{ (2.0)}}
\faketreewidth{SS}
\tree \hspace{1in}
&
\begin{array}{rcl}
{[1, S, 5]}^*  &=& 2.0 \\ 
{[1, A, 3]}^*  &=& 0.5 \\
{[1, E, 3]}^*  &=& 0.25 \\
{[1, F, 3]}^*  &=& 0.25 \\
{[3, B, 5]}^*  &=& 0.5 \\
{[3, C, 5]}^*  &=& 0.25 \\
{[3, D, 5]}^*  &=& 0.25 \\
\end{array}
\end{array} 
$$
The key element is $[1, S, 5]^*$, which is derived using the binary
labels rule instantiated with:
$$
\shortinfer{R(S, B, C, 1.0))\rspace [1, B, 3]^*\rspace [3, C, 5]^*}
{[1, S, 5]^*}{} 
$$
which, computed in the arctic semiring where the multiplicative
operator is addition, has the value 1.0 + 0.5 + 0.5 = 2.0.

%  Furthermore, we have
%glossed over proofs of correctness for either the pseudocode or
%item-based description, but these correctness proofs are simpler for
%the item-based description.  Is this true?

\section{Bracketed Recall Parsing}
\label{sec:brackconspars}

The Labelled Recall algorithm maximizes the expected number of
correct labelled constituents.  However, many commonly used evaluation
metrics, such as the Consistent Brackets Recall (Crossing Brackets)
rate, ignore labels.  Similarly, some grammar induction algorithms,
such as those used by \newcite{Pereira:92a} do not produce meaningful
labels.  In particular, the Pereira and Schabes method induces a
grammar from the brackets in the treebank, ignoring the labels in the
treebank.  While the grammar they induce has labels, these labels are
not related to those in the treebank.  Thus, while the Labelled Recall
algorithm could be used with these grammars, perhaps maximizing a
criterion that is more closely tied to the task will produce better
results.  Ideally, we would maximize the Consistent Brackets Recall rate
directly.  However, since it is time-consuming to deal with Consistent
Brackets, as described in Section \ref{sec:submaxmetrics}, we instead use the closely related Bracketed Recall rate.

For the Bracketed Recall algorithm, we find the parse that
maximizes the expected Bracketed Recall rate, $B/N_C$.  (Remember that
$B$ is the number of brackets that are correct, and $N_C$ is the
number of constituents in the correct parse.)
\begin{equation}
T_G=\arg \max_T E(B/N_C)
\label{eqn:maxbracket}
\end{equation}
Following a derivation similar to that used for the Labelled Recall
algorithm, we can rewrite Equation \ref{eqn:maxbracket} as
$$
T_G=\arg \max_T \sum_{\constituenttwo{i}{j} \in T} \sum_{X}
P(S \derives w_1 ... w_{i-1} X w_{j} ... w_n | w_1 ... w_n)
$$

The algorithm for Bracketed Recall parsing is almost identical to that
for Labelled Recall parsing.  The only required change is to sum,
rather than maximize, over the symbols $X$ to calculate $\mi{maxG}$,
substituting in the following line:
$$
\mi{maxG} := \sum_X g(i, X, j);
$$

\begin{figure}
$$
\begin{array}{cl}
\mbox{\bf Item form:} \\
{[i, A, j]} & \mbox{inside semiring} \\
{[i, j]^{\dagger}} & \mbox{inside semiring} \\
{[i, j]^*} & \mbox{arctic semiring, or similar} \\ \\
\mbox{\bf Primary Goal:} \\
{[1, S, n+1]^*} \\\\
\mbox{\bf Secondary Goal:} \\
{[1, S, n+1]} \\\\
\mbox{\bf Rules:} \\
\infer{R(A \rightarrow w_i)}{[i, A, i+1] }{} & \mbox{Unary} \\
\infer{R(A \rightarrow BC)\rspace [i, B, k]\rspace [k, C, j]}{[i, A,
j]}{}
& \mbox {Binary} \\
\infer{\VZVin([i, A, j],[1, S, n+1])}{[i, j]^{\dagger}}{} &
\mbox{Summation} \\
\infer{R(w_i, [i, i+1]^{\dagger})}{[i, i+1]^*}{} &
\mbox{Unary Brackets} \\
\infer{R(B, C, [i, j]^{\dagger})\rspace [i, k]^* \rspace [k, j]^*}{[i, j]^*}{} 
& \mbox {Binary Brackets}

\end{array}
$$
\caption{Bracketed Recall Description}\label{fig:bracketedrecalldesc}
\end{figure}

The Labelled Recall item-based description can be easily converted to
a Bracketed Recall item-based description, with the addition of one
new item type and one new rule.  Figure \ref{fig:bracketedrecalldesc}
gives an item-based description for the Bracketed Recall algorithm.
There are now 3 item types: $[i, A, j]$, which has the usual meaning;
$[i, j]^{\dagger}$ which has the value of $\sum_A g(i, A, j)$; and
$[i, j]^*$ which has the value of $\mi{MAXC}(i, j)$.  The
summation rule sums over items of type $[i, A, j]$ to produce items of
the type $[i, j]^{\dagger}$.  

\section{Experimental Results}
\label{sec:results1}
We describe two experiments that tested these algorithms.  The first
uses a grammar without meaningful nonterminal symbols, and compares
the Bracketed Recall algorithm to the traditional Labelled Tree
(Viterbi) algorithm.  The second uses a grammar with meaningful
nonterminal symbols and performs a three-way comparison between the
Labelled Recall, Bracketed Recall, and Labelled Tree algorithms.
These experiments show that use of an algorithm matched appropriately
to the evaluation criterion can lead to as much as a 10\% reduction in
error rate.

\subsection{Grammar Induced by Pereira and Schabes method}\label{sec:pns}
We duplicated the experiment of \newcite{Pereira:92a}.  Pereira and
Schabes trained a grammar from a bracketed form of the TI section of
the ATIS corpus\footnote{Most researchers throw out a few sentences
because of problems aligning the part of speech and parse files, or
because of labellings of discontinuous constituents, which are not
usable with the Crossing Brackets rate.  The difficult data was
cleaned up and used in these experiments, rather than thrown out.  A
diff file between the original ATIS data and the cleaned up version,
in a form usable by the ``ed'' program, is available by anonymous FTP
from {\tt ftp://ftp.deas.harvard.edu/pub/goodman/atis-ed/
ti\_tb.par-ed} and {\tt ti\_tb.pos-ed}.  The number of changes made
was small: the diff files sum to 457 bytes, versus 269,339 bytes for
the original files, or less than 0.2\%.}
using a modified form of the inside-outside algorithm.  They then used
the Labelled Tree (Viterbi) algorithm to select the best parse for
sentences in held out test data.  We repeated the experiment, inducing
the grammar the same way.  However, during the testing phase, we ran
both the Labelled Tree and Labelled Recall algorithm for each
sentence.  In contrast to previous research, we repeated the
experiment ten times, with different random splits of the data into
training set and test set, and different random initial conditions
each time.  Note that in one detail our scoring for experiments
differs from the theoretical discussion in the paper, so that we can
more closely follow convention.  In particular, constituents of length
one are not counted in recall measures, since these are trivially
correct for most criteria.

\begin{table}
\begin{center}
\begin {tabular}{|l||r|r|r|r|r|} \hline
       Criteria & Min & Max & Range & Mean & StdDev \\ \hline \hline
\multicolumn{6}{|c|} {Labelled Tree Algorithm} \\ \hline Cons Brack
Rec &86.06\% &93.27\% & 7.20\% &90.13\% & 2.57\% \\ \hline Cons Brack
Tree &51.14\% &77.27\% &26.14\% &63.98\% & 7.96\% \\ \hline Brack Rec
&71.38\% &81.88\% &10.50\% &75.87\% & 3.18\% \\ \hline \hline
\multicolumn{6}{|c|} {Bracketed Recall Algorithm} \\ \hline Cons Brack
Rec &88.02\% &94.34\% & 6.33\% &91.14\% & 2.22\% \\ \hline Cons Brack
Tree &53.41\% &76.14\% &22.73\% &63.64\% & 7.82\% \\ \hline Brack Rec
&72.15\% &80.69\% & 8.54\% &76.03\% & 3.14\% \\ \hline \hline
\multicolumn{6}{|c|} {Bracketed Recall - Labelled Tree} \\ \hline Cons
Brack Rec &-1.55\% & 2.45\% & 4.00\% & 1.01\% & 1.07\% \\ \hline Cons
Brack Tree &-3.41\% & 3.41\% & 6.82\% &-0.34\% & 2.34\% \\ \hline
Brack Rec &-1.34\% & 2.02\% & 3.36\% & 0.17\% & 1.20\% \\ \hline
\end{tabular}
\end{center}
\caption{Labelled Tree (Viterbi) versus Bracketed Recall for P\&S}
\label{tab:maxbrackcompare}
\end{table}

In three test sets there were sentences with terminals not
present in the matching training set.  The four sentences (out of 880)
containing these terminals could not be parsed, and in the following
analysis were assigned right branching, period high structure.  This
of course affects the Labelled Recall and Labelled Tree algorithms
equally.

Table \ref{tab:maxbrackcompare} shows the results of running this
experiment, giving the minimum, maximum, mean, range, and standard
deviation for three criteria, Consistent Brackets Recall, Consistent
Brackets Tree, and Bracketed Recall.  We also computed, for each split
of the data, the difference between the Bracketed Recall algorithm and
the Labelled Tree algorithm on each criterion.  Notice that on each
criterion the minimum of the differences is negative, meaning that on
some data set the Labelled Tree algorithm worked better, and the
maximum of the differences is positive, meaning that on some data set
the Bracketed Recall algorithm worked better.  The only criterion for
which there was a statistically significant difference between the
means of the two algorithms is the Consistent Brackets Recall rate,
which was significant to the 2\% significance level (paired t-test).
Thus, use of the Bracketed Recall algorithm leads to a 10\% reduction
in error rate.

\begin{figure}
\psfig{figure=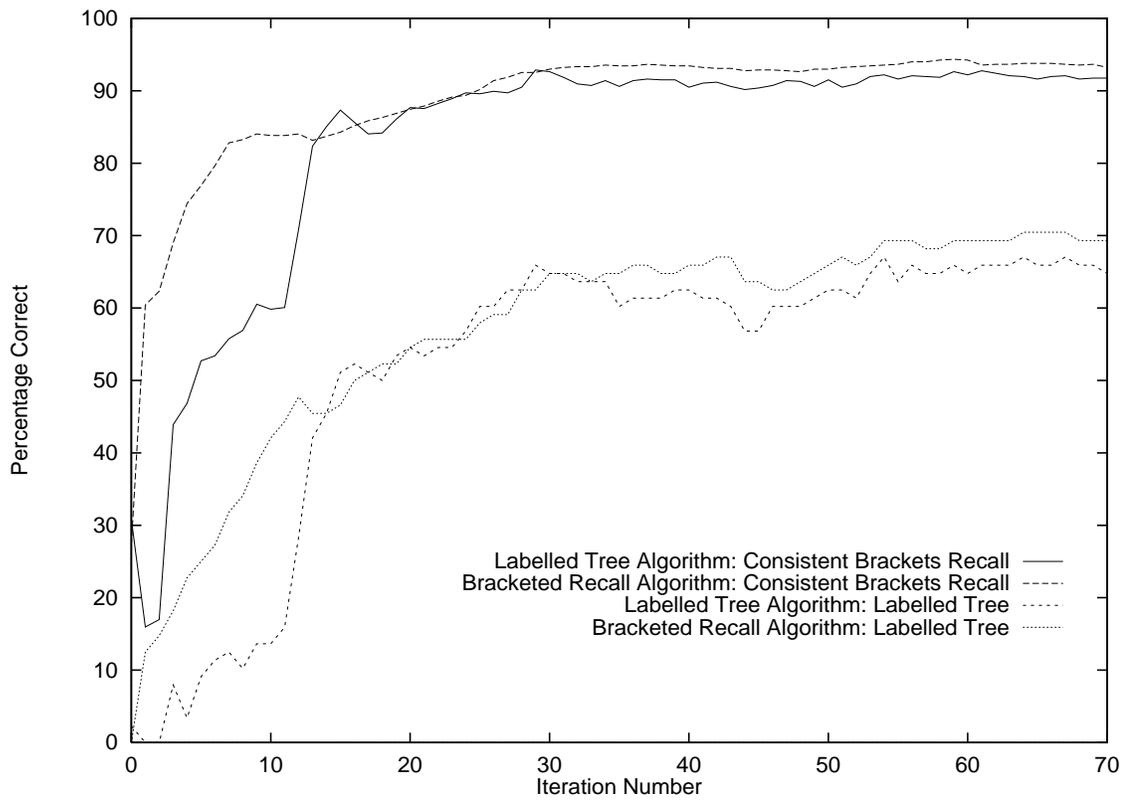,width=6in}

\caption{Labelled Tree versus Bracketed Recall in Pereira and Schabes
Grammar} \label{fig:typpns}
\end{figure}

In addition, the performance of the Bracketed Recall algorithm was
also qualitatively more appealing.  Figure \ref{fig:typpns} shows
typical results.  Notice that the Bracketed Recall algorithm's
Consistent Brackets rate (versus iteration) is smoother and more
nearly monotonic than the Labelled Tree algorithm's. The Bracketed
Recall algorithm also gets off to a much faster start, and is generally
(although not always) above the Labelled Tree level.  For the Labelled
Tree rate, the two are usually very comparable.

\subsection{Grammar Induced by Counting}
\label{sec:countgrammar}
The replication of the Pereira and Schabes experiment was useful for
testing the Bracketed Recall algorithm.  However, since that
experiment induces a grammar with nonterminals not comparable to
those in the training, a different experiment is needed to evaluate
the Labelled Recall algorithm, one in which the nonterminals in
the induced grammar are the same as the nonterminals in the test set.

\subsubsection{Grammar Induction by Counting}
For this experiment, a very simple grammar was induced by counting,
using the Penn Tree Bank, version 0.5.
% Should I cite someone here?  Who?
In particular, the trees were first made binary branching, by removing
epsilon productions, collapsing singleton productions, and by
converting n-ary productions ($n > 2$), as in Figure \ref{fig:binary}.
The resulting trees were treated as the ``Correct'' trees in the
evaluation.

\begin{figure}
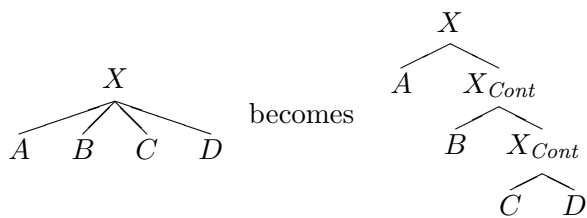

\begin{center}
\hspace{-0.3in}
\mbox{
\leaf{$A$}
\leaf{$B$} \faketreewidth{SS}
\leaf{$C$} \faketreewidth{SS}
\leaf{$D$}
\branch{4}{$X$}
\tree
\hspace{-0.1in}
} becomes
\mbox{
\leaf{$A$}
\leaf{$B$}
\leaf{$C$} \faketreewidth{SS}
\leaf{$D$} \faketreewidth{SS}
\branch{2}{$X_{\smi{Cont}}$}
\branch{2}{$X_{\smi{Cont}}$}
\branch{2}{$X$}
\hspace{-0.5in}
\tree
\hspace{-0.3in}
}
\end{center}
\caption{Conversion of Productions to Binary Branching}
\label{fig:binary}
\end{figure}

A grammar was then induced in a straightforward way from these trees,
simply by giving one count for each observed production.  No smoothing
was done.  There were 1805 sentences and 38610 nonterminals in the
test data.  (The resulting grammar undergenerated somewhat, being
unable to parse approximately 9\% of the test data.  The unparsable
data were assigned a right branching structure with their right-most
element attached high.  Notice that the Labelled Tree (Viterbi),
Labelled Recall, and Bracketed Recall algorithms all fail on exactly
the same sentences (since the inside-outside calculation fails exactly
when the Labelled Tree calculation fails).  Thus, this default
behavior affects all sentences equally.

\subsubsection{Results}
\begin{table}
\begin{center}
\begin{tabular}{|l||r|r|r||r|r|} \cline{2-6}
\multicolumn{1}{c}{} &\multicolumn{5}{|c|} {Criterion} \\ \hline
               &    Label &    Label  &    Brack   &Cons Brack&Cons Brack\\ 
Algorithm      &    Tree  &    Recall &    Recall  &Recall    &Tree      \\ \hline\hline
Label Tree     &\em 4.54\%&    48.60\%&    60.98\% &66.35\%   &12.07\%   \\ \hline
Label Recall   &    3.71\%&\em 49.66\%&    61.34\% &68.39\%   &11.63\%   \\ \hline
Bracket Recall &          &           &\em 61.63\% &68.17\%   &11.19\%
\\ \hline
\end{tabular}
\end{center}
\caption{Grammar Induced by Counting: Three Algorithms 
Evaluated on Five Criteria} \label{tab:counting}
\end{table}

Table \ref{tab:counting} shows the results of running all three
algorithms, evaluating against five criteria.  Notice that for
each algorithm, for the criterion that it optimizes it is the best
algorithm.  That is, the Labelled Tree algorithm is the best for the
Labelled Tree rate, the Labelled Recall algorithm is
the best for the Labelled Recall rate, and the Bracketed Recall
algorithm is the best for the Bracketed Recall rate.

\section{NP-Completeness of Bracketed Tree Maximization}
\label{sec:npcomplete}

In this section, we show the NP completeness of the Bracketed
Tree Maximization problem.  Bracketed Tree Maximization is the problem
of finding that bracketing which has the highest expected score on the
Bracketed Tree criterion.  In other words, it is that tree which has
the highest probability of being exactly correct, according to the
model, ignoring nonterminal labels.  Formally, the Bracketed Tree
Maximization problem is to compute
$$
\arg \max_{T_G} E\left(\bigoneif{B = N_C}\right)
$$
The Bracketed Tree criterion can be distinguished from the Labelled
Tree (Viterbi) criterion in the following example:
$$
\begin{array}{rcll}
S & \rightarrow & L\;x & 0.3 \\
S & \rightarrow & M\;x & 0.3 \\
S & \rightarrow & x\;R & 0.4 \\
L, M, R & \rightarrow & x\;x & 1.0
\end{array}
$$
Here, given an input of $xxx$, the Labelled Tree parse is
\leaf{$x$} \faketreewidth{AA}
\leaf{$x$} \faketreewidth{AA}
% Use faketreewidth to work around a bug
\faketreewidth{AA}
\leaf{$x$} \faketreewidth{AA}
% Use faketreewidth to work around a bug
\faketreewidth{AA}
\branch{2}{$R$}
\branch{2}{$S$}
\begin{center}
\tree
\end{center}
and thus the Labelled Tree bracketing is $[x [x x]]$, with a 40\%
chance of being correct, assuming that the string was produced by the
model.  On the other hand, the bracketing $[[x x] x]$ has a 60\%
chance of being correct.  If our goal is to maximize the probability
that the bracketing is exactly correct, then we would like an
algorithm that returns the second bracketing, rather than the first.
(Our previous algorithm, the Bracketed Recall algorithm, may return
bracketings with 0 probability: it does not solve this problem.)  We
will show that maximizing Bracketed Recall is NP-Complete.  It will be
easier to prove this if we first prove a related theorem about Hidden
Markov Models (HMMs).  The proof we give here is similar to one of
\newcite{Sima'an:96b}, that maximizing the Labelled Tree rate for a
Stochastic Tree Substitution Grammar is NP-Complete.

We note that for binary branching trees, there is a crossing
bracket in any tree that does not exactly match.  Thus, for binary
branching trees, Bracketed Tree Maximization is equivalent to Zero
Crossing Brackets Rate Maximization.  Thus we will implicitly also be
showing that Zero Crossing Brackets Rate Maximization is NP-Complete.

\subsection{NP-Completeness of HMM Most Likely String}

\def\NPCOMPLETE#1#2#3 {\noindent {\bf \sc {#1}}

                       \noindent {\sc Instance: } #2 

                       \noindent {\sc Question: } #3 }

\NPCOMPLETE {HMM Most Likely String (HMM-MLS)} {An HMM specification
           (with probabilities specified as fractions
           $\frac{x}{2^y}$), a length $n$ in unary, and a probability
           $p$.}  {Is there some string of length $n$ such that the
           specified HMM produces the string with probability at least
           $p$?}

The proof that Bracketed Tree Maximization is NP-Complete is easier to
understand if we first show the NP-Completeness of a simpler problem,
namely finding whether the most likely string of a given length $n$
output by a Hidden Markov Model has probability at least $p$.  (A
different problem, that of finding the most likely state sequence of
an HMM, and the corresponding most likely output, can of course be
solved easily.  However, since the same string can be output by
several state sequences, the most likely string problem is much
harder.)

We note that a related problem, Most Likely String Without Length,
in which the length of the string is not prespecified, can be shown
NP-hard by a very similar proof.  However, it cannot be shown
NP-Complete by this proof, since the most likely string could be
exponentially long, and thus the generate and test method cannot be
used to show that the problem is in NP.

We show the NP-Completeness of HMM Most Likely String by a reduction
of the NP-Complete problem 3-Sat.  The problem 3-Sat is whether or not
there is a way to satisfy a
% Cite Hopcroft and Ullman (Introduction to Automata Theory,
% Languages, and Computation), page 330-331.
% or Garey and Johnson, pages 46-49.
formula of the form $F_1 \wedge F_2\wedge ...  \wedge F_n$ where
$F_i$ is of the form $(\alpha_{i1} \vee \alpha_{i2} \vee
\alpha_{i3})$.  Here, the $\alpha_{ij}$ represent literals, either
$x_k$ or $\neg x_k$.  An example of 3-Sat is the question of whether
the following formula is satisfiable:

$$
(\neg x_1 \vee \neg x_2 \vee x_4) \wedge (x_1 \vee x_3 \vee \neg x_4)
\wedge (\neg x_4 \vee x_3 \vee x_1)
$$

We will show how to construct, for any 3-Sat formula, an HMM that has
a high probability output if and only if the formula is satisfiable.

The most obvious way to pursue this reduction is to create a network
whose output corresponds to satisfying assignments of the clauses, e.g.
a string of $T$'s and $F$'s, one letter for each variable, denoting
whether that variable is true or false.  There will be one section of
the network for each clause, and the probabilities will be assigned
in such a way that for the output probability to sum sufficiently
high, for every clause the maximal string must have a path through
the corresponding subnetwork.

The problem with this technique is that such a string may be output by
multiple paths through the subnetworks corresponding to some of the
clauses, and zero paths through the parts of the network leading to
other clauses.  What we must do is to ensure that there is exactly one
path through each clause's subnetwork.  In order to do this, we
augment the output with an initial sequence of the digits 1, 2, and 3.
Each digit specifies which variable (the first, second, or third) of
each clause satisfied that clause.  This strategy allows construction
of subnetworks for each clause with at most one path.  Then, if there
is a string whose probability indicates that it has $n$ paths through
the HMM, we know that the string satisfies every clause, and therefore
satisfies the formula.  Any string that does not correspond to a
satisfying assignment will have fewer than $n$ paths, and will not
have a sufficiently high probability.

\begin{mytheorem}
The HMM Most Likely String problem is NP-Complete
\end{mytheorem}
{\em Proof \hspace{3em}} To show that HMM Most Likely String is in NP
is trivial: guess a most likely string of the given length $n$; then
compute its probability by well known polynomial-time dynamic
programming techniques (the forward algorithm); then verify that the
probability is at
% cite Rabiner?
least $p$.

To show NP-Completeness, we show how to reduce a 3-Sat formula to an
HMM Most Likely String problem.  Let $f$ represent the number of
disjunctions and $v$ represent the number of variables in the formula
to satisfy.  Create an HMM with $3f(f+v)+2$ states, as follows.  The
states will have the following names: $\understackrel{END}{\bigcirc}$,
$\understackrel{START}{\bigcirc}$, and $\understackrel{h, i,
j}{\bigcirc}$.  $h$ ranges from 1 to $f+v$; $i$ ranges from 1 to $f$;
and $j$ ranges from 1 to 3.  Two of the states are a start and an end
state, with an equiprobable epsilon transition from the start state to
the $3f$ states of the form $\understackrel{1, i, j}{\bigcirc}$.

\begin{figure}
\begin{center}
\mbox{\psfig{figure=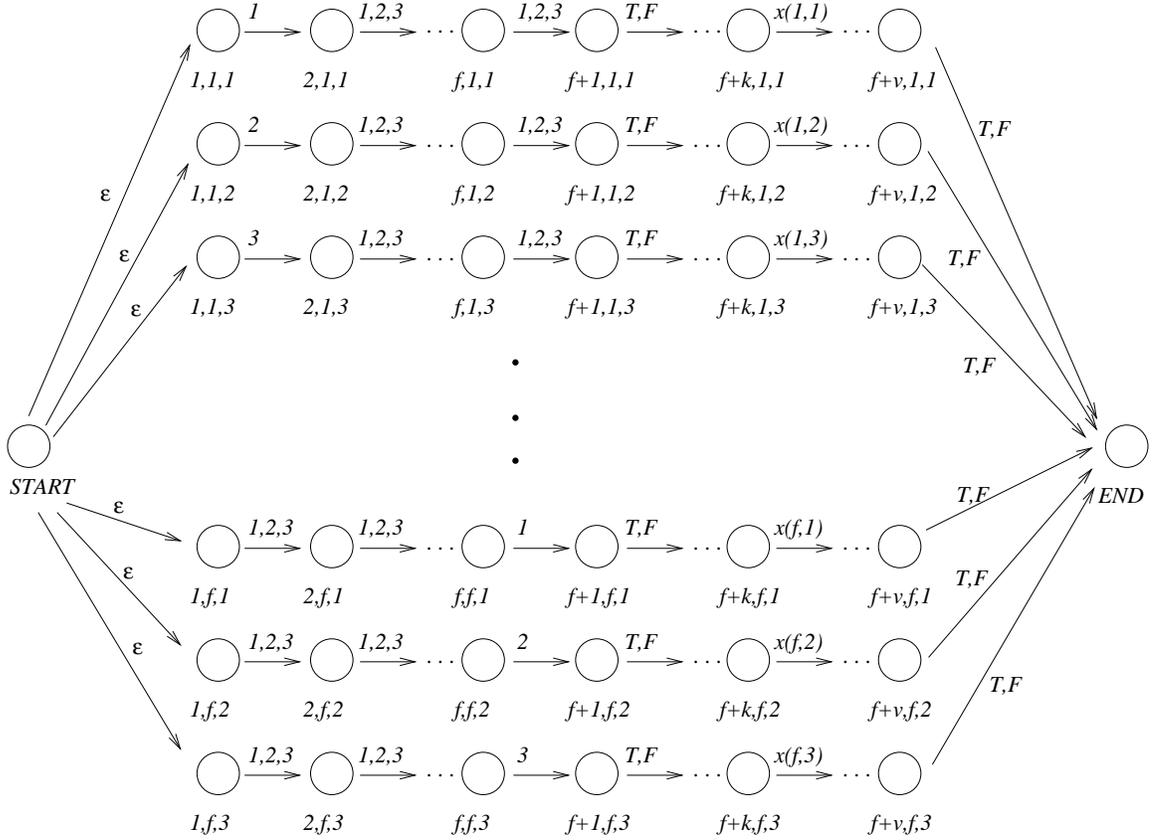,width=6in}}
\end{center}
\caption{Portion of HMM corresponding to a 3-Sat formula} \label{fig:hmmnp}
\end{figure}

The basic form of the network is
$$
\understackrel{START}{\bigcirc} \stackrel{\epsilon}{\rightarrow}
\understackrel{1,i,j}{\bigcirc} \stackrel{1,2,3}{\rightarrow}
\cdots
\understackrel{i,i,j}{\bigcirc} \stackrel{j}{\rightarrow}
\cdots
\understackrel{f,i,j}{\bigcirc} \stackrel{1,2,3}{\rightarrow}
\understackrel{f+1,i,j}{\bigcirc} \stackrel{T,F}{\rightarrow}
\cdots
\understackrel{f+k,i,j}{\bigcirc} \stackrel{x(i,j)}{\rightarrow}
\cdots
\understackrel{f+v,i,j}{\bigcirc} \stackrel{T,F}{\rightarrow}
\understackrel{END}{\bigcirc}
$$
where $x(i, j)$ is $T$ if $\alpha_{ij}$ is of the form $x_k$, and $F$ if
it is of the form $\neg x_k$.
Figure \ref{fig:hmmnp} shows a section of the HMM, giving the parts
that correspond to the first and last clause.  The notation $\bigcirc
\stackrel{1,2,3}{\rightarrow} \bigcirc$ indicates a state transition
with equal probabilities of emitting the symbols 1, 2, or 3.
Similarly $\bigcirc \stackrel{T, F}{\rightarrow} \bigcirc$ indicates a
state transition with equal probabilities of emitting $T$ or $F$.

We now show that if the given 3-Sat formula is satisfiable, then
the constructed HMM has a string with probability $\frac{1}{3^f}
\times \frac{1}{2^{v-1}}$, whereas if it is not satisfiable, the most
likely string will have lower probability.  The proof is as follows.
First, assume that there is some way to satisfy the 3-Sat formula.
Let $X_i$ denote $T$ if the variable $x_i$
takes on the value true, and let it denote $F$ otherwise, under some
assignment of values to variables that satisfies the formula.  Also,
let $J_i$ denote the number 1, 2, or 3 respectively depending on
whether for formula $i$, $\alpha_{i1}$, $\alpha_{i2}$, or
$\alpha_{i3}$ respectively satisfies formula $i$.  If $J_i$ could take
on more than one value by this definition (that is, if $F_i$ is
satisfied in more than one way) then arbitrarily, we assign it to the
lowest of these three.  Since we are assuming here that the formula
{\em is} satisfied under the assignment of values to variables, at
least one of $\alpha_{i1}$, $\alpha_{i2}$, or $\alpha_{i3}$ must
satisfy each clause.

Then, the string
$$
J_1J_2 \cdots J_fX_{1}X_{2} \cdots X_{v}
$$
will be a most probable string.  There may be other most probable
strings, but all will have the same probability and satisfy the
formula.  (Example: the string $131FTFT$ would be the most likely
string output by an HMM corresponding to the sample formula.)

This string will have probability $\frac{1}{3^f} \times
\frac{1}{2^{v-1}}$ because there will be routes
emitting that string through exactly $f$ of the $3f$ subnetworks, and
each route will have probability $\frac{1}{f} \times \frac{1}{3^f} \times
\frac{1}{2^{v-1}}$.

On the other hand, assume that the formula is not satisfiable.  Now,
the most likely string will have a lower probability.  Obviously, the
string could not have higher probability, since the best path through
any subnetwork has probability $\frac{1}{3^{f-1}} \times
\frac{1}{2^{v-1}}$ and it is not possible to have routes through more
than $f$ of the $3f$ subnetworks.  Therefore, the probability can be
at most $\frac{1}{3^{f}} \times \frac{1}{2^{v-1}}$.  However, it
cannot be this high, since if it were, then the string would have a
path through $f$ of the $3f$ networks, and would therefore correspond
to a satisfaction of the formula.  Thus, its probability must be
lower.  $\Box$

\subsection {Bracketed Tree Maximization is NP-Complete}

Having shown the NP-Completeness of the HMM Most Likely String
problem, we can show the NP-Completeness of Bracketed Tree
Maximization.  To phrase this as a language problem, we rephrase it as
the question of whether or not the best parse has an expected score of
at least $p$, according to the Bracketed Tree rate criterion.

\NPCOMPLETE {Bracketed Tree Maximization (BTM)}
           {A Probabilistic Context-Free Grammar $G$, a string 
            $w_1 ... w_n$, and a probability $p$.}
           {Is there a $T_G$ such that $E\left(\bigoneif{B =
            N_C}\right) \geq p$?}

\begin{mytheorem}
The Bracketed Tree Maximization problem is NP-Complete.
\end{mytheorem}
{\em Proof \hspace{3em}} Using the generate and test method, it is
clear that this problem is in NP.  To show completeness, we show that
for every HMM, there is an equivalent PCFG that produces bracketings
with the same probability that the HMM produces symbols.  We do this
by mapping states of the HMM to nonterminals of the grammar; we map
each output symbol of the HMM to a small subtree with a unique bracket
sequence.

\begin{figure}
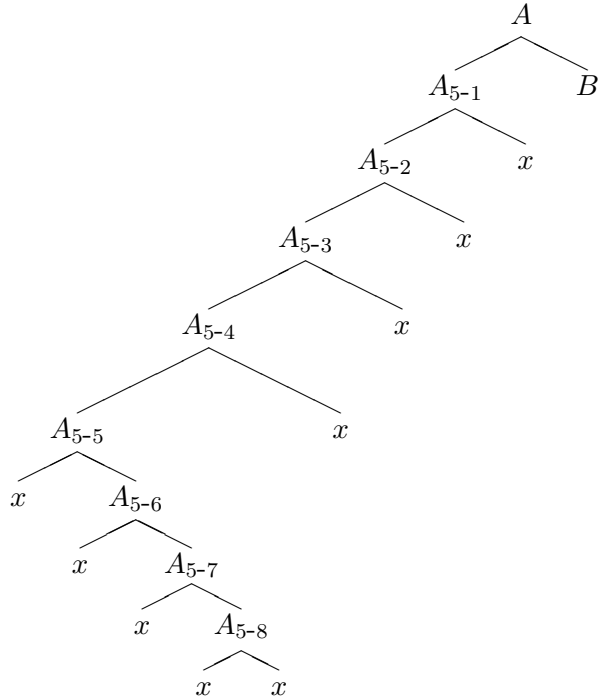

\leaf{$x$} \faketreewidth{AA}
\leaf{$x$} \faketreewidth{AA}
\leaf{$x$} \faketreewidth{AA}
\leaf{$x$} \faketreewidth{AA}
\leaf{$x$} \faketreewidth{AA}
\branch{2}{$A_{5\mbox{-}8}$}
\branch{2}{$A_{5\mbox{-}7}$}
\branch{2}{$A_{5\mbox{-}6}$}
\branch{2}{$A_{5\mbox{-}5}$}

\leaf{$x$} \faketreewidth{AA}
\branch{2}{$A_{5\mbox{-}4}$}
\leaf{$x$} \faketreewidth{AA}
\branch{2}{$A_{5\mbox{-}3}$}
\leaf{$x$} \faketreewidth{AA}
\branch{2}{$A_{5\mbox{-}2}$}
\leaf{$x$} \faketreewidth{AA}
\branch{2}{$A_{5\mbox{-}1}$}
\leaf{$B$} \faketreewidth{AA}
\branch{2}{$A$}

\begin{center}
\tree
\end{center}
\caption{Tree corresponding to an output of symbol 5 from 8 symbol
alphabet}
\label{fig:treefiveeight}
\end{figure}

The proof is as follows.  Number the output symbols of the HMM 1 to
$k$.  Assign the start state of the HMM to the start symbol of the
PCFG.  For each state $A$ with a transition emitting symbol $i$ to
state $B$ with probability $p$, include rules in the grammar that
first have $k-i+2$ symbols on right branching nodes, followed by $i-1$
symbols on left branching nodes.  For instance, for $i=5$ and $k=8$,
the tree would look like the tree in Figure \ref{fig:treefiveeight}:

Formally, we can write this as.
\begin{eqnarray*}
A_i & \rightarrow & {\overbrace{[ [ [ ... [ [}^{i} x \overbrace{[ x [
x... [x}^{k-i} x \overbrace{] ] ... ]}^{k-i+1} \overbrace{x ] x ]
... x]}^{i-1} B} \mbox{\hspace{3ex}}(p)
\end{eqnarray*}
Also, for each final state A, include a rule of the form
\begin{eqnarray*}
A & \rightarrow & \epsilon
\end{eqnarray*}

\begin{figure}
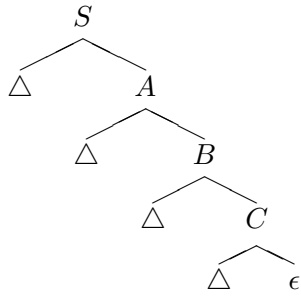

\begin{center}
\leaf{$\bigtriangleup$} \faketreewidth{AA} % S-left
\leaf{$\bigtriangleup$} \faketreewidth{AA} % A-left
\leaf{$\bigtriangleup$} \faketreewidth{AA} % B-left
\leaf{$\bigtriangleup$} \faketreewidth{AA} % C-left
\leaf{$\epsilon$} \faketreewidth{AA} % C-right
\branch{2}{$C$} % B-right
\branch{2}{$B$} % A-right
\branch{2}{$A$} % S-right
\branch{2}{$S$} % top
\tree
\end{center}
\caption{Tree corresponding to state sequence $SABC$}\label{fig:sequenceABC}
\end{figure}

If an HMM had a state sequence $SABC$, then the corresponding parse
tree would look like Figure \ref{fig:sequenceABC}, where
$\bigtriangleup$ represents a subtree corresponding to the HMM's
output.

There will be a one-to-one correspondence between parse trees in this
grammar, and state-sequence/outputs pairs in the HMM.  If we throw
away nonterminal information, leaving only brackets, there will be a
one-to-one correspondence between bracketed parse trees and HMM
outputs.  The Bracketed Tree Maximization of a string in this grammar
will correspond to the most likely output of the corresponding HMM.
Thus, if we could solve the Bracketed Tree Maximization problem, we
could also solve the HMM Most Likely String of length $n$ problem.
Therefore, Bracketed Tree Maximization is NP-Complete. $\Box$

\begin{mycorollary}
Consistent Brackets Tree (Zero Crossing Brackets) Maximization for
Binary Branching Trees is NP-Complete.
\end{mycorollary}

{\em Proof \hspace{3em}} Bracketed Tree is equivalent to Consistent
Brackets Tree if both $T_C$ and $T_G$ are binary branching.  $\Box$

We note that for n-ary branching trees, discussed in Section \ref{sec:nary},
Consistent Brackets Tree maximization is typically achieved by simply
placing parentheses at the top-most level, and nowhere else.  Thus, it
is the binary branching case that is of interest.

\section{General Recall Algorithm}
\label{sec:generalalg}

The Labelled Recall algorithm and Bracketed Recall algorithm are both
special cases of a more general algorithm, called the General Recall
algorithm.  In fact, the General Recall algorithm was developed first,
for parsing Stochastic Tree Substitution Grammars (STSGs).  In this
section, we first define STSGs, and then use them to motivate the
General Recall algorithm.  Next, we show how the other two algorithms
reduce to the general algorithm, and finally give some examples of
other cases in which one might wish to use the General Recall
algorithm.

There are two ways to define a STSG: either as a Stochastic Tree
Adjoining Grammar \cite{Schabes:92a} restricted to substitution
operations, or as an extended PCFG in which entire trees may occur on
the right hand side, instead of just strings of terminals and
non-terminals.  A PCFG then, is a special case of an STSG in which the
trees are limited to depth 1.  In the STSG formalism, for a given
parse tree, there may be many possible derivations.  Thus, variations
on the Viterbi algorithm, which find the best {\em derivation} may not
be the most appropriate.  What is really desired is to find the best
parse.  But best according to what criterion?  If the criterion used
is the Labelled Tree rate, then the problem is NP-Complete
\cite{Sima'an:96b}, and the only known approximation algorithm is a
Monte Carlo algorithm, due to \newcite{Bod:93a}.\footnote{We will show
in Section \ref{sec:timing} that this algorithm has a serious problem:
either its accuracy decreases exponentially with sentence length, or
its runtime increases exponentially.}  On the other hand, if the
criterion used is the Labelled Recall rate, then there is an
algorithm, the General Recall algorithm.

\newcommand{\probsym}[2]{$#1$\makebox[0in][l]{ $(#2)$}}

\begin{figure}
\begin{center}
\hspace{-.5in}
\begin{tabular}{ccccc}
\leaf{$x$} \faketreewidth{AA}
\leaf{$x$} \faketreewidth{AA}
\branch{2}{$B$}
\leaf{$C$}
\branch{2}{\probsym{A}{0.3}}
\hspace{-.25in} \tree &
\leaf{$B$}
\leaf{$x$} \faketreewidth{AA}
\leaf{$x$} \faketreewidth{AA}
\branch{2}{$C$}
\branch{2}{\probsym{A}{0.3}}
\hspace{-.25in} \tree &
\leaf{$x$} \faketreewidth{AA}
\leaf{$x$} \faketreewidth{AA}
\branch{2}{\probsym{B}{1.0}}
\hspace{-.25in} \tree &
\leaf{$x$} \faketreewidth{AA}
\leaf{$x$} \faketreewidth{AA}
\branch{2}{\probsym{C}{1.0}}
\hspace{-.25in} \tree &
\leaf{$C$} \faketreewidth{AA}
\leaf{$B$} \faketreewidth{AA}
\branch{2}{\probsym{A}{0.4}}
\hspace{-.25in} \tree
\end{tabular}
\medskip
\begin{tabular}{|l||c|c|}
\hline
Parse Tree &
\leaf{$x$} \faketreewidth{AA}
\leaf{$x$} \faketreewidth{AA}
\branch{2}{$B$}
\leaf{$x$} \faketreewidth{AA}
\leaf{$x$} \faketreewidth{AA}
\branch{2}{$C$}
\branch{2}{$A$}
\tree &
\leaf{$x$} \faketreewidth{AA}
\leaf{$x$} \faketreewidth{AA}
\branch{2}{$C$}
\leaf{$x$} \faketreewidth{AA}
\leaf{$x$} \faketreewidth{AA}
\branch{2}{$B$}
\branch{2}{$A$}
\tree 
\\ \hline
Derivation&&\\Probability & 0.3 & 0.4 \\ \hline
Parse&&\\Probability & 0.6 & 0.4 \\ \hline
\end{tabular}
\end{center}
\caption{Example Stochastic Tree Substitution Grammar and two parses} \label{fig:multistsg}
\end{figure}

In Figure \ref{fig:multistsg} we give a sample STSG grammar, and two
different parses of the string $xxxx$, demonstrating that the most
likely derivation differs from the most likely parse.

\begin{figure}
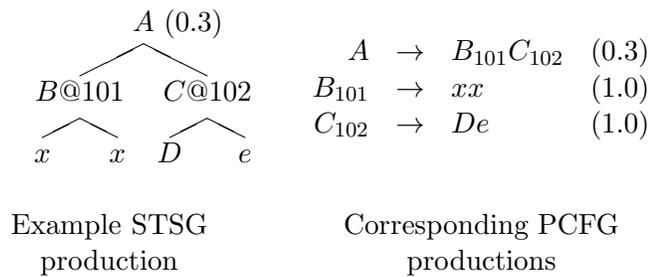

\begin{center}
\begin{tabular}{cc}
\leaf{$x$} \faketreewidth{AA}
\leaf{$x$} \faketreewidth{AA}
\branch{2}{$B@101$}
\leaf{$D$} \faketreewidth{AA}
\leaf{$e$} \faketreewidth{AA}
\branch{2}{$C@102$}
\branch{2}{\probsym{A}{0.3}}
\hspace{-.25in} \tree &
$\begin{array}{rcll}
A &\rightarrow &B_{101} C_{102} & (0.3) \\
B_{101}& \rightarrow &xx&(1.0) \\
C_{102}& \rightarrow &De&(1.0) 
\end{array}$
\\\\
Example STSG & Corresponding PCFG \\
production & productions
\end{tabular}
\end{center}
\caption{STSG to PCFG conversion example}\label{fig:STSGtoPCFG}
\end{figure}

To use the General Recall algorithm for parsing STSGs, we must first
convert the STSG to a PCFG.  This conversion is done in the usual way.
That is, assign to every internal node of every subtree of the STSG a
unique label, which we will indicate with an @ sign.  Leave the root
and leaf nodes unlabelled, or, equivalently, labelled with a null
label.  Now, for each root $A$ of a subtree with probability $p$, and
children $B@k$, $C@l$, create a PCFG rule
$$
\begin{array}{rcll}
A & \rightarrow &B_k \; C_l & (p) \\
\end{array}
$$
For each internal node $A@j$ of a subtree with children $B@k$, $C@l$,
create a PCFG rule
$$
\begin{array}{rcll}
A_j & \rightarrow & B_k \; C_l & (1.0)
\end{array}
$$
Here, the notation $X_j$ denotes a new nonterminal, created for this
node.  $k$ or $l$ may be null if the corresponding node is a leaf
node.  The number in parentheses indicates the probability associated
with a given production.  Figure \ref{fig:STSGtoPCFG} illustrates this
conversion.

\def\MAPSTSG{\mathop{\rm MAP\_STSG}\nolimits}

Now, if we were to apply the Viterbi algorithm directly to this PCFG,
rather than returning the most likely parse tree, it would return a
parse tree corresponding to the most likely derivation.  Not only
that, but the labels in this parse tree would not even be the same as
the labels in the original grammar.  Of course, we could solve this
latter problem by creating a function $\MAPSTSG$  that would map
from symbols in the PCFG to symbols in the STSG.
\begin{eqnarray*}
\MAPSTSG(X)   & = & X \\
\MAPSTSG(X_j) & = & X 
\end{eqnarray*}
In other words, $\MAPSTSG$ removes subscripts.  

One way to use this function would be to use the Viterbi algorithm or
Labelled Recall algorithms with the PCFG to find the best tree, and
then use the function to map each node's label to the original name.
But we would presumably get better results following a more principled
approach, such as trying to solve the following optimization problem:
$$
\arg\max_{T_G} E \left( \sum_{\constituent{j}{X}{k} \in T_C} \bigoneif{\constituent{j}{\MAPSTSG(X)}{k} \in T_G} \right)
$$

\def\MAPIDENT{\mathop{\rm MAP\_IDENT}\nolimits}
\def\MAPBRACK{\mathop{\rm MAP\_BRACK}\nolimits}

Here, $T_C$ is a parse tree with symbols in the PCFG, while $T_G$ is a
parse tree with symbols in the original STSG grammar.  This
maximization is exactly parallel to the maximization performed by the
Labelled Recall  algorithm, except that the symbol names are 
transformed with $\MAPSTSG$ before the maximization is
performed.  In fact, if we define two other mapping functions,
$\MAPIDENT$ (the identity mapping function) and $\MAPBRACK$ (the
function that maps everything to the same symbol), then the
similarity between the three maximizations becomes more clear.

The $\MAPIDENT$ function
\begin{eqnarray}
\MAPIDENT(X) & = & X 
\end{eqnarray}
can be used to define the Labelled Recall maximization,
$$
\max_{T_G} E \left( \sum_{\constituent{j}{X}{k} \in T_C} \bigoneif{\constituent{j}{\MAPIDENT(X)}{k} \in T_G} \right)
$$
while the $\MAPBRACK$ function 
$$
\MAPBRACK(X) = S
$$
can be used to define the Bracketed Recall maximization.
$$
\max_{T_G} E \left(\sum_{\constituent{j}{X}{k} \in T_C}
\bigoneif{\constituent{j}{\MAPBRACK(X)}{k} \in T_G }\right)
$$

%OOPS!  the mapped form isn't in T_G.  I need to map T_G as well.

If we substitute
$$
\mi{maxG} := \max_X \sum_{Y\vert X =\mi{map}(Y)}g(i, Y, j);
$$
into the Labelled Recall algorithm of Figure \ref{fig:maxcons}, we get
the General Recall algorithm.  $\mi{map}$ is a function that maps from
nonterminals in the relevant PCFG to nonterminals relevant for the
output.  It could be any of $\MAPSTSG$, $\MAPIDENT$, or $\MAPBRACK$,
among others.

\def\MAPSEM{\mathop{\rm MAP\_SEM}\nolimits}

Given this general algorithm, other uses for this technique become
apparent.  For instance, in the latest version of the Penn Treebank,
%todo: cite Marcus!
nonterminals annotated with grammatical function are included.  That
is, rather than using simple nonterminals such as $\mi{NP}$, more complex
nonterminals, such as $\mi{NP-SBJ}$ and $\mi{NP-PRD}$ are used (to indicate subject
and predicate).  Depending on the application, one might choose to use
the annotated nonterminals in a PCFG, but to return only the simpler,
unannotated nonterminals (\newcite{Collins:97a} uses a version of the
grammatically annotated nonterminals internally, parsing with the
Labelled Tree (Viterbi) algorithm, and then stripping the
annotations.)  So, we could define a mapping function $\MAPSEM$.
\begin{eqnarray*} 
\MAPSEM(X)   & = & X \\
\MAPSEM(X\mbox{-}SY) & = & X 
\end{eqnarray*}
We could then use the General Recall algorithm with this function.  If
we used a simpler scheme, such as parsing using the Viterbi algorithm,
and then mapping the resulting parse, problems might occur.  For
instance, since noun phrases are divided into classes, while verb
phrases are not, an inadvertent bias against noun phrases will have
been added into the system.  Using the General Recall algorithm
removes such biases, since all the different kinds of noun-phrases are
summed over, and is theoretically better motivated.

\begin{figure}
$$
\begin{array}{cl}
\mbox{\bf Item form:} \\
{[i, A, j]} & \mbox{inside semiring} \\
{[i, A, j]^{\dagger}} & \mbox{inside semiring} \\
{[i, A, j]^*} & \mbox{arctic semiring, or similar} \\ \\
\mbox{\bf Primary Goal:} \\
{[1, S, n+1]^*} \\\\
\mbox{\bf Secondary Goal:} \\
{[1, S, n+1]} \\\\
\mbox{\bf Rules:} \\
\infer{R(A \rightarrow w_i)}{[i, A, i+1] }{} & \mbox{Unary} \\
\infer{R(A \rightarrow BC)\rspace [i, B, k]\rspace [k, C, j]}{[i, A,
j]}{}
& \mbox {Binary} \\
\infer{\VZVin([i, A, j],[1, S, n+1])}{[i, B, j]^{\dagger}}{\mi{map}(A)=B} &
\mbox{Summation} \\
\infer{R(A, w_i, [i, i+1]^{\dagger}, [1, S, n+1])}{[i, i+1]^*}{} &
\mbox{Unary Labels} \\
\infer{R(A, B, C, [i, j]^{\dagger})\rspace [i, k]^* \rspace [k, j]^*}{[i, j]^*}{} 
& \mbox {Binary Labels}

\end{array}
$$
\caption{General Recall Description}\label{fig:generalrecalldesc}
\end{figure}

An item-based description of the General Recall algorithm is given in
Figure \ref{fig:generalrecalldesc}.  The General Recall description is
the same as the Bracketed Recall description, except that we now add
nonterminals to the items, and use the function $\mi{map}$.  In the
next chapter, we will show how to use the General Recall algorithm to
greatly speed up Data-Oriented Parsing \cite{Bod:92a}.

\section{N-Ary Branching Parse Trees}
\label{sec:nary}
Previously, we have been discussing binary branching parse trees.
However, in practice, it is often useful to have trees with more than
two branches.  In this section, we discuss n-ary branching parse trees
(denoted by the symbol $\stackrel{n}{T}$).  In general, n-ary
branching parse trees can have any number of branches, including zero
or one, but the discussion in this chapter will be limited to those
trees where every node has at least two branches.

We begin by discussing why, with n-ary branching parse trees,
different evaluation metrics should be used from those for binary
branching ones.  In particular, we discuss Consistent Brackets
Recall (Crossing Brackets) and Consistent Brackets Tree (Zero Crossing
Brackets) versus Bracketed or Labelled Precision and Recall.  We
then go on to discuss approximations for Bracketed and Labelled
Precision and Recall, the Bracketed Combined rate and Labelled
Combined rate.  Next, we show an algorithm for maximizing these
two rates.  Finally, we give results using this new algorithm.

\subsection{N-Ary Branching Evaluation Metrics}
\label{sec:narybranch}
If we wish to model n-ary branching trees, rather than just binary
branching ones, we could continue to use the same metrics as before.
For instance, we could use the Consistent Brackets Recall rate or
Consistent Brackets Tree rate.  However, both of these have a problem,
that simply by returning trees like
\begin{center}
\leaf{$w_1$}
\leaf{$w_2$}
\leaf{$...$}
\leaf{$w_{n-1}$}
\leaf{$w_n$}
\branch{5}{$S$}
\tree
\end{center}
i.e., trees that have only one node and all terminals as children of
that node, a 100\% score can be achieved.  The problem here is that
there is no reward for returning correct constituents, only a penalty
for returning incorrect ones.  The Labelled Recall rate has the
opposite problem: there is no penalty for including too many nodes,
e.g. returning a binary branching tree.

What is needed is some criterion that combines both the penalty and
the reward.  The traditional solution has been to use a pair of
criteria, the Bracketed Recall rate and the Bracketed Precision rate,
or more common recently, the Labelled Recall rate and the Labelled
Precision rate.  The Bracketed Recall rate gives the reward, yielding
higher scores for answers with more correct constituents, while the
Bracketed Precision rate gives the penalty, giving lower scores when
there are more incorrect constituents.  These Precision Rates are
defined as follows, and the Recall rates are included for comparison:

\begin{itemize}

\item {\em Labelled Precision Rate} = $L/N_G$.

\item {\em Bracketed Precision Rate} = $B/N_G$.

\item {\em Labelled Recall Rate} = $L/N_C$.

\item {\em Bracketed Recall Rate} = $B/N_C$.
\end{itemize}

Ideally, we would now develop an algorithm that maximizes some
weighted sum of a Recall rate and a Precision rate.  However, it is
relatively time-consuming to maximize Precision, because the
denominator is $N_G$, the number of nodes in the guessed tree.  Thus,
the relative penalty for making a mistake differs, depending on how
many constituents total are returned, slowing dynamic programming
algorithms.  On the other hand, we can create metrics closely related
to the Precision metrics, which we will call Mistakes and define as
follows:
\begin{itemize}

\item {\em Labelled Mistakes Rate} = $(N_G-L)/N_C$.

\item {\em Bracketed Mistakes Rate} = $(N_G-B)/N_C$.

\end{itemize}
$N_G$ is the number of guessed constituents, and $L$ is the number of
correctly labelled constituents, so $N_G-L$ is the number of
constituents that are not correct, i.e. mistakes.  We normalize by
dividing through by $N_C$, the number of constituents in the guessed
parse.  While this factor is less intuitive than $N_G$, it will remain
constant across parse trees, making the cost of each mistake
independent of the size of the guessed tree.

Now, we can define combinations of Mistakes and Recall, using a
weighting factor $\lambda$:
\begin{itemize}

\item {\em Labelled Combined Rate} = $L/N_C - \lambda (N_G-L)/N_C$

\item {\em Bracketed Combined Rate} = $B/N_C - \lambda (N_G-B)/N_C$

\end{itemize}
The positive term, $L/N_C$, provides a reward for correct
constituents, and the negative term $\lambda (N_G-L)/N_C$ provides a
penalty for incorrect constituents.

\subsection{Combined Rate Maximization}
We can now state the Labelled Combined Rate Maximization problem, as
\newcommand{\TNG}{{\stackrel{n}{T}}_G}
\newcommand{\TNC}{{\stackrel{n}{T}}_C}
$$
\arg \max_{\TNG} E(L/N_C - \lambda (N_G-L)/N_C)
$$
Using manipulations similar to those used previously, we can rewrite
this as
\begin{equation}
\label{eqn:labcom1}
\arg \max_{\TNG} \sum_{\constituent{i}{X}{j} \in \TNG} g(i, X, j)-\lambda(1-g(i,X,j)))
\end{equation}

The preceding expression says that we want to find that parse tree
which maximizes our score, where our score is one point for each
correct constituent and $-\lambda$ points for each incorrect one.  

The preceding expression does not address an
interesting question.  In general, if we are inducing n-ary branching
trees, it will be because we are interested in grammars with other
than two expressions on the right hand sides of their productions.
While grammars with zero or one elements on their right hand side are
interesting, they are in general significantly harder to parse, so we
will only be concerned in this chapter with grammars with at least two
nonterminals on the right side of
productions.\footnote{\label{note:Collinsbug} Furthermore,
even though using the techniques of Chapter \ref{ch:semi} we can parse
grammars that have loops from unary or epsilon productions, it becomes
more complicated to define precision and recall appropriately for
these grammars.  For instance, given our definitions, recall scores
above 100\% are possible by returning trees of the form
\begin{center}
\leaf{$S$}
\branch{1}{$S$}
\branch{1}{$S$}
\branch{1}{$S$}
\tree
\end{center}
because this could lead to $L$ or $B$ exceeding $N_C$, since the
repeated $S$ constituents were all counted as correct.  In fact, when
we examined scoring code used by \newcite{Collins:96a}, we
found this problem; while Collins (personal communication) says that
his parser could not return trees of this form, it illustrates the
problems that begin to surface in scoring trees with unary branches.}

Grammars with at least two elements on the right can easily be
converted to grammars with exactly two elements on the right, using a
trick best known from Earley's algorithm, in which new nonterminals
%cite Earley?
are created by dotting.  A rule such as 
$$
\begin{array}{rcll}
A & \rightarrow & B \;\; C\;\;D\;\;E & (p)
\end{array}
$$
is converted into three rules of the form
$$
\begin{array}{rcll}
A & \rightarrow & B \;\; B \!\bullet\! CDE & (p) \\
B \!\bullet\! CDE & \rightarrow & C \;\; BC \!\bullet\! DE & (1) \\
BC \!\bullet\! DE & \rightarrow & D \;\; E & (1)
\end{array}
$$
This works fine in that it produces a grammar which produces the same
strings with the same probabilities, and which is amenable to simple
chart parsing.  However, the parse trees produced are binary
branching.

Let us define an operator, $\mi{nodot}(X)$ that is true if and only
if the label does not have a dot in it, and then modify Expression
\ref{eqn:labcom1} to be
\begin{equation}
\arg \max_{\TNG} \sum_{\constituent{i}{X}{j} \in \TNG \vert \smi{nodot}(X)}
    g(i, X, j)-\lambda(1-g(i, X, j))
\label{eqn:naivenary}
\end{equation}
The modified expression does not count the value of any dotted nodes.
There is one more complication: for any given constituent, even those
without a dot, it may contribute a negative score, even though its
children contribute positive scores.  We thus want to be able to omit
any constituent which contributes negatively, while keeping the
children.  The final expression we maximize, which returns a binary
branching tree, is
\begin{equation}
\label{eqn:labcom4}
\arg\max_{{T}_G} \sum_{\constituent{i}{X}{j} \in T_G \vert
\smi{nodot}(X)} \max\left(0, g(i, X, j)-\lambda(1-g(i, X, j))\right)
\end{equation}
We take the binary branching tree which maximizes this expression, and
remove all dotted nodes, and all nodes which contribute a negative
score, while leaving their children in the tree.  The resulting tree
is the optimal n-ary branching tree.  Because expression
\ref{eqn:labcom4} returns a binary branching tree, we can use our
previous CKY-style parsing algorithms to efficiently perform the
maximization, and remove the dummy and negative nodes as a
post-processing step.

A similar maximization can be used for the bracketed case.  We first
define $g(i,j)$ as
$$
g(i,j) = \sum_{X|\smi{nodot}(X)} g(i, X, j)
$$
Now we can give the expression for Bracketed Combined maximization:
$$
\arg \max_{T_G} \sum_{\constituent{i}{X}{j} \in {T_G}} \max \left(0, g(i, j)-\lambda(1-g(i,j)) \right)
$$

Using these expressions, we could modify the Labelled Recall algorithm
to maximize either the Labelled Combined rate or the Bracketed
Combined rate, or more generally, we could modify the General Recall
algorithm.  If we substitute
$$
\mi{bestG} := \max_X \sum_{Y\vert X =\mi{map}(Y)}g\constituent{i}{Y}{j};
$$
$$
\mi{maxG} := \max \left(0, \mi{bestG} - \lambda\times(1-\mi{bestG}\right);
$$
into the Labelled Recall algorithm of Figure \ref{fig:maxcons}, we get
the general algorithm for n-ary branching parse trees, which we call
the General Combined algorithm.

\begin{figure}
$$
\begin{array}{cl}
\mbox{\bf Item form:} \\
{[i, A, j]} & \mbox{inside semiring} \\
{[i, A, j]^*} & \mbox{arctic semiring, or similar} \\ \\
\mbox{\bf Primary Goal:} \\
{[1, S, n+1]^*} \\\\
\mbox{\bf Secondary Goal:} \\
{[1, S, n+1]} \\\\
\mbox{\bf Rules:} \\
\infer{R(A \rightarrow w_i)}{[i, A, i+1] }{} & \mbox{Unary} \\
\infer{R(A \rightarrow BC)\rspace [i, B, k]\rspace [k, C, j]}{[i, A,
j]}{}
& \mbox {Binary} \\
\infer{R(A, w_i, \mi{combined}(i, A, i+1))}{[i, A,i+1]^*}{\mi{combined}(i, A,i+1) > 0} &
\mbox{Unary Labels} \\
\infer{R(A, B, C, \mi{combined}(i, A, j))\rspace [i, B, k]^*\rspace [k, C, j]^*}
{[i, A, j]^*}{\mi{combined}(i, A, j) > 0} 
& \mbox {Binary Labels} \\
\infer{R(\bullet, w_i, \mi{combined}(i, A, i+1))}{[i, A,i+1]^*}{\mi{combined}(i, A, i+1) = 0} &
\mbox{Unary Omit} \\
\infer{R(\bullet, B, C, \mi{combined}(i, A, j))\rspace [i, B, k]^*\rspace [k, C, j]^*}
{[i, A, j]^*}{\mi{combined}(i, A, j) = 0} 
& \mbox {Binary Omit}

\end{array}
$$
\caption{N-ary Labelled Recall Description}\label{fig:nlabelledrecalldesc}
\end{figure}

In Figure \ref{fig:nlabelledrecalldesc} we give an item-based
description for the Labelled Combined  algorithm, very similar to
the Labelled Recall description.  This algorithm uses the function
$\mi{combined}$ to give the score of constituents:
$$
\mi{combined}(i, A, j) = 
\left\{ \begin{array}{ll}
\max\left(0, \begin{array}{l}
  \VZVin([i, A, j], [1, S, n+1]) \\
  -\lambda(1-\VZVin([i, A, j], [1, S, n+1])) 
  \end{array} \right)& 
  \mbox{\it if }\mi{nodot}(X) \\
0 & \mi{otherwise}
\end{array}  \right.
$$
Rather than just using the unary labels rules, the description also
contains unary omit and binary omit rules, which are triggered for
items whose expected contribution is 0.  These items then use rules
with a $\bullet$ on the left hand side in the derivation, which can
be used to reconstruct an appropriate n-ary branching tree.

\subsection{N-Ary Branching Experiments}

\begin{figure}
\psfig{figure=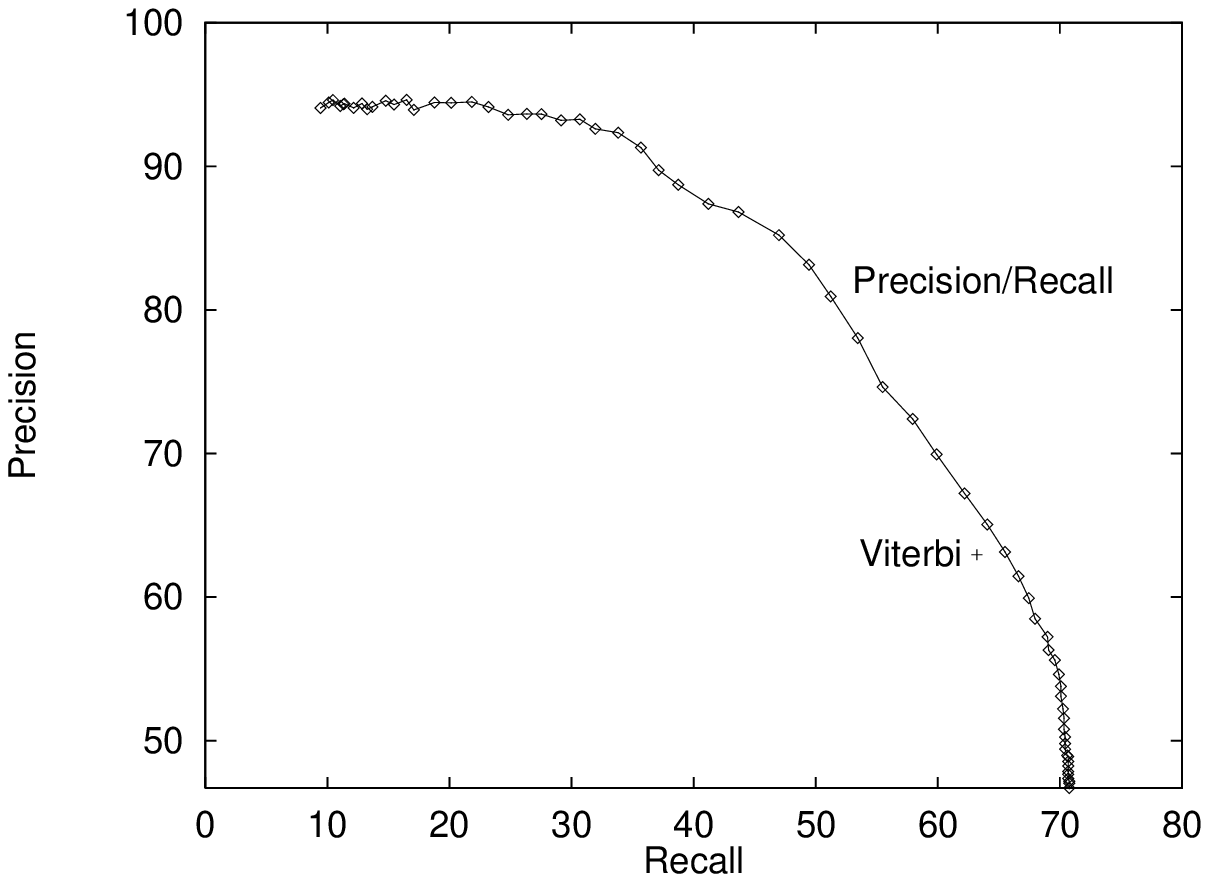,width=7in,height=6in}
\caption{Labelled Combined Algorithm vs. Labelled Tree Algorithm}
\label{fig:combinedrate}
\end{figure}

We performed a simple experiment using the Penn Treebank, version
II, sections 2-21 for training, section 23 for test, extracting a
grammar in the same way as in Section \ref{sec:countgrammar}.  We
varied the precision-recall tradeoff over a range of 65 values.

Figure \ref{fig:combinedrate} gives the results.  As can be seen, the
Labelled Combined algorithm not only produced a smooth tradeoff
between Labelled Precision and Labelled Recall, it also worked better
than the Labelled Tree (Viterbi) algorithm on both measures
simultaneously.  The algorithm asymptotes vertically at about 70\%
Labelled Recall.  This asymptote corresponds to such a strong weight
on recall that the resulting tree is binary branching.  The horizontal
asymptote is at 94\% Labelled Precision, corresponding to a single
constituent containing the entire sentence.  The reason the asymptote
does not approach 100\% is that we measured Labelled Precision;
because of our scoring mechanism, not all top nodes in the Penn
Treebank were labelled the same, leading to occasional mistakes in the
top node label.\footnote{In the Penn Treebank version II, the top
bracket is always unlabelled and unary branching.  
%see ~/cellar/wsj2/doc/manual/punct.tex, first paragraph
We collapsed unary branches before scoring, since this algorithm could
not produce unary branches.  This lead to a top node that was usually
an {\it S} but could be something else, such as {\it S-INV} for
questions.}

%timings from profile.  Tried timing using time command, but timings
%were unreliable.  Divided some times by 2, because they were both
%labels and brackets.
%
%                                  called/total       parents 
%index  %time    self descendents  called+self    name    	index
%                                  called/total       children
%
%                0.14     1635.07       1/1           _start [3]
%[1]     90.0    0.14     1635.07       1         main [1]
%              602.71        0.00      86/86          ParseOut__FPiiR10CNFGrammarPPPd [4]
%              419.91        0.05      86/86          Parse__FPiiR10CNFGrammarPd [5]
%              301.71        0.00      86/86          ParseIn__FPiiR10CNFGrammarPd [6]
%                0.10      184.54   39920/39920       GetNextSentence__Fb [7]
%                0.07       71.61   11266/11266       nextMatch__8ScoreSetP4NodeN21iPi [18]
%               11.71        6.83   11180/11180       ParseMax2__FiPdfPP15ParseCellSymbol [30]
%                1.52        8.82   39918/51184       NodeCNF__FP4Node [35]
%                1.01        8.13   39832/39832       NodeMakeContinuations__FP4Node [39]
%                6.18        0.00     172/172         ParseMax1__FPiiR10CNFGrammarPPPdT3b [53]
%                2.03        3.99       1/1           NodeMakeGrammar__FPP4Nodei [56]
%                2.77        0.00   91188/231528      _$_4Node [50]
%                1.22        0.00      86/86          ParseEFSum__FiPPPdT1 [74]
%                0.00        0.04      86/86          NodeSymbols__FP4NodePi [101]

\begin{table}
\begin{center}
\begin{tabular}{|l|r|} \hline 
Algorithm  & Time \\ \hline \hline
Viterbi algorithm & 420 \\ \hline
Inside algorithm & 302 \\ \hline
Outside algorithm & 603 \\ \hline
Find 1 Combined tree & 3 \\ \hline
Find 65 Combined trees & 10 \\ \hline
\end{tabular}
\end{center}
\caption{Algorithm Timings in Seconds} \label{fig:timings}
\end{table}

Table \ref{fig:timings} gives the timings of the various algorithms in
seconds.  The Viterbi algorithm and inside algorithm take
approximately the same amount of time, while the outside algorithm
takes about twice as long.  Once the inside and outside probabilities
are computed, the Labelled Combined trees can be computed very quickly.  A
portion of the work in computing combined trees, computing the $g$
values, needs to be done only once per sentence, which is why the time
to find 65 trees is not 65 times the time of finding 1 tree.  It is
very significant that almost all of the work to compute a combined
tree needs only to be done once: this means that if we are going to
compute one tree, we might as well compute the entire ROC
(precision-recall) curve.

The fact that we can compute a precision-recall curve makes it much
easier to compare parsing algorithms.  If there are two parsing
algorithms, and one gets a better score on Labelled Precision, and the
other gets a better score on Labelled Recall, we cannot determine
which is better.  However, if for one or both algorithms, an ROC curve
exists, then we can determine which is the better algorithm, unless of
course the curves cross.  But even in the crossing case, we have
learned the useful fact that one algorithm is better for some things,
and the other algorithm is better for other things.

\section{Conclusions}

\begin{table}
\begin{center}
\begin{tabular}{|l||l|l|} \hline
                & Bracketed        & Labelled \\ \hline \hline
Tree            & (NP-Complete)    & Labelled Tree \\ \hline 
Recall          & Bracketed Recall & Labelled Recall \\ \hline
\end{tabular}
\end{center}
\caption{Metrics and corresponding algorithms} \label{tab:ressum}
\end{table}

Matching parsing algorithms to evaluation criteria is a powerful
technique that can be used to achieve better performance than standard
algorithms.  In particular, the Labelled Recall algorithm has better
performance than the Labelled Tree algorithm on the Consistent
Brackets Recall, Labelled Recall, and Bracketed Recall rates.
Similarly, the Bracketed Recall algorithm has better performance than
the Labelled Tree algorithm on Consistent Brackets and Bracketed
Recall rates.  Thus, these algorithms improve performance not only on
the measures that they were designed for, but also on related
criteria.  For n-ary branching trees, we have shown that the Labelled
Combined algorithm can lead to improvements on both precision and
recall, and can allow us to trade precision and recall off against
each other, or even to quickly produce the curve showing this
tradeoff.

Of course, there are limitations to the approach.  For instance, the
problem of maximizing the Bracketed Tree rate (equivalent to Zero
Crossing Brackets rate in the case of binary branching data) is
NP-Complete: not all criteria can be optimized directly.

In the next chapter, we will see that these techniques can also in
some cases speed parsing.  In particular, we will introduce
Data-Oriented Parsing \cite{Bod:92a}, and show that while we cannot
efficiently maximize the Labelled Tree rate, we can use the General
Recall algorithm to maximize the Labelled Recall rate in time
$O(n^3)$, leading to significant speedups.

{\beginappendix
\appendixsection{Proof of Crossing Brackets Theorem}
\label{sec:crossproof} 

\begin{figure}[hpb] %looks funny when at top, preceding appendix...
\begin{center}
\mbox{\psfig{figure=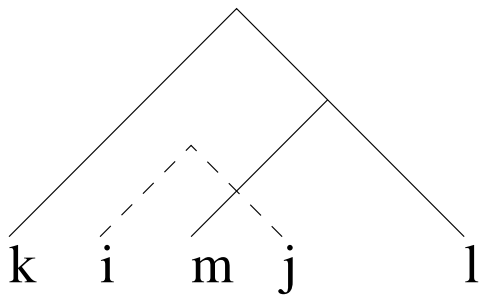,height=1.5in}}
\caption{Crossing trees}\label{fig:crosstree}
\end{center}
\end{figure}

In this appendix, we prove that if the correct parse tree is binary
branching, then Consistent Brackets and Bracketed Match are identical.
\newcite{Pereira:92a} make reference to an equivalent observation, but
without proof.  The proof is broken into two parts.  In the first
part, we show that any match does not cross.  In the second part, we
show that any constituent $\constituenttwo{i}{j}$ that does not have a
Bracketed Match does cross.

To show that any element which does match does not cross, we note that
by a simple induction, the left descendants of any constituent can
never cross the right descendants of that constituent.  Now, assume
that an element which did match, also crossed.  Find the lowest common
ancestor of the match and the crossing element.  One of them must be a
descendant of the left child, and the other must be a descendant of
the right child.  But then, by the lemma, they would not cross, and we
have a contradiction, so our assumption must be wrong.

To show that any element $\constituenttwo{i}{j}$ that does not match
must have a crossing element, we find the smallest element
$\constituenttwo{k}{l}$ containing $\constituenttwo{i}{j}$ (that is,
$k \leq i$ and $l \geq j$).  Now, since $\constituenttwo{i}{j}$
doesn't match $\constituenttwo{k}{l}$, one of these inequalities must
be strict (that is, $k < i$ or $l > j$).  Assume without loss of
generality that $l > j$.  Since we assumed that $T_C$ is binary
branching, $\constituenttwo{k}{l}$ has two children.  Let
$\constituenttwo{m}{l}$ represent the right child.  This configuration
is illustrated in Figure \ref{fig:crosstree}.  We know that $m > i$,
since otherwise $\constituenttwo{m}{l}$ would be a constituent smaller
than $\constituenttwo{k}{l}$ containing $\constituenttwo{i}{j}$.
Similarly, we know that $m < j$, since otherwise $\constituenttwo{k}{m}$ would be a
constituent smaller than $\constituenttwo{k}{l}$.  Thus, we have $i < m < j < l$,
meeting the definition for a cross.

\appendixsection{Glossary}
\label{sec:glossary}
\begin{description}
\item[$B$] Number of correctly bracketed constituents; see Section \ref{sec:evalmet}.
\item[Bracketed Mistakes rate] $(N_G-B)/N_C$.  Approximation to
unlabelled precision.  See Section \ref{sec:narybranch}.
\item[Bracketed Combined rate] $B/N_C - \lambda (N_G-B)/N_C$.  The
weighted difference of Bracketed Recall and Bracketed Mistakes.  See
Section \ref{sec:narybranch}.
\item[Bracketed Precision rate] $B/N_G$.  A score which penalizes
incorrect guesses.  See Section \ref{sec:narybranch}.
\item[Bracketed Recall algorithm] Algorithm maximizing Bracketed
Recall rate.  See Section \ref{sec:brackconspars}.
\item[Bracketed Recall rate] $B/N_C$.  Closely related to Consistent
Brackets rate.  See Section \ref{sec:evalmet}.
\item[Bracketed match] $\constituent{i}{X}{j}$ bracketed matches
$\constituent{i}{Y}{j}$.  See Section \ref{sec:evalmet}.
\item[Bracketed Tree rate] $\bigoneif{B = N_C}$  See Section \ref{sec:evalmet}.
\item[$C$] Number of constituents that do not cross a correct constituent; see Section \ref{sec:evalmet}.
\item[Consistent Brackets Recall rate] $C/N_G$.  Often called Crossing
Brackets rate.  When the parses are binary branching, the same as the
Bracketed Recall rate.
\item[Consistent Brackets match] $\constituent{i}{X}{j}$ matches if
there is no $\constituent{k}{Y}{l}$ crossing it.
\item[Consistent Brackets Tree rate] $\bigoneif{C = N_G}$  Closely
related to Bracketed Tree rate.  When the parses are binary branching,
the two metrics are the same.  Also called the Zero Crossing Brackets
rate. See Section \ref{sec:evalmet}.
\item[Crossing Brackets rate] Conventional name for Consistent
Brackets rate.  
\item[$E$] Expected value function.
\item[$\mi{inside}(i, X, j)$] Inside value.
\item[Exact Match rate] Conventional name for Labelled Tree rate.
\item[$\mi{outside}(i, X, j)$] Outside value.
\item[$g(i, X, j)$] Normalized inside-outside value.
\item[$L$] Number of correct constituents; see Section \ref{sec:evalmet}.
\item[Labelled Mistakes rate] $(N_G-L)/N_C$.  An approximation to
Labelled Precision.  See Section \ref{sec:narybranch}.
\item[Labelled Combined rate] $L/N_C - \lambda (N_G-L)/N_C$.  The
weighted difference of Labelled Recall and Labelled Mistakes.  See Section \ref{sec:narybranch}.
\item[Labelled Precision rate] $L/N_G$.  A score which penalizes
incorrect guesses.  See Section \ref{sec:narybranch}.
\item[Labelled Recall algorithm] Algorithm for maximizing Labelled
Recall rate.  See Section \ref{sec:labconspars}. 
\item[Labelled Recall rate] $L/N_C$.  See Section \ref{sec:evalmet}.
\item[Labelled match] $\constituent{i}{X}{j}$ occurs in both correct
and guessed parse trees.
\item[Labelled Tree algorithm] Algorithm for maximizing Labelled Tree
rate.  Also called Viterbi algorithm.
\item[Labelled Tree rate] $\bigoneif{L = N_C}$  This metric is also called 
the Viterbi criterion or the Exact Match rate.
\item[$N_C$] Number of constituents in tree in treebank.
\item[$N_G$] Number of constituents in tree output by parser.
\item[$T_C$] Correct parse tree -- tree in treebank.
\item[$T_G$] Guessed parse tree -- tree output by parser.
\item[Viterbi algorithm] Well-known CKY style algorithm for maximizing
Labelled Tree rate.
\item[$w_i$] Word $i$ of input sentence.
\item[Zero Crossing Brackets rate] Conventional name for Consistent
Brackets Tree rate.
\end{description}

}%end secappendix

% -*- mode: latex; -*-

\chapter{Data-Oriented Parsing} \label{ch:DOP}

In this chapter we describe techniques for parsing the Data-Oriented
Parsing (DOP) model 500 times faster than with previous parsers.  This
work \cite{Goodman:96b} represents the first replication of the DOP
model, and calls into question the source of the previously reported
extraordinary performance levels.  The results in this chapter rely
primarily on two techniques: an efficient conversion of the DOP model
to a PCFG, and the General Recall algorithm of the preceding chapter.

\section{Introduction}
The Data-Oriented Parsing (DOP) model has an interesting and
controversial history.  It was introduced by Remko Scha
\shortcite{Scha:90a} and was then studied by Rens Bod.  Bod
\shortcite{Bod:93a,Bod:92a} was not able to find an efficient exact
algorithm for parsing using the model; however he did discover and
implement Monte Carlo approximations.  He tested these algorithms on a
cleaned up version of the ATIS corpus \cite{Hemphill:90a}, in which
inconsistencies in the data had been removed by hand.  Bod achieved
some very exciting results, reportedly getting 96\% of his test set
exactly correct, a huge improvement over previous results.  For
instance, Bod \shortcite{Bod:93b} compares these results to Schabes
\shortcite{Schabes:93a}, in which, for short sentences, 30\% of the
sentences have no crossing brackets (a much easier measure than exact
match).  Thus, Bod achieves an extraordinary 8-fold error rate
reduction.

Other researchers attempted to duplicate these results, but because of
a lack of details of the parsing algorithm in his publications, were
unable to confirm the results (Magerman, Lafferty, personal
communication).  Even Bod's thesis \cite{Bod:95a} does not contain
enough information to replicate his results.

Parsing using the DOP model is especially difficult.  The model can be
summarized as a special kind of Stochastic Tree Substitution Grammar
(STSG): given a bracketed, labelled training corpus, let {\em every}
subtree of that corpus be an elementary tree, with a probability
proportional to the number of occurrences of that subtree in the
training corpus.  Unfortunately, the number of trees is in general
exponential in the size of the training corpus trees, producing an
unwieldy grammar.

In this chapter, we introduce a reduction of the DOP model to an exactly
equivalent Probabilistic Context-Free Grammar (PCFG) that is linear in
the number of nodes in the training data.  Next, we show that the
General Recall algorithm (introduced in Section \ref{sec:generalalg}),
which uses the inside-outside probabilities, can be used to
efficiently parse the DOP model in time $O(Tn^3)$, where $T$ is
training data size.  This polynomial run time is especially
significant given that \newcite{Sima'an:96b} showed that computing the
most probable parse of an STSG is NP-Complete.  We also give a random
sampling algorithm equivalent to Bod's that runs in time $O(Gn^2)$
rather than Bod's $O(Gn^3)$ per sample, where $G$ is the grammar size.
We use the reduction and the two parsing algorithms to parse held out
test data, comparing these results to a replication of Pereira and
Schabes \shortcite{Pereira:92a} on the same data.  These results are
disappointing: both the Monte Carlo parser and the General Recall
parser applied to the DOP model perform about the same as the Pereira
and Schabes method.  We present an analysis of the runtime of our
algorithm and Bod's.  Finally, we analyze Bod's data, showing that
some of the difference between our performance and his is due to a
fortuitous choice of test data.

This work was the first published replication of the full DOP model,
i.e. using a parser that sums over derivations.  It also contains
algorithms implementing the model with significantly fewer resources
than previously needed.  Furthermore, for the first time, the DOP
model is compared to a competing model on the same data.

\section{Previous Research}
\label{sec:previous}
The DOP model itself is extremely simple and can be described as
follows: for every sentence in a parsed training corpus, extract every
subtree.  In general, the number of subtrees will be very large,
typically exponential in sentence length.  Now, use these trees to
form a Stochastic Tree Substitution Grammar (STSG).\footnote{STSGs
were described in Section \ref{sec:generalalg}.}  Each tree is
assigned a number of counts, one count for each time it occurred as a
subtree of a tree in the training corpus.  Each tree is then assigned a
probability by dividing its number of counts by the total number of
counts of trees with the same root nonterminal.

\begin{figure}
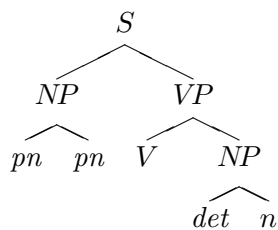


% Here is the main tree
\begin{center}
\leaf{$\mi{pn}$}
\leaf{$\mi{pn}$}
\branch{2}{$N\!P$}
\leaf{$V$}
\leaf{$\mi{det}$}
\leaf{$n$}
\branch{2}{$N\!P$}
\branch{2}{$V\!P$}
\branch{2}{$S$}
\tree
\end{center}
\caption{Training corpus tree for DOP example}
\label{fig:maintree}
\end{figure}

Given the tree of Figure \ref{fig:maintree}, we can use the DOP model
to convert it into the STSG of Figure \ref{fig:sampleDOP}.  The
numbers in parentheses represent the probabilities.  To give one
example, the tree
$$
\leaf{$\mi{det}$}
\leaf{$n$}
\branch{2}{$N\!P$} \tree
$$
is assigned probability $0.5$ because the tree occurs once in the
training corpus, and there are two subtrees in the training corpus
that are rooted in $N\!P$, so $\frac{1}{2}=0.5$.  The resulting STSG
can be used for parsing.

\begin{figure*}
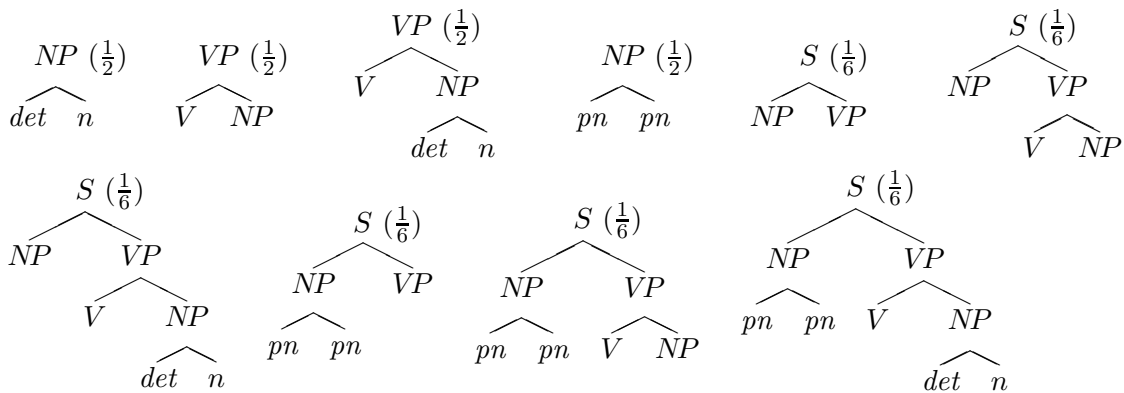

% Here we begin subtrees of S
% First, are the three with NP-N 
\leaf{$\mi{pn}$}
\leaf{$\mi{pn}$}
\branch{2}{$N\!P$}
\leaf{$V$}
\leaf{$\mi{det}$}
\leaf{$n$}
\branch{2}{$N\!P$}
\branch{2}{$V\!P$}
\branch{2}{$S$\makebox[0in][l]{ $(\frac{1}{6})$}}

\leaf{$\mi{pn}$}
\leaf{$\mi{pn}$}
\branch{2}{$N\!P$}
\leaf{$V$}
\leaf{$N\!P$}
\branch{2}{$V\!P$}
\branch{2}{$S$\makebox[0in][l]{ $(\frac{1}{6})$}}

\leaf{$\mi{pn}$}
\leaf{$\mi{pn}$}
\branch{2}{$N\!P$}
\leaf{$V\!P$}
\branch{2}{$S$\makebox[0in][l]{ $(\frac{1}{6})$}}

%Now are the three subtrees without N
\leaf{$N\!P$}
\leaf{$V$}
\leaf{$\mi{det}$}
\leaf{$n$}
\branch{2}{$N\!P$}
\branch{2}{$V\!P$}
\branch{2}{$S$\makebox[0in][l]{ $(\frac{1}{6})$}}

\leaf{$N\!P$}
\leaf{$V$}
\leaf{$N\!P$}
\branch{2}{$V\!P$}
\branch{2}{$S$\makebox[0in][l]{ $(\frac{1}{6})$}}

\leaf{$N\!P$}
\leaf{$V\!P$}
\branch{2}{$S$\makebox[0in][l]{ $(\frac{1}{6})$}}

%Now are the subtrees of the left NP
\leaf{$\mi{pn}$}
\leaf{$\mi{pn}$}
\branch{2}{$N\!P$\makebox[0in][l]{ $(\frac{1}{2})$}}

%Now are the subtrees of VP
\leaf{$V$}
\leaf{$\mi{det}$}
\leaf{$n$}
\branch{2}{$N\!P$}
\branch{2}{$V\!P$\makebox[0in][l]{ $(\frac{1}{2})$}}

\leaf{$V$}
\leaf{$N\!P$}
\branch{2}{$V\!P$\makebox[0in][l]{ $(\frac{1}{2})$}}

%Now are the subtrees of right NP
\leaf{$\mi{det}$}
\leaf{$n$}
\branch{2}{$N\!P$\makebox[0in][l]{ $(\frac{1}{2})$}}

\tree \tree \tree \tree \tree \tree \tree \tree \tree \tree 
\caption{Sample STSG Produced from DOP Model}
\label{fig:sampleDOP}
\end{figure*}

In theory, the DOP model has several advantages over other models.
Unlike a PCFG, the use of trees allows capturing large contexts,
making the model more sensitive.  Since every subtree is included,
even trivial ones corresponding to rules in a PCFG, novel sentences
with unseen contexts may still be parsed.

Because every subtree is included, the number of subtrees is huge;
therefore Bod randomly samples 5\% of the subtrees, throwing away the
rest.  This 95\% reduction in grammar size significantly speeds up
parsing.

As we discussed in Section \ref{sec:generalalg}, there are three ways
to parse a STSG and thus to parse DOP.  The two ways Bod considered
were the most probable derivation, and the most probable parse (best
according to the Labelled Tree criterion).  The most probable
derivation and the most probable parse may differ when there are
several derivations of a given parse, as previously illustrated in
Figure \ref{fig:multistsg}.  The third way to parse a STSG is to use
the General Recall algorithm to find the best Labelled Recall parse.

Bod \shortcite{Bod:93a} shows how to approximate the most probable
parse using a Monte Carlo algorithm.  The algorithm
randomly samples possible derivations, then finds the tree with the
most sampled derivations.  Bod shows that the most probable parse
yields better performance than the most probable derivation on the
exact match criterion.

\newcite{Sima'an:96a} implemented a version of the DOP model,
which parses efficiently by limiting the number of trees used and by
using an efficient most probable derivation model.  His experiments
differed from ours and Bod's in many ways, including his use of a
different version of the ATIS corpus; the use of word strings, rather
than part of speech strings; and the fact that he did not parse
sentences containing unknown words, effectively throwing out the most
difficult sentences.  Furthermore, Sima'an limited the number of
substitution sites for his trees, effectively using a subset of the
DOP model.  \newcite{Sima'an:96b} shows that computing the most
probable parse of a STSG is NP-Complete.

\section{Reduction of DOP to PCFG}
\label{sec:reduction}

Bod's reduction to a STSG is extremely expensive, even when throwing
away 95\% of the grammar.  However, it is possible to find an
equivalent PCFG that contains at most eight PCFG rules for each node
in the training data; thus it is $O(n)$.  Because this reduction is so
much smaller, we do not discard any of the grammar when using it.  The
PCFG is equivalent in two senses: first it generates the same strings
with the same probabilities; second, using an isomorphism defined
below, it generates the same trees with the same probabilities,
although one must sum over several PCFG trees for each STSG tree.

\begin{figure}
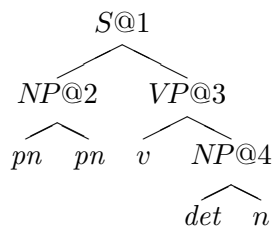

\begin{center}
\mbox{
\leaf{$\mi{pn}$}
\leaf{$\mi{pn}$}
\branch{ 2}{$N\!P@2$}
\leaf{$v$}
\leaf{$\mi{det}$}
\leaf{$n$}
\branch{2}{$N\!P@4$}
\branch{2}{$V\!P@3$}
\branch{2}{$S@1$}
\hspace{-4em}
\tree
}
\end{center}
\caption{Example tree with addresses} \label{fig:addressnodes}
\end{figure}

To show this reduction and equivalence, we must first define some
terminology.  We assign every node in every tree a unique number,
which we will call its address.  Let $A@k$ denote the node at address
$k$, where $A$ is the non-terminal labeling that node.  Figure
\ref{fig:addressnodes} shows the example tree augmented with
addresses.  We will need to create one new non-terminal for each node
in the training data.  We will call this non-terminal $A_k$.  We will
call non-terminals of this form ``interior'' non-terminals, and the
original non-terminals in the parse trees ``exterior.''

Let $a_j$ represent the number of nontrivial subtrees headed by the
node $A@j$.  Let $a$ represent the number of nontrivial subtrees headed by nodes
with non-terminal $A$, that is $a=\sum_j a_j$.

Consider a node $A@j$ of the form:
\begin{center}
\leaf{$B@k$} \faketreewidth{AA}
\leaf{$C@l$} \faketreewidth{AA}
\branch{2}{$A@j$}
\tree
\end{center}
How many nontrivial subtrees does it have?  Consider first the
possibilities on the left branch.  There are $b_k$ non-trivial
subtrees headed by $B@k$, and there is also the trivial case where the
left node is simply $B$.  Thus there are $b_k+1$ different
possibilities on the left branch.  Similarly, for the right branch
there are $c_l+1$ possibilities.  We can create a subtree by choosing
any possible left subtree and any possible right subtree.  Thus, there
are $a_j=(b_k+1)(c_l+1)$ possible subtrees headed by $A@j$.  In our
example tree of Figure \ref{fig:addressnodes}, both noun phrases have
exactly one subtree: $np_4=np_2=1$; the verb phrase has 2 subtrees:
$vp_3=2$; and the sentence has 6: $s_1=6$.  These numbers correspond
to the number of subtrees in Figure \ref{fig:sampleDOP}.

\begin{figure}
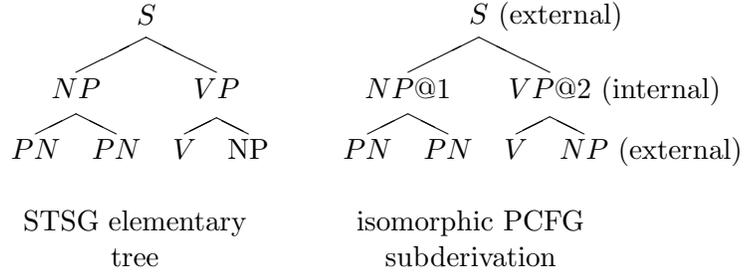

\begin{center}
\begin{tabular}{cc}
\leaf{$PN$}
\leaf{$PN$}
\branch{2}{$NP$}
\leaf{$V$}
\leaf{NP}
\branch{2}{$VP$}
\branch{2}{$S$}
\hspace{-0.3in}\tree &
\leaf{$PN$}
\leaf{$PN$}
\branch{2}{$NP@1$}
\leaf{$V$}
\leaf{$NP$\makebox[0in][l]{ (external)}}
\branch{2}{$VP@2$\makebox[0in][l]{ (internal)}}
\branch{2}{$S$\makebox[0in][l]{ (external)}} 
\hspace{-0.3in}\tree \\
\\
STSG elementary & isomorphic PCFG \\
tree & subderivation 
\end{tabular} 
\end{center}
\caption{STSG elementary tree isomorphic to a PCFG subderivation}
\label{fig:subderivation}
\end{figure}

We will call a PCFG subderivation isomorphic to a STSG elementary tree
if the subderivation begins with an external non-terminal, uses
internal non-terminals for intermediate steps, and ends with external
non-terminals.  Figure \ref{fig:subderivation} gives an example of an
STSG elementary tree taken from Figure \ref{fig:sampleDOP}, and an
isomorphic PCFG subderivation.  
%$S \Rightarrow\ \:NP@1\ \:VP@2 \Rightarrow PN\ \:PN\ \:VP@2
%\Rightarrow PN\ \:PN\ \:V\ \:NP$.

We will give a simple small PCFG with the following surprising
property: for every subtree in the training corpus headed by $A$, the
grammar will generate an isomorphic subderivation with probability
$1/a$.  In other words, rather than using the large, explicit STSG, we
can use this small PCFG that generates isomorphic derivations, with
identical probabilities.

The construction is as follows.  For a node such as
\begin{center}
\leaf{$B@k$} \faketreewidth{AA}
\leaf{$C@l$} \faketreewidth{AA}
\branch{2}{$A@j$}
\tree
\end{center}
we will generate the following eight
PCFG rules, where the number in parentheses following a rule is its
probability.
\begin{equation}
\begin{array}{llll}
A_j \rightarrow B C & (1/a_j) &
A \rightarrow B C & (1/a) \\
A_j \rightarrow B_k C & (b_k/a_j) &
A \rightarrow B_k C & (b_k/a) \\
A_j \rightarrow B C_l & (c_l/a_j) &
A \rightarrow B C_l & (c_l/a) \\
A_j \rightarrow B_k C_l & (b_k c_l/a_j) &
A \rightarrow B_k C_l & (b_k c_l/a) \\
\end{array}
\end{equation}  

\begin{mytheorem}
Subderivations headed by $A$ with external non-terminals at the roots
and leaves and internal non-terminals elsewhere have probability
$1/a$.  Subderivations headed by $A_j$ with external non-terminals
only at the leaves and internal non-terminals elsewhere, have probability
$1/a_j$.
\end{mytheorem}
\startproof The proof is by induction on the depth of the trees.  For
trees of depth 1, there are two cases:
\begin{center}
\leaf{$B$} \faketreewidth{AA}
\leaf{$C$} \faketreewidth{AA}
\branch{2}{$A$}
\tree
\leaf{$B$} \faketreewidth{AA}
\leaf{$C$} \faketreewidth{AA}
\branch{2}{$A@j$}
\tree
\end{center}
Trivially, these trees have the required probabilities.

Now, assume that the theorem is true for trees of depth $n$ or less.
We show that it holds for trees of depth $n+1$.  There are eight
cases, one for each of the eight rules.  We show two of them.  Let
$\begin{array}{c} \scriptstyle B@k\\ \scriptstyle \vdots\end{array}$ represent a
tree of at most depth $n$ with external leaves, headed by $B@k$, and
with internal intermediate non-terminals.  Then, for trees such as
\begin{center}
\leaf{$\begin{array}{c}B@k\\ \vdots\end{array}$} \faketreewidth{AA}
\leaf{$\begin{array}{c}C@l\\ \vdots\end{array}$} \faketreewidth{AA}
\branch{2}{$A@j$}
\tree
\end{center}
the probability of the tree is $\frac{1}{b_k}\frac{1}{c_l}\frac{b_k
c_l}{a_j}=\frac{1}{a_j}$.  Similarly, for another case, trees headed by
\begin{center}
\leaf{$B@k$} \faketreewidth{AA}
\leaf{$C$} \faketreewidth{AA}
\branch{2}{$A$}
\tree
\end{center}
the probability of the tree is
$\frac{1}{b_k}\frac{b_k}{a}=\frac{1}{a}$.  The other six cases follow
trivially with similar reasoning. $\Box$

\begin{figure}
\begin{center}
\begin{tabular}{c}
PCFG derivation \\
4 productions \\
\leaf{$PN$}
\leaf{$PN$}
\branch{2}{$NP@3$}
\leaf{$V$}
\leaf{$DET$}
\leaf{$N$}
\branch{2}{$NP$}
\branch{2}{$VP@1$}
\branch{2}{$S$}
$\tree$ \\

\medskip\\
STSG derivation \\
2 subtrees \\
\leaf{$PN$}
\leaf{$PN$}
\branch{2}{$NP$}
\leaf{$V$}
\leaf{$DET$}
\leaf{$N$}
\branch{2}{$\begin{array}{cc}NP\\ \vspace{0in} \\ NP\end{array}$}
\branch{2}{$VP$}
\branch{2}{$S$}
$\tree$ \\

\end{tabular}
\end{center}
\caption{Example of Isomorphic Derivation}
\label{fig:isomorphic}
\end{figure}

We call a PCFG derivation isomorphic to a STSG derivation if for every
substitution in the STSG there is a corresponding subderivation in the
PCFG.  Figure \ref{fig:isomorphic} contains an example of isomorphic
derivations, using two subtrees in the STSG and four productions in
the PCFG.

We call a PCFG tree isomorphic to a STSG tree if they are identical
when internal non-terminals are changed to external non-terminals.

\begin{mytheorem}
This construction produces PCFG trees isomorphic to the STSG trees
with equal probability.
\end{mytheorem}
\startproof If every subtree in the training corpus occurred exactly
once, the proof would be trivial.  For every STSG subderivation, there
would be an isomorphic PCFG subderivation, with equal probability.
Thus for every STSG derivation, there would be an isomorphic PCFG
derivation, with equal probability.  Thus every STSG tree would be
produced by the PCFG with equal probability.

However, it is extremely likely that some subtrees, especially
trivial ones like
\begin{center}
\leaf{$NP$} \faketreewidth{AA}
\leaf{$VP$} \faketreewidth{AA}
\branch{2}{$S$}
\tree
\end{center}
will occur repeatedly.    

If the STSG formalism were modified slightly, so that trees could
occur multiple times, then our relationship could be made one-to-one.
Consider a modified form of the DOP model, in which the counts of
subtrees which occurred multiple times in the training corpus were not
merged: both identical trees would be added to the grammar.  Each of
these trees will have a lower probability than if their counts were
merged.  This would change the probabilities of the derivations;
however the probabilities of parse trees would not change, since there
would be correspondingly more derivations for each tree.  Now, the
desired one-to-one relationship holds: for every derivation in the new
STSG there is an isomorphic derivation in the PCFG with equal
probability.  Thus, summing over all derivations of a tree in the STSG
yields the same probability as summing over all the isomorphic
derivations in the PCFG.  Thus, every STSG tree would be produced by
the PCFG with equal probability.

It follows trivially from this that no extra trees are produced by the
PCFG.  Since the total probability of the trees produced by the STSG
is 1, and the PCFG produces these trees with the same probability, no
probability is ``left over'' for any other trees. $\Box$

\section{Parsing Algorithms}
\label{sec:algorithm}
As we discussed in Chapter \ref{ch:max}, there are several different
evaluation metrics one could use for finding the best parse.  The
three most interesting are the most probable derivation (which can be
found using the Viterbi algorithm); the most probable parse, which can
be found by random sampling; and the
Labelled Recall parse, which can be found using the General Recall
algorithm of Section \ref{sec:generalalg}.

Bod \shortcite{Bod:93c,Bod:95a} shows that the most probable
derivation does not perform as well as the most probable parse for the
DOP model, getting 65\% exact match for the most probable derivation,
versus 96\% exact match for the most probable parse.  This performance
difference is not surprising, since each parse tree can be derived by
many different derivations; the most probable parse criterion takes
all possible derivations into account.  Similarly, the Labelled Recall
parse is also derived from the sum of many different derivations.
Furthermore, although the Labelled Recall parse should not do as well
on the exact match criterion, it should perform even better on the
Labelled Recall rate and related criteria such as the Crossing
Brackets rate.  In the preceding chapter, we performed a detailed
comparison between the most likely parse (the Labelled Tree parse) and
the Labelled Recall parse for PCFGs; we showed that the two have very
similar performance on a broad range of measures, with at most a 10\%
difference in error rate (i.e., a change from 10\% error rate to 9\%
error rate.)  We therefore think that it is reasonable to use the
General Recall algorithm (which for STSGs can compute the Labelled
Recall parse) to parse the DOP model, especially since our comparisons
will be on the Crossing Brackets rate, where we expect from both
theoretical and empirical considerations that an algorithm maximizing
the Labelled Recall rate will outperform one maximizing the exact
match (Labelled Tree) rate.  \newcite{Bod:95c} complains that if we
use the General Recall algorithm, we are not really parsing the DOP
model, since our parser does not return a most probable parse.  As we
will show in the results section, our parser performs at least as well
as a parser that does return the most probable parse, so this
objection is immaterial.

Although the General Recall algorithm was described in detail in the
previous chapter, we review it here.  First, for each potential
constituent, where a constituent is a non-terminal, a start position,
and an end position, the algorithm uses the inside-outside values to
find the probability that that constituent is in the correct parse.
After that, the algorithm uses dynamic programming to put the most
likely constituents together to form an output parse tree; the output
parse tree maximizes the expected Labelled Recall rate.

Recall from Section \ref{sec:labelledalgorithm} that the run time of
the General Recall algorithm is dominated by the time to compute the
inside and outside probabilities.  For a grammar with $r$ rules, this
is $O(rn^3)$.  Now, since there are at most eight rules for each node
in the training data, if the training data is of size $T$, the overall
run time is $O(Tn^3)$.

\subsection{Sampling Algorithms}

\begin{figure}
\begin{tabbing}
\verb|   |\=\verb|   |\=\verb|   |\=\verb|   |\=\verb|   |\=\verb|   |\=\verb|   |\=\verb|   |\=\verb|   |\=\verb|   |\= \kill
repeat until the standard error and the mpp error are smaller than a
threshold \\
 \> sample a random derivation from the derivation forest\\
 \> store the parse generated by the sampled derivation\\
 \> mpp := parse with maximal frequency\\
 \> calculate the standard error of the mpp and the mpp error
\end{tabbing}
\caption{Monte Carlo parsing algorithm}\label{fig:Monte}
\end{figure}

\begin{figure}
\begin{tabbing}
\verb|   |\=\verb|   |\=\verb|   |\=\verb|   |\=\verb|   |\=\verb|   |\=\verb|   |\=\verb|   |\=\verb|   |\=\verb|   |\= \kill
\alfor $\mi{length} := 1$ \alto $n$ \\
 \>  \alfor $\mi{start} := 1$ \alto $n-\mi{length} +1$ \\
 \>  \> \alforeach node $X \in \mi{chart}[\mi{start}, \mi{start}
+\mi{length}]$ \\
 \>  \>  \>  select at random a subderivation of $X$; \\
 \>  \>  \>  eliminate the other subderivations;
\end{tabbing}
\caption{Bod's $O(Gn^3)$ sampling algorithm}\label{fig:BodSample}
\end{figure}

\begin{figure}
\begin{tabbing}
\verb|   |\=\verb|   |\=\verb|   |\=\verb|   |\=\verb|   |\=\verb|   |\=\verb|   |\=\verb|   |\=\verb|   |\=\verb|   |\= \kill
\alFunction $\mi{fastsample}(i, X, j)$ \\
 \> \alif $i = j+1$ \\
 \> \> \alreturn $\mi{leaf}(X)$; \\
 \> \alelse \\
 \>  \> select at random a subderivation of $X$: $i, Y, k$ and $k, Z, j$; \\
 \>  \> \alreturn $\mi{tree}(X, \mi{fastsample}(i, Y, k),\mi{fastsample}(k, Z, j))$;
\end{tabbing}
\caption{Faster $O(Gn^2)$ sampling algorithm}\label{fig:MySample}
\end{figure}

\newcite[p. 56]{Bod:95a} gives the simple algorithm of Figure
\ref{fig:Monte} for finding the most probable parse (mpp) (i.e. the
parse that maximizes the expected Exact Match rate).  Essentially, the
algorithm is to randomly sample parses from the derivation forest, and
pick the maximal frequency parse.  In practice, rather than compute
standard errors, Bod simply ran the outer loop 100 times.  The
algorithm Bod used for sampling a random derivation is given in Figure
\ref{fig:BodSample}; Bod analyzes the run time of the algorithm for
computing one sample as $O(Gn^3)$, although by using tables, it can be
efficiently approximated in time $O(Gn^2)$ (Bod, personal
communication).
% Should be $Vn^2$, not Gn^2, but V here is the number of
% nonnterminals + the number of internal nodes of the trees, which for
% almost any STSG (except PCFGs) is essentially G.

We used a different, but mathematically equivalent sampling algorithm.
Rather than a bottom-up algorithm, we used a top-down algorithm, as
shown in Figure \ref{fig:MySample}.  The run time for our sampling
algorithm, if naively implemented (as we did), is at worst $O(Gn^2)$,
although using the same trick that Bod used, it can be implemented in
time $O(n)$.

\section{Experimental Results and Discussion}
\label{sec:results}
\begin{table*}
\begin{center}
\begin {tabular}{|l||r|r|r|r|r|} \hline
       Criteria & Min    & Max    & Range  & Mean  & StdDev  \\ \hline
Cross Brack DOP &86.53\% &96.06\% & 9.53\% &90.15\% & 2.65\%  \\ \hline
Cross Brack P\&S &86.99\% &94.41\% & 7.42\% &90.18\% & 2.59\%  \\ \hline
Cross Brack DOP$-$P\&S &-3.79\% & 2.87\% & 6.66\% &-0.03\% & 2.34\%  \\ \hline
Zero Cross Brack DOP &60.23\% &75.86\% &15.63\% &66.11\% & 5.56\%  \\ \hline
Zero Cross Brack P\&S &54.02\% &78.16\% &24.14\% &63.94\% & 7.34\%  \\ \hline
Zero Cross Brack DOP$-$P\&S &-5.68\% &11.36\% &17.05\% & 2.17\% & 5.57\%  \\ \hline
\end{tabular}
\end{center}
\caption{DOP Labelled Recall versus Pereira and Schabes on Minimally Edited ATIS}
\label{tab:results}
\end{table*}

\begin{table*}
\begin{center}
\begin {tabular}{|l||r|r|r|r|r|} \hline
       Criteria & Min    & Max    & Range  & Mean  & StdDev  \\ \hline 
Cross Brack DOP &95.63\% &98.62\% & 2.99\% &97.16\% & 0.93\%  \\ \hline 
Cross Brack P\&S &94.08\% &97.87\% & 3.79\% &96.11\% & 1.14\%  \\ \hline 
Cross Brack DOP$-$P\&S &-0.16\% & 3.03\% & 3.19\% & 1.05\% & 1.04\%  \\ \hline 
Zero Cross Brack DOP &78.67\% &90.67\% &12.00\% &86.13\% & 3.99\%  \\ \hline 
Zero Cross Brack P\&S &70.67\% &88.00\% &17.33\% &79.20\% & 5.97\%  \\ \hline 
Zero Cross Brack DOP$-$P\&S &-1.33\% &20.00\% &21.33\% & 6.93\% & 5.65\%  \\ \hline 
Exact Match DOP &58.67\% &68.00\% & 9.33\% &63.33\% & 3.22\%  \\ \hline 
\end{tabular}
\end{center}
\caption{DOP Labelled Recall versus Pereira and Schabes on Bod's Data}
\label{tab:Bodresults}
\end{table*}

\begin{table}
%grid.perl pns-out44 parser-out100-summary
\begin{center}
\begin{tabular}{|l||c|c||c|c|} \hline
& Labelled Recall & Most Probable & Pereira and & Significant \\
& Parse & Parse & Schabes & \\ \hline
Cross Brack &90.1&90.0&90.2 &\\ \hline
0 Cross Brack &66.1&65.9&63.9 &\\ \hline
Exact Match &40.0&39.2&&\\ \hline
\end{tabular} 
\end{center}
\caption{Three way comparison on minimally edited ATIS data}
\label{fig:threewaymin}
\end{table}

\begin{table}
%getparsercums.perl  pns-out44 parser-out100-summary

%Exact Match    MaxCons-Monte &-1.15\% & 4.55\% & 5.69\% & 0.80\% & 1.52\% \\\hline
%This is on the edge of statistical significance, between .1 and .05,
%one sided, but not quite significant.
%
%grid.perl pns-out47 parser-out98-summary
\begin{center}
\begin{tabular}{|l||c|c||c|c|} \hline
& Labelled Recall & Most Probable & Pereira and & Significant \\
& Parse & Parse & Schabes & \\ \hline
Cross Brack &97.2&97.1&96.1&$\surd $ \\ \hline
0 Cross Brack &86.1&86.1&79.2& $\surd$\\ \hline
Exact Match &63.3&63.1&& \\ \hline
\end{tabular}
\end{center}
\caption{Three way comparison on ATIS data edited by Bod}
\label{fig:threewaybod}
\end{table}

We are grateful to Bod for supplying us with data edited for his
experiments \cite{Bod:95b,Bod:95a,Bod:93a}, although it appears not to
have been exactly the data he used.  We have been unable to obtain the
exact same data he used, and since we cannot get it, we use the data
Bod gave us.\footnote{The data Bod gave us contained no epsilon
productions (traces), while Bod's \shortcite{Bod:95c} data apparently
did contain epsilons as explicit part-of-speech tags in the data.  We
note that this is an unconventional way to handle epsilon productions,
since real data would typically not contain traces annotated in this
way, although it might be possible to train a tagger to produce them.
We have asked Bod for the correct data, but we have never received it.
We note that soon after pointing out to us that the data he had given
us was incorrect (since it did not contain epsilons) Bod mailed
another researcher, John Maxwell, data without epsilons.  Strangely,
the data Maxwell received is different from the data we received.  In
particular, the data Maxwell received is a subset of the data we
received, with some lines repeated to reach the same total number of
lines.}

The original ATIS data from the Penn Treebank, version 0.5, is very
noisy; it is difficult to even automatically read this data, due to
inconsistencies between files.  Researchers are thus left with the
difficult decision as to how to clean up the data.  For this chapter, we
conducted two sets of experiments: one using a minimally cleaned up
set of data, the same as described in Section \ref{sec:results1},
making our results comparable to previous results; the other using the
ATIS data prepared by Bod, which contained much more significant
revisions.

Ten data sets were constructed by randomly splitting minimally edited
ATIS sentences into a 700 sentence training set, and an 88 sentence
test set, then discarding sentences of length $> 30$.  For each of the
ten sets, the Labelled Recall parse, the sampling algorithm given in
Figure \ref{fig:MySample} (equivalent to but faster than Bod's), and
the grammar induction experiment of \newcite{Pereira:92a} were run.
All sentences output by the parser were made binary branching using
the Continued transformation, as described in Section
\ref{sec:analysis}, since otherwise the crossing brackets measures are
meaningless \cite{Magerman:94a}.  A few sentences were not parsable;
these were assigned right branching period high structure, a good
heuristic \cite{Brill:93a}.  Crossing brackets, zero crossing
brackets, and the paired differences between Labelled Recall and
Pereira and Schabes are presented in Table \ref{tab:results}.  The
results are disappointing.  In absolute value, the results are
significantly below the 96\% exact match reported by Bod.  In relative
value, they are also disappointing: while the DOP results are slightly
higher on average than the Pereira and Schabes results, the
differences are small.

We also ran experiments using Bod's data, 75 sentence test sets, and
no limit on sentence length.  However, while Bod provided us with
data, he did not provide us with a split into test and training data;
as before, we used ten random splits.  The DOP results, while better,
are still disappointing, as shown in Table \ref{tab:Bodresults}.  They
continue to be noticeably worse than those reported by Bod, and again
comparable to the Pereira and Schabes algorithm.  Even on Bod's data,
the 86\% DOP achieves on the zero crossing brackets criterion is not
close to the 96\% Bod reported on the much harder exact match
criterion.  It is not clear what exactly accounts for these
differences.
It is also noteworthy that the results are much better on Bod's data
than on the minimally edited data: crossing brackets rates of 96\% and
97\% on Bod's data versus 90\% on minimally edited data.  Thus it
appears that part of Bod's extraordinary performance can be explained
by the fact that his data is much cleaner than the data used by other
researchers.

DOP does do slightly better on most measures.  We performed a
statistical analysis using a $t$-test on the paired differences
between DOP and Pereira and Schabes performance on each run.  On the
minimally edited ATIS data, the differences were statistically
insignificant, while on Bod's data the differences were statistically
significant beyond the 98'th percentile.  Our technique for finding
statistical significance is more strenuous than most: we assume that
since all test sentences were parsed with the same training data, all
results of a single run are correlated.  Thus we compare paired
differences of entire runs, rather than of sentences or constituents.
This makes it harder to achieve statistical significance.

Notice also the minimum and maximum columns of the ``DOP$-$P\&S'' lines,
constructed by finding for each of the paired runs the difference
between the DOP and the Pereira and Schabes algorithms.  Notice that
the minimum is usually negative, and the maximum is usually positive,
meaning that on some tests DOP did worse than Pereira and Schabes and
on some it did better.  It is important to run multiple tests,
especially with small test sets like these, in order to avoid
misleading results.

Tables \ref{fig:threewaymin} and \ref{fig:threewaybod} show a three-way
comparison between all the algorithms; the sampling algorithm and the
Labelled Recall algorithm perform almost identically.  In the next
section, we will show that the sampling algorithm's performance
probably does not scale well to longer sentences.

\section{Timing Analysis}
\label{sec:timing}

In this section, we examine the empirical runtime of the General
Recall algorithm, and analyze the runtime of Bod's Monte Carlo algorithm.
%  We note that modifications to Bod's algorithm
%could significantly speed it up.
We also note that Bod's algorithm
will probably be particularly inefficient on longer sentences.

It takes about 6 seconds per sentence to run our algorithm on an HP
9000/715, versus 3.5 hours  to run Bod's algorithm on a Sparc 2
\cite{Bod:95b}.  Factoring in that the HP is roughly four
times faster than the Sparc,
% analysis from ftp.nosc.mil/pub/aburto/flops_4.tbl, about eight flops
% for sun 2, 30 for HP.
% same file gives Khalil's Sparc 10 between 9 and 30 flops, depending
% on configuration.  He requires 14 to 19 seconds per sentence,
% so our algorithm is at least as fast as his probably.
the new algorithm is about 500 times faster.  Of course, some of this
difference may be due to differences in implementation, so this
estimate is approximate.

Furthermore, we believe Bod's analysis of his parsing algorithm is
flawed.  Letting $G$ represent grammar size, and $\epsilon$ represent
maximum estimation error, Bod correctly analyzes his runtime as
$O(Gn^3\epsilon^{-2})$.  However, Bod then neglects analysis of this
$\epsilon^{-2}$ term, assuming that it is constant.  Thus he concludes
that his algorithm runs in polynomial time.  However, for his
algorithm to have some reasonable chance of finding the most probable
parse, the number of times he must sample his data is, as a
conservative estimate, inversely proportional to the conditional
probability of that parse.  For instance, if the maximum probability
parse had probability $1/50$, then he would need to sample at least
$50$ times to be reasonably sure of finding that parse.

%Now, we note that the probability of the most probable parse tree will
%in general decline exponentially with sentence length.  Consider a
%sentence with 10 words, ``John and Mary want to visit the small toy
%factory.''  This sentence has one significant ambiguity: small toys or
%small factory. This causes the probability of the most probable parse
%to be around $1/2$.  Now consider a sentence with 20 words, and say it
%has two significant ambiguities.  Then the probability of the most
%probable parse will be around $1/4$.  Similarly, with 30 words, the
%probability will be around $1/8$.  Thus we see that the probability of
%the most probable parse will decline exponentially with sentence
%length.

Now, we note that the conditional probability of the most probable
parse tree will in general decline exponentially with sentence length.
We assume that the number of ambiguities in a sentence will increase
linearly with sentence length; if a five word sentence has on average
one ambiguity, then a ten word sentence will have two, etc.  A linear
increase in ambiguity will lead to an exponential decrease in
probability of the most probable parse.

Since the probability of the most probable parse decreases
exponentially in sentence length, the number of random samples needed
to find this most probable parse increases exponentially in sentence
length.  Thus, when using the Monte Carlo algorithm, one is left with
the uncomfortable choice of exponentially decreasing the probability
of finding the most probable parse, or exponentially increasing the
runtime.

We admit that this argument is somewhat informal.  Still, the Monte
Carlo algorithm has never been tested on sentences longer than those
in the ATIS corpus; there is good reason to believe the algorithm will
not work as well on longer sentences.  We note that our algorithm has
true runtime $O(Tn^3)$, as shown previously.

\section{Analysis of Bod's Data}
\label{sec:analysis}

In the DOP model, a sentence cannot be given an exactly correct parse
unless all productions in the correct parse occur in the training set.
Thus, we can get an upper bound on performance by examining the test
corpus and finding which parse trees could not be generated using only
productions in the training corpus.  As mentioned in
Section \ref{sec:results}, the data Bod provided us with may not have
been the data he used for his experiments; furthermore, the data was
not divided into test and training.  Nevertheless, we analyze this
data to find an upper bound on average case performance.

In our paper on DOP \cite{Goodman:96b}, we performed an analysis of
Bod's data based on the following lines from his thesis
\cite[p. 64]{Bod:95b}:
\begin{quote}
It may be relevant to mention that the parse {\em coverage} was
99\%.  This means that for 99\% of the test strings the perceived [test
corpus] parse  was in the derivation forest generated by the system.
\end{quote} 
Using this 99\% figure, we were able to achieve a strong bound on the
likelihood of achieving Bod's results.  However, Bod later informed us
(personal communication) that
\begin{quote}
The 99\% coverage refers to the percentage of sentences for which a
parse was found. I did not check whether the ``appriopriate'' [{\it sic}] parse was
among the found parses. I just assumed that this would have been the
case, but probably it wasn't.
\end{quote}
We can still use the fact that Bod got a 96\% exact match
rate to aid us, although this leads to a weaker upper bound than that
in the original paper.

\begin{table*}
\begin{center}
\begin{tabular}{|c|c|c|c|}
\hline
Original & Correct & Continued & Simple \\  \hline
\leaf{$B$}
\leaf{$C$}
\leaf{$D$}
\leaf{$E$}
\branch{4}{$A$}
\hspace{-.3in}\tree
&
\leaf{$B$}
\leaf{$C$}
\leaf{$D$}
\leaf{$E$}
\branch{2}{$*\_DE$}
\branch{2}{$*\_CDE$}
\branch{2}{$A$}
\hspace{-.3in}\tree
&
\leaf{$B$}
\leaf{$C$}
\leaf{$D$}
\leaf{$E$}
\branch{2}{$A\_*$}
\branch{2}{$A\_*$}
\branch{2}{$A$}
\hspace{-.3in}\tree
&
\leaf{$B$}
\leaf{$C$}
\leaf{$D$}
\leaf{$E$}
\branch{2}{$A$}
\branch{2}{$A$}
\branch{2}{$A$}
\hspace{-.3in}\tree
\\ \hline
\end{tabular}
\end{center}
\caption{Transformations from $N$-ary to Binary Branching Structures} 
\label{tab:convert}
\end{table*}

Bod randomly split his corpus into test and training.  From the 96\%
exact match rate of his parser, we conclude that only three of his 75
test sentences had a correct parse that could not be generated from
the training data.  This small number turns out to be very surprising.
An analysis of Bod's data shows that at least some of the difference
in performance between his results and ours must be due to a
fortuitous choice of test data, or to the data he used being even
easier than the data he sent us (which was significantly easier than
the original ATIS data).  Bod did examine versions of DOP that
smoothed, allowing productions that did not occur in the training set;
however his exact match rate and his reference to coverage are both
with respect to a version that does no smoothing.

\begin{table*}
\begin{center}
\begin{tabular}{|l||lr|lr|lr|}
\hline 

% old values -- exactly 3 ungeneratable
% & Correct & Continued & Simple
%no unary & 0.78 & 0.0000167  &  0.88 & 0.0147029  &  0.90 & 0.0417259\\  \hline
%unary    & 0.80 & 0.0000629  &  0.90 & 0.0391041  &  0.92 & 0.0950151\\  \hline
% New values: 3 or fewer ungeneratable
 & \multicolumn{2}{c|}{Correct} & \multicolumn{2}{c|}{Continued}&\multicolumn{2}{c|}{Simple}\\
\hline
no unary & 0.78 & 0.0000195  &  0.88 & 0.0202670  &  0.90 & 0.0620323\\  \hline
unary    & 0.80 & 0.0000744  &  0.90 & 0.0577880  &  0.92 & 0.1568100\\  \hline

\end{tabular}
\end{center}
\caption{Probabilities of test data
with ungeneratable sentences}
\label{tab:under}
\end{table*}

In order to perform our analysis, we must determine certain details of
Bod's parser that affect the probability of having most sentences
correctly parsable.  When using a chart parser, as Bod did, three
problematic cases must be handled: $\epsilon$ productions, unary
productions, and $n$-ary ($n > 2$) productions.  The first two kinds
of productions can be handled with a probabilistic chart parser, but
large and difficult matrix manipulations are required
\cite{Stolcke:93a}; these manipulations would be especially difficult
given the size of Bod's grammar.  In the data Bod gave us there were
no epsilon productions; in other data, Bod (personal communication)
treated epsilons the same as other part of speech tags, a strange
strategy.  We also assume that Bod made the same choice we did and
eliminated unary productions, given the difficulty of correctly
parsing them.  Bod himself does not know which technique he used for
$n$-ary productions, since the chart parser he used was written by a
third party (Bod, personal communication).
%
% Bod's data example 9 contains a unary branching example
%

The $n$-ary productions can be parsed in a straightforward manner, by
converting them to binary branching form; however, there are at least
three different ways to convert them, as illustrated in Table
\ref{tab:convert}.  In method ``Correct'', the $n$-ary branching
productions are converted in such a way that no overgeneration is
introduced.  A set of special non-terminals is added, one for each
partial right hand side.  In method ``Continued'', a single new
non-terminal is introduced for each original non-terminal.  Because
these non-terminals occur in multiple contexts, some overgeneration is
introduced.  However, this overgeneration is constrained, so that
elements that tend to occur only at the beginning, middle, or end of
the right hand side of a production cannot occur somewhere else.  If
the ``Simple'' method is used, then no new non-terminals are
introduced; using this method, it is not possible to recover the
$n$-ary branching structure from the resulting parse tree, and
significant overgeneration occurs.

Table \ref{tab:under} shows the undergeneration probabilities for each
of these possible techniques for handling unary productions and
$n$-ary productions.\footnote{A perl script for analyzing Bod's data
is available by anon\-y\-mous FTP from
ftp://ftp.das.harvard.edu/pub/goodman/analyze.perl} The first column
gives the probability that a sentence contains a production found only
in that sentence, and the second column contains the probability that
a random set of 75 test sentences would contain at most three such
sentences:\footnote{Actually, this is a slight overestimate for a few reasons,
including the fact that the 75 sentences are drawn without
replacement.  Also, consider a sentence with a production that occurs
only in one other sentence in the corpus; there is some probability
that both sentences will end up in the test data, causing both to be
ungeneratable.}
$$
\frac{p^{72} \times (1-p)^3 \times 75!}{72!3!} +
\frac{p^{73} \times (1-p)^2 \times 75!}{73!2!} +
\frac{p^{74} \times (1-p)^1 \times 75!}{74!1!} +
\frac{p^{75} \times (1-p)^0 \times 75!}{75!0!}
$$

The table is arranged from least generous to most generous: in the
upper left hand corner is a technique Bod might reasonably have used;
in that case, the probability of getting the test set he described is
less than one in 50,000.  In the lower right corner we give Bod the
absolute maximum benefit of the doubt: we assume he used a parser
capable of parsing unary branching productions, that he used a very
overgenerating grammar, and that he used a loose definition of ``Exact
Match.''  Even in this case, there is only about a 15\% chance of
getting the test set Bod describes.

\section{Conclusion}

We have given efficient techniques for parsing the DOP model.  These
results are significant since the DOP model has perhaps the best
reported parsing accuracy; previously the full DOP model had not been
replicated due to the difficulty and computational complexity of the
existing algorithms.  We have also shown that previous results were
partially due to heavy cleaning of the data, which reduced the
difficulty of the task, and partially due to an unlikely choice of
test data -- or to data even easier than that which Bod gave us.

Of course, this research raises as many questions as it answers.  Were
previous results due only to the choice of test data, or are
differences in implementation partly responsible?  In that case, there
is significant future work required to understand which differences
account for Bod's exceptional performance.  This will be complicated
by the fact that sufficient details of Bod's implementation are not
available.  However, based on the fact that further extraordinary DOP
results have not been reported since the conference version of this
work was published, it appears that problems in our implementation are
not the source of the discrepancy.

This research also shows the importance of testing on more than one
small test set, as well as the importance of not making cross-corpus
comparisons; if a new corpus is required, then previous algorithms
should be duplicated for comparison.

The speedups we achieved were critical to the success of this chapter.
Running even one experiment without the 500 times speedup we achieved
would have been difficult, never mind running the ten experiments that
allowed us to compute accurate average performance and to compute
statistical significance.  These speedups were made possible by the
combination of our efficient equivalent grammar and our use of the
General Recall algorithm, which depends on the inside-outside
probabilities.

% -*- Mode: latex; -*-

\chapter{Thresholding} \label{ch:thresh}

In this chapter, we show how to efficiently threshold Probabilistic
Context-Free Grammars and Probabilistic Feature Grammars, using three
new thresholding algorithms: beam thresholding with the prior; global
thresholding; and multiple pass thresholding \cite{Goodman:97b}.  Each
of these algorithms approximates the inside-outside probabilities.  We
also give an algorithm that uses the inside probabilities to
efficiently optimize the settings of all of the parameters
simultaneously.

\section{Introduction}

In this chapter, we examine thresholding techniques for statistical
parsers.  While there exist theoretically efficient ($O(n^3G)$)
algorithms for parsing Probabilistic Context-Free Grammars (PCFGs),
$n^3G$ can still be fairly large in practice.  Sentence lengths of 30
words ($n^3 = 27,000$) are common, and large grammars are increasingly
frequent.  For instance, \newcite{Charniak:96a} used a 10,000 rule
grammar built by simply reading rules off a tree bank.  The grammar
size of more recent, lexicalized grammars, such as Probabilistic
Feature Grammars (PFGs), described in the next chapter, is effectively
much larger.  The product of $n^3$ and $G$ can quickly become very
large; thus, practical parsing algorithms usually make use of pruning
techniques, such as beam thresholding, for increased speed.

We introduce two novel thresholding techniques, global thresholding
and multiple-pass parsing, and one significant variation on
traditional beam thresholding; each of these three techniques uses a
different approximation to the inside-outside probabilities to improve
thresholding.  We examine the value of these techniques when used
separately, and when combined.  In order to examine the combined
techniques, we also introduce an algorithm for optimizing the settings
of multiple thresholds, using the inside probabilities.  When all
three thresholding methods are used together, they yield very
significant speedups over traditional beam thresholding alone, while
achieving the same level of performance.

We apply our techniques to CKY chart parsing, one of the most commonly
used parsing methods in natural language processing, as described in
Section \ref{sec:CFG}.  Recall that in a CKY chart parser, a
two-dimensional matrix of cells, the chart, is filled in.  Each cell
in the chart corresponds to a span of the sentence, and each cell of
the chart contains the nonterminals that could generate that span.
The parser fills in a cell in the chart by examining the nonterminals
in lower, shorter cells, and combining these nonterminals according to
the rules of the grammar.  The more nonterminals there are in the
shorter cells, the more combinations of nonterminals the parser must
consider.

\begin{figure}
\begin{center}
\begin{tabular}{c}
\psfig{figure=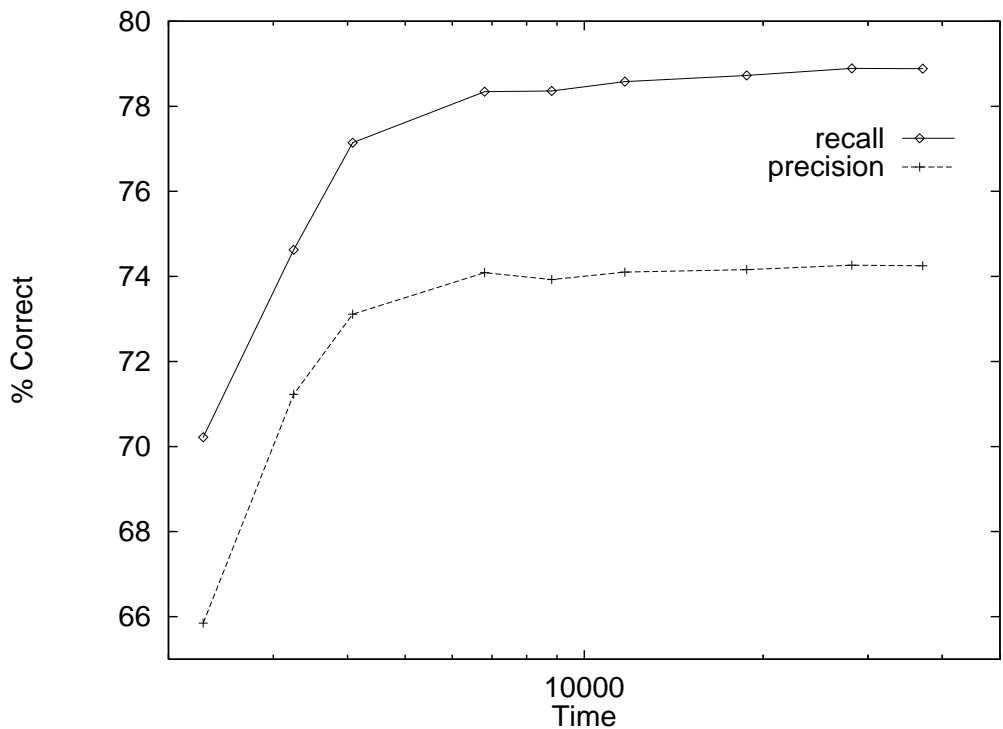,width=5.5in} 
\end{tabular}
\end{center}
\caption{Precision and Recall versus Time in Beam Thresholding}
%from ~/attach/dop/emnlp97/smoothness-1000.plot
%which uses general-out182-dt
\label{fig:tradeoff}
\end{figure}

In some grammars, such as PCFGs and PFGs, probabilities are associated
with the grammar rules.  This introduces problems, since in many
grammars, almost any combination of nonterminals is possible, perhaps
with some low probability.  The large number of possibilities can
greatly slow parsing.  On the other hand, the probabilities also
introduce new opportunities.  For instance, if in a particular cell in
the chart there is some nonterminal that generates the span with high
probability, and another that generates that span with low
probability, then we can remove the less likely nonterminal from the
cell.  The less likely nonterminal will probably not be part of either
the correct parse or the tree returned by the parser, so removing it
will do little harm.  This technique is called {\em beam
thresholding}.

If we use a loose beam threshold, removing only those nonterminals
that are much less probable than the best nonterminal in a cell, our
parser will run only slightly faster than with no thresholding, while
performance measures such as precision and recall will remain
virtually unchanged.  On the other hand, if we use a tight threshold,
removing nonterminals that are almost as probable as the
best nonterminal in a cell, then we can get a considerable speedup,
but at a considerable cost.  Figure \ref{fig:tradeoff} shows the
tradeoff between accuracy and time, using a beam threshold that ranged
from .2 to .0002.

When we beam threshold, we remove less likely nonterminals from the
chart.  There are many ways to measure the likelihood of a
nonterminal.  The ideal measure would be the normalized inside-outside
probability, which would give the probability that the nonterminal was
correct, given the whole sentence.  However, we cannot compute
the outside probability of a nonterminal until we are finished
computing all of the inside probabilities, so this technique cannot be
used in practice.

We can, though, approximate the inside-outside probability, in
several different ways.  In this chapter, we will consider three
different kinds of thresholding, using three different approximations
to the inside-outside probability.  In traditional beam search, only
the inside probability is used, the probability of the nonterminal
generating the terminals of the cell's span.  We have found that a
minor variation, introduced in Section \ref{sec:beamthresh}, in which
we also consider the average outside probability of the nonterminal
(which is proportional to its prior probability of being part of the correct parse)
can lead to nearly an order of magnitude improvement.

The problem with beam search is that it only compares nonterminals to
other nonterminals in the same cell.  Consider the case in which a
particular cell contains only bad nonterminals, all of roughly equal
probability.  We cannot threshold out these nodes, because even though
they are all bad, none is much worse than the best.  Thus, what we
want is a thresholding technique that uses some global information for
thresholding, rather than just using information in a single cell.
The second kind of thresholding we consider is a novel technique, {\em
global thresholding}, described in Section \ref{sec:globalthresh}.
Global thresholding uses an approximation to the outside probability
that uses all nonterminals not covered by the constituent, allowing
inside-outside probabilities of nonterminals covering different spans
to be compared.

The last technique we consider, {\em multiple-pass parsing}, is
introduced in Section \ref{sec:multithresh}.  The basic idea is that
we can use inside-outside probabilities from parsing with one grammar
as approximations to inside-outside probabilities in another.  We run
two passes with two different grammars.  The first grammar is fast and
simple.  We compute the inside-outside probabilities using this first
pass grammar, and use these probabilities to avoid considering
unlikely constituents in the second pass grammar.  The second pass is
more complicated and slower, but also more accurate.  Because we have
already eliminated many low probability nodes using the inside-outside
probabilities from the first pass, the second pass can run much
faster, and, despite the fact that we have to run two passes, the
added savings in the second pass can easily outweigh the cost of the
first one.

Experimental comparisons of these techniques show that they lead to
considerable speedups over traditional thresholding, when used
separately.  We also wished to combine the thresholding techniques;
this is relatively difficult, since searching for the optimal
thresholding parameters in a multi-dimensional space is potentially
very time consuming.  Attempting to optimize performance measures such
as precision and recall using gradient descent is not feasible,
because these measures are too noisy.  However, we found that the
inside probability was monotonic enough to optimize, and designed a
variant on a gradient descent search algorithm to find the optimal
parameters.  Using all three thresholding methods together, and the
parameter search algorithm, we achieved our best results, running an
estimated 30 times faster than traditional beam search, at the same
performance level.

\section{Beam Thresholding}

\label{sec:beamthresh}

The first, and simplest, technique we will examine is beam
thresholding.  While this technique is used as part of many search
algorithms, beam thresholding with PCFGs is most similar to beam
thresholding as used in speech recognition.  Beam thresholding is often
used in statistical parsers, such as that of \newcite{Collins:96a}.

Consider a nonterminal $X$ in a cell covering the span of terminals
$w_j...w_{k-1}$.  We will refer to this as {\em node} $\constituent{j}{X}{k}$,
since it corresponds to a potential node in the final parse tree.  In
beam thresholding, we compare nodes $\constituent{j}{X}{k}$ and $\constituent{j}{Y}{k}$
covering the same span.  If one node is much more likely than the
other, then it is unlikely that the less probable node will be part of
the correct parse, and we can remove it from the chart, saving time
later.

\begin{figure}
\begin{center}
\begin{tabbing}
\verb|    |\=\verb|    |\=\verb|    |\=\verb|    |\=\verb|    |\=\verb|    |\=\verb|    |\=\verb|    |\=\verb|    |\=\verb|    |\= \kill
\alforeach start $s$ \\
 \> \alforeach rule $A \rightarrow w_s$ \\
 \>  \> $\mi{chart}[s, A, s+1] := P(A \rightarrow w_s); $ \\
\alforeach length $l$, shortest to longest \\
 \> \alforeach start $s$ \\
 \>  \> \alforeach split length $t$ \\
 \>  \>  \> \alforeach $B$ s.t. $\mi{chart}[s, B, s+t] > 0$ \\
 \>  \>  \>  \> \alforeach $C$ s.t. $\mi{chart}[s+t, C, s+l] > 0$ \\
 \>  \>  \>  \>  \> \alforeach rule $A \rightarrow BC\;\in R$ \\
 \>  \>  \>  \>  \>  \> $\mi{chart}[s, A, s+l] := \mi{chart}[s,A, s+l] +$ \\
 \>  \>  \>  \>  \>  \>  \> $P(A \rightarrow BC) \times \mi{chart}[s, B, s+t] \times \mi{chart}[s+t, C, s+l]) $; \\
 \>  \>  $\mi{best} := \max_A \mi{chart}[s, A, s+l];$\\
 \>  \>  \alforeach $A$ \\
 \>  \>  \> \alif $\mi{chart}[s, A, s+l] < T_B \times \mi{best}$\\
 \>  \>  \>  \> $\mi{chart}[s, A, s+l] := 0;$ \\
\alreturn $\mi{chart}[1, S, n+1]$
\end{tabbing}
\end{center}
\caption{Inside Parser with Beam Thresholding} \label{fig:beamparser}
\end{figure}

There is some ambiguity about what it means for a node
$\constituent{j}{X}{k}$ to be more likely than some other node.
According to folk wisdom, the best way to measure the likelihood of a
node $\constituent{j}{X}{k}$ is to use the inside probability,
$\mi{inside}(j, X, k) =P(X \derives w_j...w_{k-1})$.  Figure
\ref{fig:beamparser} shows a PCFG parser for computing inside
probabilities that uses this traditional beam thresholding.  This
parser is just a conventional PCFG parser with two changes.  The most
important change is that after parsing a given span, we find the most
probable nonterminal in that span.  Then, given a thresholding factor
$T_B\leq 1$, we find all nonterminals less probable than the best by a
factor of $T_B$, and set their probabilities to 0.  The other change
from the way we have specified PCFG parsers elsewhere is that we loop
over child symbols $B, C$ with non-zero probabilities, and then find
rules consistent with these children.  Elsewhere, we looped over all
rules; but doing that here would mean that beam thresholding would not
lead to any reduction in the number of rules examined. %
%\footnote{A possibly more efficient
%implementation would modify the loop over right child symbols $C$ to
%be constrained to symbols such that for some $A$, $A \rightarrow BC$,
%but this is not the implementation we used.}

Recall that the outside probability of a node $\constituent{j}{X}{k}$
is the probability of that node given the surrounding terminals of the
sentence, i.e. $\mi{outside}(j, X, k) = P(S \derives w_1...w_{j-1} X
w_{k} ... w_n)$.  Ideally, we would multiply the inside probability by
the outside probability, and normalize, computing
$$
\frac{\mi{inside}(j, X, k) \times \mi{outside}(j, X, k)}{\mi{inside}(1, S, n\!+\!1)} =
P(S\derives w_1... w_{j-1} Xw_k... w_n \derives w_1... w_n |S \derives w_1... w_n)
$$
This expression would give us the overall probability that the node is
part of the correct parse, which would be ideal for thresholding.
However, there is no good way to quickly compute the outside
probability of a node during bottom-up chart parsing (although it can
be efficiently computed afterwards).  One simple approximation to the
outside probability of a node $\constituent{j}{X}{k}$ is just the
average outside probability of the nonterminal $X$ across the
language:
\begin{eqnarray*}
\sum_{j, k \geq j, n \geq k, w_1...w_n} \mi{outside}(i, X, j)
\times P(S\derives w_1... w_n) = \\
\sum_{j, k \geq j, n \geq k, w_1...w_n} P(S \derives w_1...w_{j-1}Xw_{k}...w_n | S
\derives w_1...w_n) \times P(S \derives w_1...w_n)
\end{eqnarray*}
We will show that the average outside probability of a nonterminal $X$
is proportional to the prior probability of $X$, where by prior
probability we mean the probability that a random nonterminal of a
random parse tree will be $X$.  Formally, letting $C(D, X)$ denote the
number of occurrences of nonterminal $X$ in a derivation $D$, we can
write the prior probability as
\begin{equation}
\label{eqn:prior}
P(X) =
\frac{\sum_{\mbox{$\scriptstyle D$ \small a derivation}} P(D) \times C(D,X)}
{\sum_Y \sum_{\mbox{$\scriptstyle D$ \small a derivation}} P(D) \times C(D,Y)}
\end{equation}
Now,
\begin{eqnarray*}
\sum_{j, k \geq j, n \geq k, w_1...w_n} P(S \derives w_1...w_{j-1}Xw_{k}...w_n | S
\derives w_1...w_n) \times P(S \derives w_1...w_n) = \\
\sum_{j, k \geq j, n \geq k, w_1...w_n} P(S \derives
w_1...w_{j-1}Xw_{k}...w_n) \times P(X \derives w_j...w_{j-1}) = \\
\sum_{\mbox{$\scriptstyle D$ \small a derivation}} P(D) \times C(D,X)
\end{eqnarray*}
Thus, the average outside probability of a nonterminal $X$ is the same
as the prior probability of $X$, except for a factor equal to the
normalization term of Expression \ref{eqn:prior}:
$$
\frac{1} {\sum_Y \sum_{\mbox{$\scriptstyle D$ \small a derivation}}
P(D) \times C(D,Y)}
$$
Since it is easier to compute the prior probability than the average
outside probability, and since all of our values will have the same
normalization factor, we use the prior probability rather than the
average outside probability.

\begin{figure}
$$
\begin{array}{cl}
\mbox{\bf Item form:} \\
{[i, A, j]} & \mbox{inside or Viterbi} \\
{[i, j]} & \mbox{inside or Viterbi} \\ \\
\mbox{\bf Goal:} \\
{[1, S, n+1]} \\\\
\mbox{\bf Rules:} \\
\infer{R(A \rightarrow w_i)}{[i, A, i+1] }{} & \mbox{Unary} \\
\infer{R(A \rightarrow BC)\rspace [i, B, k]\rspace [k, C, j]}{[i, A,
j]}
{\begin{array}{l}
R(B)\times[i, B, k] \geq T_B \times [i, k] \wedge \\
R(C)\times[k, C, j] \geq T_B \times [k, j]
\end{array}
}
& \mbox{Binary} \\
\infer{R(A)\rspace[i, A, j]}{[i, j]}{} & \mbox{Thresholding}
\end{array}
$$
\caption{Beam thresholding item-based description}\label{fig:beamCKY}
\end{figure}

Our final thresholding measure then is $P(X) \times \mi{inside}(j, X, k)$;
the algorithm of Figure \ref{fig:beamparser} is modified to read:

{\renewcommand{\baselinestretch}{1} \normalsize %normalsize to
					        %recalculate baselineskip
\begin{tabbing}
\renewcommand{\baselinestretch}{1}
\verb|    |\=\verb|    |\=\verb|    |\=\verb|    |\=\verb|    |\=\verb|    |\=\verb|    |\=\verb|    |\=\verb|    |\=\verb|    |\= \kill
 \>  \>  $\mi{best} := \max_A \mi{chart}[s, A, s+l] \times P(A);$\\
 \>  \>  \alforeach $A$ \\
 \>  \>  \> \alif $\mi{chart}[s, A, s+l]\times P(A) < T_B \times \mi{best}$\\
 \>  \>  \>  \> $\mi{chart}[s, A, s+l] := 0;$ \\
\end{tabbing}}

We can also give an item-based description for the CKY algorithm with
beam thresholding with the prior, as shown in Figure
\ref{fig:beamCKY}.  The item-based descriptions in this chapter can be
skipped for those not familiar with semiring parsing, as described in
Chapter \ref{ch:semi}; the descriptions in this chapter use minor
extensions to Chapter \ref{ch:semi} described in Section
\ref{sec:itemextend}.  The thresholding algorithm is the same as the
usual CKY algorithm, except that there is an added item form, $[i, j]$
containing the probability of the most probable nonterminal in the
span, and an added side condition on the binary rule, which ensures
that both children are sufficiently probable.  For the item-based
description, where we indicate rule values with the function $R$, we
have used the notation $R(X)$ to indicate the prior probability of
nonterminal $X$.

In Section \ref{sec:beamexp}, we will show experiments comparing
inside-probability beam thresholding to beam thresholding using the
inside probability times the prior.  Using the prior can lead to a
speedup of up to a factor of 10, at the same performance level.

\begin{figure}
\begin{center}
\begin{tabular}{c}
\psfig{figure=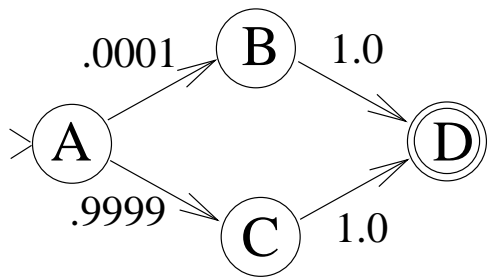}
\end{tabular}
\end{center}
\caption{Example Hidden Markov Model} \label{fig:simplehmm}
\end{figure}

To the best of our knowledge, using the prior probability in beam
thresholding is new, although not particularly insightful on our part.
Collins (personal communication) independently observed the usefulness
of this modification, and \newcite{Caraballo:96a} used a related
technique in a best-first parser.  We think that the main reason this
technique was not used sooner is that beam thresholding for PCFGs is
derived from beam thresholding in speech recognition using Hidden
Markov Models (HMMs), and in HMMs, this extra factor is almost never
needed.

Consider the simple HMM of Figure \ref{fig:simplehmm}.  We have not
annotated any output symbols: all states output the same symbol.
After a single input symbol, we are in state $B$ with probability
0.0001 and in state $C$ with probability 0.9999.  If we are using
thresholding, we should threshold out state $B$.  After one more
symbol, we arrive in the final state, $D$, and we are done.  Now,
consider the same process backwards. HMMs can be run backwards,
starting from the final state, and the last time, and moving towards
the start state and beginning time.  The total probability of any
string will be the same computed backwards as it was forwards.
However, notice what happens to the thresholding: state $B$ and state
$C$ are both equally likely, with value 1, when we move backwards, and
no thresholding occurs.  When moving forwards, as long as every state
in an HMM has some path to a final state -- and in essentially all
speech recognition applications, this is the case -- every state in an
HMM has probability 1 of eventually, perhaps after transitioning
through many other states, reaching the final state.  When processed
backwards, the probability of reaching the start state can be much
less than one, and in order to perform optimal thresholding, it is
necessary to factor in the prior probability of reaching each state
from the start.

In speech recognition, where beam thresholding was developed,
processing is usually done forwards, and this extra factor is not
needed.\footnote{In the cases where HMM processing is done backwards,
typically the forward probabilities are available, and techniques more
sophisticated than beam thresholding can be used.}  In contrast, in
parsing, the processing is usually bottom up, corresponding to a
backwards processing from end states (terminals) to the start state
(the start symbol $S$).  It is because of this bottom-up, backwards
processing that we need the extra factor that indicates the
probability of getting from the start symbol to the nonterminal in
question, which is proportional to the prior probability.  As we
noted, this can be very different for different nonterminals.

\section{Global Thresholding}

\label{sec:globalthresh}

As mentioned earlier, the problem with beam thresholding is that it
can only threshold out the worst nodes of a cell.  It cannot threshold
out an entire cell, even if there are no good nodes in it.  To remedy
this problem, we introduce a novel thresholding technique, global
thresholding, that uses an approximation to the outside probability
which takes into account all terminals not covered by the span under
consideration.  This allows nonterminals in different cells to be
compared to each other.

\begin{figure}
\begin{center}
\begin{tabular}{c}
\psfig{figure=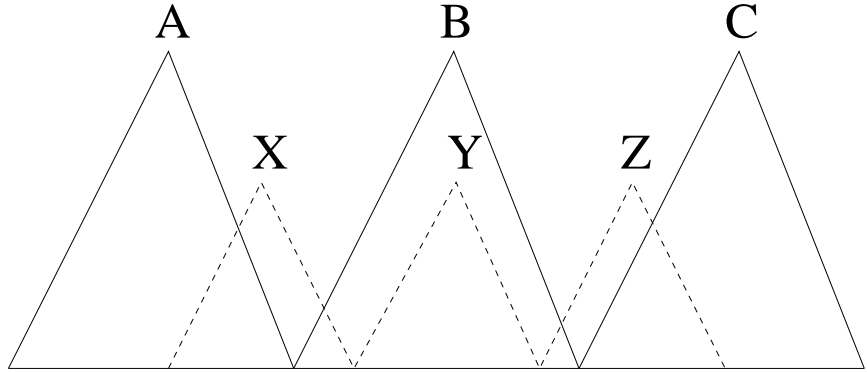,width=4.5in}
\end{tabular}
\end{center}
\caption{Global Thresholding Motivation} \label{fig:globalthreshold}
\end{figure}

The key insight of global thresholding is due to \newcite{Rayner:96a}.
Rayner and Carter noticed that a particular node cannot be part of
the correct parse if there are no nodes in adjacent cells.  In
fact, it must be part of a sequence of nodes stretching from the start
of the string to the end.  In a probabilistic framework where almost
every node will have some (possibly very small) probability, we can
rephrase this requirement as being that the node must be part of a
reasonably probable sequence.  

Figure \ref{fig:globalthreshold} shows an example of this insight.
Nodes A, B, and C will not be thresholded out, because each is part of
a sequence from the beginning to the end of the chart.  On the other
hand, nodes X, Y, and Z will be thresholded out, because none is part
of such a sequence.

Rayner and Carter used this insight for a hierarchical, non-recursive
grammar, and only used their technique to prune after the first level
of the grammar.  They computed a score for each sequence as the
minimum of the scores of each node in the sequence, and computed a
score for each node in the sequence as the minimum of three scores:
one based on statistics about nodes to the left, one based on nodes to
the right, and one based on unigram statistics.

We wanted to extend the work of Rayner and Carter to general PCFGs,
including those that were recursive.  Our approach therefore differs
from theirs in many ways.  Rayner and Carter ignore the inside
probabilities of nodes.  While this approach may work after processing
only the first level of a grammar, when the inside probabilities will
be relatively homogeneous, it could cause problems after other levels,
when the inside probability of a node will give important information
about its usefulness.  On the other hand, because long nodes will tend
to have low inside probabilities, taking the minimum of all scores
strongly favors sequences of short nodes.  Furthermore, their
algorithm requires time $O(n^3)$ to run just once.  This runtime is
acceptable if the algorithm is run only after the first level, but
running it more often would lead to an overall run time of $O(n^4)$.
Finally, we hoped to find an algorithm that was somewhat less
heuristic in nature.

Our global thresholding technique thresholds out node
$\constituent{j}{X}{k}$ if the ratio between the most probable
sequence of nodes including node $\constituent{j}{X}{k}$ and the
overall most probable sequence of nodes is less than some threshold,
$T_G$.  Formally, denoting sequences of nodes by $L$, we threshold
node $\constituent{j}{X}{k}$ if
\begin{equation}
T_G \max_{L} P(L) > \max_{L | \constituent{j}{X}{k} \in L} P(L)
\label{eqn:globalmax}
\end{equation}

Now, the hard part is determining $P(L)$, the probability of a node
sequence.  There is no way to do this efficiently as part of the
intermediate computation of a bottom-up chart parser.  Thus, we will
approximate $P(L)$ as follows:
$$
P(L) = \prod_{i} P(L_i|L_1...L_{i-1}) \approx \prod_{i} P(L_i)
$$
That is, we assume independence between the elements of a sequence.
The probability of node $L_i=\constituent{j}{X}{k}$ is just its prior probability
times its inside probability, as before.

Another way to look at global thresholding is as an approximation to
the un-normalized inside-outside probability.  In particular,
\begin{eqnarray}
P(S\derives w_1... w_{j-1} Xw_k... w_n\derives w_1... w_n) \approx \nonumber\\
\sum_{l, m,A_1... A_l, B_1... B_m} P(S \derives A_1... A_lXB_1... B_m
\derives w_1...w_{j-1} X w_k...w_n \derives w_1...w_n ) = \nonumber \\
\sum_{L | \constituent{j}{X}{k} \in L} P(L) \label{eqn:globalsum}
\end{eqnarray}
%
% The preceding equations are actually basically correct, although they skip a bunch
% of steps.  In particular, we need to have notation that makes it
% clear that it is the X after A_l that derives w_j...w_k, with 
% an intermediate step like S \derives A_1... A_lXB_1... B_m \derives
% w_1...w_{j-1} X B_1...B_m \derives w_1...w_{j-1} X w_k...w_n 
%
%\sum_{l, m,A_1... A_l, B_1... B_m} P(S \derives A_1... A_lXB_1... B_m
%\derives w_1_w_{j-1} X B_1...B_m 
%\derives w_1...w_{j-1} X w_k...w_n \derives w_1...w_n ) = \nonumber \\
%\sum_{l, m,A_1... A_l, B_1... B_m} \sum{\mbox{\small all ways for A_1
%through A_l to derive w_1...w_{j-1} \sum{\mbox{\small all ways for B_1
%through B_m to derive w_k...w_{n} P(S \derives A_1... A_lXB_1... B_m
%\derives w_1_w_{j-1} X B_1...B_m 
%\derives w_1...w_{j-1} X w_k...w_n \derives w_1...w_n ) \mbox{(in the
%specified way)}= \nonumber \\
%\sum_{L | \constituent{j}{X}{k} \in L} P(L) \label{eqn:globalsum}
%
%We won't bother to do this correctly, because it would only confuse things
%
%

Unlike Expression \ref{eqn:globalmax}, Equation \ref{eqn:globalsum}
uses a summation rather than a maximum; in practice we haven't found a
performance difference using either form.  This approximation is not a
very good one, since it will sum most derivations repeatedly.  For
instance, if we have
$$
S \Rightarrow A X \Rightarrow A_1 A_2 X \derives w_1...w_n
$$
then we will sum both
$$
P(S \derives A X \derives w_1...w_{j-1} X \derives w_1...w_n)
$$
and 
$$
P(S \derives A_1 A_2 X \derives w_1...w_{j-1} X \derives w_1...w_n)
$$
However, we speculate that each node $\constituent{j}{X}{k}$ is affected more or
less equally by this approximation, and the effects cancel out.  Next,
we make the same independence assumption as before, that the
probability of a sequence of nonterminals is equal to the product of
the prior probabilities of the nonterminals:
$$
P(S \derives A_1... A_lXB_1... B_m)\approx P(A_1)\times\cdots\times
P(A_l)\times P(X)\times P(B_1)\times\cdots\times P(B_m)
$$
Using this expression we can approximate the inside-outside
probabilities of any node $\constituent{j}{X}{k}$, and, compare it to the best
inside-outside probability of the sentence.

The most important difference between global thresholding and beam
thresholding is that global thresholding is global: any node in the
chart can help prune out any other node.  In stark contrast, beam
thresholding only compares nodes to other nodes covering the same
span.  Beam thresholding typically allows tighter thresholds since
there are fewer approximations, but does not benefit from global
information.

\subsection{Global Thresholding Algorithm}

\begin{figure}

\begin{tabbing}
\verb|    |\=\verb|    |\=\verb|    |\=\verb|    |\=\verb|    |\=\verb|    |\=\verb|    |\=\verb|    |\=\verb|    |\=\verb|    |\= \kill
\alfloat $f[1..n\!+\!1] := \{1, 0, 0, ..., 0\}$; \\
\alfor $\mi{end}$ := 2 to $n+1$ \\
 \> $f[\mi{end}] := \max_{\smi{start}<\smi{end}, X}
f[\mi{start}] \times \mi{inside}(\mi{start}, X, \mi{end}) \times P(X)$; \\
\\
\alfloat $b[1..n\!+\!1] := \{0, ..., 0, 0, 1\}$; \\
\alfor $\mi{start}$ := $n$ downto 1 \\
 \> $b[\mi{start}] := \max_{\smi{end}>\smi{start}, X} 
\mi{inside}(\mi{start}, X, \mi{end}) \times P(X)  \times b[\mi{end}]$; \\
\\
$\mi{bestProb} := f[n\!+\!1]$; \\
\alforeach node $\constituent{\mi{start}}{X}{\mi{end}}$ \\
 \> $\mi{total} := f[\mi{start}] \times \mi{inside}(\mi{start}, X, \mi{end}) \times P(X) \times b[\mi{end}]$; \\
 \> $\mi{active}[\mi{start}, X, \mi{end}] := 
\left\{\begin{array}{ll}
\mi{TRUE} & \mi{if } \mi{total} > \mi{bestProb} \times T_G \\
\mi{FALSE} & \mi{otherwise}
\end{array} \right. $ 
\end{tabbing}
\caption{Global Thresholding Algorithm} \label{fig:globalg}
\end{figure}

Global thresholding is performed in a bottom-up chart parser
immediately after each length is completed.  It thus runs $n$ times
during the course of parsing a sentence of length $n$.

We use the simple dynamic programming algorithm in Figure
\ref{fig:globalg}.  There are $O(n^2)$ nodes in the chart, and each
node is examined exactly three times, so the run time of this
algorithm is $O(n^2)$.  The first section of the algorithm works
forwards, computing, for each $i$, $f[i]$, which contains the score of
the best sequence covering terminals $w_1...w_{i-1}$.  Thus
$f[n\!+\!1]$ contains the score of the best sequence covering the
whole sentence, $\max_{L} P(L)$.  The algorithm works analogously to
the Viterbi algorithm for HMMs.  The second section is analogous, but
works backwards, computing $b[i]$, which contains the score of the
best sequence covering terminals $w_i...w_n$. 

Once we have computed the preceding arrays, computing $\max_{L |
\constituent{j}{X}{k} \in L} P(L)$ is straightforward.  We simply want
the score of the best sequence covering the nodes to the left of $j$,
$f[j]$, times the score of the node itself, times the score of the
best sequence of nodes from $k$ to the end, which is just $b[k]$.
Using this expression, we can threshold each node quickly.

Since this algorithm is run $n$ times during the course of parsing,
and requires time $O(n^2)$ each time it runs, the algorithm requires
time $O(n^3)$ overall.  Experiments will show that the time it saves
easily outweighs the time it uses.

\begin{figure}
$$
\begin{array}{cl}
\mbox{\bf Item form:} \\
{[i, A, j]} & \mbox{inside or Viterbi} \\
{[i]_j} & \mbox{inside or Viterbi} \\ \\
\mbox{\bf Primary Goal:} \\
{[1, S, n\!+\!1]} \\\\
\mbox{\bf Secondary Goals:} \\
{[n\!+\!1]_j} \\\\
\mbox{\bf Rules:} \\
\infer{R(A \rightarrow w_i)}{[i, A, i\!+\!1] }{} & \mbox{Unary} \\
\infer{R(A\! \rightarrow\! BC)\!\!\!\rspace [i, B, k]\!\!\!\rspace [k, C, j]}{[i, A,
j]}
{\begin{array}{l}
\VZVin([i, B, k], [n\!+\!1]_{j-i-1}) \geq T_G \wedge \\
\VZVin([k, C, j], [n\!+\!1]_{j-i-1}) \geq T_G
\end{array}
}
%{
%V_{\smi{in}}([i]_{j-i-1})V_{\smi{in}}([i, B, k])Z_{\smi{in}}([k]_{j-i-1}) \geq
%\mi{thresh} \times V_{\smi{in}}([n\!+\!1]_{j-i-1})
%\wedge 
%V_{\smi{in}}([k]_{j-i-1})V_{\smi{in}}([k, C, j])Z_{\smi{in}}([j]_{j-i-1}) \geq
%\mi{thresh} \times V_{\smi{in}}([n\!+\!1]_{j-i-1})
%}
& \mbox{Binary} \\
\infer{}{[1]_k}{} & \mbox{Initialization} \\
\infer{[i]_k [i, A, j]}{[j]_k}{k \leq j-i} & \mbox {Extension} \\
\end{array}
$$
\caption{Global thresholding item-based description}\label{fig:multiglobalthresh}
\end{figure}

In Figure \ref{fig:multiglobalthresh} we give an item-based description for
a global thresholding parser.  The algorithm is, again, very similar
to the CKY algorithm.  We have unary and binary rules, as usual.
However, there is now a side condition on the binary rules: both the
left and the right child must be part of a reasonably likely sequence.

There are two item types.  The first, $[i, A, j]$ has the usual
meaning.  The second item type, $[i]_j$, can be deduced if there is a
sequence of items $[1, A, m],[m, B, n],...,[o, C, i-1]$ where each
item has length at most $j$, where the length of an item $[i, A, k]$
is $k-i$: the number of words it covers.

The pseudocode of Figure \ref{fig:globalg} contains code only for
thresholding, which would be run after each length was processed in
the main parser.  The item-based description of Figure
\ref{fig:multiglobalthresh} gives a description for the complete
parser.  In the procedural version, the value of $f[i]$ when it is
computed after length $j$ would equal the forward value of
$[i]_j$ in the item-based description, and the value of $b[i]$ would
equal the reverse value of $[i]_j$.

There are two rules for deducing items of type $[i]_j$: initialization
and extension.  Initialization simply states that there is a zero
length sequence starting at word 1.  Extension states that if there is
a sequence of items of length at most $k$ covering words 1 through
$i-1$, and there is an item covering words $i$ through $j-1$, of
length at most $k$, then there is a sequence of items of length at
most $k$ covering words 1 through $j-1$.  The trickiest rule is the
binary rule, which has a fairly complicated side condition.  It checks
for both $[i, B, k]$ and $[k, C, j]$ that there is a reasonably likely
sequence covering the sentence, using items of length at most $j-i$
and including the item.

We should note that an earlier version of global thresholding using a
standard pseudo-code specification contained a subtle
bug:\footnote{Thanks to Michael Collins for catching it.}  the reverse
values for items $[i]_j$ were incorrectly computed.  The item-based
description of Figure \ref{fig:multiglobalthresh} makes it clearer
what is going on than the procedural definition of Figure
\ref{fig:globalg} can, and makes a bug of this form much less likely.

\section{Multiple-Pass Parsing}

\label{sec:multithresh}

In this section, we discuss a novel thresholding technique,
multiple-pass parsing.  We show that multiple-pass parsing techniques
can yield large speedups.  Multiple-pass parsing is a variation on a
new technique in speech recognition, multiple-pass speech recognition
\cite{Zavaliagkos:94a}, which we introduce first.  

\subsection{Multiple-Pass Speech Recognition}

The basic idea behind multiple-pass speech recognition is that we can
use the normalized forward-backward probabilities of one HMM as
approximations to the normalized forward-backward probabilities of
another HMM.  In an idealized multiple-pass speech recognizer, we
first run a simple, fast first pass HMM, computing the forward and
backward probabilities.  We then use these probabilities as
approximations to the probabilities in the corresponding states of a
slower, more accurate second pass HMM.  We don't need to examine
states in this second pass HMM that correspond to low inside-outside
probability states in the first pass.  The extra time of running two
passes is more than made up for by the time saved in the second pass.

The mathematics of multiple-pass recognition is fairly simple.  In the
first simple pass, we record the forward probabilities,
$\mi{forward}(t, i)$, and backward probabilities, $\mi{backward}(t, i)$, of each
state $i$ at each time $t$.  Now, $\frac{\smi{forward}(t, i) \times
\smi{backward}(t, i)}{\smi{forward}(T, \smi{final})}$ gives the overall
probability of being in state $i$ at time $t$ given the acoustics.
Our second pass will use an HMM whose states are analogous to the
first pass HMM's states.  If a first pass state at some time is
unlikely, then the analogous second pass state is probably also
unlikely, so we can threshold it out.

There are a few complications to multiple-pass recognition.  First,
storing all the forward and backward probabilities can be expensive.
Second, the second pass is more complicated than the first, typically
meaning that it has more states.  So the mapping between states in the
first pass and states in the second pass may be non-trivial.  To
solve both these problems, only states at word transitions are saved.
That is, from pass to pass, only information about where words are
likely to start and end is used for thresholding.

\subsection{Multiple-Pass Parsing}

We can use an analogous algorithm for multiple-pass parsing.  In this
case, we will use the normalized inside-outside probabilities of one
grammar as approximations to the normalized inside-outside
probabilities of another grammar.  We first compute the normalized
inside-outside probabilities of a simple, fast first pass grammar.  Next, we run
a slower, more accurate second pass grammar, ignoring constituents
whose corresponding first pass inside-outside probabilities are too
low.

Of course, for our second pass to be more accurate, it will probably
be more complicated, typically containing an increased number of
nonterminals and productions.  Thus, we create a mapping function from
each first pass nonterminal to a set of second pass nonterminals, and
threshold out those second pass nonterminals that map from low-scoring
first pass nonterminals.  We call this mapping function the {\em
elaborations function}.\footnote{In this chapter, we will assume that
each second pass nonterminal is an elaboration of at most one first
pass nonterminal in each cell.  The grammars used here have this
property.  If this assumption is violated, multiple-pass parsing is
still possible, but some of the algorithms need to be changed.}

There are many possible examples of first and second pass
combinations.  For instance, the first pass could use regular
nonterminals, such as $\mathit{NP}$ and $\mathit{VP}$ and the second
pass could use nonterminals augmented with head-word information.  The
elaborations function then appends the possible head words to the first
pass nonterminals to get the second pass ones.

\begin{figure}
\begin{tabbing}
\verb|  |\=\verb|  |\=\verb|  |\=\verb|  |\=\verb|  |\=\verb|  |\=\verb|  |\=\verb|  |\=\verb|  |\=\verb|  |\= \kill
\alfor $\mi{length} := 2$ \alto $n$ \\
 \> \alfor $\mi{start} := 1$ \alto $n-\mi{length}+1$ \\
 \> \> \alfor $\mi{leftLength} := 1$ \alto $\mi{length}-1$ \\
 \> \>  \> $\mi{LeftPrev} := \mi{PrevChart}[\mi{leftLength}][\mi{start}]$; \\
 \> \>  \> \alforeach $\mi{LeftNodePrev} \in \mi{LeftPrev}$ \\
 \> \>  \>  \> \alforeach non-thresholded production instance $\mi{Prod}$ from \\
 \> \>  \>  \>  \>  \>  \> $\mi{LeftNodePrev}$ of size $\mi{length}$ \\
 \> \>  \>  \>  \> \alforeach elaboration $L$ of $\mi{Prod}_{\smi{Left}}$ \\
 \> \>  \>  \>  \>  \> \alforeach elaboration $R$ of $\mi{Prod}_{\smi{Right}}$ \\
 \> \>  \>  \>  \>  \>  \> \alforeach elaboration $P$ of $\mi{Prod}_{\smi{Parent}}$ \\
 \> \>  \>  \>  \>  \>  \>  \>  \>  \> such that $P \rightarrow L\; R$ \\
 \> \>  \>  \>  \>  \>  \>  \> add $P$ to $\mi{Chart}[\mathit{length}][\mi{start}]$; \\
% \>  \>  \>  \>  \>  \>  \> add $P$ to
%$\mi{Elaborations}[\mi{length}][\mi{start}][\mi{Prod}_{\smi{Parent}}]$; \\
\end{tabbing}
\caption{Second Pass Parsing Algorithm} \label{fig:secondpass}
\end{figure}

Even though the correspondence between forward/backward and
inside/outside probabilities is very close, there are important
differences between speech-recognition HMMs and natural-language
processing PCFGs.  In particular, we have found that in some cases it
is more important to threshold productions than nonterminals.  That
is, rather than just noticing that a particular nonterminal
$\mathit{VP}$ spanning the words ``killed the rabbit'' is very likely,
we also note that the production $ \mathit{VP} \rightarrow
V\;\mathit{NP}$ (and the relevant spans) is likely.

Both the first and second pass parsing algorithms are simple
variations on CKY parsing.  In the first pass, we now keep track of
each {\em production instance} associated with a node,
i.e. $\constituent{i}{X}{j} \rightarrow
\constituent{i}{Y}{k}\;\constituent{k}{Z}{j}$, computing the inside
and outside probabilities of each.  We remove all constituents and
production instances from the first pass whose normalized
inside-outside probability is too small.  The second pass requires
more changes.  Let us denote the elaborations of nonterminal $X$ by
$X_1...X_x$.  In the second pass, for each production of the form
$\constituent{i}{X}{j} \rightarrow \constituent{i}{Y}{k} \;
\constituent{k}{Z}{j}$ in the first pass that was not thresholded out
by multi-pass, beam, or global thresholding, we consider every
elaborated production instance, that is, all those of the form
$\constituent{i}{X_p}{j} \rightarrow \constituent{i}{Y_q}{k} \;
\constituent{k}{Z_r}{j}$, for appropriate values of $p, q, r$.  This
algorithm is given in Figure \ref{fig:secondpass}, which uses a
current pass matrix $\mi{Chart}$ to keep track of nonterminals in the
current pass, and a previous pass matrix, $\mi{PrevChart}$ to keep
track of nonterminals in the previous pass.  We use one additional
optimization, keeping track of the elaborations of each nonterminal in
each cell in $\mi{PrevChart}$ that are in the corresponding cell of
$\mi{Chart}$.

We tried multiple-pass thresholding in two different ways.  In the
first technique we tried, production-instance thresholding, we remove
from consideration in the second pass the elaborations of all
production instances whose combined inside-outside probability falls
below a threshold.  In the second technique, node thresholding, we
remove from consideration the elaborations of all nodes whose
inside-outside probability falls below a threshold.  In our pilot
experiments, we found that in some cases one technique works slightly
better, and in some cases the other does.  We therefore ran our
experiments using both thresholds together.

\begin{figure}
$$
\begin{array}{cl}
\mbox{\bf Item form:} \\
{[i, A, j]_x} \\ \\
\mbox{\bf Primary Goal:} \\
{[1, S, n\!+\!1]_p} \\\\
\mbox{\bf Secondary Goals:} \\
{[1, S, n\!+\!1]_x \rspace x < p}  \\\\
\mbox{\bf Rules:} \\
\infer{R_x(A \rightarrow w_i)}{[i, A, i\!+\!1]_x }
{
\begin{array}{l}
x = 1\vee \\
\VZVin([i, \mi{map}_x(A), i\!+\!1]_{x-1}, [1, S, n\!+\!1]_{x-1})\geq T_x
\end{array}
} & \mbox{Unary} \\
\infer{R_x(A\! \rightarrow\! BC)\smallrspace [i, B, k]_x\smallrspace [k, C, j]_x}{[i, A,
j]_x}{\begin{array}{l}
x = 1\vee \\
\VZVin([i, \mi{map}_x(A), j]_{x-1}, [1, S, n\!+\!1]_{x-1})\geq T_x \\
\end{array}}
& \mbox {Binary}
\end{array}
$$
\caption{Multiple-Pass Parsing Description}\label{fig:threshmulti}
\end{figure}

The item-based description format of Chapter \ref{ch:semi} allows us
to describe multiple pass parsing very succinctly; Figure
\ref{fig:threshmulti} is such a description.  This description only
does node thresholding; production thresholding could be implemented in a
similar way.  The description is identical to the CKY item-based
description with the following changes.  First, every item is
annotated with a subscript indicating the pass of that item.  We have
also annotated the rule value function with a subscript, indicating
the pass for that rule.  Finally, each deduction rule contains an
additional side condition of the form
$$
x = 1\vee \VZVin([i, \mi{map}_x(A), j]_{x-1}, [1,S,n\!+\!1]_{x-1})\geq
T_x 
$$
indicating that the deduction rule should trigger only if we are
either on the first pass, or the equivalent item from the previous
pass (derived using the $\mi{map}_x$ function) was within the
threshold, $T_x$.  Notice that the first $p-1$ passes use the inside
semiring, but that the final pass can use any semiring.

One nice feature of multiple-pass parsing is that under special
circumstances, it is an {\em admissible} search technique, meaning
that we are guaranteed to find the best solution with it.  In
particular, if we parse using no thresholding, and our grammars have
the property that for every non-zero probability parse in the second
pass, there is an analogous non-zero probability parse in the first
pass, then multiple-pass search is admissible.  Under these
circumstances, no non-zero probability parse will be thresholded out,
but many zero probability parses may be removed from consideration.
While we will almost always wish to parse using thresholds, it is nice
to know that multiple-pass parsing can be seen as an approximation to
an admissible technique, where the degree of approximation is
controlled by the thresholding parameter.

\section{Multiple Parameter Optimization}

\label{sec:descent}

The use of any one of these techniques does not exclude the use of the
others.  There is no reason that we cannot use beam thresholding,
global thresholding, and multiple-pass parsing all at the same time.
In general, it would not make sense to use a technique such as
multiple-pass parsing without other thresholding techniques; our
first pass would be overwhelmingly slow without some sort of
thresholding.

There are, however, some practical considerations.  To optimize a
single threshold, we could simply sweep our parameters over a one
dimensional range, and pick the best speed versus performance
tradeoff.  In combining multiple techniques, we need to find optimal
combinations of thresholding parameters.  Rather than having to
examine ten values in a single dimensional space, we might have to
examine one hundred combinations in a two dimensional space.  Later,
we show experiments with up to six thresholds.  Since we do not have
time to parse with one million parameter combinations, we need a
better search algorithm.

Ideally, we would simply run some form of gradient descent algorithm,
optimizing a weighted sum of performance and time.  There are two
problems with this approach.  First is that most measures of
performance are too noisy.  It is important when doing gradient
descent that the performance measure be smooth and monotonic enough
that the numerical derivative can be accurately measured.  If the
performance measure is noisy, then we must run a large number of
sentences in order to accurately measure the derivative.  However, if
the number of sentences is too large, then the algorithm will be too
slow.  Thus, it is important to have a smooth performance measure.  We
will show that the inside probability is much smoother and more
monotonic than conventional performance measures.  Because it is so
smooth, we can use a relatively small number of sentences when
determining derivatives, allowing the optimization algorithm to run in
a reasonable amount of time.

The second problem with using a simple gradient descent algorithm is
that it does not give us much control over the solution we arrive at.
Ideally, we would like to be able to pick a performance level (in
terms of either entropy or precision and recall) and find the best set
of thresholds for achieving that performance level as quickly as
possible.  If an absolute performance level is our goal, then a normal
gradient descent technique will not work, since we cannot use such a
technique to optimize one function of a set of variables (time as a
function of thresholds) while holding another one constant
(performance).  We could use gradient descent to minimize a weighted
sum of time and performance, but we would not know at the beginning
what performance level we would have at the end.  If our goal is to
have the best performance we can while running in real time, or to
achieve a minimum acceptable performance level with as little time as
necessary, then a simple gradient descent function would not work as
well as the algorithm we will give.

We will show that the inside probability is a good performance measure
to optimize.  We need a metric of performance that will be sensitive
to changes in threshold values.  In particular, our ideal metric would
be strictly increasing as our thresholds loosened, so that every
loosening of threshold values would produce a measurable increase in
performance.  The closer we get to this ideal, the fewer sentences we
need to test during parameter optimization.

\begin{table}
\begin{center}
\begin{tabular}{r|lll}
Metric & decrease & same & increase \\ \hline
              Inside &    7 &   65 & 1625 \\
             Viterbi &    6 & 1302 &  389 \\
       Cross Bracket &  132 & 1332 &  233 \\
  Zero Cross Bracket &   18 & 1616 &   63 \\
           Precision &  132 & 1280 &  285 \\
              Recall &  126 & 1331 &  240 \\
\end{tabular}
\end{center}
\caption{Monotonicity of various metrics} \label{tab:metricdirection}
\end{table}

We tried an experiment in which we ran beam thresholding with a tight
threshold, and then a loose threshold, on all sentences of section 0
of length at most $40$.  For this experiment only, we discarded those
sentences that could not be parsed with the specified setting of the
threshold, rather than retrying with looser thresholds.  For any given
sentence, we would generally expect that it would do better on most
measures with a loose threshold than with a tight one, but of course
this will not always be the case.  For instance, there is a very good
chance that the number of crossing brackets or the precision and
recall for any particular sentence will not change at all when we move
from a tight to a loose threshold.  There is even some chance, for any
particular sentence that the number of crossing brackets, or the
precision or the recall, will even get worse.  We calculated, for each
measure, how many sentences fell into each of the three categories:
increased score, same score, or decreased score.  Table
\ref{tab:metricdirection} gives the results.  As can be seen, the
inside score was by far the most nearly strictly increasing metric.
Furthermore, as we will show in Figure \ref{fig:smoothit}, this metric
also correlates well with precision and recall, although with less
noise.  Therefore, we should use the inside probability as our metric
of performance; however inside probabilities can become very close to
zero, so instead we measure entropy, the negative logarithm of the
inside probability.

\begin{figure}
\begin{tabbing}
\verb|  |\=\verb|  |\=\verb|  |\=\verb|  |\=\verb|  |\=\verb|  |\=\verb|  |\=\verb|  |\=\verb|  |\=\verb|  |\= \kill
\alwhile \alnot $\mi{Thresholds} \in \mi{ThresholdsSet}$ \\
 \> add $\mi{Thresholds}$ to $\mi{ThresholdsSet};$ \\
 \> $(\mi{BaseE}_T, \mi{BaseTime}) :=$ ParseAll$(\mi{Thresholds})$; \\
 \> \alforeach $\mi{Threshold} \in \mi{Thresholds}$ \\
 \>  \> if $\mi{BaseE}_T > \mi{TargetE}_T$ \\
 \>  \>  \> tighten $\mi{Threshold}$; \\
 \>  \>  \> $(\mi{NewE}_T, \mi{NewTime}) :=$ ParseAll$(\mi{Thresholds})$; \\
 \>  \>  \> $\mi{Ratio} := (\mi{BaseTime}-\mi{NewTime}) \; / $ \\
 \>  \>  \>  \>  \>  \>  $(\mi{BaseE}_T-\mi{NewE}_T)$; \\
 \>  \> \alelse \\
 \>  \>  \> loosen $\mi{Threshold}$; \\
 \>  \>  \> $(\mi{NewE}_T, \mi{NewTime}) :=$ ParseAll$(\mi{Thresholds})$; \\
 \>  \>  \> $\mi{Ratio} := (\mi{BaseE}_T-\mi{NewE}_T)\;/$ \\
 \>  \>  \>  \>  \>  \> $(\mi{BaseTime}-\mi{NewTime})$; \\
 \> change $\mi{Threshold}$ with best $\mi{Ratio}$; \\
\end{tabbing}

\caption{Gradient Descent Multiple Threshold Search} \label{fig:multisearch}
\end{figure}

%In Section \ref{sec:expmeasure} we discuss the reasons for choosing to
%optimize entropy instead of precision and recall in more detail, and
%give experimental results that show that entropy is indeed a good
%measure.

We implemented a variation on a steepest descent search technique.  We
denote the entropy of the sentence after thresholding by $E_T$.  Our
search engine is given a target performance level $E_T$ to search for,
and then tries to find the best combination of parameters that works
at approximately this level of performance.  At each point, it finds
the threshold to change that gives the most ``bang for the buck.''  It
then changes this parameter in the correct direction to move towards
$E_T$ (and possibly overshoot it).  A simplified version of the
algorithm is given in Figure \ref{fig:multisearch}.

\begin{figure}
\begin{center}
\begin{tabular}{cc}
\psfig{figure=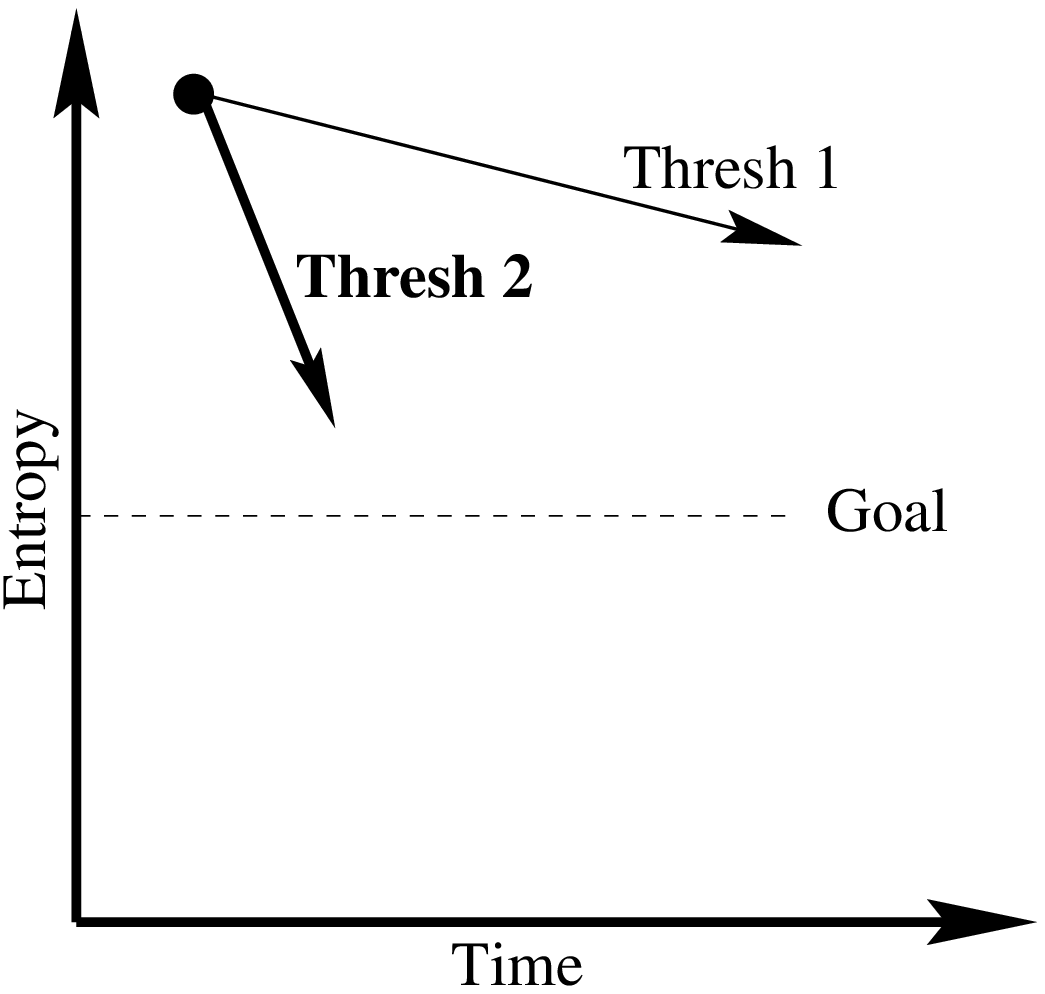,width=2.8in}  &
\psfig{figure=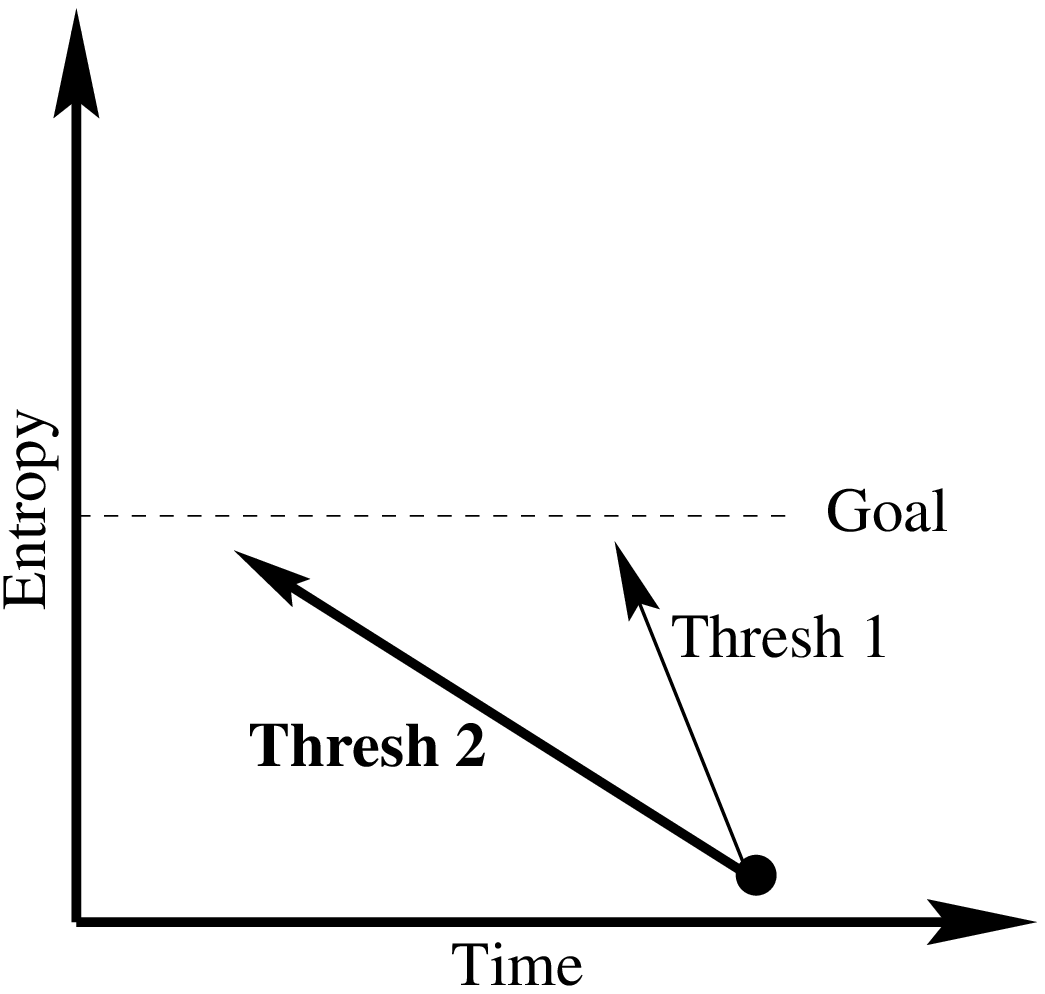,width=2.8in} \\
Optimizing for & Optimizing for \\
{\bf Lower Entropy}: &  {\bf Faster Speed}: \\
{\bf Steeper} is Better &  {\bf Flatter} is Better \\
\end{tabular}
\caption{Optimizing for Lower Entropy versus Optimizing for Faster
Speed} \label{fig:optimpic}
\end{center}
\end{figure}

Figure \ref{fig:optimpic} shows graphically how the algorithm works.
There are two cases.  In the first case, if we are currently above the
goal entropy, then we loosen our thresholds, leading to slower speed\footnote{For this algorithm (although not for most
experiments), our measurement of time was the total number of
productions searched, rather than cpu time; we wanted the greater
accuracy of measuring productions.}
and lower entropy.  We then wish to get as much entropy reduction as
possible per time increase; that is, we want the steepest slope possible.
On the other hand, if we are trying to increase our entropy, we want
as much time decrease as possible per entropy increase; that is, we want
the flattest slope possible.  Because of this difference, we need to
compute different ratios depending on which side of the goal we are on.

There are several subtleties when thresholds are set very tightly.
When we fail to parse a sentence because the thresholds are too tight,
we retry the parse with lower thresholds.  This can lead to conditions
that are the opposite of what we expect; for instance, loosening
thresholds may lead to faster parsing, because we don't need to parse
the sentence, fail, and then retry with looser thresholds.  The full
algorithm contains additional checks that our thresholding change had
the effect we expected (either increased time for decreased entropy or
vice versa).  If we get either a change in the wrong direction, or a
change that makes everything worse, then we retry with the reverse
change, hoping that that will have the intended effect.  If we get a
change that makes both time and entropy better, then we make that
change regardless of the ratio.

Also, we need to do checks that the denominator when computing $Ratio$
is not too small.  If it is very small, then our estimate may be
unreliable, and we do not consider changing this parameter.  Finally,
the actual algorithm we used also contained a simple ``annealing
schedule'', in which we slowly decreased the factor by which we changed
thresholds.  That is, we actually run the algorithm multiple times to
termination, first changing thresholds by a factor of 16.  After a
loop is reached at this factor, we lower the factor to 4,
then 2, then 1.414, then 1.15.

We note that this algorithm is fairly task independent.  It can be used
for almost any statistical parsing formalism that uses thresholds, or
even for speech recognition.

\section{Comparison to Previous Work}

Beam thresholding is a common approach.  While we do not know of other
systems that have used exactly our techniques, our techniques are
certainly similar to those of others.  For instance,
\newcite{Collins:96a} uses a form of beam thresholding that differs
from ours only in that it does not use the prior probability of
nonterminals as a factor, and \newcite{Caraballo:96a} use a version
with the prior, but with other factors as well.

Much of the previous related work on thresholding is in the similar
area of priority functions for agenda-based parsers.  These parsers
try to do ``best first'' parsing, with some function akin to a
thresholding function determining what is best.  The best comparison
of these functions is due to Caraballo and Charniak
\shortcite{Caraballo:96a,Caraballo:97a}, who tried various
prioritization methods.  Several of their techniques are similar to
our beam thresholding technique, and one of their techniques, not yet
published \cite{Caraballo:97a}, would probably work better.

The only technique that Caraballo and Charniak
\shortcite{Caraballo:96a} give that took into account the scores of
other nodes in the priority function, the ``prefix model,'' required
$O(n^5)$ time to compute, compared to our $O(n^3)$ system.  On the
other hand, all nodes in the agenda parser were compared to all other
nodes, so in some sense all the priority functions were global.

We note that agenda-based PCFG parsers in general require more than
$O(n^3)$ run time, because, when better derivations are discovered,
they may be forced to propagate improvements to productions that they
have previously considered.  For instance, if an agenda-based system
first computes the probability for a production $S \rightarrow
\mathit{NP}\;\mathit{VP}$, and then later computes some better
probability for the $\mathit{NP}$, it must update the probability for
the $S$ as well.  This could propagate through much of the chart.  To
remedy this, Caraballo \etal only propagated probabilities that
caused a large enough change \cite{Caraballo:97a}.  Also, the question
of when an agenda-based system should stop is a little discussed
issue, and difficult since there is no obvious stopping criterion.
Because of these issues, we chose not to implement an agenda-based
system for comparison.

As mentioned earlier, \newcite{Rayner:96a} describe a system that is
the inspiration for global thresholding.  Because of the limitation of
their system to non-recursive grammars, and the other differences
discussed in Section \ref{sec:globalthresh}, global thresholding represents
a significant improvement.

\newcite{Collins:96a} uses two thresholding techniques.  The first of
these is essentially beam thresholding without a prior.  In the second
technique, there is a constant probability threshold.  Any nodes with
a probability below this threshold are pruned.  If the parse fails,
parsing is restarted with the constant lowered.  We attempted to
duplicate this technique, but achieved only negligible performance
improvements.  Collins (personal communication) reports a 38\% speedup
when this technique is combined with loose beam thresholding, compared to
loose beam thresholding alone.  Perhaps our lack of success is due to
differences between our grammars, which are fairly different
formalisms.  When Collins began using a formalism somewhat closer to
ours, he needed to change his beam thresholding to take into account
the prior, so this hypothesis is not unlikely.  Hwa (personal
communication) using a model similar to PCFGs, Stochastic Lexicalized
Tree Insertion Grammars, also was not able to obtain a speedup using
this technique.

There is previous work in the speech recognition community on
automatically optimizing some parameters \cite{Schwartz:92a}.
However, this previous work differed significantly from ours both in
the techniques used, and in the parameters optimized.  In particular,
previous work focused on optimizing weights for various components,
such as the language model component.  In contrast, we optimize
thresholding parameters.  Previous techniques could not be used for or
easily adapted to thresholding parameters.

\section{Experiments}

\subsection{Data}

\begin{figure}
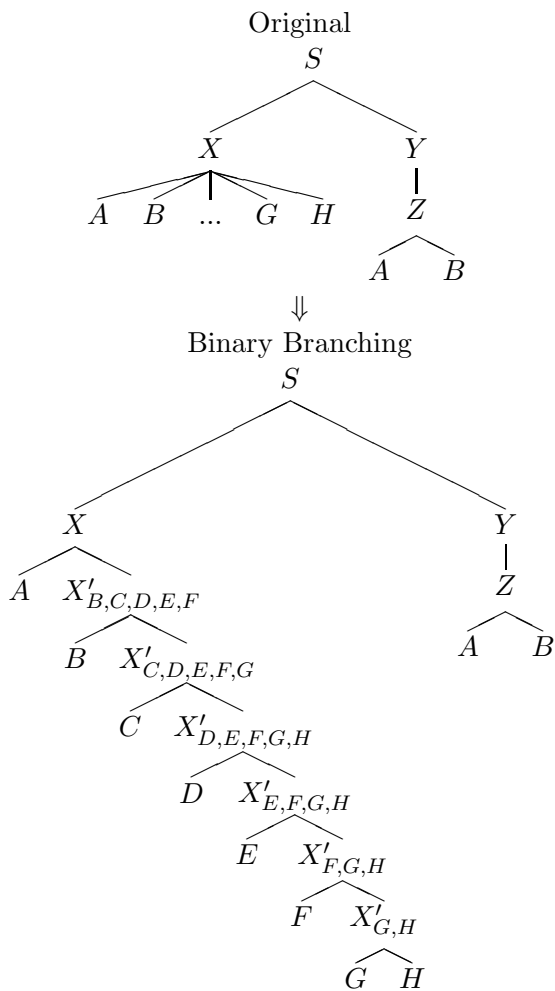

\begin{center}
\begin{tabular}{c}
Original \\
\leaf{$A$}
\leaf{$B$}
\leaf{$...$}
\leaf{$G$}
\leaf{$H$}
\branch{5}{$X$}
\leaf{$A$}\faketreewidth{AA}
\leaf{$B$}\faketreewidth{AA}
\branch{2}{$Z$}
\branch{1}{$Y$}
\branch{2}{$S$}
\hspace{-.8in}\tree \\
$\Downarrow$ \\
Binary Branching \\
\leaf{$A$}
\leaf{$B$}
\leaf{$C$}
\leaf{$D$}
\leaf{$E$}
\leaf{$F$}
\leaf{$G$}
\leaf{$H$}
\branch{2}{$X^{\prime}_{G,H}$}
\branch{2}{$X^{\prime}_{F,G,H}$}
\branch{2}{$X^{\prime}_{E,F,G,H}$}
\branch{2}{$X^{\prime}_{D,E,F,G,H}$}
\branch{2}{$X^{\prime}_{C,D,E,F,G}$}
\branch{2}{$X^{\prime}_{B,C,D,E,F}$}
\branch{2}{$X$}
\leaf{$A$}\faketreewidth{AA}
\leaf{$B$}\faketreewidth{AA}
\branch{2}{$Z$}
\branch{1}{$Y$}
\branch{2}{$S$}
\hspace {-1.2in}\tree
\end{tabular}
\end{center}
\caption{Converting to Binary Branching}
\label{fig:threshbinary}
\end{figure}

All experiments were trained on sections 2-18 of the Penn Treebank,
version II.  A few were tested, where noted, on the first 200
sentences of section 00 of length at most 40 words.  In one
experiment, we used the first 15 of length at most 40, and in the
remainder of our experiments, we used those sentences in the first
1001 of length at most 40.  Our parameter optimization algorithm
always used the first 31 sentences of length at most 40 words from
section 19.  We ran some experiments on more sentences, but there were
three sentences in this larger test set that could not be parsed with
beam thresholding, even with loose settings of the threshold; we
therefore chose to report the smaller test set, since it is difficult
to compare techniques that did not parse exactly the same sentences.

\subsection{The Grammar}

We needed several grammars for our experiments so that we could test
the multiple-pass parsing algorithm.  The grammar rules, and their
associated probabilities, were determined by reading them off of the
training section of the treebank, in a manner very similar to that
used by \newcite{Charniak:96a}.  The main grammar we chose was
essentially of the following form:%
\footnote{In Chapter \ref{ch:PFG}, we describe Probabilistic Feature
Grammars.  This grammar can be more simply described as a PFG with six
features: the continuation feature,  child1, child2,..., child5.  No
smoothing was done, and some dependencies that could have been
captured, such as the dependence between the left child's child1
feature and the parent's child2 feature, were ignored, in order to
capture only the dependencies described here.  Unary branches were
handled as described in Section \ref{sec:unary}.}

\begin{center}
\renewcommand{\arraystretch}{0.5}
\begin{tabular}{rcl}
$X$                  & $\Rightarrow$ & $A \; X^{\prime}_{B,C,D,E,F} $ \\
$X^{\prime}_{A,B,C,D,E}$      & $\Rightarrow$ & $A \; X^{\prime}_{B,C,D,E,F} $ \\
$X$                  & $\Rightarrow$ & $A$ \\
$X$                  & $\Rightarrow$ & $A \; B$ \\
\end{tabular}
\end{center}

That is, productions were all unary or binary branching.  There were
never more than five subscripted symbols for any nonterminal, although
there could be fewer than five if there were fewer than five symbols
remaining on the right hand side.  Thus, our grammar was a kind of
6-gram model on symbols in the grammar.

Figure \ref{fig:threshbinary} shows an example of how we converted
trees to the form of our grammar.  We refer to this grammar as the
{\em 6-gram grammar}.  The terminals of the grammar were the
part-of-speech symbols in the treebank.  Any experiments that do not
mention which grammar we used were run with the 6-gram grammar.

\begin{figure}
\begin{center}
\begin{tabular}{c}
Original \\
\leaf{$\mathit{det}$}
\leaf{$\mathit{adj}$}
\leaf{$\mathit{noun}$}
\branch{3}{$\mathit{NP}$}
\leaf{$\mathit{verb}$}
\branch{1}{$\mathit{VP}$}
\branch{2}{$\mathit{S}$}
\hspace{-.6in} \tree  \\ \\ \\
Terminal \\
\leaf{$\mathit{det}$}
\leaf{$\mathit{adj}$}
\leaf{$\mathit{noun}$}
\branch{2}{$\mathit{ADJ}$}
\branch{2}{$\mathit{DET}$}
\leaf{$\mathit{verb}$}
\branch{1}{$\mathit{VERB}$}
\branch{2}{$\mathit{DET}$}
\hspace{-.55in} \tree \\ \\ \\
Terminal-Prime \\
\leaf{$\mathit{det}$}
\leaf{$\mathit{adj}$}
\leaf{$\mathit{noun}$}
\branch{2}{$\mathit{ADJ'}$}
\branch{2}{$\mathit{DET}$}
\leaf{$\mathit{verb}$}
\branch{1}{$\mathit{VERB}$}
\branch{2}{$\mathit{DET}$}
\hspace{-.55in} \tree 
\\
\end{tabular}
\end{center}
\caption{Converting to Terminal and Terminal-Prime Grammars 
\label{fig:maketerm}}
\end{figure}

For a simple grammar, we wanted something that would be very fast.
The fastest grammar we can think of we call the {\em terminal}
grammar, because it has one nonterminal for each terminal symbol in
the alphabet.  The nonterminal symbol indicates the first terminal in
its span.  The parses are unary and binary branching in the same way
that the 6-gram grammar parses are.  Figure \ref{fig:maketerm} shows
how to convert a parse tree to the terminal grammar.  Since there is
only one nonterminal possible for each cell of the chart, parsing is
quick for this grammar.  For technical and practical
reasons, we
actually wanted a marginally more complicated grammar, which included
the ``prime'' symbol of the 6-gram grammar, indicating that a cell is
part of the same constituent as its parent.%
\footnote{Our parser is the PFG parser of Chapter \ref{ch:PFG}.  Many
of the decisions that parser makes depend on the value of the
continuation feature, and so the PFG parser cannot run without that
feature.  Furthermore, keeping information for each constituent about
whether it was an internal, ``primed'' feature or not seemed like it
would provide useful information to the first pass.}
Therefore, we doubled the size of
the grammar so that there would be both primed and non-primed versions
of each terminal; we call this the {\em terminal-prime} grammar, and
also show how to convert to it in Figure \ref{fig:maketerm}.  This
grammar is the one we actually used as the first pass in our
multiple-pass parsing experiments.\footnote{This grammar can be more
simply described as a PFG with two features: the continuation feature,
which corresponds to the prime, and a feature for the first terminal.
No smoothing was done.}
%I think simplest.gnr is the file I used.  Looking at it, it appears
%to capture all possible dependencies, but it also appears that this
%is exactly what is specified here. 

\subsection{What we measured}

\label{sec:expmeasure}

The goal of a good thresholding algorithm is to trade off correctness
for increased speed.  We must thus measure both correctness and speed,
and there are some subtleties to measuring each.

The traditional way of measuring correctness is with metrics such as
precision and recall, which were described in Chapter
\ref{sec:narybranch}.  There are two problems with these measures.
First, they are two numbers, neither useful without the other.
Second, they are subject to considerable noise.  In pilot experiments,
we found that as we changed our thresholding values monotonically,
precision and recall changed non-monotonically (see Figure
\ref{fig:smoothit}).  We attribute this to the fact that we must
choose a single parse from our parse forest, and, as we tighten a
thresholding parameter, we may threshold out either good or bad
parses.  Furthermore, rather than just changing precision or recall by
a small amount, a single thresholded item may completely change the
shape of the resulting tree.  Thus, precision and recall are only
smooth with very large sets of test data.  However, because of the
large number of experiments we wished to run, using a large set of
test data was not feasible.  Thus, we looked for a surrogate measure,
and decided to use the total inside probability of all parses, which,
with no thresholding, is just the probability of the sentence given
the model.  If we denote the total inside probability with no
thresholding by $I$ and the total inside probability with thresholding
by $I_T$, then $\frac{I_T}{I}$ is the probability that we did not
threshold out the correct parse, given the model.  Thus, maximizing
$I_T$ should maximize correctness.  Since probabilities can become
very small, we instead minimize entropies, the negative logarithm of
the probabilities.  Figure \ref{fig:smoothit} shows that with a large
data set, entropy correlates well with precision and recall, and that
with smaller sets, it is much smoother.  Entropy is smoother because
it is a function of many more variables: in one experiment, there were
about 16000 constituents that contributed to precision and recall
measurements, versus 151 million productions potentially contributing
to entropy.  Thus, we choose entropy as our measure of correctness for
most experiments.  When we did measure precision and recall, we used
the metric as defined by \newcite{Collins:96a}.

\begin{figure*}
\begin{center}
\begin{tabular}{cc}
Precision/Recall & Total Entropy \\
15 Sentences & 15 Sentences \\
\psfig{figure=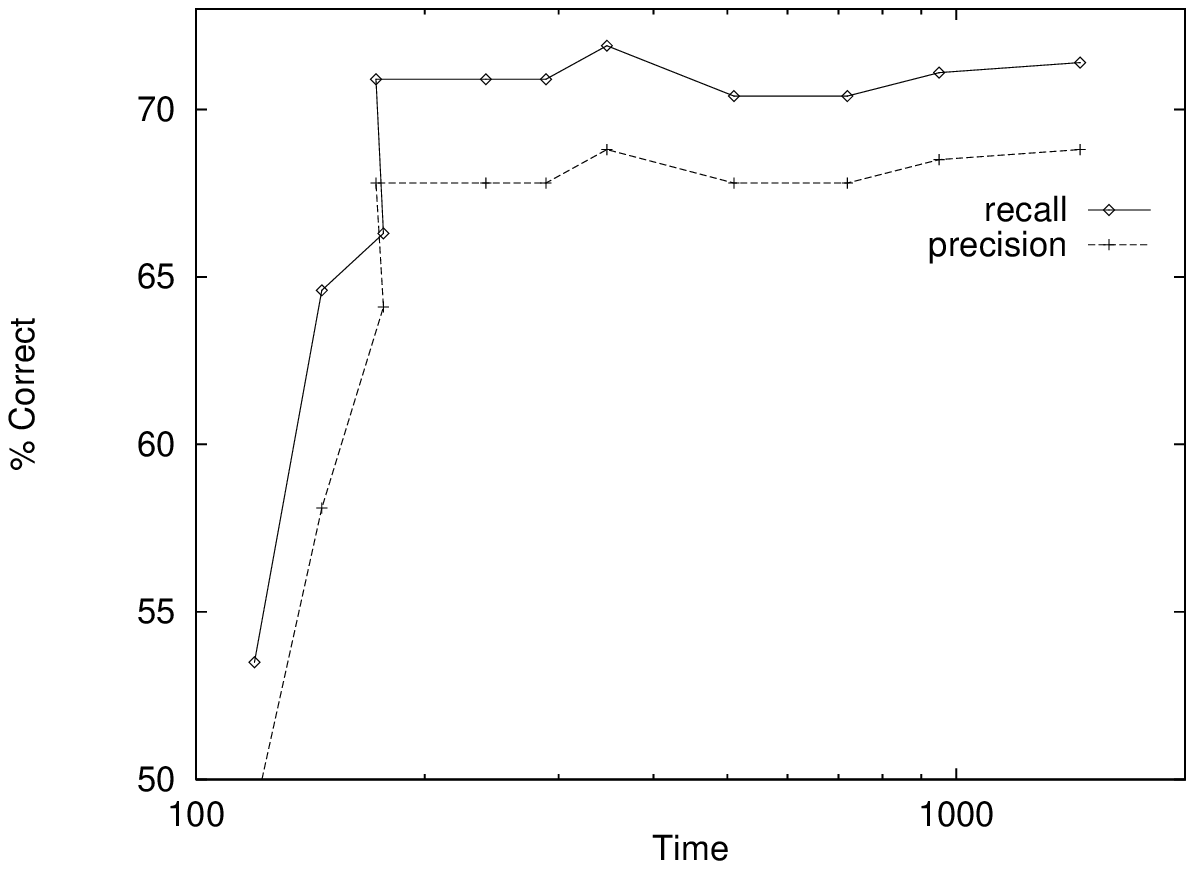,width=3.1in} &
\psfig{figure=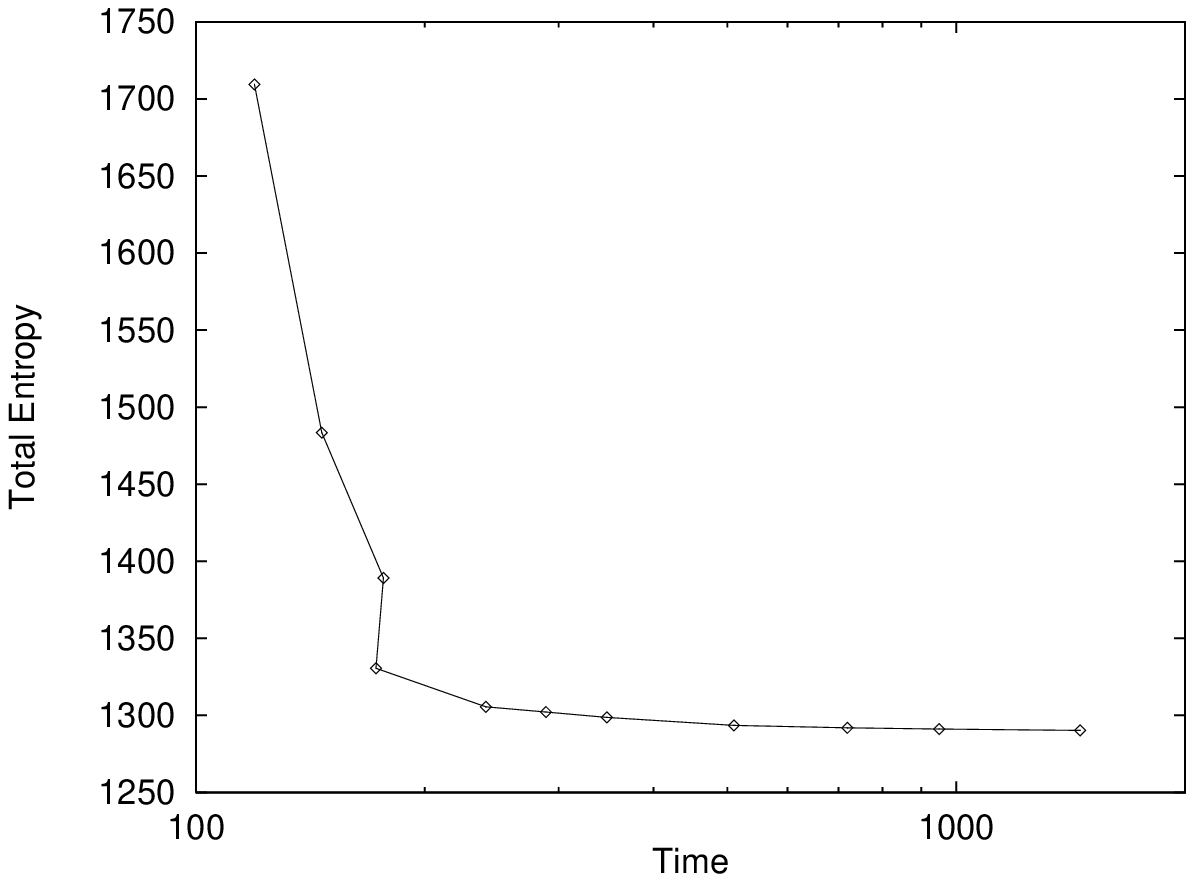,width=3.1in} \\
Precision/Recall & Total Entropy \\
200 Sentences & 200 Sentences \\
\psfig{figure=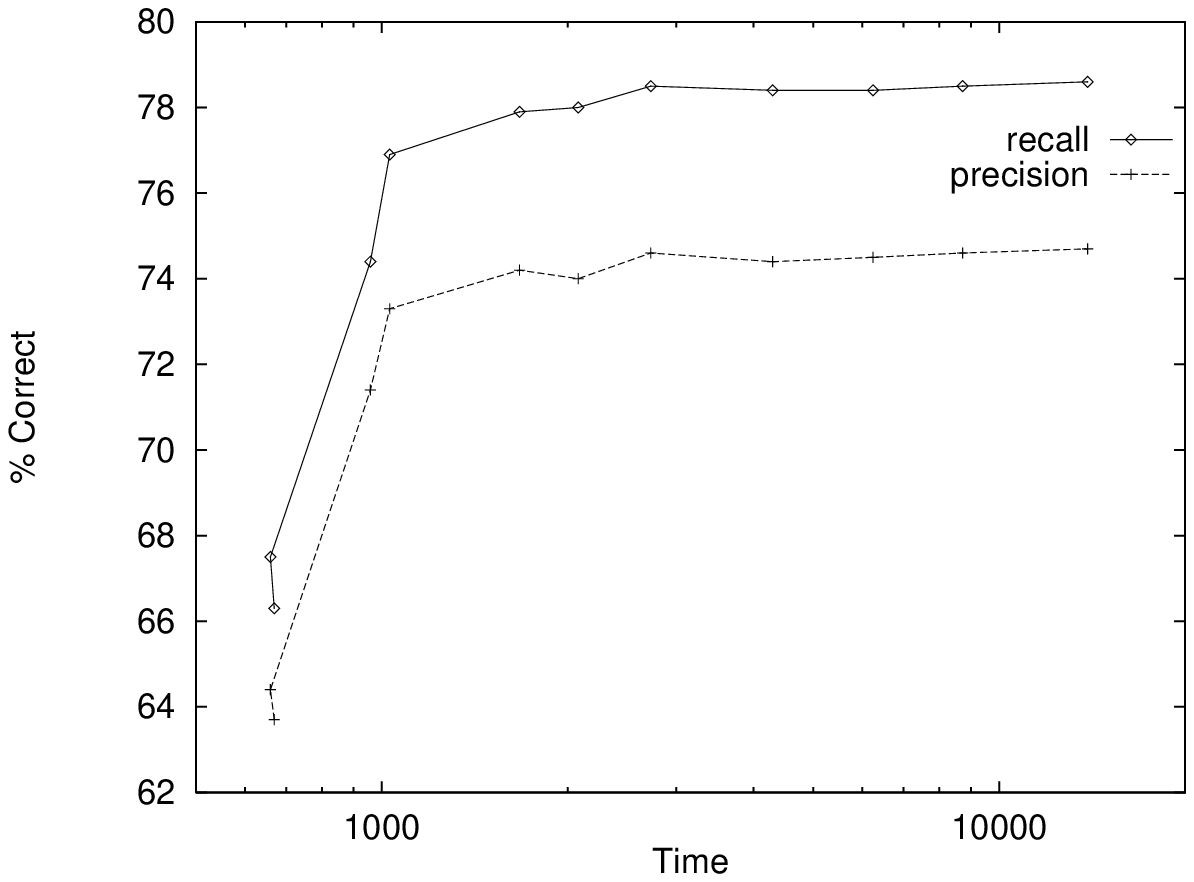,width=3.1in} &
\psfig{figure=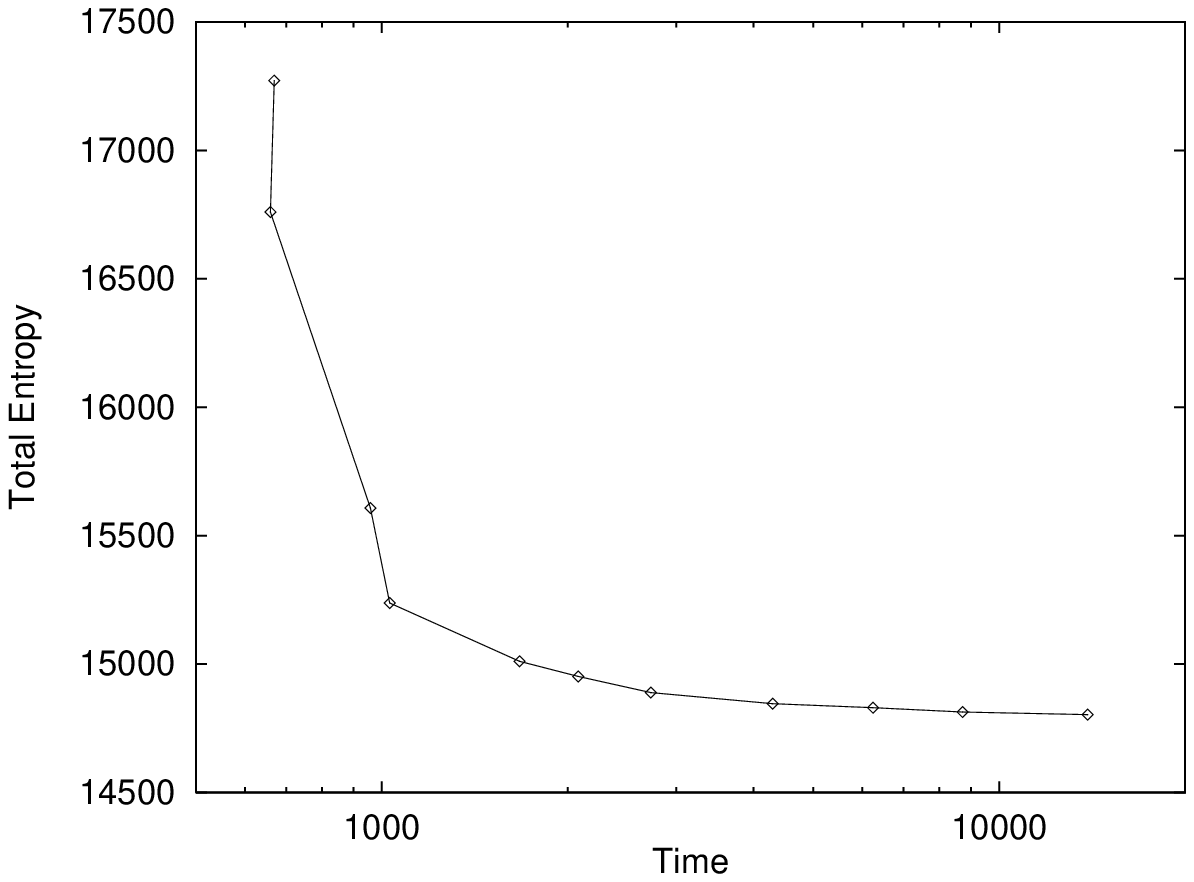,width=3.1in} \\
\end{tabular}
\caption{Smoothness for Precision and Recall versus Total Inside for
Different Test Data Sizes} \label{fig:smoothit}
\end{center}
\end{figure*}

The fact that entropy changes smoothly and monotonically is
critical for the performance of the multiple parameter optimization
algorithm.  Furthermore, we may have to run quite a few iterations of
that algorithm to get convergence, so the fact that entropy is smooth
for relatively small numbers of sentences is a large help.  Thus, the
discovery that entropy (or, equivalently, the log of the inside
probability) is a good surrogate for precision and recall is
non-trivial.  The same kinds of observations could be extended to
speech recognition to optimize multiple thresholds there (the typical
modern speech system has quite a few thresholds), a topic for future
research.

For some sentences, with too tight thresholding, the parser will fail
to find any parse at all.  We dealt with these cases by restarting the
parser with all thresholds lowered by a factor of 5, iterating this
loosening until a parse could be found.  This restarting is why for
some tight thresholds, the parser may be slower than with looser
thresholds: the sentence has to be parsed twice, once with tight
thresholds, and once with loose ones.

\begin{figure}
% Figure produced from general-out108-5
\begin{center}
\begin{tabular}{c}
\psfig{figure=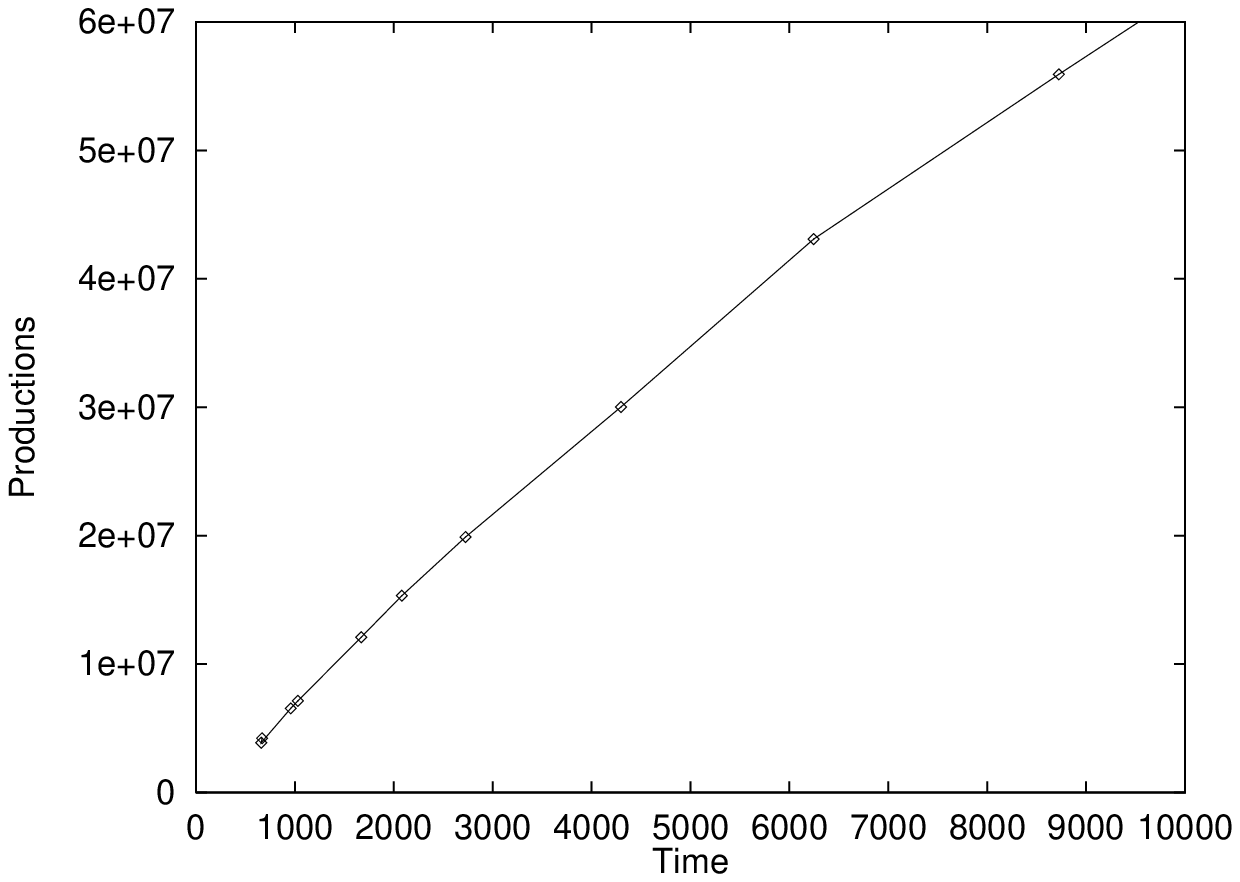,width=6in}
\end{tabular}
\end{center}
\caption{Productions versus Time} \label{fig:timevsprod}
\end{figure}

Next, we needed to choose a measure of time.  There are two obvious
measures: amount of work done by the parser, and elapsed time.  If we
measure amount of work done by the parser in terms of the number of
productions with non-zero probability examined by the parser, we have
a fairly implementation-independent, machine-independent measure of
speed.  On the other hand, because we used many different thresholding
algorithms, some with a fair amount of overhead, this measure seems
inappropriate.  Multiple-pass parsing requires use of the outside
algorithm; global thresholding uses its own dynamic programming
algorithm; and even beam thresholding has some per-node overhead.
Thus, we will give most measurements in terms of elapsed time, not
including loading the grammar and other $O(1)$ overhead.  We did want to
verify that elapsed time was a reasonable measure, so we did a beam
thresholding experiment to make sure that elapsed time and number of
productions examined were well correlated, using 200 sentences and an
exponential sweep of the thresholding parameter.  The results, shown
in Figure \ref{fig:timevsprod}, clearly indicate that time is a good
proxy for productions examined.

\subsection{Experiments in Beam Thresholding}

\label{sec:beamexp}

\begin{figure*}
\begin{tabular}{cc}
\psfig{figure=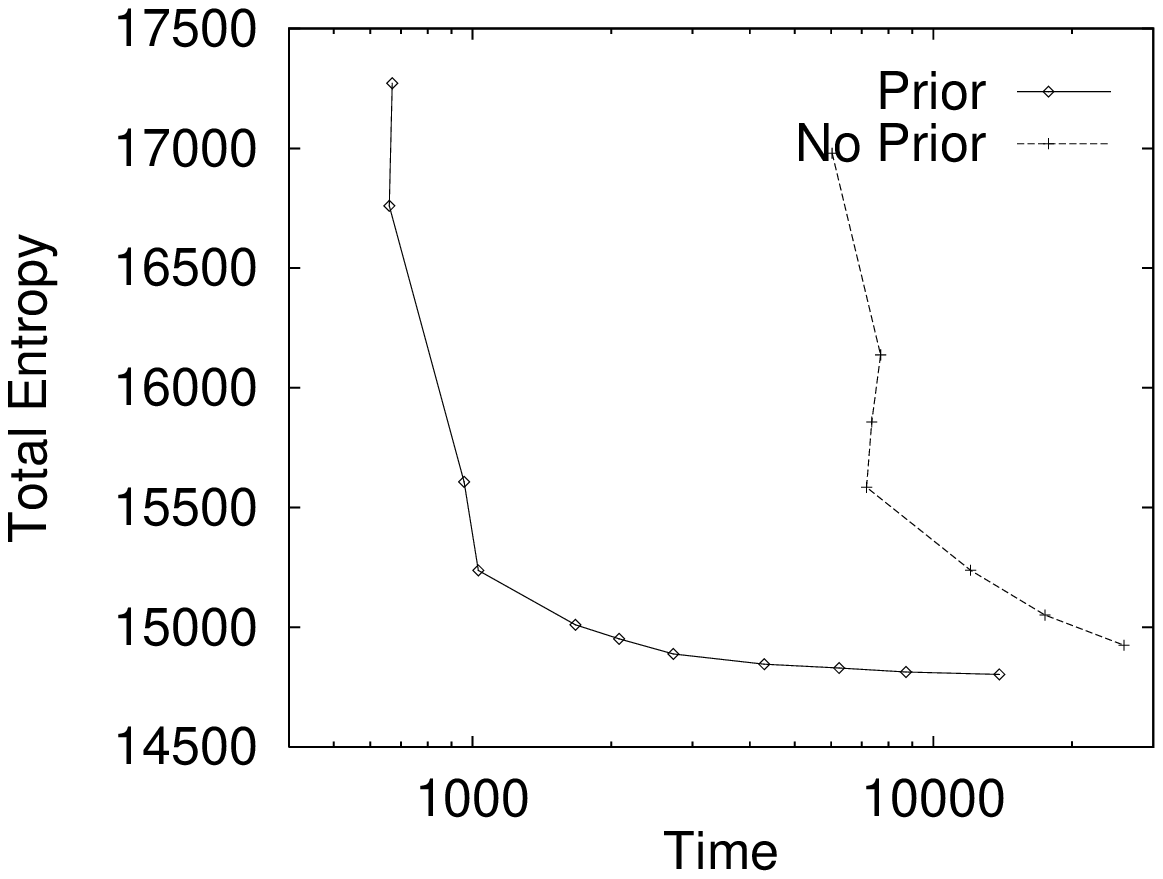,width=3.1in} &
\psfig{figure=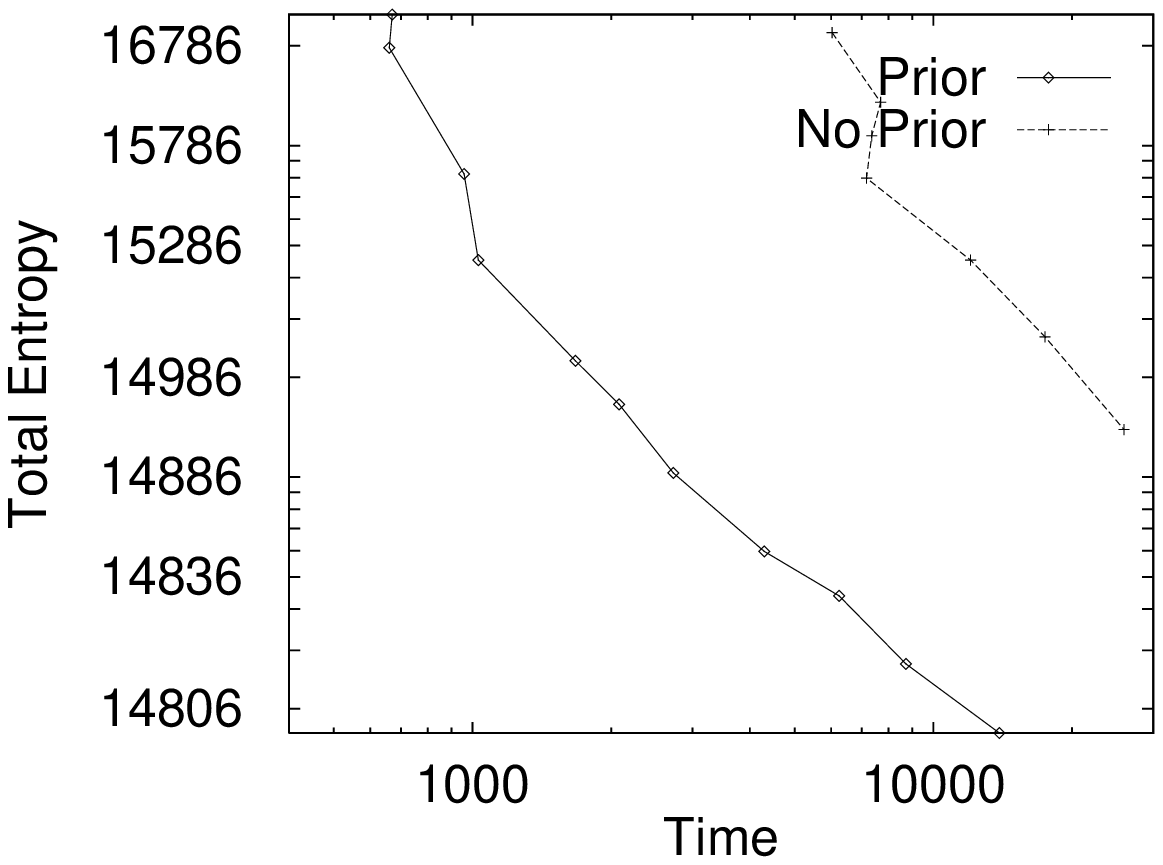,width=3.1in} \\ 
%\multicolumn{1}{l}{\hspace{.7in}$X$ axis: $\mathit{log(time)}$} &
%\multicolumn{1}{l}{\hspace{.7in}$X$ axis: $\mathit{log(time)}$} \\
%\multicolumn{1}{l}{\hspace{.7in}$Y$ axis: $\mathit{entropy}$} &
%\multicolumn{1}{l}{\hspace{.7in}$Y$ axis: $\mathit{log}(\mathit{entropy}-\mathit{asymptote})$}
\hspace{.3in}$X$ axis: $\mathit{log(time)}$ &
\hspace{.3in}$X$ axis: $\mathit{log(time)}$ \\
\hspace{.3in}$Y$ axis: $\mathit{entropy}$ &
\hspace{.3in}$Y$ axis: $\mathit{log}(\mathit{entropy}-\mathit{asymptote})$
\\
\end{tabular}
\caption{Beam Thresholding with and without the Prior Probability,
Two Different Scales}
\label{fig:priorvsno}
\end{figure*}

Our first goal was to show, at least informally, that entropy is a
good surrogate for precision and recall.  We thus tried two
experiments: one with a relatively large test set of 200 sentences,
and one with a relatively small test set of 15 sentences.  Presumably,
the 200 sentence test set should be much less noisy, and fairly
indicative of performance.  We graphed both precision and recall, and
entropy, versus time, as we swept the thresholding parameter over a
sequence of values.  The results are in Figure \ref{fig:smoothit}.  As
can be seen, entropy is significantly smoother than precision and
recall for both size test corpora.  In Section \ref{sec:descent}, we
gave a more rigorous discussion of the monotonicity of entropy versus
precision and recall with the same conclusion.

Our second goal was to check that the prior probability is indeed
helpful.  We ran two experiments, one with the prior and one without.
The results, shown in Figure \ref{fig:priorvsno}, indicate that the
prior is a critical component.  This experiment was run on 200
sentences of test data.

Notice that as the time increases, the data tends to approach an
asymptote, as shown in the left hand graph of Figure
\ref{fig:priorvsno}.  In order to make these small asymptotic changes
more clear, we wished to expand the scale towards the asymptote.  The
right hand graph was plotted with this expanded scale, based on
$\mathit{log}(\mathit{entropy}-\mathit{asymptote})$, a slight
variation on a normal log scale.  We use this scale in all the
remaining entropy graphs.  A normal logarithmic scale is used for the
time axis.  The fact that the time axis is logarithmic is especially
useful for determining how much more efficient one algorithm is than
another at a given performance level.  If one picks a performance
level on the vertical axis, then the distance between the two curves
at that level represents the ratio between their speeds.  There is
roughly a factor of 8 to 10 difference between using the prior and not
using it at all graphed performance levels, with a slow trend towards
smaller differences as the thresholds are loosened.  Because of the
large difference between using the prior and not using it, all other
beam thresholding experiments included the prior.

\subsection{Experiments in Global Thresholding}

We tried an experiment comparing global thresholding to beam
thresholding.  Figure \ref{fig:beamjob} shows the results of this
experiment, and later experiments.  In the best case, global
thresholding works twice as well as beam thresholding, in the sense
that to achieve the same level of performance requires only half as
much time, although smaller improvements were more typical.

We have found that, in general, global thresholding works better on
simpler grammars.  In the complicated grammars of Chapter \ref{ch:PFG}
there were systematic, strong correlations between nodes, which
violated the independence approximation used in global thresholding.
This prevented us from using global thresholding with these grammars.
In the future, we may modify global thresholding to model some of
these correlations.

\subsection{Experiments combining Global Thresholding and Beam Thresholding}

\begin{figure*}
\psfig{figure=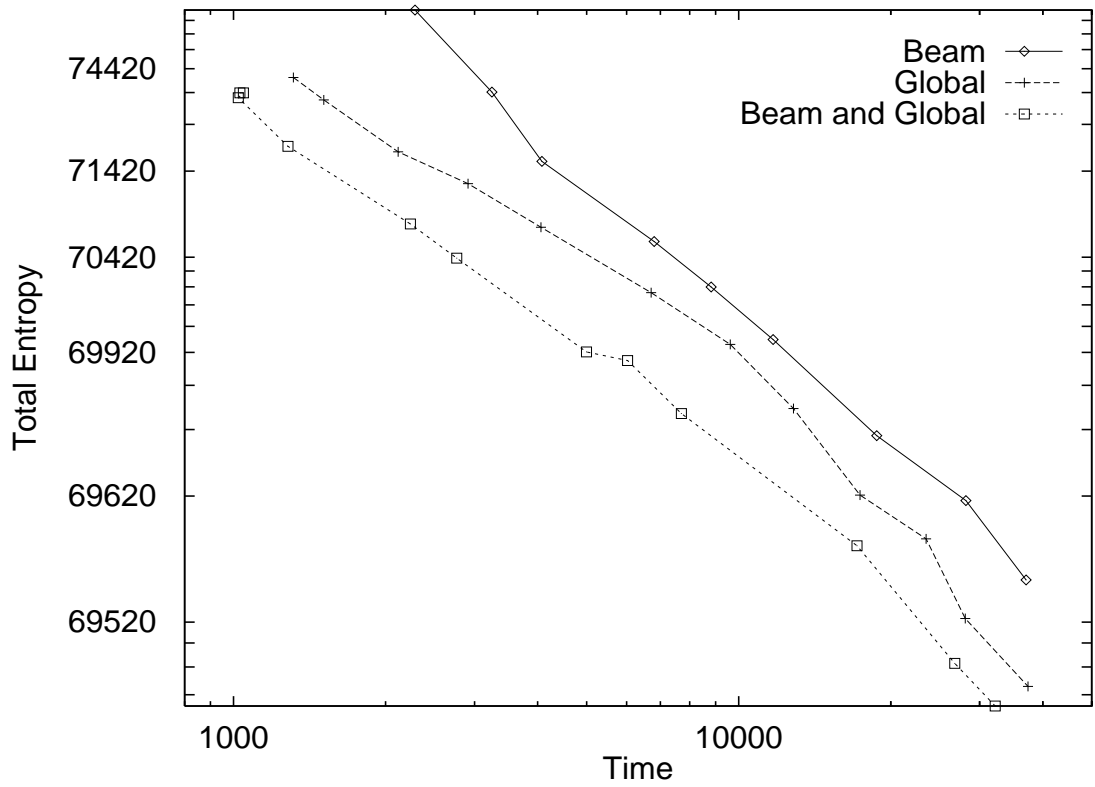,width=6in} 
\caption{Combining Beam and Global Search}
\label{fig:beamjob}
\end{figure*}

While global thresholding works better than beam thresholding in
general, each has its own strengths.  Global thresholding can
threshold across cells, but because of the approximations used, the
thresholds must generally be looser.  Beam thresholding can only
threshold within a cell, but can do so fairly tightly.  Combining the
two offers the potential to get the advantages of both.  We ran a
series of experiments using the thresholding optimization algorithm of
Section \ref{sec:descent}.  Figure \ref{fig:beamjob} gives the
results.  The combination of beam and global thresholding together is
clearly better than either alone, in some cases running 40\% faster
than global thresholding alone, while achieving the same performance
level.  The combination generally runs twice as fast as beam
thresholding alone, although up to a factor of three.

\subsection{Experiments in Multiple-Pass Parsing}

\label{sec:expmulti}

Multiple-pass parsing improves even further on our experiments
combining beam and global thresholding.  In addition to multiple-pass
parsing, we used both beam and global thresholding for both the first
and second pass in these experiments.  The first pass grammar was the
very simple terminal-prime grammar, and the second pass grammar was
the usual 6-gram grammar.

Our goal throughout this chapter has been to maximize precision and
recall, as quickly as possible.  In general, however, we have measured
entropy rather than precision and recall, because it correlates well
with those measures, but is much smoother.  While this correlation has
held for all of the previous thresholding algorithms we have tried, it
turns out not to hold in some cases for multiple-pass parsing.  In
particular, in the experiments conducted here, our first and second
pass grammars were very different from each other.  For a given parse
to be returned, it must be in the intersection of both grammars, and
reasonably likely according to both.  Since the first and second pass
grammars capture different information, parses that are likely
according to both are especially good.  The entropy of a sentence
measures its likelihood according to the second pass, but ignores the
fact that the returned parse must also be likely according to the
first pass.  Thus, in these experiments, entropy does not correlate
nearly as well with precision and recall.  We therefore give precision
and recall results in this section.  We still optimized our
thresholding parameters using the same 31 sentence held out corpus,
and minimizing entropy versus number of productions, as before.

\begin{figure*}
\begin{center}
\begin{tabular}{c}
\psfig{figure=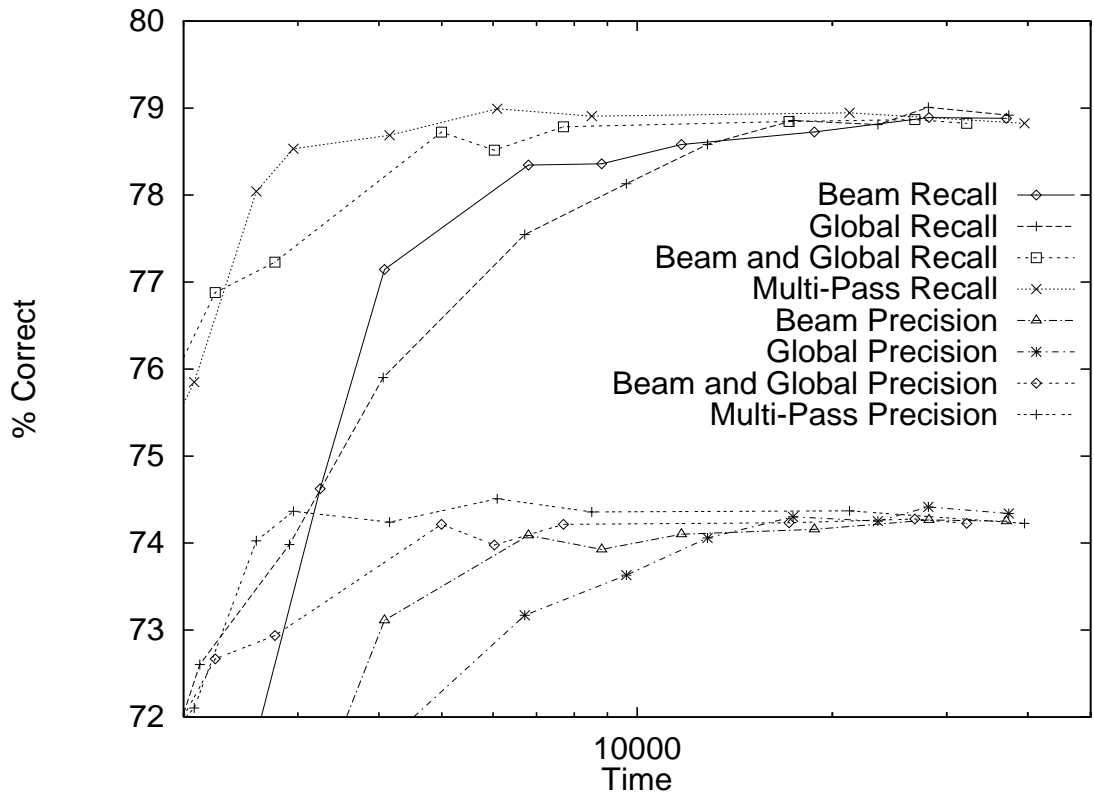,width=6in} 
\end{tabular}
\end{center}
\caption{Multiple Pass Parsing vs. Beam and Global vs. Beam}
\label{fig:multiexp}
\end{figure*}

We should note that when we used a first pass grammar that captured a
strict subset of the information in the second pass grammar, we have
found that entropy is a very good measure of performance.  As in our
earlier experiments, it tends to be well correlated with precision and
recall but less subject to noise.  It is only because of the grammar
mismatch that we have changed the evaluation.

Figure \ref{fig:multiexp} shows precision and recall curves for single
pass versus multiple pass experiments.  As in the entropy curves, we
can determine the performance ratio by looking across horizontally.
For instance, the multi-pass recognizer achieves a 74\% recall level
using 2500 seconds, while the best single pass algorithm requires
about 4500 seconds to reach that level.  Due to the noise resulting
from precision and recall measurements, it is hard to exactly quantify
the advantage from multiple pass parsing, but it is generally about
50\%.

\section{Future Work and Conclusion}

\subsection{Future Work}

In this chapter, we only considered applying multiple-pass and global
thresholding techniques to parsing probabilistic context-free
grammars.  However, just about any probabilistic grammar formalism for
which inside and outside probabilities can be computed can benefit
from these techniques.  For instance, Probabilistic Link Grammars
\cite{Lafferty:92a} could benefit from our algorithms.  We have
however had trouble using global thresholding with grammars that
strongly violated the independence assumptions of global thresholding.

One especially interesting possibility is to apply multiple-pass
techniques to formalisms that require greater than $O(n^3)$ parsing
time, such as Stochastic Bracketing Transduction Grammar (SBTG)
\cite{Wu:96a} and Stochastic Tree Adjoining Grammars (STAG)
\cite{Resnik:92a,Schabes:92a}.  SBTG is a context-free-like formalism
designed for translation from one language to another; it uses a four
dimensional chart to index spans in both the source and target
language simultaneously.  It would be interesting to try speeding up
an SBTG parser by running an $O(n^3)$ first pass on the source
language alone, and using this to prune parsing of the full SBTG.

The STAG formalism is a mildly context-sensitive formalism, requiring
$O(n^6)$ time to parse.  Most STAG productions in practical grammars
are actually context-free.  The traditional way to speed up STAG
parsing is to use the context-free subset of an STAG to form a
Stochastic Tree Insertion Grammar (STIG) \cite{Schabes:94a}, an
$O(n^3)$ formalism, but this method has problems, because the STIG
undergenerates since it is missing some elementary trees.  A different
approach would be to use multiple-pass parsing.  We could first find a
context-free covering grammar for the STAG, and use this as a first
pass, and then use the full STAG for the second pass.

There is also future work that could be done on further improvements
to thresholding algorithms.  One potential improvement that should be
tried is modifying beam thresholding with the prior, or global
thresholding, so that each compares only nonterminals which are
similar in some way.  Both of these algorithms make certain
approximations.  The more similar two nonterminals are, the smaller
the relative error from these approximations will be.  Thus, comparing
only similar nonterminals will allow tighter thresholding.  Obviously,
these modified algorithms should be combined with the original
algorithms, using the multiple parameter search technique.

\subsection{Conclusions}

The grammars described here are fairly simple, presented for purposes
of explication, and to keep our experiments simple enough to
replicate easily.  In Chapter \ref{ch:PFG}, we use significantly more
complicated grammars, Probabilistic Feature Grammars (PFGs).  For some
PFGs, the improvements from multiple-pass parsing are even more
dramatic: single pass experiments are simply too slow to run at all.

We have also found the automatic thresholding parameter optimization
algorithm to be very useful.  Before writing the parameter
optimization algorithm, we had developed a complicated PFG grammar and
the multiple-pass parsing technique and ran a series of experiments
using hand optimized parameters.  We thereafter ran the optimization
algorithm and reran the experiments, achieving a factor of two speedup
with no performance loss.  While we had not spent a great deal of time
hand optimizing these parameters, we are very encouraged by the
optimization algorithm's practical utility.

This chapter introduces four new techniques: beam thresholding with
priors, global thresholding, multiple-pass parsing, and automatic
search for the parameters of combined algorithms.  Beam thresholding
with priors can lead to almost an order of magnitude improvement over
beam thresholding without priors.  Global thresholding can be up to
two times as efficient as the new beam thresholding technique,
although the typical improvement is closer to 50\%.  When global
thresholding and beam thresholding are combined, they are usually two
to three times as fast as beam thresholding alone.  Multiple-pass
parsing can lead to up to an additional 50\% improvement with the
grammars in this chapter.  We expect the parameter optimization
algorithm to be broadly useful.

% -*- mode: latex; -*-

\chapter{Probabilistic Feature Grammars} \label{ch:PFG}

This chapter introduces Probabilistic Feature Grammars (PFGs), a
relatively simple and elegant formalism that is the first
state-of-the-art formalism for which the inside and outside
probabilities can be computed \cite{Goodman:97b}.  Because we can
compute the inside and outside probabilities, we can use the efficient
thresholding algorithms of Chapter \ref{ch:thresh} when parsing PFGs.

\section{Introduction}

Recently, many researchers have worked on statistical parsing
techniques which try to capture additional context beyond that of
simple probabilistic context-free grammars (PCFGs), including work by
\newcite{Magerman:95a}, Charniak
\shortcite{Charniak:96a,Charniak:97a}, Collins
\shortcite{Collins:96a,Collins:97a}, Black \etal \shortcite{Black:92a},
\newcite{Eisele:94a} and \newcite{Brew:95a}.  Each researcher has
tried to capture the hierarchical nature of language, as typified by
context-free grammars, and to then augment this with additional
context sensitivity based on various {\em features} of the input.
However, none of these works combines the most important benefits of
all the others, and most lack a certain elegance.  We have therefore
tried to synthesize these works into a new formalism, {\em
probabilistic feature grammar} (PFG).  PFGs have several important
properties.  First, PFGs can condition on features beyond the
nonterminal of each node, including features such as the head word or
grammatical number of a constituent.  Also, PFGs can be parsed using
efficient polynomial-time dynamic programming algorithms, and learned
quickly from a treebank.  Finally, unlike most other formalisms, PFGs
are potentially useful for language modeling or as one part of an
integrated statistical system (e.g. Miller \smalletal, 1996)
\nocite{Miller:96a} or for use with algorithms requiring outside
probabilities.  Empirical results are encouraging: our best parser is
comparable to those of \newcite{Magerman:95a} and
\newcite{Collins:96a} when run on the same data.  When we run using
part-of-speech (POS) tags alone as input, we perform significantly
better than comparable parsers.

\section{Motivation}

\label{sec:motivation}

PFG can be regarded in several different ways: as a way to make
history-based grammars \cite{Magerman:95a} more context-free, and thus
amenable to dynamic programming; as a way to generalize the work of
\newcite{Black:92b}; as a way to turn Collins' parser
\cite{Collins:96a} into a generative probabilistic language model; or
as an extension of language-modeling techniques to stochastic
grammars.  The resulting formalism is relatively simple and elegant.
In Section \ref{sec:pfgcomparison}, we will compare PFGs to each of
the systems from which it derives, and show how it integrates their
best properties.

\newcommand{\NT}[1]{\mbox{\em #1}}

Consider the following simple parse tree for the sentence ``The man
dies'':
\begin{center}
\leaf{\NT{the}}
\leaf{\NT{man}}
\branch{2}{\NT{NP}}
\leaf{\NT{dies}}
\branch{1}{\NT{VP}}
\branch{2}{\NT{S}}
\tree
\end{center}
\newlength{\Xheight}
\settoheight{\Xheight}{X}
\newlength{\singularwidth}
\settowidth{\singularwidth}{singular}
\newcommand{\Xstrut}{\rule{0pt}{\Xheight}}
\newcommand{\featuretrip}[3]{\mbox{
\renewcommand{\arraystretch}{0.7}
\begin{tabular}{|c|}\hline
\Xstrut \makebox[\singularwidth]{\NT{#1}} \\ \Xstrut \NT{#2}\\ \Xstrut \NT{#3} \\ 
\hline \end{tabular}}}
\begin{figure}
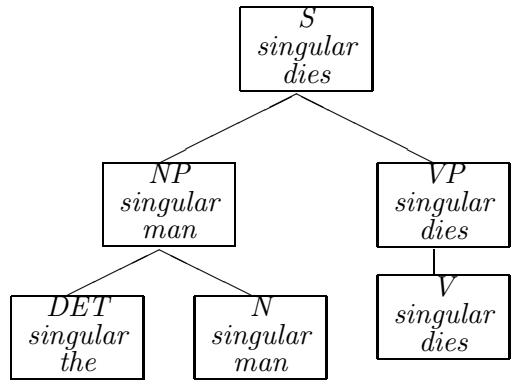

\begin{center}
\leaf{\featuretrip{DET}{singular}{the}}
\leaf{\featuretrip{N}{singular}{man}}
\branch{2}{\featuretrip{NP}{singular}{man}}
\leaf{\featuretrip{V}{singular}{dies}}
\branch{1}{\featuretrip{VP}{singular}{dies}}
\branch{2}{\featuretrip{S}{singular}{dies}}
\tree
\end{center}
\caption{Example tree with features}\label{fig:examplefeaturetree}
\end{figure}
While this tree captures the simple fact that sentences are composed
of noun phrases and verb phrases, it fails to capture other important
restrictions.  For instance, the \NT{NP} and \NT{VP} must have the same number,
both singular, or both plural.  Also, a man is far more likely to die
than spaghetti, and this constrains the head words of the
corresponding phrases.  This additional information can be captured in
a parse tree that has been augmented with features, such as the
category, number, and head word of each constituent, as is
traditionally done in many feature-based formalisms, such as HPSG,
LFG, and others.  Figure \ref{fig:examplefeaturetree} shows a parse
tree that has been augmented with these features.

While a normal PCFG has productions such as 
$$
\NT{S} \rightarrow \NT{NP}\;\NT{VP}
$$
we will write these augmented productions as, for instance,
$$
(\NT{S}, \NT{singular}, \NT{dies}) \rightarrow 
(\NT{NP}, \NT{singular}, \NT{man}) 
(\NT{VP}, \NT{singular}, \NT{dies})
$$
In a traditional probabilistic context-free grammar, we could augment
the first tree with probabilities in a simple fashion.  We estimate
the probability of $\NT{S}\rightarrow\NT{NP}\;\NT{VP}$ using a tree
bank to determine $\displaystyle
\frac{C(\NT{S}\rightarrow\NT{NP}\;\NT{VP})}{C(\NT{S})}$, the number of
occurrences of $\NT{S}\rightarrow\NT{NP}\;\NT{VP}$ divided by the
number of occurrences of \NT{S}.  For a reasonably large treebank,
probabilities estimated in this way would be reliable enough to be
useful \cite{Charniak:96a}.  On the other hand, it is not unlikely
that we would never have seen any counts at all of
$$
\frac{C((\NT{S}, \NT{singular}, \NT{dies}) \rightarrow 
(\NT{NP}, \NT{singular}, \NT{man}) 
(\NT{VP}, \NT{singular}, \NT{dies}))}
{C((\NT{S}, \NT{singular}, \NT{dies}))}
$$
which is the estimated probability of the corresponding production in
our grammar augmented with features.

The introduction of features for number and head word has created a
data sparsity problem.  Fortunately, the data-sparsity problem is well
known in the language-modeling community, and we can use their
techniques, $n$-gram models and smoothing, to help us.  Consider the
probability of a five word sentence, $w_1 ... w_5$:
$$
P(w_1 w_2 w_3 w_4 w_5) = 
P(w_1) \times P(w_2|w_1) \times P(w_3| w_1 w_2) \times P(w_4|w_1 w_2
w_3) \times P(w_5|w_1 w_2 w_3 w_4) 
$$
While exactly computing $ P(w_5|w_1 w_2 w_3 w_4) $ is difficult, a
good approximation is $
P(w_5|w_1 w_2 w_3 w_4) \approx  P(w_5| w_3 w_4) $.

Let $C(w_3 w_4 w_5)$ represent the number of occurrences of the
sequence $w_3 w_4 w_5$ in a corpus.  We can then empirically
approximate $P(w_5 |w_3 w_4)$ with $\frac{C(w_3 w_4 w_5)}{C(w_4
w_5)}$.  However, this approximation alone is not enough; there
may still be many three word combinations that do not occur in the
corpus, but that should not be assigned zero probabilities.  So we
{\it smooth} this approximation, for instance by using
$$
P(w_5 | w_3 w_4) \approx 
\lambda_1 \frac{C(w_3 w_4 w_5)} {C(w_3 w_4)} \; + \;
(1-\lambda_1) \left( \lambda_2 \frac{C(w_4 w_5)} {C(w_4)} \; + \;
                     (1- \lambda_2) \frac{C(w_5)} {\sum_w C(w)} \right)
$$
where $\lambda_1, \lambda_2$ are numbers between 0 and 1 that
determine the amount of smoothing.

Now, we can use these same approximations in PFGs.  Let us assume that
our PFG is binary branching and has $g$ features, numbered $1...g$; we
will call the parent features $a_i$, the left child features $b_i$,
and the right child features $c_i$.  In our earlier example, $a_1$
represented the parent nonterminal category; $a_2$ represented the
parent number (singular or plural); $a_3$ represented the parent head
word; $b_1$ represented the left child category; etc.  We can write a
PFG production as $(a_1, a_2, ..., a_g) \rightarrow (b_1, b_2, ...,
b_g) (c_1, c_2, ..., c_g)$.  If we think of the set of features for a
constituent $A$ as being the random variables $A_1,...,A_g$, then the
probability of a production is the conditional probability
$$
P(B_1=b_1,...,B_g=b_g, C_1=c_1, ..., C_g=c_g | A_1=a_1, ..., A_g = a_g) 
$$
We write $a_i$ as shorthand for $A_i=a_i$, and $a_1^k$ to represent
$A_1=a_1,...,A_k=a_k$.  We can then write this conditional probability
as
$$
P(b_1^g, c_1^g|a_1^g)
$$
This joint probability can be factored as the product of a set of
conditional probabilities in many ways.  One simple way is to
arbitrarily order the features as $b_1,...,b_g,c_1,...,c_g$.  We then
condition each feature on the parent features and all features earlier
in the sequence.
$$
P(b_1^g,c_1^g|a_1^g) = P(b_1|a_1^g) \times P(b_2|a_1^g, b_1^1)
\times P(b_3|a_1^g, b_1^2) \times \cdots \times
P(c_g|a_1^g,b_1^g,c_1^{g-1})
$$

\newcommand{\probpair}[2]{{
\begin{tabular}{c}{#1}\\{$\approx$}\\{#2}\end{tabular}}}

\begin{figure}
\begin{tabular}{cc}
\leaf{\featuretrip{}{}{}}
\leaf{\featuretrip{}{}{}}
\branch{2}{\featuretrip{NP}{singular}{man}}
\tree& 
\\
$\downarrow$ &  
\\
\leaf{\featuretrip{DET}{}{}}
\leaf{\featuretrip{}{}{}}
\branch{2}{\featuretrip{NP}{singular}{man}}
\tree &
\probpair{$P(B_1$=\NT{DET}$\vert A_1=$\NT{NP}, $A_2$=\NT{singular}, $A_3$=\NT{man}$)$}
{$P(B_1$=\NT{DET}$\vert A_1=$\NT{NP}, $A_2$=\NT{singular}$)$ }
\\
$\downarrow$ &  
\\
\leaf{\featuretrip{DET}{singular}{}}
\leaf{\featuretrip{}{}{}}
\branch{2}{\featuretrip{NP}{singular}{man}}
\tree &
\probpair{$P(B_2$=\NT{singular}$\vert A_1=$\NT{NP}, $A_2$=\NT{singular}, $A_3$=\NT{man}, $B_1$=\NT{DET}$)$}
         {$P(B_2$=\NT{singular}$\vert A_2$=\NT{singular}, $B_1$=\NT{DET}$)$}
\\
$\downarrow$ & 
\\
\leaf{\featuretrip{DET}{singular}{the}}
\leaf{\featuretrip{}{}{}}
\branch{2}{\featuretrip{NP}{singular}{man}}
\tree &
\probpair{$P(B_3$=\NT{the}$\vert A_1=$\NT{NP}, $A_2$=\NT{singular},..., $B_2$=\NT{singular}$)$}
         {$P(B_3$=\NT{the}$\vert B_1=$\NT{DET}, $A_3$=\NT{man}$)$ }
\\
\vdots & 
\\
\leaf{\featuretrip{DET}{singular}{the}}
\leaf{\featuretrip{N}{singular}{man}}
\branch{2}{\featuretrip{NP}{singular}{man}}
\tree &
\probpair{$P(C_3$=\NT{man}$\vert A_1=$\NT{NP}, $A_2$=\NT{singular},..., $C_2$=\NT{singular}$)$}
         {$P(C_3$=\NT{man}$\vert A_3=$\NT{man}, $A_1$=\NT{NP}$)$}
\end{tabular}
\caption{Producing {\em the man}, one feature at a time} \label{fig:themanone}
\end{figure}

We can now approximate the various terms in the factorization by
making independence assumptions.  For instance, returning to the concrete example above,
consider feature $c_1$, the right child nonterminal or terminal
category.  The following approximation should work fairly well in
practice:
$$
P(c_1|a_1^g b_1^g) \approx P(c_1 | a_1, b_1)
$$
That is, the category of the right child is well determined by the
category of the parent and the category of the left child.  Just as
$n$-gram models approximate conditional lexical probabilities by
assuming independence of words that are sufficiently distant, here we
approximate conditional feature probabilities by assuming independence
of features that are sufficiently unrelated.  Furthermore, we can use
the same kinds of backing-off techniques that are used in smoothing
traditional language models to allow us to condition on relatively
large contexts.  In practice, a grammarian determines the order of the
features in the factorization, and then for each feature, which
features it depends on and the optimal order of backoff, possibly
using experiments on development test data for feedback.  It might be
possible to determine the factorization order, the best independence
assumptions, and the optimal order of backoff automatically, a subject
of future research.

Intuitively, in a PFG, features are produced one at a time.  This
order corresponds to the order of the factorization.  The probability
of a feature being produced depends on a subset of the features in a
local context of that feature.  Figure \ref{fig:themanone} shows an
example of this feature-at-a-time generation for the noun phrase ``the
man.''  In this example, the grammarian picked the simple feature
ordering $b_1, b_2, b_3, c_1, c_2, c_3$.  To the right of the figure,
the independence assumptions made by the grammarian are shown.

\section{Formalism}

In a PCFG, the important concepts are the terminals and nonterminals,
the productions involving these, and the corresponding probabilities.
In a PFG, a vector of features corresponds to the terminals and
nonterminals.  PCFG productions correspond to PFG {\em events} of the
form $(a_1,...,a_g) \rightarrow (b_1,...,b_g)(c_1,...,c_g)$, and our
PFG rule probabilities correspond to products of conditional
probabilities, one for each feature that needs to be generated.

\subsection{Events and EventProbs}

There are two kinds of PFG events of immediate interest.  The
first is a {\em binary event}, in which a feature set $a_1^g$
(the parent features) generates features $b_1^gc_1^g$ (the child
features).  Figure \ref{fig:themanone} is an example of a single such
event.  Binary events generate the whole tree
except the start node, which is generated by a {\em start event}.

\newcommand{\calK}{{\mathcal{K}}}
\newcommand{\barN}{\overline{N}}
\newcommand{\calE}[1]{{{\mathcal E}_{#1}}}
\newcommand{\barF}{\overline{F}}

The probability of an event is given by an {\em EventProb}, which
generates each new feature in turn, assigning it a
conditional probability given all the known features.  For instance,
in a binary event, the EventProb assigns probabilities to each of the
child features, given the parent features and any child features that
have already been generated.

\newcommand{\seq}[1]{\langle #1 \rangle}

Formally, an EventProb $\calE{}$ is a 3-tuple $\seq{\calK, \barN,
\barF }$, where $\calK$ is the set of conditioning features (the Known
features), $\barN = N_1, N_2, ..., N_n$ is an ordered list of
conditioned features (the New features), and $\barF = f_1, f_2, ...,
f_n$ is a parallel list of functions.  Each function $f_i(n_i,
k_1,...,k_k,n_1,n_2,...,n_{i-1})$ returns $P(N_i = n_i \vert K_1 = k_1,
...K_k = k_k, N_1=n_1, N_2=n_2, ..., N_{i-1}=n_{i-1})$, the
probability that feature $N_i = n_i$ given all the known features and
all the lower indexed new features.

For a binary event, we may have $\calE{B} = \seq{ \{a_1, a_2,
...,a_g\}, \seq{b_1,...,b_g,c_1,...,c_g}, \barF_{B} }$; that is, the
child features are conditioned on the parent features and earlier
child features.  For a start event we have $\calE{S} = \seq{ \{\},
\seq{a_1, a_2, ..., a_g}, \barF_{S} }$; i.e. the parent features are
conditioned only on each other in sequence.

\subsection{Terminal Function, Binary PFG, Alternating PFG}

We need one last element: a function $T$ from a set of $g$ features to
$\seq{T, N}$ which tells us whether a part of an event is terminal or
nonterminal: the {\em terminal function}.  A Binary PFG is then a
quadruple $\seq{g, \calE{B}, \calE{S}, T}$: a number of features, a binary
EventProb, a start EventProb, and a terminal function.

Of course, using binary events allows us to model $n$-ary branching
grammars for any fixed $n$: we simply add additional features for
terminals to be generated in the future, as well as a feature for
whether or not this intermediate node is a ``dummy'' node (the
continuation feature).  We demonstrate how to do this in detail in
Section \ref{sec:features}.

\label{sec:unary}

On the other hand, it does not allow us to handle unary branching
productions.  In general, probabilistic grammars that allow an
unbounded number of unary branches are very difficult to deal with
\cite{Stolcke:93a}.  There are a number of ways we could have handled
unary branches.  The one we chose was to enforce an alternation
between unary and binary branches, marking most unary branches as
``dummies'' with the continuation feature, and removing them before
printing the output of the parser.

To handle these unary branches, we add one more EventProb, $\calE{U}$.
Thus, an {\em Alternating PFG} is a quintuple of $\{g, \calE{B},
\calE{S}, \calE{U}, T\}$.

It is important that we allow only a limited number of unary branches.
As we discussed in Chapter \ref{ch:semi}, unlimited unary branches
lead to infinite sums.  For conventional PCFG-style grammar
formalisms, these infinite sums can be computed using matrix
inversion, which is still fairly time-consuming.  For a formalism such
as ours, or a similar formalism, the effective number of nonterminals
needed in the matrix inversion is potentially huge, making such
computations impractical.  Thus, instead we simply limit the number of
unary branches, meaning that the sum is finite, for computing both the
inside and the outside values.  Two competing formalisms, those of
\newcite{Collins:97a} and \newcite{Charniak:97a} allow unlimited unary
branches, but because of this, can only compute Viterbi probabilities,
not inside and outside probabilities.

\section{Comparison to Previous Work}
\label{sec:pfgcomparison}

PFG bears much in common with previous work, but in each case has
at least some advantages over previous formalisms.

Some other models \cite{Charniak:96a,Brew:95a,Collins:96a,Black:92a}
use probability approximations that do not sum to 1, meaning that they
should not be used either for language modeling, e.g. in a speech
recognition system, or as part of an integrated model such as that of
\newcite{Miller:96a}.  Some models \cite{Magerman:95a,Collins:96a}
assign probabilities to parse trees conditioned on the strings, so
that an unlikely sentence with a single parse might get probability 1,
making these systems unusable for language modeling.  PFGs use joint
probabilities, so can be used both for language modeling and as part of
an integrated model.

Furthermore, unlike all but one of the comparable systems
\cite{Black:92b}, PFGs can compute outside probabilities, which
are useful for grammar induction, as well as for the parsing
algorithms of Chapter \ref{ch:max}, and the thresholding algorithms of
Chapter \ref{ch:thresh}.

{\it Bigram Lexical Dependency Parsing.}  \newcite{Collins:96a}
introduced a parser with extremely good performance.  From this
parser, we take many of the particular conditioning features that we
will use in PFGs.  As noted, this model cannot be used for language
modeling.  There are also some inelegancies in the need for a separate
model for Base-NPs, and the treatment of punctuation as inherently
different from words.  The model also contains a non-statistical rule
about the placement of commas.  Finally, Collins' model uses memory
proportional to the sum of the squares of each training sentence's
length.  PFGs in general use memory that is only linear.

{\it Generative Lexicalized Parsing.}  \newcite{Collins:97a} worked
independently from us to construct a model that is very similar to
ours.  In particular, Collins wished to adapt his previous parser
\cite{Collins:96a} to a generative model.  In this he succeeded.
However, while we present a fairly simple and elegant formalism, which
captures all information as features, Collins uses a variety of
different techniques.  First, he uses variables, which are analogous
to our features.  Next, both our models need a way to determine when
to stop generating child nodes; like everything else, we encode this
in a feature, but Collins creates a special STOP nonterminal.  For
some information, Collins modifies the names of nonterminals, rather
than encoding the information as additional features.  Finally, all
information in our model is generated top-down.  In Collins' model,
most information is generated top-down, but distance information is
propagated bottom-up.  Thus, while PFGs encode all information as
top-down features, Collins' model uses several different techniques.
This lack of homogeneity fails to show the underlying structure of the
model, and the ways it could be expanded.  While Collins' model could
not be encoded exactly as a PFG, a PFG that was extremely similar
could be created.

Furthermore, our model of generation is very general.  While our
implementation captures head words through the particular choice of
features, Collins' model explicitly generates first the head phrase,
then the right children, and finally the left children.  Thus, our
model can be used to capture a wider variety of grammatical theories,
simply by changing the choice of features.

{\it Simple PCFGs.}  \newcite{Charniak:96a} showed that a simple PCFG formalism in which
the rules are simply ``read off'' of a treebank can perform very
competitively.  Furthermore, he showed that a simple modification, in
which productions at the right side of the sentence have their
probability boosted to encourage right branching structures, can
improve performance even further.  PFGs are a superset of PCFGs, so we
can easily model the basic PCFG grammar used by Charniak, although the
boosting cannot be exactly duplicated.  However, we can use more
principled techniques, such as a feature that captures whether a
particular constituent is at the end of the sentence, and a feature
for the length of the constituent.  Charniak's boosting strategy means
that the scores of constituents are no longer probabilities, meaning
that they cannot be used with the inside-outside algorithm.
Furthermore, the PFG feature-based technique is not extra-grammatical,
meaning that no additional machinery needs to be added for parsing or
grammar induction.

{\it PCFG with Word Statistics.} \newcite{Charniak:97a} uses a grammar
formalism which is in many ways similar to the PFG model, with several
minor differences, and one important one.  The main difference is that
while we binarize trees, and encode rules as features about which
nonterminal should be generated next, Charniak explicitly uses rules,
in the style of traditional PCFG parsing, in combination with other
features.  This difference is discussed in more detail in Section
\ref{sec:features}.

{\it Stochastic HPSG.}  \newcite{Brew:95a} introduced a
stochastic version of HPSG.  In his formalism, in some cases even if
two features have been constrained to the same value by unification,
the probabilities of their productions are assumed independent.  The
resulting probability distribution is then normalized so that
probabilities sum to one.  This leads to problems with grammar
induction pointed out by \newcite{Abney:96a}.  Our formalism, in
contrast, explicitly models dependencies to the extent possible given
data sparsity constraints.

{\it IBM Language Modeling Group.}  Researchers in the IBM
Language Modeling Group developed a series of successively more
complicated models to integrate statistics with features.

The first model \cite{Black:93a,Black:92a} essentially tries to
convert a unification grammar to a PCFG, by instantiating the values
of the features.  Because of data sparsity, however, not all features can
be instantiated.  Instead, they create a grammar where many features
have been instantiated, and many have not; they call these partially
instantiated features sets {\em mnemonics}.  They then create a PCFG
using the mnemonics as terminals and nonterminals.  Features
instantiated in a particular mnemonic are generated probabilistically,
while the rest are generated through unification.  Because no
smoothing is done, and because features are grouped, data sparsity
limits the number of features that can be generated probabilistically,
whereas because we generate features one at a time and smooth, we are
far less limited in the number of features we can use.  Their
technique of generating some features probabilistically, and the rest
by unification, is somewhat inelegant; also, for the probabilities to
sum to one, it requires an additional step of normalization, which
they appear not to have implemented.

In their next model \cite{Black:92b}, which
strongly influenced our model, five attributes are associated with
each nonterminal: a syntactic category, a semantic category, a rule,
and two lexical heads.  The rules in this grammar are the same as the
mnemonic rules used in the previous work, developed by a grammarian.
These five attributes are generated one at a time, with backoff
smoothing, conditioned on the parent attributes and earlier
attributes.  Our generation model is essentially the same as this.
Notice that in this model, unlike ours, there are two kinds of
features: those features captured in the mnemonics, and the five
categories; the categories and mnemonic features are modeled very
differently.  Also, notice that a great deal of work is required by a
grammarian, to develop the rules and mnemonics.

The third model \cite{Magerman:94a}, extends the second model to capture
more dependencies, and to remove the use of a grammarian.  Each
decision in this model can in principal depend on any previous
decision and on any word in the sentence.  Because of these
potentially unbounded dependencies, there is no dynamic programming
algorithm: without pruning, the time complexity of the model is
exponential.  One motivation for PFG was to capture similar
information to this third model, while allowing dynamic programming.
This third model uses a more complicated probability model: all
probabilities are determined using decision trees; it is an area for
future research to determine whether we can improve our performance by
using decision trees.

{\it Probabilistic LR Parsing with Unification Grammars.}
Briscoe and Carroll describe a formalism
\cite{Briscoe:93a,Carroll:92a} similar in many ways to the first IBM
model.  In particular, a context-free covering grammar of a
unification grammar is constructed.  Some features are captured by the
covering grammar, while others are modeled only through unifications.
Only simple plus-one-style smoothing is done, so data sparsity is
still significant.  The most important difference between the work of
\newcite{Briscoe:93a} and that of \newcite{Black:93a} is that Briscoe
\etal associate probabilities with the (augmented) transition matrix
of an LR Parse table; this gives them more context sensitivity than
Black \etal  However, the basic problems of the two approaches are
the same: data sparsity; difficulty normalizing probabilities; and
lack of elegance due to the union of two very different approaches.

\section{Parsing}
\label{sec:pfgparsing}
\begin{figure}
\begin{tabbing}
\verb|   |\=\verb|   |\=\verb|   |\=\verb|   |\=\verb|   |\=\verb|   |\=\verb|   |\=\verb|   |\=\verb|   |\=\verb|   |\= \kill
\alforeach length $l$, shortest to longest \\
 \> \alforeach start $s$ \\
 \>  \> \alforeach split length $t$ \\
 \>  \>  \> \alforeach $b_1^g$ s.t. $\mi{chart}[s, s+t, b_1^g] \neq 0$\\
 \>  \>  \>  \> \alforeach $c_1^g$ s.t. $\mi{chart}[s+t, s+l, c_1^g] \neq 0$\\
 \>  \>  \>  \>  \> \alforeach $a_1$ consistent with $b_1^gc_1^g$\\
 \>  \>  \>  \>  \>  \>  \> $\vdots$ \\
 \>  \>  \>  \>  \>  \> \alforeach $a_g$ consistent with $b_1^gc_1^ga_1^{g-1}$\\
 \>  \>  \>  \>  \>  \>  \> $\mi{chart}[s, s+l, a_1^g] +\!= \calE{B}(a_1^g\rightarrow b_1^gc_1^g)$\\
\alreturn ${\displaystyle \sum_{a_1^g}} \calE{S}(a_1^g) \times \mi{chart}[1, n+1, a_1^g])$
\end{tabbing}
\caption{PFG Inside Algorithm}\label{fig:PFGalg}
\end{figure}

The parsing algorithm we use is a simple variation on probabilistic
versions of the CKY algorithm for PCFGs, using feature vectors instead
of nonterminals \cite{Baker:79b,Lari:90a}.  The parser computes inside
probabilities (the sum of probabilities of all parses, i.e. the
probability of the sentence) and Viterbi probabilities (the
probability of the best parse), and, optionally, outside
probabilities.  In Figure \ref{fig:PFGalg} we give the inside
algorithm for PFGs.  Notice that the algorithm requires time $O(n^3)$
in sentence length, but is potentially exponential in the number of
features, since there is one loop for each parent feature, $a_1$
through $a_g$.

When parsing a PCFG, it is a simple matter to find for every right and
left child what the possible parents are.  On the other hand, for a
PFG, there are some subtleties.  We must loop over every possible
value for each feature.  At first, this sounds overwhelming, since it
requires guessing a huge number of feature sets, leading to a run time
exponential in the number of features.  In practice, most values of
most features will have zero probabilities, and we can avoid
considering these; only a small number of values will be consistent
with previously determined features.  For instance, features such as the length of a
constituent take a single value per cell.  Many other features take on
very few values, given the children.  For example, we arrange our
parse trees so that the head word of each constituent is dominated by
one of its two children.  This means that we need consider only two
values for this feature for each pair of children.  The single most
time consuming feature is the Name feature, which corresponds to the
terminals and non-terminals of a PCFG.  For efficiency, we keep a list
of the parent/left-child/right-child name triples that have non-zero
probabilities, allowing us to hypothesize only the possible values for
this feature given the children.  Careful choice of features helps
keep parse times reasonable.

\subsection{Pruning}

We use two pruning methods to speed parsing.  The first is beam
thresholding with the prior, as described in Section
\ref{sec:beamthresh}.  Within each cell in the parse chart, we
multiply each entry's inside probability by the prior probability of
the parent features of that entry, using a special EventProb,
$\calE{P}$.  We then remove those entries whose combined probability
is too much lower than the best entry of the cell.

The other technique we use is multiple-pass parsing, described in
Section \ref{sec:multithresh}.  Recall that in multiple-pass parsing,
we use a simple, fast grammar for the first pass, which approximates
the later pass.  We then remove any events whose combined
inside-outside product is too low: essentially those events that are
unlikely given the complete sentence.  The technique is particularly
natural for PFGs, since for the first pass, we can simply use a
grammar with a superset of the features from the previous pass.  The
features we used in our first pass were Name, Continuation, and two
new features especially suitable for a fast first pass, the length of
the constituent and the terminal symbol following the constituent.
Since these two features are uniquely determined by the chart cell of
the constituent, they work particularly well in a first pass, since
they provide useful information without increasing the number of
elements in the chart.  However, when used in our second pass, these
features did not help performance, presumably because they captured
information similar to that captured by other features.  Multiple-pass
techniques have dramatically sped up PFG parsing.

\section{Experimental Results}
\label{sec:featureresults}

The PFG formalism is an extremely general one that has the capability
to model a wide variety of phenomena, and there are very many possible
sets of features that could be used in a given implementation.  We
will, on an example set of features, show that the formalism can be
used to achieve a high level of accuracy.

\subsection{Features}
\label{sec:features}

\newcommand{\widestar}{\makebox[.86em]{$\star$}}

In this section, we will describe the actual features used by our
parser.  The most interesting and most complicated features are those
used to encode the rules of the grammar, the child features.  We will
first show how to encode a PCFG as a PFG.  The PCFG has some maximum
length right hand side, say five symbols.  We would then create a PFG
with six features.  The first feature would be $N$, the nonterminal
symbol.  Since PFGs are binarized, while PCFGs are n-ary branching, we
will need a feature that allows us to recover the n-ary branching
structure from a binary branching tree.  This feature, which we
call the continuation feature, C, will be 0 for constituents that are
the top of a rule, and 1 for constituents that are used internally.
The next 5 features
%,$F^1...F^5$, 
describe the
future children of the nonterminal.  We will use the symbol $\star$ to
denote an empty child.  To encode a PCFG rule such as $A \rightarrow
BCDEF$, we would assign the following probabilities to the features
sets:
$$
\begin{array}{r@{}c@{}c@{}rcl}
\calE{B}((A,0,B,C,D,E,F) & \rightarrow & (B,0,X_1,...,X_5) & (A, 1, C, D, E, F,
\widestar)) &=& P(B \rightarrow X_1...X_5) \\
\calE{B}((A,1,C,D,E,F,\widestar) & \rightarrow & (C,0,X_1,...,X_5) & (A, 1, D, E, F,
\widestar, \widestar)) &=& P(C \rightarrow X_1...X_5) \\
\calE{B}((A,1,D,E,F,\widestar,\widestar) & \rightarrow & (D,0,X_1,...,X_5) & (A, 1, E, F, \widestar,
\widestar, \widestar)) &=& P(D \rightarrow X_1...X_5) \\
\calE{B}((A,1,E,F,\widestar,\widestar,\widestar) & \rightarrow & (E,0,X_1,...,X_5) & (F, 0,
Y_1,...,Y_5)) &=& P(E \rightarrow X_1...X_5) \times \\
&&&&&P(F \rightarrow Y_1...Y_5) \\
\end{array}
$$
It should be clear that we can define probabilities of individual
features in such as way as to get the desired probabilities for the
events.
%
%For instance, to get that 
%$$
%\calE{B}((A,0,B,C,D,E,F) \rightarrow (B,0,X_1,...,X_5) (A, 1, C, D, E, F,
%\widestar)) &=& P(B \rightarrow X_1...X_5) \\
%$$
%we set 
%$$
%\begin{array}{rcl}
%P(N_L = F^1_P)
%\end{array}
%$$
%
We also need to define a distribution over the start symbols:
$$
\begin{array}{rl}
\calE{S}((S,0,A,B,C,D,E) \rightarrow (A,0,X_1,...,X_5) (S, 1, B, C, D, E,
\widestar)) = & P(S \rightarrow ABCDE) \times \\
& P(A \rightarrow X_1...X_5)
\end{array}
$$
A quick inspection will show that this assignment of probabilities to
events leads to the same probabilities being assigned to analogous
trees in the PFG as in the original PCFG.

Now, imagine if rather than using the raw probability estimates from
the PCFG, we were to smooth the probabilities in the same way we
smooth all of our feature probabilities.  This will lead to a kind of
smoothed 5-gram model on right hand sides.  Presumably, there is some
assignment of smoothing parameters that will lead to performance which
is at least as good, if not better, than unsmoothed probabilities.
Furthermore, what if we had many other features in our PFG, such as
head words and distance features.  With more features, we might have
too much data sparsity to make it really worthwhile to keep all five
of the children as features.  Instead, we could keep, say, the first
two children as features.  After each child is generated as a left
nonterminal, we could generate the next child (the third, fourth,
fifth, etc.) as the Child2 feature of the right child.  Our new
grammar probabilities might look like:
$$
\begin{array}{r@{}c@{}c@{}rcl}
\calE{B}((A,0,B,C) & \rightarrow & (B,0,X_1,X_2) & (A, 1, C, D) &=& P(B \rightarrow X_1X_2...) \\
\calE{B}((A,1,C,D) & \rightarrow & (C,0,X_1,X_2) & (A, 1, D, E)) &=& P(C \rightarrow X_1X_2...) \\
\calE{B}((A,1,D,E) & \rightarrow & (D,0,X_1,X_2) & (A, 1, E, F)) &=& P(D \rightarrow X_1X_2...) \\
\calE{B}((A,1,E,F) & \rightarrow & (E,0,X_1,X_2) & (F, 0, Y_1,Y_2) &=& P(E \rightarrow X_1X_2...) \times P(F \rightarrow Y_1Y_2...) \\
\end{array}
$$
where $P(B \rightarrow X_1X_2...) = \sum_{\alpha}P(B \rightarrow
X_1X_2\alpha)$.  This assignment will produce a good approximation to
the original PCFG, using a kind of trigram model on right hand sides,
with fewer parameters and fewer features than the exact PFG.  Fewer
child features will be very helpful when we add other features to the
model.  We will show in Section \ref{sec:pfgcontribution} that with
the set of features we use, 2 children is optimal.

\newcommand{\featureten}[9]{\mbox{
\renewcommand{\arraystretch}{0.7}
\begin{tabular}{|rc|}\hline
\Xstrut N & \makebox[\singularwidth]{\NT{#1}} \\ \Xstrut C & \NT{#2} \\ 
\Xstrut 1 & \NT{#3} \\ \Xstrut 2 & \NT{#4} \\ \Xstrut H & \NT{#5} \\ 
\Xstrut P & \NT{#6} \\ \Xstrut W & \NT{#7} \\ 
\Xstrut D$^L$ & \NT{#8} \\ \Xstrut D$^R$ & 000 \\ \Xstrut D$^B$ & \NT{#9} \\ 
\hline \end{tabular}}}
\newcommand{\featureterm}[3]{\featureten{#1}{0}{}{}{#1}{#1}{#2}{#3}{000}}

\begin{table}
\begin{description}
\item[{\makebox[3em]{N}} Name] Corresponds to the terminals and nonterminals of a PCFG. 
\item[{\makebox[3em]{C}} Continuation] Tells whether we are generating 
modifiers to the right or the left, and whether it is time to generate
the head node.
\item[{\makebox[3em]{1}} Child1] Name of first child to be generated. 
\item[{\makebox[3em]{2}} Child2] Name of second child to be generated.
In combination with Child1, this allows us to simulate a second order
Markov process on nonterminal sequences.
\item[{\makebox[3em]{H}} Head name] Name of the head category. 
\item[{\makebox[3em]{P}} Head pos] Part of speech of head word. 
\item[{\makebox[3em]{W}} Head word] Actual head word.  Not used in POS only model. 
\item[{\makebox[3em]{D$^L$}} $\Delta$ left] Count of punctuation, verbs, words to left of head. 
\item[{\makebox[3em]{D$^R$}} $\Delta$ right] Counts to right of head. 
\item[{\makebox[3em]{D$^B$}} $\Delta$ between] Counts between parent's and child's heads. 
\end{description}
\caption{Features Used in Experiments}\label{fig:features}
\end{table}

We can now describe the features actually used in our experiments.  We
used two PFGs, one that used the head word feature, and one otherwise
identical grammar with no word based features, only POS tags.  The
grammars had the features shown in Table \ref{fig:features}.  A sample
parse tree with these features is given in Figure
\ref{fig:featuretree}.

Recall that in Section \ref{sec:pfgparsing} we mentioned that in order
to get good performance, we made our parse trees binary branching in
such a way that every constituent dominates its head word.  To achieve
this, rather than generating children strictly left to right, we first
generate all children to the left of the head word, left to right,
then we generate all children to the right of the head word, right to
left, and finally we generate the child containing the head word.  It
is because of this that the root node of the example tree in Figure
\ref{fig:featuretree} has $N\!P$ as its first child, and $V\!P$ as its
second, rather than the other way around.

The feature D$^L$ is a 3-tuple, indicating the number of
punctuation characters, verbs, and words to the left of a
constituent's head word.  To avoid data sparsity, we do not count
higher than 2 punctuation characters, 1 verb, or 4 words.  So, a value
of D$^L = 014$ indicates that a constituent has no punctuation, at
least one verb, and 4 or more words to the left of its head word.
D$^R$ is a similar feature for the right side.  Finally, D$^B$ gives
the numbers between the constituent's head word, and the head word of
its other child.  Words, verbs, and
punctuation are counted as being to the left of themselves: that is, a
terminal verb has one verb and one word on its left.

\begin{figure}
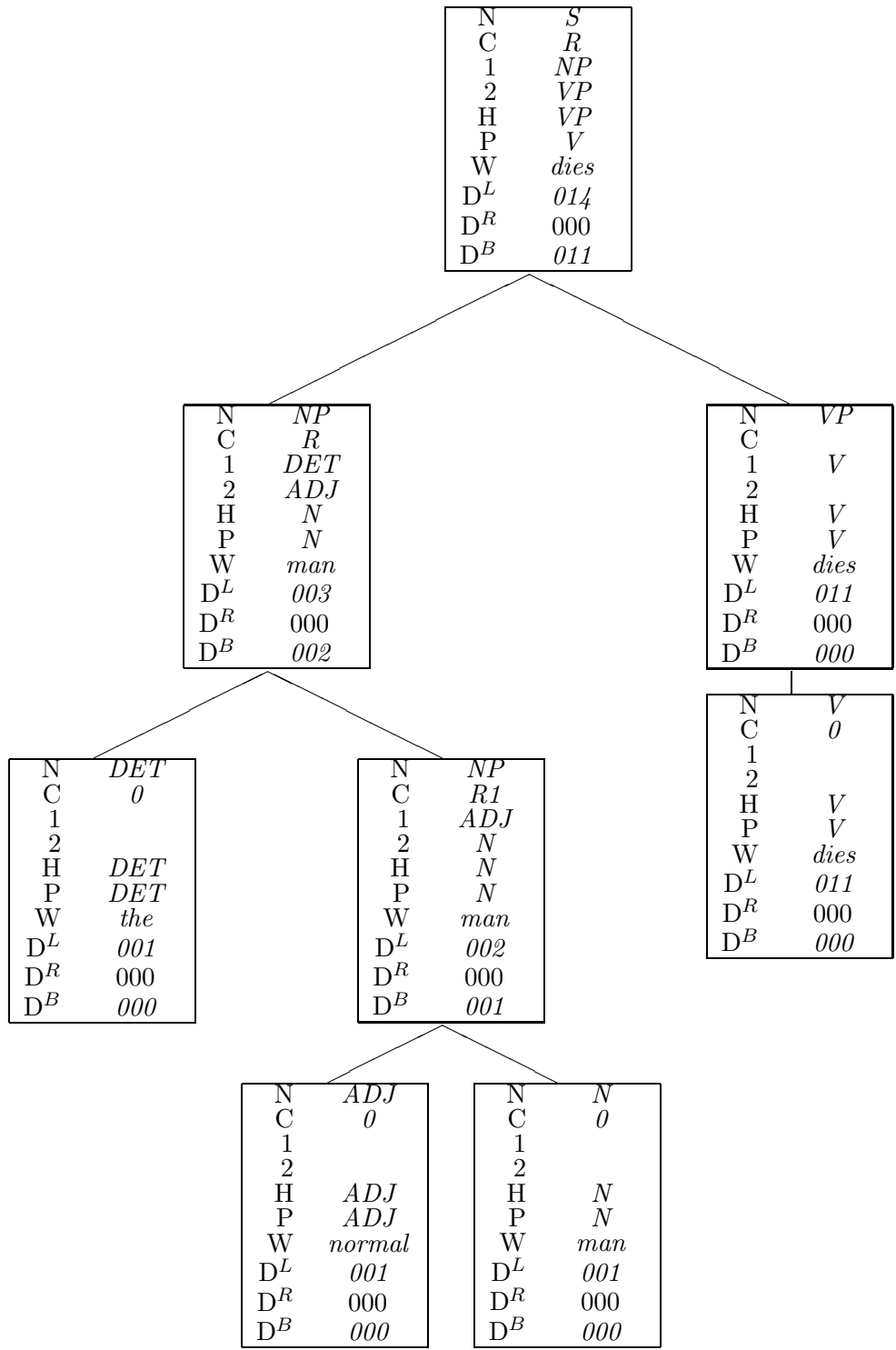

\begin{center}
\leaf{\featureterm{DET}{the}{001}}
\leaf{\featureterm{ADJ}{normal}{001}}
\leaf{\featureterm{N}{man}{001}}
\branch{2}{\featureten{NP}{R1}{ADJ}{N}{N}{N}{man}{002}{001}}
\branch{2}{\featureten{NP}{R}{DET}{ADJ}{N}{N}{man}{003}{002}}
\leaf{\featureterm{V}{dies}{011}}
\branch{1}{\featureten{VP}{}{V}{}{V}{V}{dies}{011}{000}}
\branch{2}{\featureten{S}{R}{NP}{VP}{VP}{V}{dies}{014}{011}}
\tree
\end{center}
\caption{Example tree with features: The normal man dies} \label{fig:featuretree}
\end{figure}

Notice that the continuation of the lower \NT{NP} in Figure
\ref{fig:featuretree} is R1, indicating that it
inherits its child from the right, with the 1 indicating that it is a
``dummy'' node to be left out of the final tree.

\subsection{Experimental Details}

The probability functions we used were similar to those of Section
\ref{sec:motivation}, but with three important differences.  The first
is that in some cases, we can compute a probability exactly.  For
instance, if we know that the head word of a parent is ``man'' and
that the parent got its head word from its right child, then we know
that with probability 1, the head word of the right child is ``man.''
In cases where we can compute a probability exactly, we do so.
Second, we smoothed slightly differently.  In particular, when
smoothing a probability estimate of the form
$$
p(a|bc) \approx \lambda \frac{C(abc)}{C(bc)} + (1-\lambda)p(a|b) 
$$
we set $\lambda = \frac{C(bc)}{k+C(bc)}$, using a separate $k$ for
each probability distribution.  Finally, we did additional smoothing
for words, adding counts for the unknown word.

The actual tables that show for each feature the order of backoff for
that feature are given in Appendix \ref{sec:backofftables}.  In this
section, we simply discuss a single example, the order of backoff for
the $2_R$ feature, the category of the second child of the right
feature set.  The most relevant features are first: $N_R, C_R, H_R,
1_R, N_P, C_P, N_L, C_L, P_R, W_R $.  We back off from the head word
feature first, because, although relevant, this feature creates
significant data sparsity.  Notice that in general, features from the
same feature set, in this case the right features, are kept longest;
parent features are next most relevant; and sibling features are
considered least relevant.  We leave out entirely features that are
unlikely to be relevant and that would cause data sparsity, such as
$W_L$, the head word of the left sibling.

\subsection{Results}

\begin{table}
\begin{center}
{
\begin{tabular}{|l||c|c|c|c|c|} \hline
Model & Labeled  & Labeled   & Crossing & 0 Crossing  & $\leq2$ Crossing \\
      & Recall   & Precision & Brackets & Brackets    & Brackets \\
\hline \hline
PFG Words        & 84.8\% & 85.3\% & 1.21 & 57.6\% & 81.4\% \\
PFG POS only     & 81.0\% & 82.2\% & 1.47 & 49.8\% & 77.7\% \\

\hline 
Collins 97 best     & 88.1\% & 88.6\% & 0.91 & 66.5\% & 86.9\%\\
Collins 96 best     & 85.8\% & 86.3\% & 1.14 & 59.9\% & 83.6\% \\
Collins 96 POS only & 76.1\% & 76.6\% & 2.26 &        &        \\
Magerman          & 84.6\% & 84.9\% & 1.26 & 56.6\% & 81.4\% \\
\hline
\end{tabular}
}
\end{center}
\caption{PFG experimental results}\label{fig:PFGresults}
\end{table}

We used the same machine-labeled data as Collins
\shortcite{Collins:96a,Collins:97a}: TreeBank II sections 2-21 for
training, section 23 for test, section 00 for development, using all
sentences of 40 words or less.\footnote{We are grateful to Michael
Collins and Adwait Ratnaparkhi for supplying us with the
part-of-speech tags.}  We also used the same scoring method
(replicating even a minor bug for the sake of comparison; see the
footnote on page \pageref{note:Collinsbug}.).

Table \ref{fig:PFGresults} gives the results.  Our results are the
best we know of from POS tags alone, and, with the head word feature,
fall between the results of Collins and Magerman, as given by
\newcite{Collins:97a}.  To take one measure as an example, we got 1.47
crossing brackets per sentence with POS tags alone, versus Collins'
results of 2.26 in similar conditions.  With the head word feature, we
got 1.21 crossing brackets, which is between Collins' .91 and
Magerman's 1.26.  The results are similar for other measures of
performance.

\subsection{Contribution of Individual Features}
\label{sec:pfgcontribution}
\begin{table}
\begin{center}
\begin{tabular}{|l|l@{}l@{}l@{}l@{}l@{}l@{}l@{}l@{}l@{}l@{}l||c|c|c|c|c|c|}
\hline
Name & \multicolumn{11}{|c||}{Features} & Label  & Label   & Cross & 0 Cross  & $\leq2$ Cross & Time \\
&&&&&&&&&&&               & Recall   & Prec & Brack & Brack    & Brack & \\
\hline \hline
Base    &H&N&C&W&P&D$^B$&D$^L$&D$^R$&1&2& & 86.4\% & 87.2\% & 1.06 & 60.8\% & 84.2\% & 40619 \\
NoW     &H&N&C& &P&D$^B$&D$^L$&D$^R$&1&2& & 82.7\% & 84.1\% & 1.36 & 52.2\% & 79.3\% & 39142 \\
NoP     &H&N&C&W& &D$^B$&D$^L$&D$^R$&1&2& & 85.8\% & 86.5\% & 1.18 & 57.5\% & 81.3\% & 46226 \\
NoD$^B$ &H&N&C&W&P&     &D$^L$&D$^R$&1&2& & 86.0\% & 84.9\% & 1.24 & 56.6\% & 81.9\% & 49834 \\
NoD     &H&N&C&W&P&     &     &     &1&2& & 85.9\% & 85.9\% & 1.23 & 59.4\% & 81.0\% & 45375 \\
No2     &H&N&C&W&P&D$^B$&D$^L$&D$^R$&1& & & 85.2\% & 86.5\% & 1.17 & 58.7\% & 81.9\% & 47977 \\
No12    &H&N&C&W&P&D$^B$&D$^L$&D$^R$& & & & 76.6\% & 79.3\% & 1.65 & 45.8\% & 74.9\% & 37912 \\
\hline											    
BaseH   & &N&C&W&P&D$^B$&D$^L$&D$^R$&1&2& & 86.7\% & 87.7\% & 1.01 & 61.6\% & 85.1\% & 52785 \\
NoWH    & &N&C& &P&D$^B$&D$^L$&D$^R$&1&2& & 82.7\% & 84.0\% & 1.38 & 51.9\% & 79.5\% & 40080 \\
NoPH    & &N&C&W& &D$^B$&D$^L$&D$^R$&1&2& & 86.1\% & 87.2\% & 1.08 & 59.5\% & 83.1\% & 38502 \\
NoD$^B$H& &N&C&W&P&     &D$^L$&D$^R$&1&2& & 86.2\% & 85.2\% & 1.19 & 57.2\% & 82.5\% & 41415 \\
NoDH    & &N&C&W&P&     &     &     &1&2& & 86.4\% & 86.4\% & 1.17 & 59.9\% & 81.8\% & 39387 \\
No2H    & &N&C&W&P&D$^B$&D$^L$&D$^R$&1& & & 82.4\% & 86.7\% & 1.11 & 56.8\% & 82.8\% & 37854 \\
No12H   & &N&C&W&P&D$^B$&D$^L$&D$^R$& & & & 65.1\% & 80.0\% & 1.69 & 41.5\% & 73.5\% & 36790 \\
NoNames & & &C&W&P&D$^B$&D$^L$&D$^R$& & & &        &        & 2.41 & 41.9\% & 64.0\% & 55862 \\
NoHPlus3& &N&C&W&P&D$^B$&D$^L$&D$^R$&1&2&3& 86.9\% & 87.3\% & 1.07 & 61.7\% & 84.0\% & 41676 \\
\hline
\end{tabular}
\end{center}
\caption{Contribution of individual features} \label{fig:individualcontribution}
\end{table}

In this subsection we analyze the contribution of features, by running
experiments using the full set of features minus some individual
feature.  The difference in performance between the full set and the
full set minus some individual feature gives us an estimate of the
feature's contribution.  Note that the contribution of a feature is
relative to the other features in the full set.  For instance, if we
had features for both head word and a morphologically stemmed head
word, their individual contributions as measured in this manner would
be nearly negligible, because both are so highly correlated.  The same
effect will hold to lesser degrees for other features, depending on
how correlated they are.  Some features are not meaningful without
other features.  For instance, the child2 feature is not particularly
meaningful without the child1 feature.  Thus, we cannot simply remove
just child1 to compute its contribution; instead we first remove
child2, compute its contribution, and then remove child1, as well.

When performing these experiments, we used the same dependencies and
same order of backoff for all experiments.  This probably causes us to
overestimate the value of some features, since when we delete a
feature, we could add additional dependencies to other features that
depended on it.  We kept the thresholding parameters constant (as
optimized by the parameter search algorithm of Chapter
\ref{ch:thresh}), but adjusted the parameters of the backoff algorithm.

In the original work on PFGs, we did not perform this feature
contribution experiment, and thus included the head name (H) feature
in all of our original experiments.  When we performed these
experiments, we discovered that the head name feature had a negative
contribution.  We thus removed the head name feature, and repeated the
feature contribution experiment without this feature.  Both sets of
results are given here.  In order to minimize the number of runs on
the final test data, all of these experiments were run on the
development test section (section 00) of the data.  

Table \ref{fig:individualcontribution} shows the results of 16
experiments removing various features.  Our first test was the
baseline.  We then removed the head word feature, leading to a fairly
significant but not overwhelming drop in performance (6.8\% combined
precision and recall).  Our results without the head word are perhaps
the best reported.  The other features -- head pos, distance between,
all distance features, and second child -- all lead to modest drops in
performance (1.3\%, 2.7\%, 1.8\%, 1.9\%, respectively, on combined
precision and recall).  On the other hand, when we removed both child1
and child2, we got a 17.7\% drop.  When we removed the head name
feature, we achieved a 0.8\% improvement.  We repeated these
experiments without the head name feature.  Without the head name
feature, the contributions of the other features are very similar: the
head word feature contributed 7.7\%.  Other features -- head pos,
distance between, all distance features, and second child --
contributed 1.1\%, 3.0\%, 1.6\%, 5.3\%.  Without child1 and child2,
performance dropped 29.3\%.  This is presumably because there were now
almost no name features remaining, except for name itself.  When the
name feature was also removed, performance dropped even further (to
about 2.41 crossing brackets per sentence, versus 1.01 for the no head
name baseline -- there are no labels here, so we cannot compute
labelled precision or labelled recall for this case.)  Finally, we
tried adding in the child3 feature, to see if we had examined enough
child features.  Performance dropped negligibly: 0.2\%.

%Note: Collins 96 implicitly captures H feature in head category
%Note: Collins 97 explicitly captures H feature in conditioning

These results are significant regarding child1 and child2.  While the
other features we used are very common, the child1 and child2 features
are much less used.  Both Magerman \shortcite{Magerman:95a} and
\newcite{Ratnaparkhi:97a} make use of essentially these features, but
in a history-based formalism.  One generative formalism which comes
close to using these features is that of \newcite{Charniak:97a}, which
uses a feature for the complete rule.  This captures the child1,
child2, and other child features.  However, it does so in a crude way,
with two drawbacks.  First, it does not allow any smoothing; and
second, it probably captures too many child features.  As we have
shown, the contribution of child features plateaus at about two
children; with three children there is a slight negative impact, and more
children probably lead to larger negative contributions.  On the other
hand, Collins \shortcite{Collins:96a,Collins:97a} does not use these
features at all, although he does use a feature for the head name.
(Comparing No12 to No12H, in Table \ref{fig:individualcontribution},
we see that without Child1 and Child2, the head name makes a
significant positive contribution.)  Extrapolating from these results,
integrating the Child1 and Child2 features (and perhaps removing the
head name feature) in a state-of-the-art model such as that of
\newcite{Collins:97a} would probably lead to improvements.

The last column of Table \ref{fig:individualcontribution}, the Time
column, is especially interesting.  Notice that the runtimes are
fairly constrained, ranging from a low of about 36,000 seconds to
a high of about 55,000 seconds.  Furthermore, the longest runtime
comes with the worst performance, and the fewest features.  Better
models often allow better thresholding and faster performance.
Features such as child1 and child2 that  we might expect to
significantly slow down parsing, because of the large number of features
sets they allow, in some cases (Base to No2) actually speed
performance.  Of course, to fully substantiate these claims would
require a more detailed exploration into the tradeoff between speed
and performance for each set of features, which would be beyond the
scope of this chapter.

We reran our experiment on section 23, the final test data, this time
without the head name feature.  The results are given here, with the
original results with the head name repeated for comparison.  The
results are disappointing, leading to only a slight improvement.  The
smaller than expected improvement might be attributable to random
variation, either in the development test data, or in the final test
data, or to some systematic difference between the two sets.
Analyzing the two sets directly would invalidate any further
experiments on the final test data, so we cannot determine which of
these is the case.
\begin{center}
{
\begin{tabular}{|l||c|c|c|c|c|} \hline
Model               & Labeled  & Labeled   & Crossing & 0 Crossing  & $\leq2$ Crossing \\
                    & Recall   & Precision & Brackets & Brackets    & Brackets \\
\hline \hline
Words, head name    & 84.8\% & 85.3\% & 1.21 & 57.6\% & 81.4\% \\
Words, no head name &84.9\% & 85.3\% & 1.19 & 58.0\% & 82.0\% \\ \hline
\end{tabular}
}
\end{center}

\section{Conclusions and Future Work}

While the empirical performance of Probabilistic Feature Grammars is
very encouraging, we think there is far more potential.  First, for
grammarians wishing to integrate statistics into more conventional
models, the features of PFG are a very useful tool, corresponding to
the features of DCG, LFG, HPSG, and similar formalisms.  TreeBank II
is annotated with many semantic features, currently unused in all but
the simplest way by all systems; it should be easy to integrate these
features into a PFG.

PFG has other benefits that we would like to explore, including the
possibility of its use as a language model, for applications such as
speech recognition.  Furthermore, the dynamic programming used in the
model is amenable to efficient rescoring of lattices output by speech
recognizers.

Another benefit of PFG is that both inside and outside probabilities
can be computed, making it possible to reestimate PFG parameters.  We
would like to try experiments using PFGs and the inside/outside
algorithm to estimate parameters from unannotated text.

Because the PFG model is so general, it is amenable to many further
improvements: we would like to try a wide variety of them.  We would
like to try more sophisticated smoothing techniques, as well as more
sophisticated probability models, such as decision trees.  We would
like to try a variety of new features, including classes of words and
nonterminals, and morphologically stemmed words, and integrating the
most useful features from recent work, such as that of
\newcite{Collins:97a}.  We would also like to perform much more
detailed research into the optimal order for backoff than the few
pilot experiments used here.

While the generality and elegance of the PFG model make these and
many other experiments possible, we are also encouraged by the very
good experimental results.  Wordless model performance is excellent,
and the more recent models with words are comparable to the state of
the art.

{\beginappendix
\appendixsection{Backoff Tables}
\label{sec:backofftables} 

\begin{table}[b!]
\begin{center}
\begin{tabular}{|l|l|} 
\hline
$C_L $&$C_P N_P 2_P 1_P D^B_P D^L_P D^R_P $ \\
$C_R $&$C_P C_L N_P 2_P 1_P D^B_P D^L_P D^R_P $ \\
$N_R $&$\vert $ \\
$N_L $&$\vert $ \\
$P_L $&$C_P N_L N_P D^B_P P_P W_P $ \\
$P_R $&$C_P N_R N_P D^B_P P_P W_P $ \\
$W_L $&$P_L C_P N_L N_P D^B_P P_P W_P $ \\
$W_R $&$P_R C_P N_R N_P D^B_P P_P W_P $ \\
$H_L $&$C_L C_P \vert N_L P_L W_L N_P N_R $ \\
$H_R $&$C_R C_P \vert N_R P_R W_R N_P N_L $ \\
$1_L $&$N_L C_L H_L N_P C_P N_R C_R P_L W_L $ \\
$2_L $&$N_L C_L H_L 1_L N_P C_P N_R C_R P_L W_L $ \\
$1_R $&$N_R C_R H_R N_P C_P N_L C_L P_R W_R $ \\
$2_R $&$N_R C_R H_R 1_R N_P C_P N_L C_L P_R W_R $ \\
$D^L_L $&$C_P C_L D^L_P D^B_P N_L N_P P_L $ \\
$D^R_R $&$C_P C_R D^R_P D^B_P N_R N_P P_R $ \\
$D^R_L $&$C_P C_L C_R D^B_P N_L N_R P_L P_R H_L H_R $ \\
$D^L_R $&$C_P C_L D^B_P D^R_L N_R N_P P_R $ \\
$D^B_L $&$\vert $ \\
$D^B_R $&$\vert $ \\
\hline
\end{tabular}
\end{center}
\caption{Binary Event Backoff} \label{tab:binarybackoff}
\end{table}

Tables \ref{tab:binarybackoff}, \ref{tab:unarybackoff}, and
\ref{tab:startbackoff} give the order of backoff used in the
experiments.  For the wordless experiments, the head word feature --
$W$ -- was simply deleted.  The order of backoff is provided here for
those who would like to duplicate these results.  However, this table
was produced mostly from intuition, and with the help of only a few
pilot experiments; there is no reason to think that this is a
particularly good set of tables.

The tables require a bit of explanation.  A basic table entry was
already described in Section \ref{sec:featureresults}.  Briefly, a
table entry gives the order of backoff of features, with the most
relevant features first.  In some cases, we wished to stop backoff
beyond a certain point.  This is indicated by a vertical bar, $\vert$.
(This feature was especially useful in constructing the 6-gram grammar
of Chapter \ref{ch:thresh}, since it allowed smoothing to be easily
turned off, making those experiments more easily replicated.)  In some
cases, it was possible to determine the value of a feature from the
value of already known features.  For instance, from the parent
continuation feature, $C_P$, the parent first child feature, $1_P$,
and the parent name feature, $N_P$, we can determine the left child
name feature, $N_L$.  Thus, no backoff is necessary; this is indicated
by a single vertical bar in the entry for $N_L$.

\begin{table}
\begin{center}
\begin{tabular}{|l|l|} 
\hline
$D^L_L $&$D^L_P $ \\
$D^R_L $&$D^R_P $ \\
$C_L $&$C_P N_P 1_P 2_P D^L_L D^R_L $ \\
$N_L $&$\vert $ \\
$W_L $&$\vert $ \\
$P_L $&$\vert $ \\
$H_L $&$N_L C_L C_P N_P D^L_L D^R_L $ \\
$1_L $&$N_L C_L N_P C_P P_L W_L D^L_L D^R_L $ \\
$2_L $&$N_L C_L 1_L N_P C_P P_L W_L D^L_L D^R_L $ \\
$D^B_L $&$C_L D^L_L D^R_L N_L P_L H_L 1_L 2_L W_L $ \\
\hline
\end{tabular}
\end{center}
\caption{Unary Event Backoff} \label{tab:unarybackoff}
\end{table}

\begin{table}
\begin{center}
\begin{tabular}{|l|l|} 
\hline
\multicolumn{2}{|l|}{Start/Prior Event Backoff}\\
$N_P $&$\vert $ \\
$C_P $&$N_P $ \\
$H_P $&$N_P C_P \vert $ \\
$1_P $&$N_P C_P H_P $ \\
$2_P $&$N_P C_P 1_P H_P $ \\
$P_P $&$N_P C_P H_P 1_P 2_P $ \\
$W_P $&$P_P N_P C_P H_P 1_P 2_P $ \\
$D^B_P $&$P_P C_P N_P H_P 1_P 2_P W_P $ \\
$D^L_P $&$P_P D^B_P C_P N_P 1_P 2_P H_P W_P $ \\
$D^R_P $&$P_P D^B_P D^L_P C_P N_P 1_P 2_P H_P W_P $ \\
\hline
\end{tabular}
\end{center}
\caption{Start/Prior Event Backoff} \label{tab:startbackoff}
\end{table}

}%endappendix

%
%	$Id: conclusion.tex,v 1.3 1998/05/19 15:43:43 goodman Exp $
%

\chapter{Conclusion} \label{ch:concl}

\begin{quote}
All this means, of course, is that I didn't solve the natural language
parsing problem.  No big surprise there (although the naive graduate
student in me is a little disappointed). \emph{David Magerman}
\end{quote}
% page v of his thesis

\section{Summary}

The goal of this thesis has been to show new uses for the
inside-outside probabilities and to provide useful tools for finding
these probabilities.  In Chapter \ref{ch:semi}, we gave tools for
finding inside-outside probabilities in various formalisms.  We then
showed in Chapters \ref{ch:max}, \ref{ch:DOP} and \ref{ch:thresh} that
these probabilities have several uses beyond grammar induction,
including more accurate parsing by matching parsing algorithms to
metrics; 500 times faster parsing of the DOP model; and 30 times
faster parsing of general PCFGs through thresholding.  Finally, in
Chaper \ref{ch:PFG} we gave a state-of-the-art parsing formalism that
can compute inside and outside probabilities and that makes a good
framework for future research.

We began by presenting in Chapter \ref{ch:semi} a general framework
for parsing algorithms, semiring parsing.  Using the item-based
descriptions of semiring parsing, it is easy to derive formulas for a
wide variety of parsers, and for a wide variety of values, including
the Viterbi, inside, and outside probabilities.  As we look towards
the future of parsing, we see a movement towards new formalisms with
an increasingly lexicalized emphasis, and a movement towards parsing
algorithms that can take advantage of the efficiencies of
lexicalization \cite{Lafferty:92a}.  Whether or not this occurs, as
long as parsing technology does not remain stagnant, the theory of
semiring parsing will simplify the development of whatever new parsers
are needed.  Furthermore, given the expense of developing
treebanks, learning algorithms like the inside-outside reestimation
algorithm will probably play a key role in the development of
practical systems.  Thus, the ease with which semiring parsing allows
the inside and outside formulas to be derived will be important.

In Chapter \ref{ch:max}, we showed that the inside and outside
probabilities could be used to improve performance, by matching
parsing algorithms to metrics.  This basic idea is powerful.  While
many researchers have worked on different grammar induction techniques
and different grammar formalisms, the only commonly used parsing
algorithm was the Viterbi algorithm.  Thus, many researchers can use
the algorithms of this chapter, or appropriate variations, to get
improvements

In Chapter \ref{ch:DOP}, we sped up Data-Oriented Parsing by 500
times, and replicated the algorithms for the first time, using the
inside-outside probabilities.  While DOP had incredible reported
results, we showed that these previous results were probably
attributable to chance or to a particularly easy corpus.  Negative
results are always disappointing, but we hope to have at least helped
steer the field in a more fruitful direction.

The normalized inside-outside probabilities are ideal for
thresholding, giving the probability that any given constituent is
correct.  The only problem with using the inside-outside probabilities
for thresholding is that the outside probability is unknown until long
after it is needed.  In Chapter \ref{ch:thresh} we showed that
approximations to the inside-outside probabilities can be used to
significantly speed up parsing.  In particular, we sped up parsing by
a factor of about 30 at the same accuracy level as traditional
thresholding algorithms.  We expect that these techniques will be
broadly useful to the statistical parsing community.

Finally, in Chapter \ref{ch:PFG} we gave a state-of-the-art parsing
formalism that could be used to compute inside and outside
probabilities.  While others \cite{Charniak:97a,Collins:97a}
independently worked in a very similar direction, we provided several
unique contributions, including: an elegant theoretical framework; a
useful way of breaking rules into pieces; an analysis of the value of
each feature; and excellent wordless results.  We think the
theoretical structure of PFGs makes a good platform for future parsing
research, allowing a variety of researchers to phrase their models in
a unified framework.

At the end of a thesis like this, the question of the future direction
of statistical parsing naturally arises.  Given that statistical
parsing has borrowed so much from speech recognition, it is only
natural to compare the progress of statistical parsing systems to the
progress of speech recognition systems.  In particular, over the last
ten or fifteen years, the progress of the speech recognition community
has been stunning.  Statistical parsing has made progress at a similar
pace, but for a much shorter period of time.  Given that a
state-of-the-art speech system requires perhaps 15 person years of
work to develop, while a state-of-the-art parsing system requires only
one or two, quite a bit of progress remains to be made, if only by
integrating everything we know how to do into a single system.  If the
most useful lessons of this thesis were combined with the most useful
lessons of others \cite[{\em inter
alia}]{Magerman:95a,Collins:97a,Ratnaparkhi:97a,Charniak:97a}, we
assume a much better parser would result.  All of the techniques used
in this thesis would fit well in such a system.

Recently, there has been work towards using statistical parsing-style
methods as the framework for relatively complete understanding systems
\cite{Miller:96a,Epstein:96a}.  As this higher level work progresses,
it seems likely that the techniques developed in this thesis will be
be applicable.  Our work on maximizing the right criteria and on
thresholding algorithms for search could both probably be applied in
higher level domains.  Our work on PFGs could even serve as the
framework for such new approaches, with features used to store or
encode the semantic representation.

The lessons learned in this thesis are clearly applicable to future
work in statistical NLP, but they are also useful more broadly within
computer science.  The general lessons include:
\begin{itemize}
\item Maximize the true criterion, or as close as you can get.
\item Maximize pieces correct, using the overall probability of each
piece; this can be easier, and sometimes more effective, than
maximizing the whole.
\item Guide search using the best approximations to future and past
you can find, including local, global, and multiple-pass techniques.
It may be possible to get better results by combining all of these
techniques together.
\end{itemize}
Our work in Chapter \ref{ch:semi} presented a general framework, which
we used for statistical parsing.  As pointed out by
\newcite{Tendeau:97b}, this kind of framework can be used to encode
almost any dynamic programming algorithm, so the techniques of
semiring parsing can actually be applied widely.

In summary, we have shown how to use the inside-outside probabilities
and approximations to them to improve accuracy and speed parsing, and
we have provided the tools to find these probabilities.  These
techniques will be useful both in future work on statistical Natural
Language Processing, and more broadly.

%
%	$Id: s-bib.tex,v 1.1 1996/02/01 21:38:02 sfc Exp $
%

\bibliography{master}

\end{document}